\documentclass{article}
\usepackage{fullpage}
\usepackage[utf8]{inputenc}
\usepackage[english]{babel}
\usepackage[toc,page, title, titletoc]{appendix}
\usepackage{wrapfig}
\usepackage{authblk} 

\usepackage{amssymb}
\usepackage{pifont} 
\usepackage{amsmath,bm}
\usepackage{bbm}
\usepackage{amsthm}
\usepackage{thmtools, thm-restate}
\usepackage[hidelinks]{hyperref}
\usepackage[noabbrev,capitalize]{cleveref}
\usepackage{aligned-overset}

\newtheorem{theorem}{Theorem}[subsection]
\newtheorem{defin}[theorem]{Definition}
\newtheorem{prop}[theorem]{Proposition}

\usepackage[table]{xcolor}
\usepackage[column=0]{cellspace}
\def \vec #1{\texorpdfstring{\bm{#1}}{#1}}
\def \mat #1{\texorpdfstring{\bm{#1}}{#1}}

\usepackage{caption}
\usepackage{subcaption}
\usepackage[export]{adjustbox}
\usepackage{graphicx}
\usepackage{rotating}
\usepackage{mwe}
\graphicspath{{images/}{{\subfix{images/}}}}
\usepackage{float} 

\usepackage{subfiles} 

\usepackage{xcolor}

\definecolor{asym-struc}{RGB}{0,254,0}
\definecolor{trans-struc}{RGB}{127,255,127}
\definecolor{asym-osc}{RGB}{0,149,254}
\definecolor{trans-osc}{RGB}{127,202,255}
\definecolor{asym-cyc}{RGB}{254,0,0}
\definecolor{trans-cyc}{RGB}{255,127,127}

\title{A Tutorial on the Spectral Theory of Markov Chains}

\author[1]{Eddie Seabrook\thanks{Email: eddie.seabrook@ini.rub.de, ORCID-ID: 0000-0002-8985-1893}}
\author[1]{Laurenz Wiskott\thanks{Email: laurenz.wiskott@ini.rub.de, ORCID-ID: 0000-0001-6237-740X}}
\affil[1]{Institut für Neuroinformatik, Ruhr-Universität Bochum}
\date{\today}

\begin{document}

\maketitle

\begin{abstract}
Markov chains are a class of probabilistic models that have achieved widespread application in the quantitative sciences. This is in part due to their versatility, but is compounded by the ease with which they can be probed analytically. This tutorial provides an in-depth introduction to Markov chains, and explores their connection to graphs and random walks. We utilize tools from linear algebra and graph theory to describe the transition matrices of different types of Markov chains, with a particular focus on exploring properties of the eigenvalues and eigenvectors corresponding to these matrices. The results presented are relevant to a number of methods in machine learning and data mining, which we describe at various stages. Rather than being a novel academic study in its own right, this text presents a collection of known results, together with some new concepts. Moreover, the tutorial focuses on offering intuition to readers rather than formal understanding, and only assumes basic exposure to concepts from linear algebra and probability theory. It is therefore accessible to students and researchers from a wide variety of disciplines.\\

\end{abstract}

\tableofcontents

\newpage
\section{Introduction}
\label{Intro}
Markov chains are a versatile tool for modelling stochastic processes, and have been applied in a wide variety of scientific disciplines, such as biology, computer science, and finance \cite{Pardoux2010}. This is unsurprising considering the number of practical advantages they offer: (i) they are easy to describe analytically, (ii) in many domains they make complex computations tractable, and (iii) they are a well understood model type, meaning that they offer some level of interpretability when used as a component of an algorithm. Furthermore, as we show in this tutorial, Markov chains are temporal processes that take place on graphs. This makes them particularly suitable for modeling data generating processes that underlie time series and graph data sets, both of which have received much attention in the fields of machine learning and data mining \cite{Aggarwal2015}.

The application of Markov chains requires the assumption that at least some aspect of the process being modelled has no memory. An important consequence of this assumption is that the process can be described in detail using a transition matrix. Furthermore, there exists a rich framework for describing distinct features of such processes based on the eigenvalues and eigenvectors of this matrix. This tutorial provides an in-depth exploration of this framework, making use of tools from probability theory, linear algebra and graph theory. Since the work is intended for readers from diverse academic backgrounds, we concentrate on providing intuition for the tools used rather than strict mathematical formalism. 

The material presented underlies multiple methods from different areas of machine learning, and instead of exploring these methods individually we focus on the general properties that make Markov chains useful across these domains. Nonetheless, so that readers can appreciate the scope of the tutorial, we now briefly summarize the methods that it is relevant to. In graph-based unsupervised learning, it is related to non-linear dimensionality reduction techniques such as \textit{Laplacian eigenmaps} \cite{Belkin2001,Belkin2003} and \textit{spectral clustering} \cite{Weiss1999,Ng2001,vonLuxburg2007}. These two closely related methods both aim to represent data sets in a way that preserves local geometry, and are traditionally formulated using graph Laplacians. However, one line of work on spectral clustering instead uses Markov chains \cite{Meila2000,Meila2001,Tishby2001,Saerens2004,Liu2011,Meila2007,Huang2006}. Furthermore, the method of \textit{diffusion maps} \cite{Coifman2005,Coifman2006} is a generalization of Laplacian eigenmaps that is based on Markov chains, and can be tuned to different length scales in a graph, thereby allowing a multiscale geometric analysis of data sets. An in-depth survey of Laplacian eigenmaps, spectral clustering, diffusion maps, as well as other related methods can be found in \cite{Ghojogh2021}. In the domain of time series analysis, the tutorial is relevant to slow feature analysis (SFA) \cite{Wiskott2002}, a dimensionality reduction technique that is based on the notion of temporal coherence and is conceptually related to Laplacian eigenmaps \cite{Sprekeler2011}. The ideas underlying Laplacian eigenmaps and spectral clustering have also been extended to classification problems, both for labelled \cite{Kamvar2003}, and partially labelled data sets \cite{Szummer2001,Joachims2003,Zhou2005}. Lastly, the material presented in this tutorial also forms the basis of various approaches to value function approximation in reinforcement learning, such as Mahadevan's proto-value functions \cite{Mahadevan2005a,Mahadevan2007,Johns2007}, Stachenfeld's work on the successor representation \cite{Stachenfeld2014,Stachenfeld2017}, and other closely related methods \cite{Petrik2007,Wu2019}. Something common to many of the applications mentioned thus far is that they assume all underlying graphs to be undirected, or equivalently that the corresponding Markov chain is reversible. This provides a number of guarantees that are crucial for these methods to work, and we explore these guarantees in-depth in this tutorial. In most cases, the extension to the directed/non-reversible setting faces a number of challenges and is still actively researched. We discuss these challenges and present various solutions that have been suggested in the literature.

The rest of the text is organized as follows. In \cref{MCs}, we give a general introduction to discrete-time, stationary Markov chains on finite state spaces and explore some specific types of chains in detail. \cref{Graphs} then gives a formal introduction to graphs in order to provide a more detailed description of Markov chains. In \cref{RWs}, random walks are presented as a canonical transformation that turns any graph into a Markov chain, and the undirected/directed cases are considered separately to better understand the types of Markov chains that they typically give rise to.
\section{Markov Chains}
\label{MCs}
\subsection{Definition}
\label{MCs-def}
\textit{Markov processes} are an elementary family of stochastic models describing the temporal evolution of an infinite sequence of random variables $\mathcal{X}=\{X_t : t\in T\}$, defined on a state space $\mathcal{S}$ and indexed by a time set $T$. Such processes respect the \textit{Markov property}, in which the future evolution is conditionally independent of the past, given the present state of the chain. In this tutorial, we focus on models for which time is discretized, i.e.\ $T=\mathbb{N}_0$, known as \textit{Markov chains}. Furthermore, we restrict our consideration to Markov chains defined on finite state spaces with $|\mathcal{S}|=N$ states. In such settings, the Markov property can be formalized in terms of transition probabilities:
\begin{equation}
    \text{Pr}(X_{t+1}=x|X_1=x_1, X_2=x_2,X_3=x_3,...,X_t=x_t)=\text{Pr}(X_{t+1}=x|X_t=x_t)
\end{equation}If these probabilities are themselves independent of time, the Markov chain is said to be \textit{homogeneous}, and its evolution can be fully described by one-step transition probabilities between pairs of states in $\mathcal{S}$: $\text{Pr}(X_{t+1}=s_j|X_{t}=s_i)=P_{ij}$. Collectively, these probabilities can be represented as an $N\times N$ right-stochastic matrix:
\begin{equation}
    \mat{P}=\left(
    \begin{array}{cccc}
       P_{11} & P_{12} & \cdots & P_{1N} \\
       P_{21} & P_{22} & \cdots & P_{2N} \\
       \vdots & \vdots & \ddots & \vdots \\
       P_{N1} & P_{N2} & \cdots & P_{NN} \\
    \end{array}
    \right)
    \label{eq:Pmat}
\end{equation}
which is called the \textit{transition matrix} of the Markov chain and has the property that the rows sum to one, i.e.\ $\sum_{j=1}^NP_{ij}=1$ $\forall i$.

\begin{wrapfigure}[12]{r}{0.3\textwidth}
\vspace{-11pt} 
  \centering
  \includegraphics[width=0.2\textwidth]{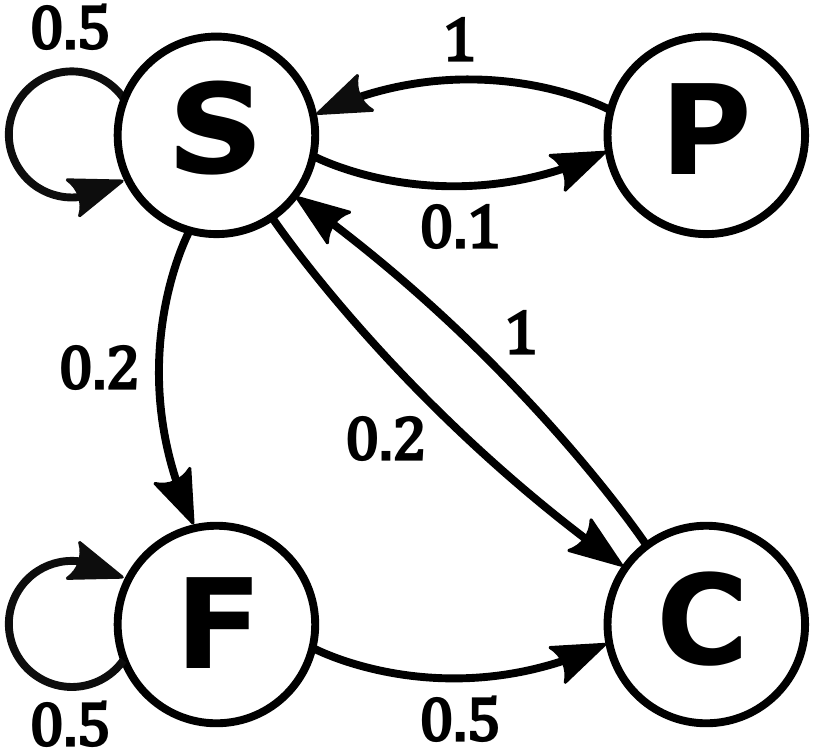}
  \caption{The transition graph of a Markov chain describing the activities of a PhD student.}
  \label{fig:TransGraph}
\end{wrapfigure}

Markov chains can also be depicted visually in the form of a graph, with the state space $\mathcal{S}$ drawn as a collection of circles and labelled arrows between these circles representing the non-zero transition probabilities $P_{ij}$. We call this diagram the \textit{transition graph} of a Markov chain. A formal introduction to the mathematics of graphs is given in \cref{Graphs}, but until then transition graphs are simply used as an illustrative tool.

As an example, imagine you are a PhD student who wants to evaluate how efficiently you work. In order to simplify your analysis, you posit that at any time of a working day you are doing one of four activities: (i) studying, (ii) speaking to your professor, (iii) eating food, and (iv) drinking coffee, which you denote as a set of states $s_1=S$, $s_2=P$, $s_3=F$ and $s_4=C$, respectively. As a further simplification, you assume that transitions between these activities are Markovian. After monitoring your activities for a few days, you come up with a set of empirical transition probabilities which you use to construct a transition graph, shown in \cref{fig:TransGraph}, and the following transition matrix:
\begin{equation}
\label{eq:TransMat}
    \mat{P}=\bordermatrix{ & S & P & F & C \cr
      S & 0.5 & 0.1 & 0.2 & 0.2\cr
      P & 1 & 0 & 0 & 0\cr
      F & 0 & 0 & 0.5 & 0.5\cr
      C & 1 & 0 & 0 & 0}
\end{equation}

With either of these two representations, it is straightforward to generate a realization of this Markov chain. To do this, we first need to pick a starting activity. Suppose that you are studying at time $t=0$, then all activities are possible at $t=1$. In order to choose from these possibilities, one must sample from a probability vector equal to the first row of $\mat{P}$, i.e.\ $\text{Pr}(X_1|X_0=S)=(0.5,0.1,0.2,0.2)$. If our sample yields $X_1=C$, then this becomes the current state and we repeat the process. Doing this iteratively can generate sequences of arbitrary length, for example:
\begin{equation*}
    S - C - S - F - F - C - S - P - S - F - ...
\label{eq:SingleTraj}
\end{equation*}
which we refer to as a \textit{trajectory} in the state space $\mathcal{S}$. As is often the case when studying a stochastic process, generating single trajectories is rather uninformative since it provides no collective description of how the process \textit{tends} to evolve. Naively, one way we could try to achieve such a description would be to perform a type of Monte Carlo sampling by generating several trajectories from the same starting state and summarizing the frequency with which future activities occur.

\begin{figure}[h]
     \centering
     \begin{subfigure}[b]{0.20\textwidth}
     \includegraphics[width=\textwidth]{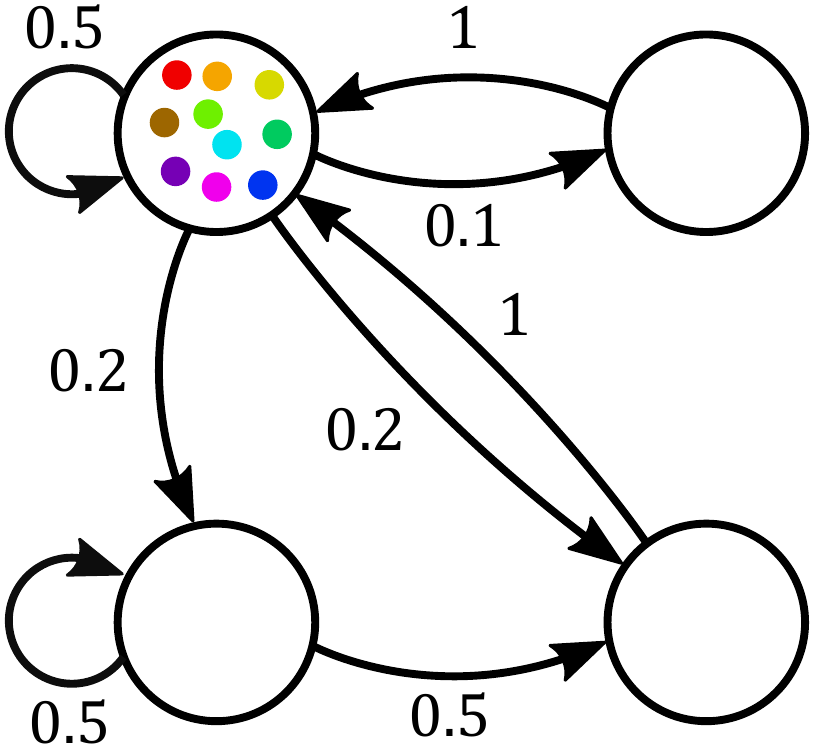}
     \caption{$n=10, t=0$}
     \end{subfigure}
     \hfill
     \begin{subfigure}[b]{0.20\textwidth}
     \centering
     \includegraphics[width=\textwidth]{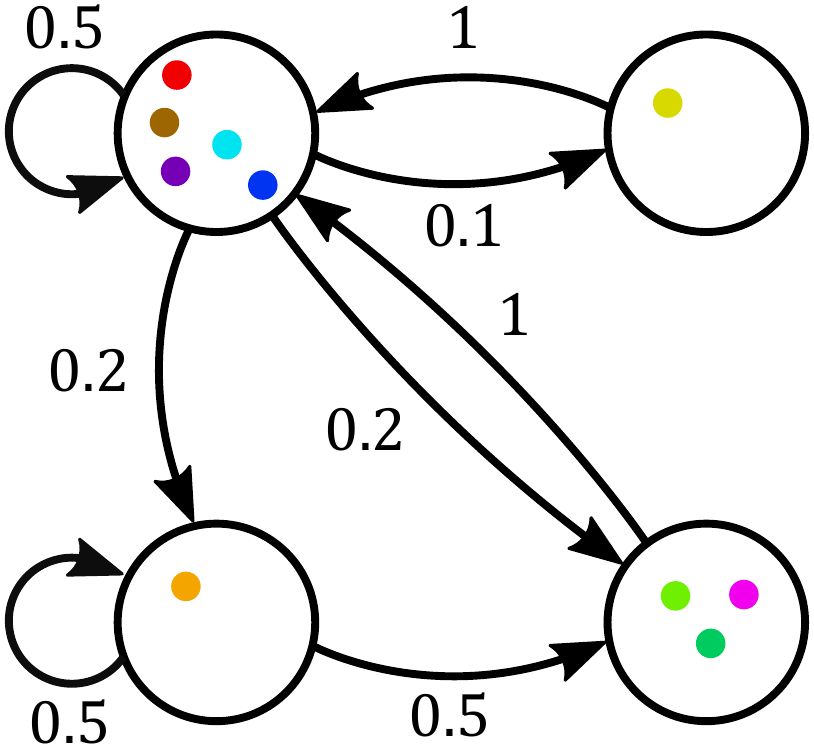}
     \caption{$n=10, t=1$}
     \end{subfigure}
     \hfill
     \begin{subfigure}[b]{0.20\textwidth}
     \centering
     \includegraphics[width=\textwidth]{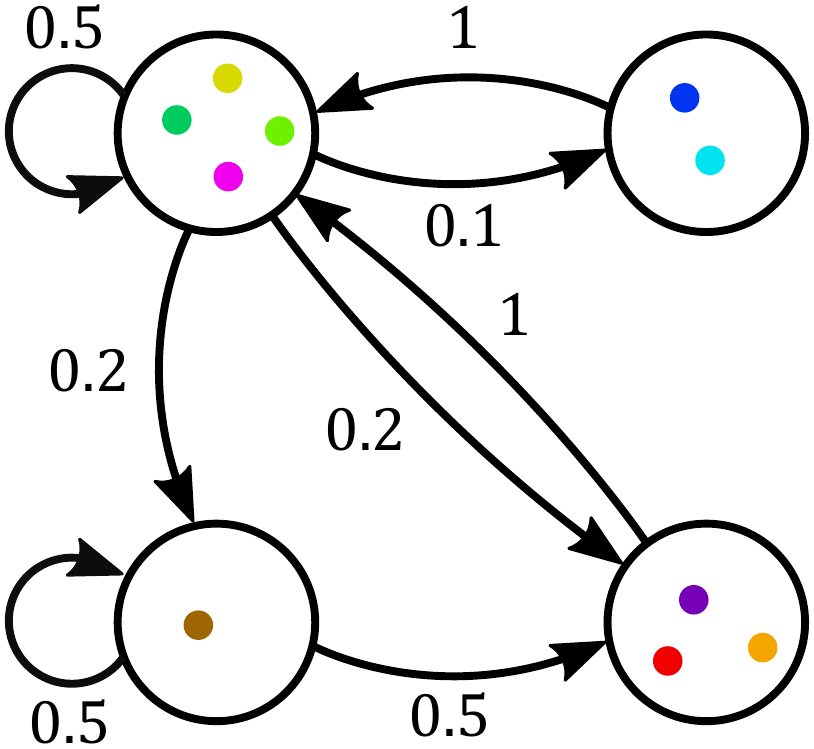}
     \caption{$n=10, t=2$}
     \end{subfigure}
     \hfill
     \begin{subfigure}[b]{0.20\textwidth}
     \centering
     \includegraphics[width=\textwidth]{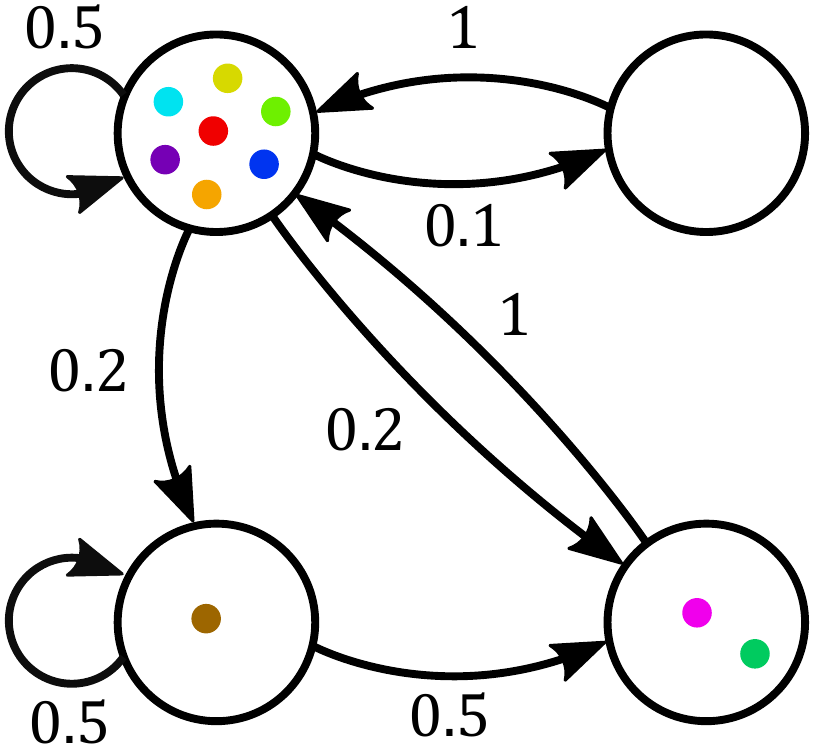}
     \caption{$n=10, t=3$}
     \end{subfigure}
     \par\bigskip
     \begin{subfigure}[b]{0.20\textwidth}
     \includegraphics[width=\textwidth]{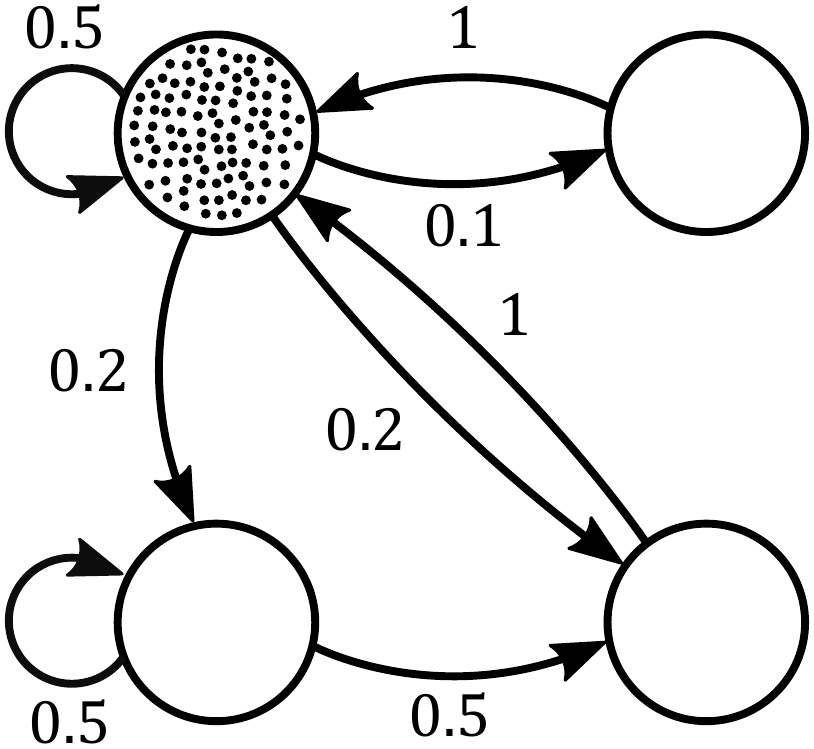}
     \caption{$n=100, t=0$}
     \end{subfigure}
     \hfill
     \begin{subfigure}[b]{0.20\textwidth}
     \centering
     \includegraphics[width=\textwidth]{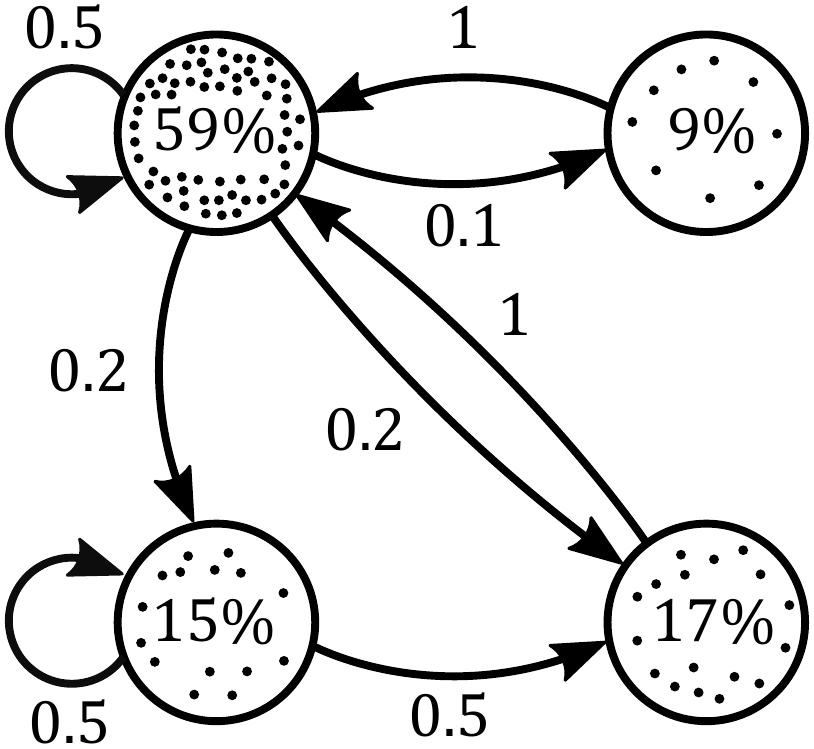}
     \caption{$n=100, t=1$}
     \end{subfigure}
     \hfill
     \begin{subfigure}[b]{0.20\textwidth}
     \centering
     \includegraphics[width=\textwidth]{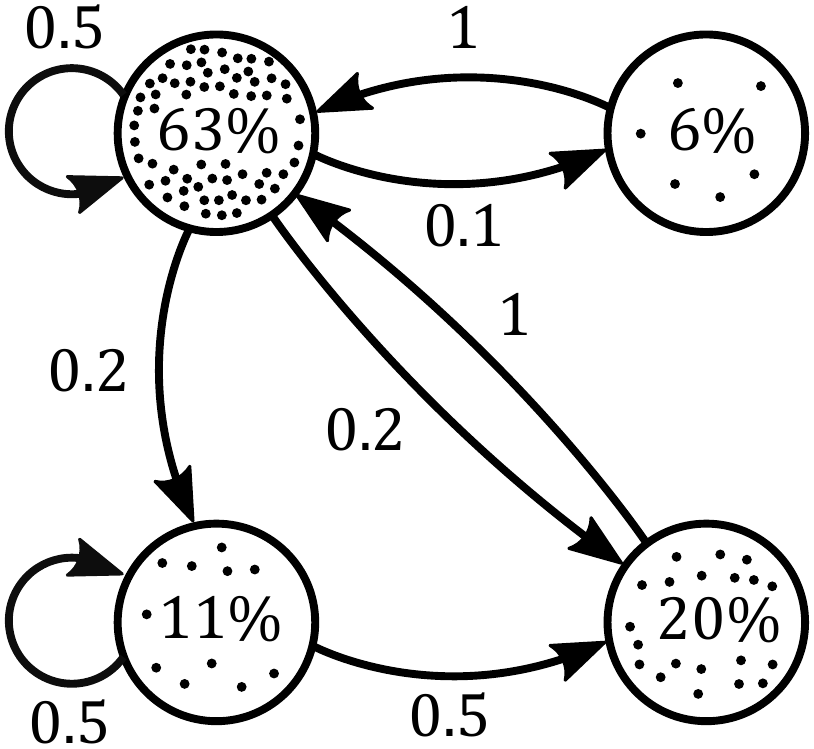}
     \caption{$n=100, t=2$}
     \end{subfigure}
     \hfill
     \begin{subfigure}[b]{0.20\textwidth}
     \centering
     \includegraphics[width=\textwidth]{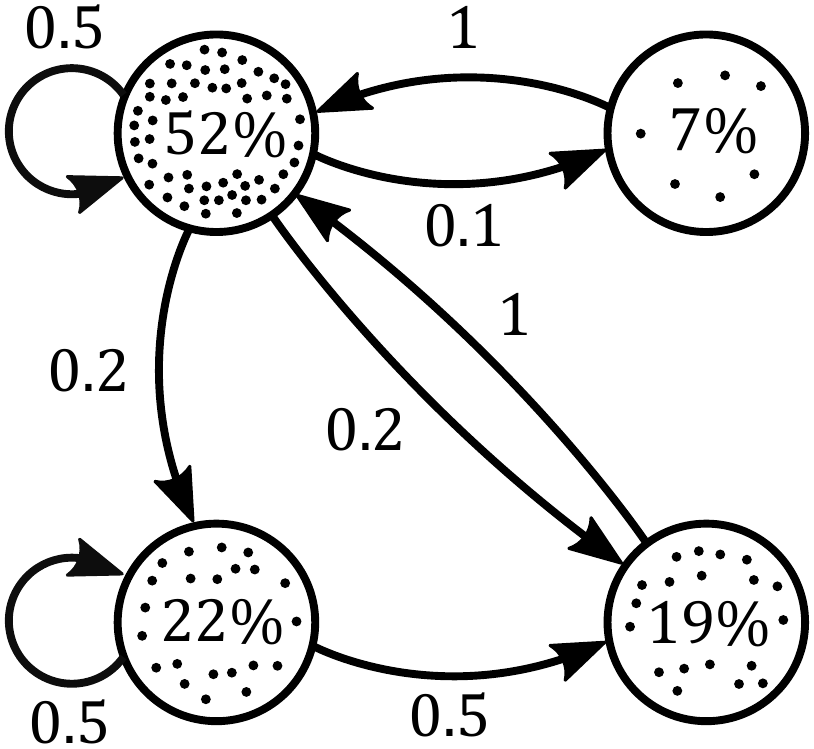}
     \caption{$n=100, t=3$}
     \end{subfigure}
     \par\bigskip
     \begin{subfigure}[b]{0.20\textwidth}
     \includegraphics[width=\textwidth]{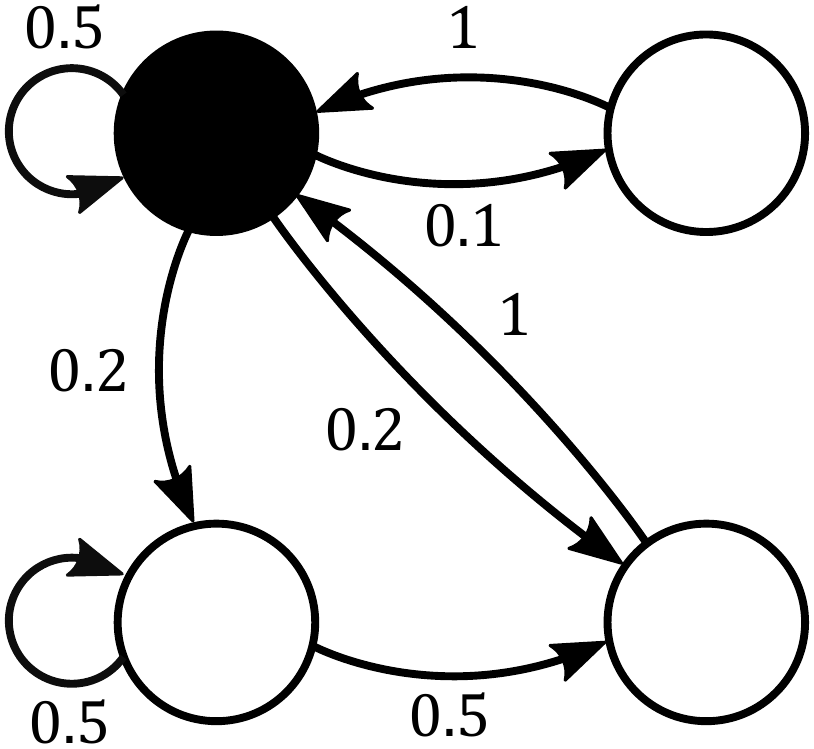}
     \caption{$n=1000, t=0$}
     \end{subfigure}
     \hfill
     \begin{subfigure}[b]{0.20\textwidth}
     \centering
     \includegraphics[width=\textwidth]{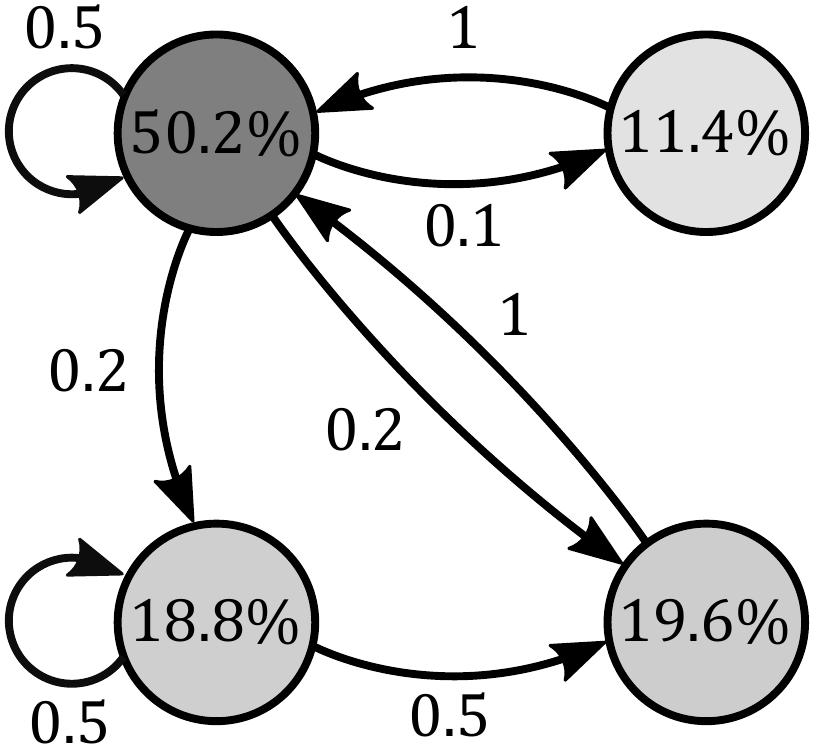}
     \caption{$n=1000, t=1$}
     \end{subfigure}
     \hfill
     \begin{subfigure}[b]{0.20\textwidth}
     \centering
     \includegraphics[width=\textwidth]{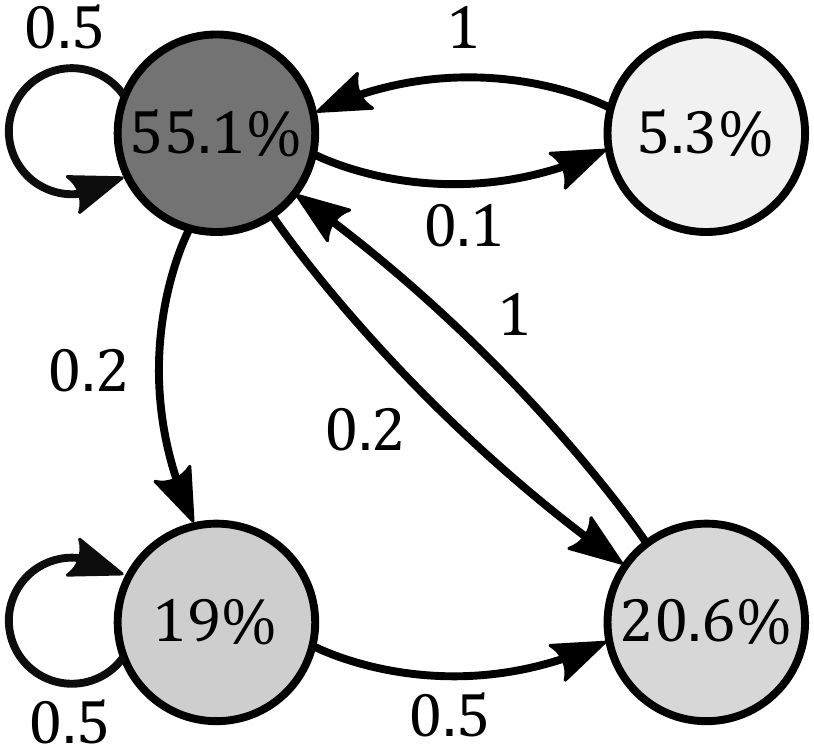}
     \caption{$n=1000, t=2$}
     \end{subfigure}
     \hfill
     \begin{subfigure}[b]{0.20\textwidth}
     \centering
     \includegraphics[width=\textwidth]{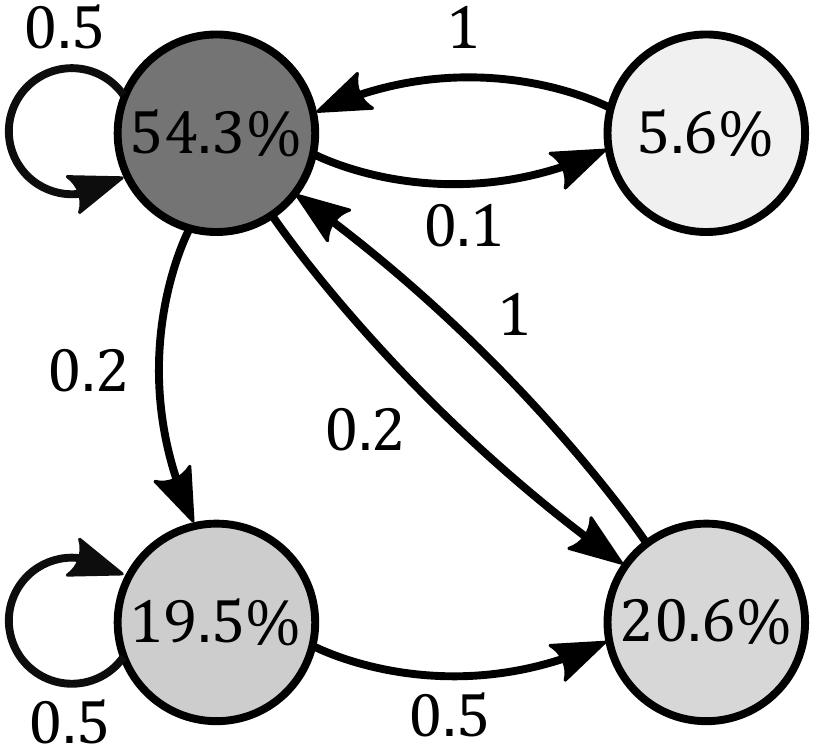}
     \caption{$n=1000, t=3$}
     \end{subfigure}
    \caption{Simulations of the chain in \cref{fig:TransGraph}, each of length $4$ and with starting state $s_1=S$. (a-d) shows $n=10$ simulations of the chain, with colors indicating each trajectory, (e-h) shows $n=100$ simulations, this time all depicted in black, and (i-l) shows the outcome of $n=1000$ simulations via gray scale shadings. In (e-h) and (i-l) percentage values indicate the relative population of each state at each time point.}
    \label{fig:Trajectories}
\end{figure}

In \cref{fig:Trajectories}(a-d), we do this for $n=10$ trajectories of length $4$, each starting with $X_0=S$. Each trajectory is depicted in a specific color, and consists of points in a transition graph plotted across $4$ time points, so that the position of each point indicates a state that one of the trajectories is in at time $t$. Other than a slight bias towards studying ($S$) at each time point, it is hard to pick out any clear patterns using only these $10$ trajectories. \cref{fig:Trajectories}(e-h) shows similar plots for $n=100$, but with all points colored black and the relative occupation of each state for $t>0$ indicated by a percentage. Finally, we increase the number of trajectories to $n=1000$ in \cref{fig:Trajectories}(i-l). Percentages are again used to indicate the relative state occupations, but instead of representing the trajectories with dots we color each state in gray-scale based on the percentage values. Comparing all the plots in this figure, one can note that in the first row it is possible to track each of the individual trajectories, whereas in the second and third rows the focus is instead on approximating the relative probability of doing each activity at each time.

The plots of \cref{fig:Trajectories} represent collections of random experiments, and therefore running them again would not yield exactly the same outcomes. However, taking a frequentist interpretation of probability, we can ask: In the limit $n\to\infty$, how often is each activity done at time $t$? The answer to this question for $t=1$ is given by the vector $\text{Pr}(X_1|X_0=S)=(0.5,0.1,0.2,0.2)$, from which we sample at each time point in a trajectory. For $t=2$, a distribution vector can be calculated by evaluating all the possible trajectories that lead up to each state after two steps. For example, first consider the probability with which we are drinking coffee at $t=2$. Clearly, there are two ways this can happen: (i) $S-S-C$ (i.e.\ $s_1-s_1-s_4$), (ii) $S-F-C$ (i.e.\ $s_1-s_3-s_4$). By combining the corresponding probabilities, we get:
\begin{align}
    \text{Pr}(X_2=s_4|X_0=s_1)&=\sum_{j=1}^4 \text{Pr}(X_1=s_j|X_0=s_1)\text{Pr}(X_2=s_4|X_1=s_j)\\
    &=\sum_{j=1}^4 P_{1j}P_{j4}\\
    &=P_{11}P_{14}+P_{12}P_{24}+P_{13}P_{34}+P_{14}P_{44}\\
    &=0.1+0+0.1+0\\
    &=0.2
\end{align}
By performing a similar calculation for the other activities at $t=2$, we get the following probability vector: $\text{Pr}(X_1|X_0=S)=(0.55, 0.05, 0.2, 0.2)$. A number of comments can be made at this stage. Firstly, while we can extend this type of calculation to $t>2$, this quickly becomes unfeasible to do by hand as the number of steps increases. It turns out that there is a simple mathematical formalism which makes these computations both more efficient and more interpretable. We introduce this formalism in the following section. Secondly, we can also apply the distribution picture at $t=0$. In the example we gave, we always started in the same state, but we can easily generalize this to the case where $X_0$ is not fully determined. For example, $\text{Pr}(X_0)=(0.5,0,0,0.5)$ indicates that studying and drinking coffee both occur at $t=0$ with probability $0.5$. Lastly, while the trajectories of a Markov chain were random, the two distributions we obtained for $t=1$ and $t=2$ were fully determined by our initial condition of $X_0=s_1=S$. Thus, moving from the trajectory perspective to the distribution perspective rather interestingly makes the evolution of our Markov chain look deterministic.

\subsection{Evolution via matrix-vector multiplication}
\label{MCs-Evol}
The structure and action of $\mat{P}$, just like any other matrix, can be evaluated using tools from linear algebra. In particular, $\mat{P}$ can either multiply column vectors from the left, or row vectors from the right. While typical conventions formulate matrix multiplication in the former way, the latter is more common for right stochastic matrices due to its semantic interpretation. Nonetheless, as both operations offer their own insight into the descriptive capacity of Markov chains, we outline both in this section.

For any vector $\vec{x}\in \mathbb{R}^N$, we see that by multiplying it in row form with $\mat{P}$ from the right gives a new vector $\vec{z}\in \mathbb{R}^N$: 
\begin{equation}
    \vec{z}^T=\vec{x}^T\mat{P}
\end{equation}
where $z_j=\sum_{i=1}^N x_iP_{ij}$. By summing over the elements of $\mat{z}$, we see that:
\begin{equation}
    \sum_{j=1}^N z_j = \sum_{j=1}^N \sum_{i=1}^N x_iP_{ij} = \sum_{i=1}^N x_i \underbrace{\sum_{j=1}^N P_{ij}}_{=1} = \sum_{i=1}^N x_i
\end{equation}
meaning that multiplying with $\mat{P}$ from the right preserves the sum over vector entries. Furthermore, note that since all entries of $\mat{P}$ are non-negative, if $\vec{x}$ is non-negative then so too is $\vec{z}$. These two properties are particularly useful when dealing with probability vectors, which by definition both sum to $1$ and have non-negative entries, since it means that a probability vector $\vec{x}$ is mapped into another probability vector when multiplied from the right by $\mat{P}$.

As we have seen, probability vectors are a suitable representation for tracking the evolution of a Markov chain, and as a shorthand we describe the distribution of a chain at time $t$ by $\vec{\mu}(t)=(\mu_1(t), \mu_2(t), ... , \mu_N(t))^T$. Such a distribution can easily be evolved into another distribution describing the chain at time $t+1$. To see this, consider the probability of being in some state $s_j$ at time $t+1$, i.e.\ $\mu_j(t+1)$. This depends both on the probability of being in each state $s_i$ at time $t$, i.e.\ $\mu_i(t)$, as well the probability of making a transition from each state $s_i$ to $s_j$, i.e.\ $P_{ij}$. By summing over all possible states $s_i$, we therefore see that:
\begin{equation}
\label{eq:Probt+1}
    \mu_j(t+1)=\sum_{i=1}^N \mu_i(t)P_{ij}
\end{equation}
Using these probabilities, we can now form the probability vector $\vec{\mu}(t+1)=(\mu_1(t+1), \mu_2(t+1), ... , \mu_N(t+1))^T$, which is described by the vector-matrix analogue of \cref{eq:Probt+1}:
\begin{equation}
    \vec{\mu}(t+1)^T=\vec{\mu}(t)^T\mat{P}
    \label{eq:Pmult}
\end{equation}
Thus, multiplying probability vectors with $\mat{P}$ from the right represents the one-step evolution of a Markov chain. Furthermore, this operation can be extended to multiple steps of evolution by raising the power of $\mat{P}$:
\begin{align}
    \vec{\mu}(t+k)^T&=\vec{\mu}(t+k-1)^T\mat{P}\\
    &=\vec{\mu}(t+k-2)^T\mat{P}\mat{P}\\
    &\;\;\vdots\\
    &=\vec{\mu}(t)^T\underbrace{\mat{P}\cdots\mat{P}}_{k\;\text{times}}\\
    &=\vec{\mu}(t)^T\mat{P}^k \label{eq:MultiMult}
\end{align}
where it is straightforward to establish that $\mat{P}^k$ is itself stochastic. The fact that the $k$-step evolution of a chain is represented simply by $\mat{P}^k$ is a result known as the Chapman-Kolmogorov equation \cite{Stewart1994}, and it tells us that the transition matrix is the only thing needed to evolve a starting distribution of a Markov chain arbitrarily far into the future.

A particularly special type of distribution for a Markov chain is one which is invariant under its evolution:

\begin{defin}[Stationary distribution]
A distribution $\vec{\pi}$ that is invariant when multiplied by $\mat{P}$ from the right, i.e.
\begin{equation}
    \label{eq:GB1}
    \vec{\pi}^T=\vec{\pi}^T \mat{P}
\end{equation}
is said to be a stationary distribution of the Markov chain, and for finite state spaces it is guaranteed that at least one such distribution exists.
\end{defin}
\noindent The set of $N$ equations implied by \cref{eq:GB1} are often referred to as the equations of \textit{global balance}. To understand why, consider the $j$-th equation in this set:
\begin{equation}
    \label{eq:GB2}
    \pi_j =\sum_{i=1}^N \pi_iP_{ij}
\end{equation}
The right-hand term in \cref{eq:GB2} represents the total \textit{flow} of probability mass into state $s_j$ from all other states (including $s_j$ itself when $P_{jj}\neq 0$). Since $\pi_j$ remains invariant under this flow, then an equal amount of probability mass must be flowing from $s_j$ to all other states in $\mathcal{S}$, hence the name global balance. An important implication of \cref{eq:GB1} and \cref{eq:GB2} is that when a chain is in a stationary distribution at time $t$, it remains there for all future time steps. We can therefore interpret such distributions as a type of \textit{steady state} of the underlying process.

We can also ask how much probability mass moves from one state to another in a given stationary distribution $\vec{\pi}$. This is described by the following matrix:

\begin{defin}[Flow matrix]
Given that a Markov chain is in one of its stationary distributions $\vec{\pi}$, the corresponding flow matrix is defined as:
\begin{equation}
    \label{eq:FlowMat}
    \mat{F}^{\vec{\pi}}:=\mat{\Pi} \mat{P}
\end{equation}
where $\mat{\Pi}:=\text{diag}(\vec{\pi})$ is a diagonal matrix consisting of the entries of $\vec{\pi}$, and $(\mat{F}^{\vec{\pi}})_{ij}=\pi_i P_{ij}$ is the stationary flow of probability mass from one state $s_i$ to another state $s_j$.
\end{defin}

So far we have only considered row vectors, and their interpretation as probability vectors was fitting due to the way they transform when multiplied by $\mat{P}$. For reasons that we outline below, this interpretation does not apply in the case of column vectors, which are multiplied by $\mat{P}$ from the left. To show this, we first assume that $\vec{x}=(x_1, x_2, ..., x_N)^T$ is any vector in $\mathbb{R}^N$. Since this vector assigns a value to each state, we can interpret it as a function on $\mathcal{S}$. If the chain is described at time $t$ by $\vec{\mu}(t)$, we can use this distribution to calculate the expected value of $\vec{x}$:
\begin{align}
    \mathbb{E}_{\mu(t)}(\vec{x})&=\sum_i \mu_i(t)x_i\\
    &=\vec{\mu}(t)^T \vec{x}
\end{align}

If we want to look $k$ time steps into the future, then expected values are calculated in the same way but with the additional evolution of $\vec{\mu}(t)$ in accordance with \cref{eq:MultiMult}:
\begin{align}
    \mathbb{E}_{\mu(t)}^k(\vec{x})&=\mathbb{E}_{\mu(t+k)}(\vec{x})\\
    &=\vec{\mu}(t+k)^T\vec{x}\\
    &=\vec{\mu}(t)^T\mat{P}^k\vec{x} \label{eq:LPR}\\
    &=\mathbb{E}_{\mu(t)}(\mat{P}^k\vec{x}) 
\end{align}
Hence, given any starting distribution $\vec{\mu}(t)$, the transition matrix can produce expected values of any function on the state space arbitrarily far into the future.

Often, the initial condition of a Markov chain has zero uncertainty, with all probability occupying a single state $s_i$. In such settings, the starting distribution is a row vector with one component equal to $1$ and all others zero - known as a \textit{one-hot vector}, and denoted by $\vec{e}_i$. If we evaluate the $k$-step expectation in \cref{eq:LPR} with $\mat{e}_i$ as a starting distribution, we get:
\begin{align}
     \mathbb{E}_{e_i}^k(\vec{x})&=\vec{e}_i^T\mat{P}^k\vec{x}\\
     &=\sum_j (\mat{P}^k)_{ij}x_j\\
     &=(\mat{P}^k\vec{x})_i
\end{align}
This tells us how a vector $\vec{x}$ is transformed when multiplied by $\mat{P}^k$ from the left - it produces a new vector $\vec{x}'=\mat{P}^k\vec{x}$, whose $i$-th element is the expected value of $\vec{x}$ after $k$ steps, conditioned on the starting state being $s_i$. Formally, this is a conditional expectation:
\begin{equation}
\label{eq:CondExp}
    (\mat{P}^k\vec{x})_i=\mathbb{E}^k(\vec{x}\; |\;X_t=s_i)
\end{equation}
which is the main operation underlying value functions in reinforcement learning \cite{Sutton2018}. 

Summing over the elements of $\vec{x}'$ gives:
\begin{equation}
    \sum_{i=1}^N x'_i=\sum_{i=1}^N\sum_{j=1}^N (\mat{P}^k)_{ij}x_j=\sum_{j=1}^Nx_j\underbrace{\sum_{i=1}^N (\mat{P}^k)_{ij}}_{\neq 1}\neq \sum_{j=1}^N x_j
\end{equation}
since the columns of $\mat{P}^k$, unlike its rows, do not sum to one. Hence, multiplying column vectors with $\mat{P}^k$ does not preserve the vector sum, which is why they are best interpreted as functions rather than distributions.

\subsection{Eigenvalues and eigenvectors}
\label{MCs-EV}

Every real matrix $\mat{A}\in\mathbb{R}^{N\times N}$ can be interpreted as a linear transformation. A central task of linear algebra is to  shed light on the relationship between the numerical properties of a matrix and various aspects of the transformation that it represents. Often, a single matrix represents a combination of several distinct transformations - e.g.\ an object in two or more dimensions can simultaneously be rotated and stretched. Finding the eigenvalues and eigenvectors of a matrix is one way to partition a linear transformation into its component parts and to reveal their relative magnitudes. In this section, we give a brief and informal summary of how this works, and apply this to transition matrices.

An eigenvector of a matrix $\mat{A}$ is a vector which is only multiplied by some number $\lambda$ when multiplied by $\mat{A}$. Like all vectors, they can either be rows, i.e.\ $\mat{l}^T\mat{A}=\lambda\mat{l}^T$, or columns, i.e.\ $\mat{A}\mat{r}=\lambda\mat{r}$, which are known as \textit{left eigenvectors} and \textit{right eigenvectors}, respectively. In both cases, the number $\lambda$ is called the \textit{eigenvalue} of the respective eigenvector. Lastly, it is worth noting that real matrices, including transition matrices, can have complex eigenvalues, i.e.\ $\lambda\in\mathbb{C}$, in which case the corresponding eigenvector is also complex, i.e.\  $\vec{v}\in\mathbb{C}^N$. However, such solutions can only occur in complex conjugate pairs, meaning that $\lambda^*$ and $\vec{v}^*$ are also guaranteed to be an eigenvalue-eigenvector pair, where $^*$ denotes the complex conjugate.

A quick inspection of \cref{eq:GB1} reveals that we have already encountered an example of an eigenvector in the case of a transition matrix: a stationary distribution $\vec{\pi}$ is a left eigenvector of $\mat{P}$ with eigenvalue $\lambda=1$, normalized such that it sums to $1$. Similarly, it is straightforward to see that $\vec{\eta}=(1,1,...,1)^T$ is a right eigenvector with eigenvalue $\lambda=1$:
\begin{equation}
    (\mat{P}\vec{\eta})_i=\sum_{j=1}^N P_{ij}\eta_j=\sum_{j=1}^N P_{ij}=1
\end{equation}
so that $\mat{P}\vec{\eta}=\vec{\eta}$.

We can similarly describe the action of a matrix $\mat{A}$ on a set of $k$ eigenvectors. Let $\mat{Y}_R\in\mathbb{R}^{n\times k}$ be a matrix whose columns are equal to $k$ right eigenvectors $\mat{A}$:
\begin{equation}
\label{eq:VeigMat}
    \mat{Y}_R=
    \begin{pmatrix}
    \vec{r}_1 & \vec{r}_2 & \cdots & \vec{r}_k
    \end{pmatrix}
\end{equation}
Then the action of $\mat{A}$ on this matrix is simply:
\begin{align}
    \mat{AY}_R&=
    \begin{pmatrix}
    \mat{A}\vec{r}_1  & \mat{A}\vec{r}_2 & \cdots & \mat{A}\vec{r}_k
    \end{pmatrix}\\
    &=\begin{pmatrix}
    \lambda_1\vec{r}_1 & \lambda_2\vec{r}_2 & \cdots & \lambda_k\vec{r}_k
    \end{pmatrix}\\
    &=\begin{pmatrix}
    \vec{r}_1 & \vec{r}_2 & \cdots & \vec{r}_k
    \end{pmatrix}
    \begin{pmatrix}
    \lambda_1 & 0 & \cdots & 0\\
    0 & \lambda_2 & \cdots & 0\\
    \vdots & \vdots & \ddots & 0 \\
    0 & 0 & 0 & \lambda_N
    \end{pmatrix}\\
    &=\mat{Y}_R\mat{\Delta} \label{eq:EigEqMat}
\end{align}
If we had alternatively used a matrix $\mat{Y}_L$ with rows equal to left eigenvectors of $\mat{A}$, then an equivalent argument can show that $\mat{Y}_L\mat{A}=\mat{\Delta}\mat{Y}_L$.

In linear algebra, one is often interested in finding \textit{linearly independent} sets of eigenvectors. Without this condition, a set of eigenvectors can contain a lot of redundancy. For example, if $\vec{r}$ is an eigenvector of $\mat{A}$ with eigenvalue $\lambda$, then so too is the vector $c\vec{r}$ for any $c\in\mathbb{C}$. Therefore, it is possible to construct an arbitrarily large set of eigenvectors using $\vec{r}$ alone, and the set of all possible vectors that can be formed in this way is known as the \textit{eigenspace} of $\mat{A}$ corresponding to $\lambda$. If we instead look for a set of linearly independent eigenvectors of $\mat{A}$, there can be at most $N$. When a set of $N$ linearly independent eigenvectors exists, they form a basis for $\mathbb{R}^{N}$ (i.e.\ the domain on which $\mat{A}$ acts). In this case, the set of eigenvectors is called an \textit{eigenbasis}, and the matrix $\mat{A}$ is said to be \textit{diagonalizable}. To justify the use of this latter term, imagine that the matrix $\mat{Y}_R$ contains $N$ linearly independent right eigenvectors. Then, by definition it is full rank, meaning that its inverse $\mat{Y}_R^{-1}$ exists. If we then multiply both sides of \cref{eq:EigEqMat} by this inverse from the left, we get:
\begin{equation}
\label{eq:DiagMat-alt}
    \mat{Y}_R^{-1}\mat{A}\mat{Y}_R=\mat{\Delta}
\end{equation}
Before interpreting this expression, it is worth reflecting on its general form. If two matrices $\mat{B}$ and $\mat{C}$ can be related via $\mat{U}^{-1}\mat{B}\mat{U}=\mat{C}$ for some invertible matrix $\mat{U}$, then they are said to be \textit{similar}. Alternatively, $\mat{C}$ is said to be a \textit{similarity transformation} on $\mat{B}$, and since we could instead write $\mat{B}=\mat{U}\mat{C}\mat{U}^{-1}$ the converse also holds. In words, similar matrices represent the same linear transformation, but expressed in two different bases, where the matrix $\mat{U}$ is known as the \textit{change-of-basis} matrix. In the case of \cref{eq:DiagMat-alt}, this means that $\mat{A}$ behaves like a diagonal matrix when acting on vectors that are expressed in its eigenbasis $\mat{Y}_R$, hence the term diagonalizable.

\cref{eq:DiagMat-alt} can easily be rearranged to get the following expression for $\mat{A}$:
\begin{equation}
\label{eq:DiagMat}
    \mat{A}=\mat{Y}_R\mat{\Delta}\mat{Y}_R^{-1}
\end{equation}
and since this only involves the eigenvalue and eigenvector matrices $\mat{\Delta}$ and $\mat{Y}_R$, it is known as the \textit{eigendecomposition} of $\mat{A}$. Furthermore, multiplying \cref{eq:DiagMat} from the left with $\mat{Y}_R^{-1}$ yields $\mat{Y}_R^{-1}\mat{A}=\mat{\Delta Y}_R^{-1}$, meaning that the rows of $\mat{Y}_R^{-1}$ are a set of $N$ linearly independent left eigenvectors of $\mat{A}$. Therefore:
\begin{align}
    \mat{A}&=\begin{pmatrix}
    \vec{r}_1 & \vec{r}_2 & \cdots & \vec{r}_N
    \end{pmatrix}
    \begin{pmatrix}
    \lambda_1 & 0 & \cdots & 0\\
    0 & \lambda_2 & \cdots & 0\\
    \vdots & \vdots & \ddots & 0 \\
    0 & 0 & 0 & \lambda_N
    \end{pmatrix}
    \begin{pmatrix}
    \vec{l}_1^T\\
    \vec{l}_2^T\\
    \vdots \\
    \vec{l}_N^T
    \end{pmatrix}\\
    &=\begin{pmatrix}
    \lambda_1\vec{r}_1 & \lambda_2\vec{r}_2 & \cdots & \lambda_N\vec{r}_N
    \end{pmatrix}
    \begin{pmatrix}
    \vec{l}_1^T\\
    \vec{l}_2^T\\
    \vdots \\
    \vec{l}_N^T
    \end{pmatrix}\\
    &=\sum_{\omega=1}^N \lambda_\omega \vec{r}_\omega\vec{l}_\omega^T\label{eq:DiagMat2}
\end{align}
where each term $\vec{r}_\omega\vec{l}_\omega^T$ in \cref{eq:DiagMat2} is a matrix of rank $1$ and is known as the \textit{outer product} between $\vec{r}_\omega$ and $\vec{l}_\omega$.\footnote{In component notation, the outer product between two vectors $\vec{x}=(x_1, x_2, \cdots, x_N)^T$ and $\vec{y}=(y_1, y_2, \cdots, y_N)^T$ is:\\
\begin{equation*}
    \vec{x}\vec{y}^T:=
    \begin{pmatrix}
    x_1 \\
    x_2 \\
    \vdots\\
    x_N
\end{pmatrix}
\begin{pmatrix}
    y_1 & y_2 & \cdots & y_N
\end{pmatrix}=
\begin{pmatrix}
    x_1 y_1 & x_1 y_2 & \cdots & x_1 y_N \\
    x_2 y_1 & x_2 y_2 & \cdots & x_2 y_N \\
    \vdots & \vdots   & \ddots & \vdots \\
    x_N y_1 & x_N y_2 & \cdots & x_N y_N
\end{pmatrix}
\end{equation*}} Writing the matrix $\mat{A}$ in this way therefore gives a good insight into how a diagonalizable matrix can be partitioned into distinct modes.

Lastly, if $\vec{l}_\omega$ and $\vec{r}_\gamma$ are left and right eigenvectors of $\mat{A}$ with distinct eigenvalues, i.e.\ $\lambda_\omega\neq\lambda_\gamma$, then:
\begin{align}
\vec{l}_\omega^T\mat{A}\vec{r}_\gamma-\vec{l}_\omega^T\mat{A}\vec{r}_\gamma&=0\\
\lambda_\omega\vec{l}_\omega^T\vec{r}_\gamma-\lambda_\gamma\vec{l}_\omega^T\mat{A}\vec{r}_\gamma&=0\\
(\lambda_\omega-\lambda_\gamma)\vec{l}_\omega^T\vec{r}_\gamma&=0 \label{eq:LeftRightOrthog}
\end{align}
Since $(\lambda_\omega-\lambda_\gamma)=0$ by assumption, this means that $\vec{l}_\omega$ and $\vec{r}_\gamma$ must be orthogonal. Thus, when all eigenvalues are distinct and the corresponding eigenvectors are normalized to have unit euclidean norm, the following relation between the columns of $\mat{Y}_R$ and the rows of $\mat{Y}_R^{-1}$ holds:
\begin{equation}
\label{eq:Biorthog}
    \vec{l}_\omega^T\vec{r}_\gamma=\delta_{\omega\gamma}=\left\{
	\begin{array}{ll}
		1 & \mbox{if } \omega=\gamma\\
		0 & \mbox{otherwise}
	\end{array}
\right.
\end{equation}
Because of this, $\mat{Y}_R^{-1}$ is sometimes called the \textit{dual basis} of $\mat{Y}_R$ and the two sets of vectors are collectively referred to as a \textit{biorthogonal system}. It is worth noting that when a diagonalizable matrix has repeated eigenvalues, there is extra freedom in the choice of left and right eigenvectors. Consequently, for such matrices \cref{eq:Biorthog} is not guaranteed to hold, however there exist certain choices of bases for which it does \cite{Denton2021}.

In the case of a Markov chain, these concepts can be applied when the transition matrix $\mat{P}$ is diagonalizable. In this case, any probability distribution $\vec{\mu}(t)$ can be expressed in terms of the eigenbasis of $\mat{P}$. Since we are treating distributions as row vectors, we do this using the left eigenvectors of $\mat{P}$:
\begin{equation}
\label{eq:ProbeigSum}
    \vec{\mu}(t)^T=\sum_{\omega=1}^N c_\omega \vec{l}_\omega^T
\end{equation}
where the components $c_\omega$ are the coordinates of $\vec{\mu}(t)$ in this basis. We can then use this to re-express the one-step evolution of the chain as:
\begin{align}
\label{eq:Eveig}
    \vec{\mu}(t+1)^T\overset{(\ref{eq:Pmult})}&{=}\vec{\mu}(t)^T\mat{P}\\
    \overset{(\ref{eq:ProbeigSum})}&{=}\bigg(\sum_{\omega=1}^N c_\omega \vec{l}_\omega^T\bigg)\mat{P}\\
    &=\sum_{\omega=1}^N c_\omega \vec{l}_\omega^T \mat{P}\\
    &=\sum_{\omega=1}^N c_\omega \lambda_\omega \vec{l}_\omega^T
\end{align}
which can easily be extended to multiple steps of evolution:
\begin{align}
    \vec{\mu}(t+k)^T\overset{(\ref{eq:MultiMult})}&{=}\vec{\mu}(t)^T\mat{P}^k\\
    &=\sum_{\omega=1}^N c_\omega \lambda_\omega^k \vec{l}_\omega^T\label{eq:Eveigk}
\end{align}
Comparing \cref{eq:MultiMult,eq:Eveigk}, one can see that by working in the eigenbasis of $\mat{P}$ we have transformed the evolution of the Markov chain from a matrix multiplication to a scalar multiplication along each basis vector $\vec{l}_\omega$ by $\lambda_\omega^k$. We have therefore improved our computational complexity from $\mathcal{O}(N^2)$ to $\mathcal{O}(N)$.

This type of manipulation can be done for any diagonalizable matrix, and does not depend on $\mat{P}$ being stochastic. However, one important result we present in \cref{Graphs-Eig} is that all eigenvalues of a transition matrix have absolute value less than or equal to $1$. Using this fact, we can order the eigenvalues of $\mat{P}$ by their absolute value and separate the terms in \cref{eq:Eveigk} corresponding to eigenvalues with $|\lambda_\omega|=1$, and those corresponding to $|\lambda_\omega|<1$:

\begin{equation}
\label{eq:EveigkPart}
    \vec{\mu}(t+k)^T=\underbrace{\sum_{\omega=1}^m c_\omega \lambda_\omega^k \vec{l}_\omega^T}_{\text{persistent}}+\underbrace{\sum_{\omega'=m+1}^Nc_{\omega'} \lambda_{\omega'}^k \vec{l}_{\omega'}^T}_{\text{transient}}
\end{equation}
where $m$ is the number of eigenvalues of $\mat{P}$ with $|\lambda_\omega|=1$. In the long time limit ($k\to\infty$) the terms with $|\lambda_{\omega}|=1$ survive whereas those with $|\lambda_{\omega}|<1$ die off, with $|\lambda|$ measuring the rate of decay in the latter case. This allows us to interpret the first and second sums in \cref{eq:EveigkPart} to represent the \textit{persistent} and \textit{transient} behavior of the Markov chain, respectively.

In the persistent case, we can partition the terms into three types based on the eigenvalues: (i) $\lambda =1$, (ii) $\lambda =-1$, and (iii) $\lambda\in\mathbb{C}$, $|\lambda|=1$. We have already seen that a stationary distribution $\vec{\pi}$ and the vector $\vec{\eta}$ are left and right eigenvectors of type (i), and in both cases these eigenvectors represent fixed structures on the state space that persist when acted on by $\mat{P}$. To capture this property, we call such eigenvectors \textit{persistent structures}. In case (ii), eigenvectors flip their sign when acted on by the transition matrix, i.e.\ $\vec{y}^T\mat{P}=-\vec{y}^T$, and return after two steps, i.e.\ $\vec{y}^T\mat{P}^2=\vec{y}^T$. Eigenvectors of this type therefore correspond to permanent oscillations of probability mass between states in $\mathcal{S}$, and we therefore refer to them as \textit{persistent oscillations}. Eigenvalues of type (iii) are explored in more depth in \cref{Graphs-Eig}, where we show that they are always complex roots of unity, i.e.\ $\lambda^k=1$ for some $k>2$. Therefore, when their corresponding eigenvectors are acted on repeatedly by $\mat{P}$ they are returned to after $k$ steps, i.e.\ $\vec{y}^T\mat{P}^k=\vec{y}^T$. Thus, in analogy to (ii) they describe permanent cycles of probability mass through the state space $\mathcal{S}$, and we therefore call such eigenvectors \textit{persistent cycles}.

\begin{wraptable}{l}{0.45\textwidth}
  \centering
   \begin{tabular}{|0c|0c|0c|}
     \hline
      & Persistent & Transient \\[2ex]
     \hline
     Structure & \cellcolor{asym-struc} $\lambda =1$ & \cellcolor{trans-struc} $\lambda \in (0, 1)$ \\[2ex]
     \hline
     Oscillation & \cellcolor{asym-osc} $\lambda=-1$ & \cellcolor{trans-osc} $\lambda \in (-1, 0)$\\[2ex]
     \hline
     Cycle & \cellcolor{asym-cyc} $\lambda \in \mathbb{C}$, $|\lambda|=1$ & \cellcolor{trans-cyc} $\lambda \in \mathbb{C}$, $|\lambda|<1$\\[2ex]
     \hline
     \end{tabular}
    \caption{Eigenvalues of type (i-iii) and (i'-iii') organized and colored based on whether they are persistent or transient, and whether they are structures, oscillations or cycles.}
    \label{tab:Eigs}
\end{wraptable}

In the transient case, an analogous categorization based on the eigenvalues can be applied, consisting of the following three types: (i') $\lambda\in[0, 1)$, (ii') $\lambda\in (-1,0)$, and (iii') $\lambda \in \mathbb{C}$, $|\lambda|<1$. For type (i'), the corresponding eigenvectors represent perturbations to the persistent behavior that decay over time, i.e.\ $\vec{y}^T\mat{P}=\lambda\vec{y}^T$ where $|\lambda\vec{y}|=|\lambda||\vec{y}|<|\vec{y}|$. We therefore call such eigenvectors \textit{transient structures}. When $|\lambda_j|\approx 1$ these structures describe  sets of states that, on average, a chain spends a long time in before converging, which are known as \textit{metastable sets} \cite{Conrad2016}. Furthermore, $\lambda=0$ can be thought of as a limiting case of type (i') in the sense that any corresponding eigenvector decays infinitely quickly, i.e.\ $\mat{P}\vec{y}=0$, and does not exhibit oscillatory or cyclic behavior. Eigenvalues of type (ii') also decay over time, but like case (ii) their negative sign means that the corresponding eigenvectors exhibit oscillatory behavior when acted on by $\mat{P}$, i.e.\ $\vec{y}^T\mat{P}=-\lambda\vec{y}^T$ where $|\lambda\vec{y}|=|\lambda||\vec{y}|<|\vec{y}|$. We refer to eigenvectors of this type as \textit{transient oscillations}. Eigenvalues of type (iii') generalize those of type (iii) to the transient case, i.e.\ $\vec{y}^T\mat{P}^k=\lambda^k\vec{y}^T$ for some $k>2$ where $|\lambda^k\vec{y}|=|\lambda^k||\vec{y}|<|\vec{y}|$, and we therefore call them \textit{transient cycles}. When $|\lambda|\approx 1$, these cycles can persist for a long time, and have been referred to as \textit{dominant cycles} \cite{Conrad2016}.

\begin{wrapfigure}[16]{r}{0.45\textwidth}
\vspace{-12pt}
  \centering
   \includegraphics[width=0.35\textwidth]{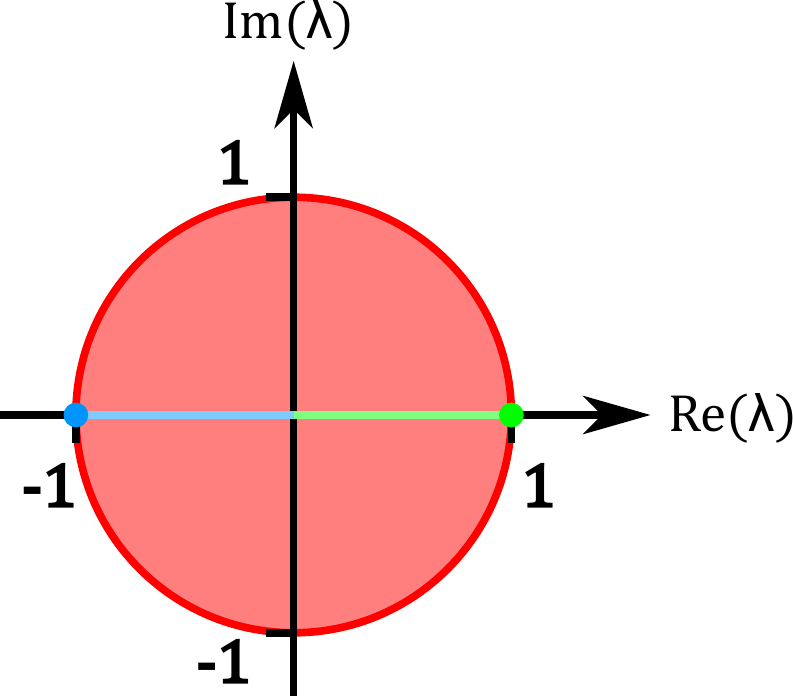}
  \caption{Eigenvalues of type (i-iii) and (i'-iii') plotted based on where they lie on the unit circle, with the same color coding as in \cref{tab:Eigs}.}
  \label{fig:EigsUnitCircle}
\end{wrapfigure}
The above categorizations are summarized in \cref{tab:Eigs}, where structures, oscillations and cycles are colored in green, blue and red, respectively, and the persistent and transient cases are shaded bright and pale, respectively. In \cref{fig:EigsUnitCircle}, we visualize the six different types of eigenvalue using this color scheme by shading the respective regions of the unit circle in which they can occur.

Before moving on, a couple of details are worth pointing out. Firstly, the above analysis is possible only if the transition matrix $\mat{P}$ is diagonalizable. One case in which this is guaranteed to hold is for symmetric matrices. However, given that transition probabilities are rarely pairwise symmetric (i.e.\ $P_{ij}=P_{ji}$), this restriction is clearly too strong. Thus, a more detailed investigation is needed in order to identify the conditions under which the above decomposition can be made, which we do in \cref{Graphs} and \cref{RWs}. For a more in-depth account of the theory of diagonalizable matrices, we recommend \cite{Meyer2000}. Secondly, a quick check reveals that all terms in \cref{eq:EveigkPart} are dependent on the components $w_j$ that describe the starting distribution (\cref{eq:ProbeigSum}). Because of this, in the most general case both the persistent and transient behavior of a Markov chain can be sensitive to initial conditions. In a later section, we consider a particular type of Markov chain for which this is not the case, and provide a simplified analysis of their evolution over time.

\subsection{Classification of states}
\label{MCs-class}
In the following sections, we explore three types of Markov chains. In order to describe each type in detail, we first need to define various properties that apply to individual states, or sets thereof, in a state space $\mathcal{S}$. This is the focus of the current section.

\subsubsection{Communicating classes}
\label{MCs-CommClass}
We start by making the following definitions related to how states in $\mathcal{S}$ are connected:

\begin{defin}[Accessibility]
For two states $s_i$, $s_j \in \mathcal{S}$, we say that $s_j$ is accessible from $s_i$, denoted $s_i \to s_j$, when it is possible to reach $s_j$ from $s_i$ in $k\geq0$ steps, i.e.\ $\exists\;k\; :(\mat{P}^k)_{ij}>0$.
\end{defin}

\begin{defin}[Communication]
Two states $s_i$, $s_j \in \mathcal{S}$, are said to communicate if $s_i \to s_j$ and $s_j \to s_i$, which is denoted $s_i \leftrightarrow s_j$.
\end{defin}

\noindent Communication is a useful property for describing states in a Markov chain, as exemplified by the following result:

\begin{prop}[Communicating class]
Communication is an equivalence relation, meaning that:
\begin{itemize}
    \item $\forall s_i \in \mathcal{S}$, $s_i\leftrightarrow s_i$, since by definition each state can reach itself in $0$ steps, i.e.\ $(\mat{P}^0)_{ii}=(\mat{\mathbbm{1}})_{ii}=1>0$.
    \item if $s_i \leftrightarrow s_j$, then $s_j \leftrightarrow s_i$.
    \item if $s_i \leftrightarrow s_j$ and $s_j \leftrightarrow s_k$,  then $s_i \leftrightarrow s_k$.
\end{itemize}
and the state space $\mathcal{S}$ can be partitioned into communicating classes, each containing states that all communicate with one another.
\end{prop}

\begin{figure}[h]
     \centering
     \begin{subfigure}[t]{0.45\textwidth}
         \centering
         \includegraphics[height=3cm]{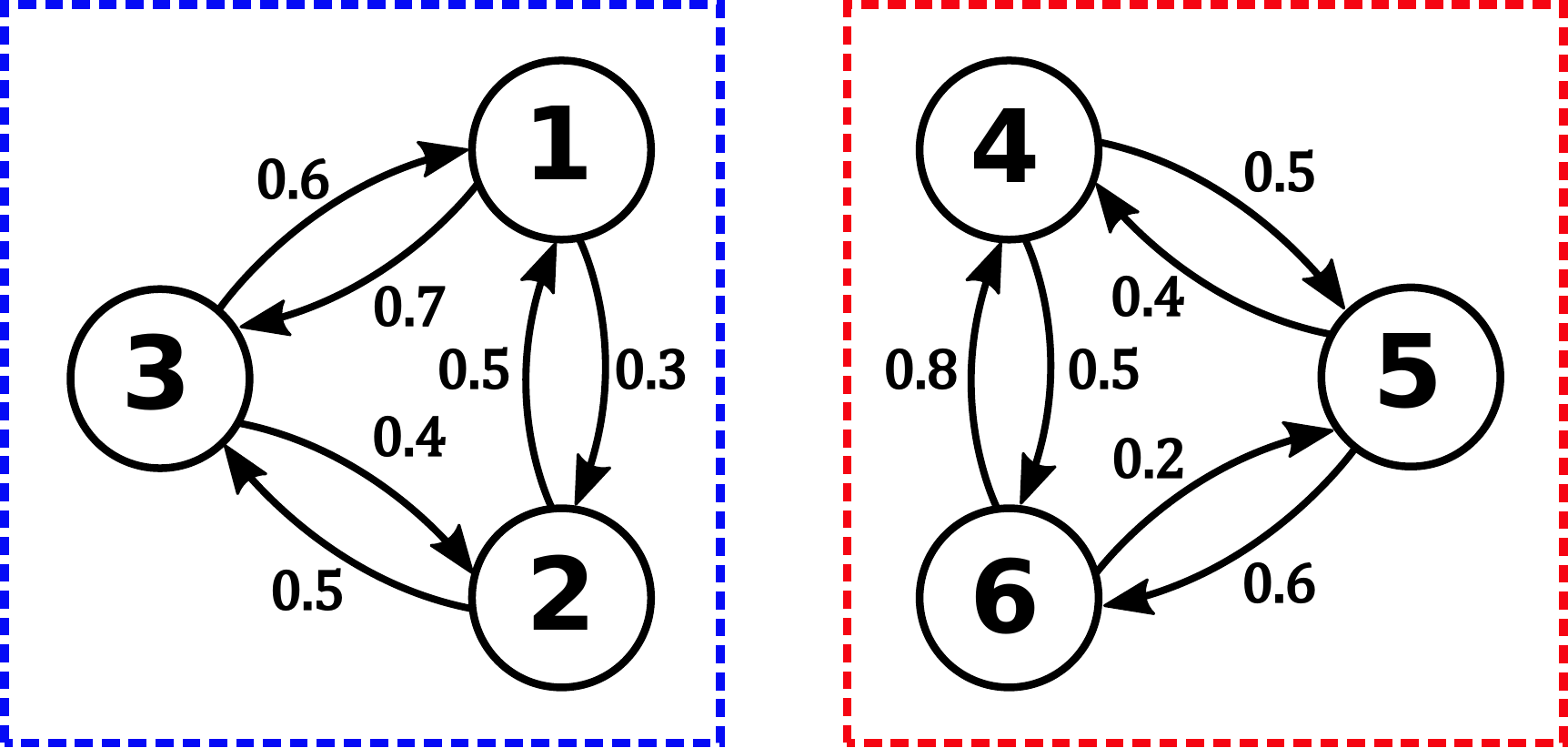}
         \caption{Reducible, recurrent}
     \end{subfigure}
     \qquad
     \begin{subfigure}[t]{0.45\textwidth}
         \centering
         \includegraphics[height=3cm]{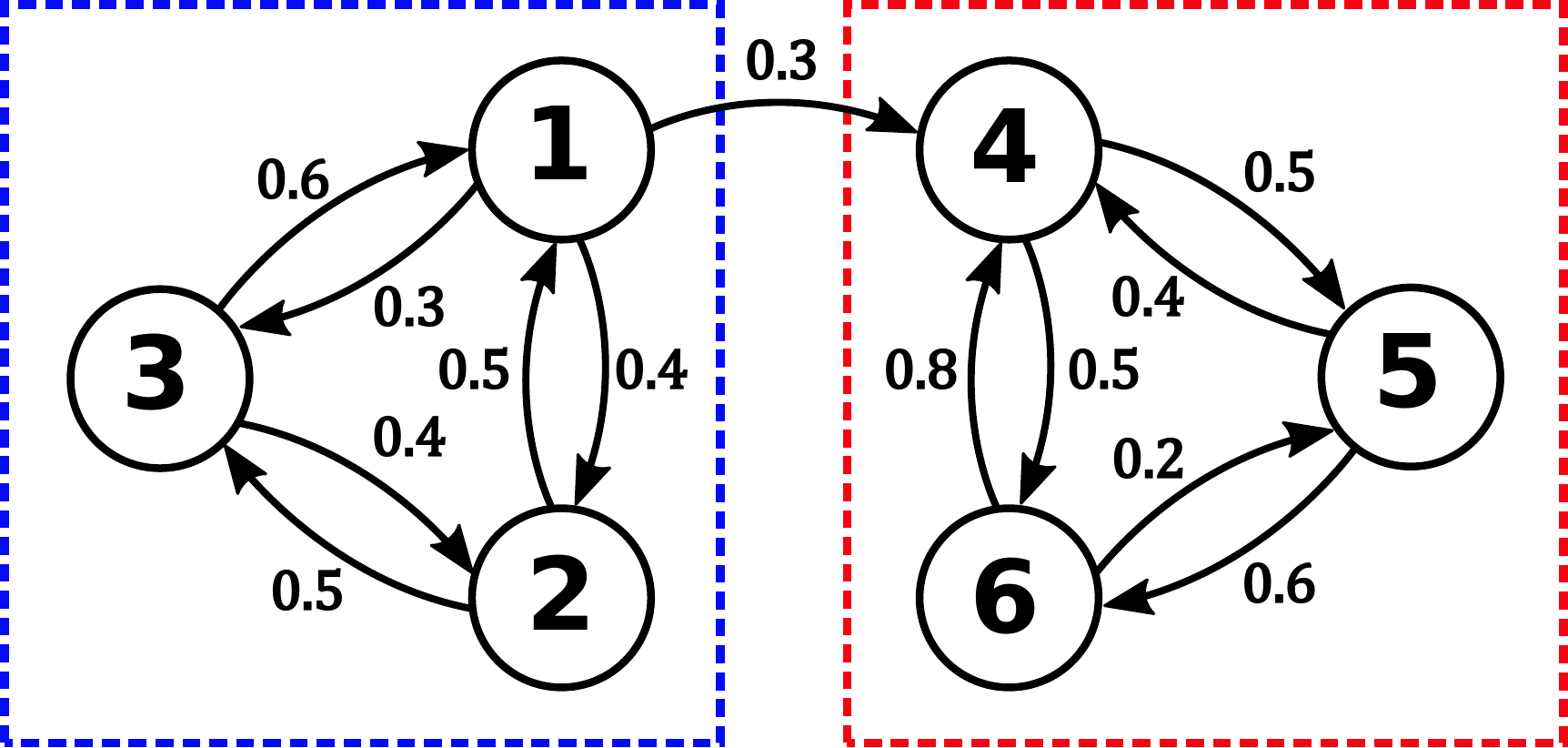}
         \caption{Reducible, not recurrent}
     \end{subfigure}
     \par\bigskip
     \begin{subfigure}[t]{0.45\textwidth}
         \centering
         \includegraphics[height=3cm]{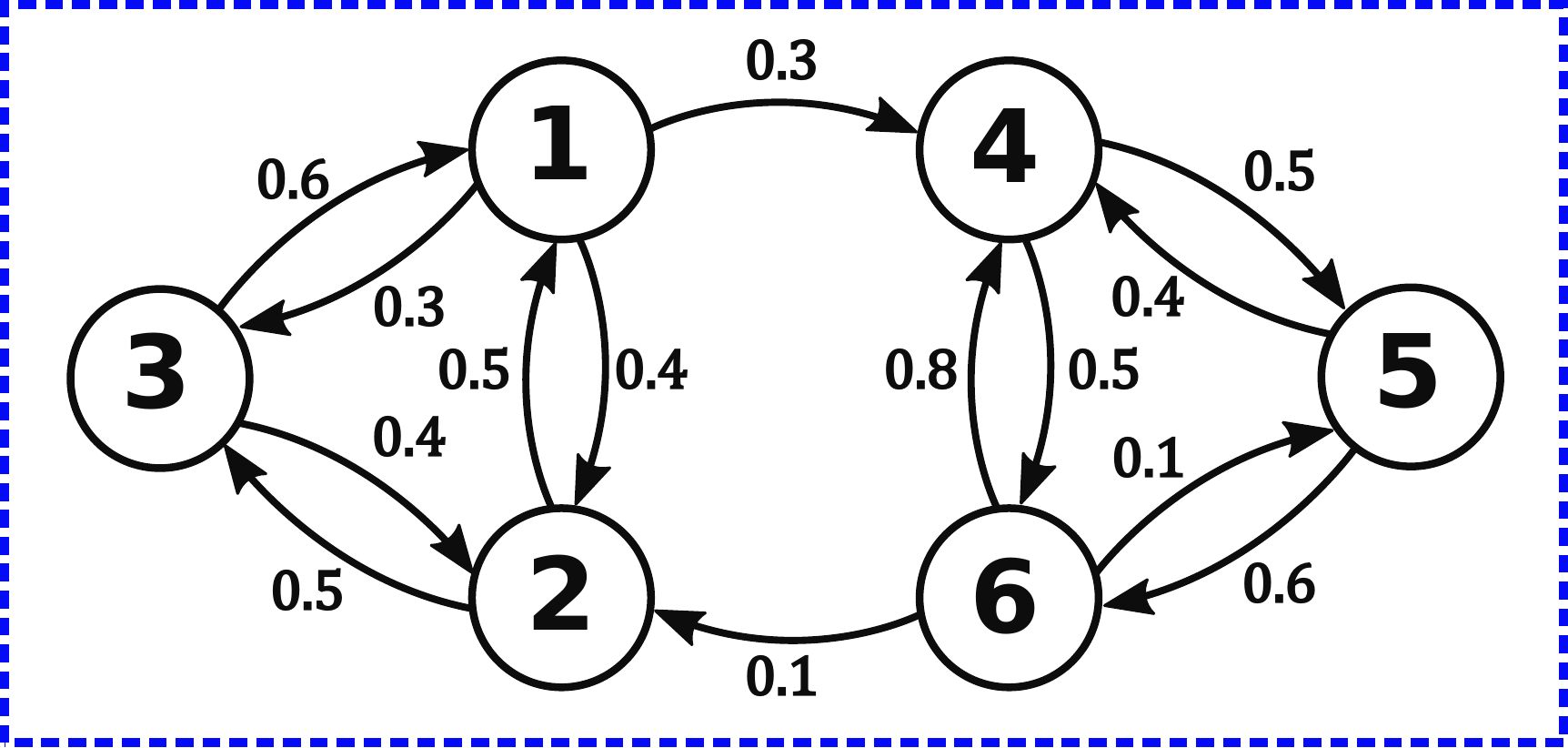}
         \caption{Irreducible}
     \end{subfigure}
        \caption{Three Markov chains with differing connectivity, each with $N=6$ states.}
        \label{fig:Comm}
\end{figure}

Furthermore, we can make a useful categorization of Markov chains based on the number of communicating classes they have:

\begin{defin}[Number of communicating classes]
Let $n$ be the number of communicating classes of a Markov chain. When $n=1$, the chain is said to be irreducible, otherwise it is reducible.
\end{defin}
\noindent In words, an irreducible Markov chain is one in which for any pair of states there exists a connecting path in both directions. In \cref{fig:Comm}, three example Markov chains are shown, with the communicating classes indicated by the dashed boxes. Take a moment to double-check why each state belongs to its communicating class, and verify that the examples in \cref{fig:Comm}(a,b) are reducible, whereas the one in \cref{fig:Comm}(c) is irreducible. Finally, observe from \cref{fig:Comm}(b) that it is possible for states in one communicating class to be accessible from states in another class (e.g.\ $s_3\to s_4$).

Irreducible Markov chains feature in a number of subsequent sections of this tutorial, due to the following result:

\begin{theorem}
\label{thm:IrredSD}
A Markov chain is irreducible if and only if it has a unique stationary distribution $\vec{\pi}$. Furthermore, this distribution has strictly positive elements, i.e.\ $\pi_i>0 \; \forall s_i \in \mathcal{S}$.
\end{theorem}

\noindent This result can be applied to the example of \cref{fig:Comm}(c) by using the observation of the previous section that stationary distributions of a given Markov chain are eigenvectors of the associated transition matrix with eigenvalue $1$. If we were to find the transition matrix of the chain in \cref{fig:Comm}(c) and compute its eigenvectors, we would indeed find a single left eigenspace of $\mat{P}$ corresponding to eigenvalue $1$, with strictly positive elements. When normalized such that the row sum is $1$, the resulting vector is $\vec{\pi}=(\pi_1, \pi_2, \pi_3, \pi_4, \pi_5, \pi_6)^T=(0.09, 0.09, 0.07, 0.31, 0.18, 0.26)^T$, which we encourage readers to check themselves. For reducible Markov chains, the guarantees of uniqueness and positivity of the stationary probabilities no longer hold. In order to describe the stationary distributions of such chains, we first introduce some new concepts for describing states.

\subsubsection{Recurrence and transience}
Each state $s_i\in\mathcal{S}$ can be categorized based on how likely it is to be revisited, given that it is currently occupied. This is formalized by the following definition:

\begin{defin}[Recurrence/transience]
Given that the system is in state $s_i$ initially, the probability of returning to $s_i$ is defined as:
\begin{equation}
    f_i:=\textup{Pr}(X_k=s_i\text{ for some }k\geq 1|X_0=s_i)
\end{equation}
States for which $f_i=1$ are called recurrent, and those for which $f_i<1$ are called transient.
\end{defin}
\noindent This is a useful way to characterize states in $\mathcal{S}$, since it generalizes to all states within a communicating class, i.e.\ for any communicating class $C\subseteq\mathcal{S}$, either all states in $C$ are recurrent or all states in $C$ are transient. Transience/recurrence are therefore examples of \textit{class properties}, and we henceforth use the terms \textit{recurrent class} and \textit{transient class} for communicating classes that contain recurrent and transient states, respectively. Furthermore, for finite state spaces a Markov chain is guaranteed to have at least one recurrent class. Building upon this, we can then apply the notion of recurrence to a Markov chain as a whole:

\begin{defin}[Recurrent chains]
A Markov chain that contains only recurrent classes is called a recurrent Markov chain.
\end{defin}

As an illustration, we apply these concepts to the reducible chains depicted in \cref{fig:Comm}. In \cref{fig:Comm}(a), there are two communicating classes and in each one there is no possibility to exit. This means that if a state in one of these classes is occupied, it is guaranteed to be revisited at some future time step, i.e.\ $f_i=1$ for all states in each class. Therefore, both classes are recurrent and the chain as a whole is recurrent. In \cref{fig:Comm}(b), the main difference is that there is now the possibility to exit the blue class without returning. For example, assuming $s_1$ is occupied, although there is a possibility that $s_1$ can be visited again later (e.g.\ $s_1\to s_2\to s_1$), as soon as a transition $s_1\to s_4$ takes place this will no longer be possible. This is why $f_i<1$ for states in the blue class. Hence, while the red class is recurrent, the blue class is transient, and because of this the chain as a whole is not recurrent. 

For an irreducible Markov chain, such as the example shown in \cref{fig:Comm}(c), every state is guaranteed to be recurrent, leading to the following proposition:

\begin{prop}[Recurrence of irreducible Markov chains]
All irreducible Markov chains are recurrent.
\end{prop}
\noindent However, from the example in \cref{fig:Comm}(a) it is clear that the converse does not hold, since it is also possible for a reducible chain to be recurrent. In fact, whether a reducible chain is recurrent or not determines certain features of the stationary distributions belonging to the chain. This is outlined by the following proposition:

\begin{prop}[Stationary distributions of Reducible chains]
\label{prop:RedSD}
For a reducible Markov chain with $r$ recurrent classes and $t$ transient classes:
\begin{itemize}
\item Any stationary distribution has probability zero for states belonging to a transient class.
\item For the $k$-th recurrent class, there exists a unique stationary distribution $\vec{\pi}_k$ with non-zero probabilities only for states in that class.
\item When the number of recurrent classes $r$ is bigger than $1$, stationary distributions can be formed via convex combinations of each distribution $\vec{\pi}_k$, i.e.
\begin{equation}
\label{eq:RecRedSD}
    \vec{\pi}_{\textup{combo}}=\sum_{k=1}^r \alpha_k \vec{\pi}_k \quad,\quad \textup{with} \quad \alpha_k\geq0 \quad \textup{and} \quad \sum_{k=1}^r \alpha_k=1
\end{equation}
meaning that there are an infinite number of stationary distributions.
\item Furthermore, when the number of transient classes $t$ is zero, or in other words when the chain is recurrent, performing the procedure above with nonzero coefficients always yields distributions that are strictly positive.
\end{itemize}
\end{prop}
We can apply these results to the two reducible chains in \cref{fig:Comm}. In the example of \cref{fig:Comm}(a), the stationary distribution associated to the blue class is $\vec{\pi}_{\text{1}}=(0.35, 0.36, 0.29, 0, 0, 0)^T$ and the one associated to the red class is $\vec{\pi}_{\text{2}}=(0, 0, 0, 0.39, 0.26, 0.35)^T$. We can then generate an arbitrary number of extra stationary distributions by taking convex combinations of $\vec{\pi}_{\text{1}}$ and $\vec{\pi}_{\text{2}}$ with coefficients $\alpha_1$ and $\alpha_2$. For example, with $\alpha_1=0.25$ and $\alpha_2=0.75$ we get $\vec{\pi}=(0.09, 0.09, 0.07, 0.29, 0.2, 0.26)^T$. Furthermore, the last bullet point in \cref{prop:RedSD} tells us that since this chain is recurrent we can be sure that any convex combination with positive coefficients yields a distribution with strictly positive entries. This property of recurrent chains is particularly relevant to our treatment of both reversible chains in \cref{MCs-Rev} and random walks in \cref{RWs}, and we henceforth use $\vec{\pi}>0$ to denote a stationary distribution with this property. In the example of \cref{fig:Comm}(b), the red class is the only recurrent class, meaning that the stationary distribution associated to this class is the only stationary distribution of the chain. Since the transition probabilities for states in this class are the same as in the example of \cref{fig:Comm}(a), this stationary distribution is $\vec{\pi}=\vec{\pi}_2$. Furthermore, in agreement with \cref{prop:RedSD}, we see that the transient states in this chain have a stationary probability of $0$.

\subsubsection{Periodicity}
The notion that states can be revisited is also meaningful in another sense. We define the following quantity, which describes how frequently such revisits can take place.

\begin{defin}[Periodic chains]
\label{defin:Period}
For each state $s_i\in \mathcal{S}$, the period is defined as:
\begin{equation}
\label{eq:MCperiod}
    d_i:=gcd\{k\geq1\; : \; (\mat{P}^k)_{ii}>0\}
\end{equation}
where gcd indicates the greatest common divisor. Then, \cref{eq:MCperiod} says that when starting in state $s_i$ it is only possible to return to $s_i$ in multiples of the period $d_i$. States for which $d_i>1$ are called \textit{periodic} and those for which $d_i=1$ are \textit{aperiodic}.
\end{defin}

Like transience/recurrence, period is also a class property, and we use $d$ to refer to the period of a whole class. This in turn allows us to define the period of a Markov chain:

\begin{defin}[Periodicity]
When all communicating classes in $\mathcal{S}$ have period $d>1$, the Markov chain is said to be periodic, with period $d$.
\end{defin}

\begin{defin}[Aperiodicity]
When all communicating classes in $\mathcal{S}$ are aperiodic, the Markov chain is said to be aperiodic.
\end{defin}

\begin{defin}[Mixed Periodicity]
If $\mathcal{S}$ contains communicating classes with different periods, the Markov chain is said to have mixed periodicity.
\end{defin}

\begin{figure}[h]
     \centering
     \begin{subfigure}[b]{0.2\textwidth}
         \centering
         \includegraphics[width=\textwidth]{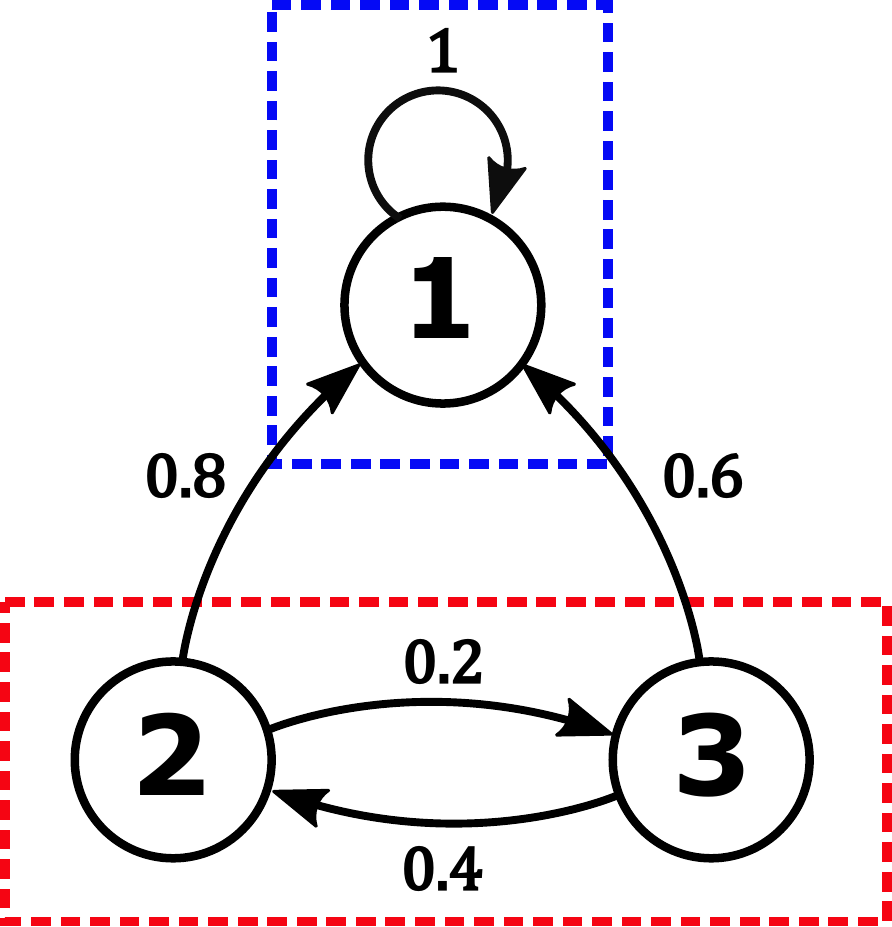}
         \caption{$d=1$, $d=2$}
     \end{subfigure}
     \hfill
     \begin{subfigure}[b]{0.2\textwidth}
         \centering
         \includegraphics[width=\textwidth]{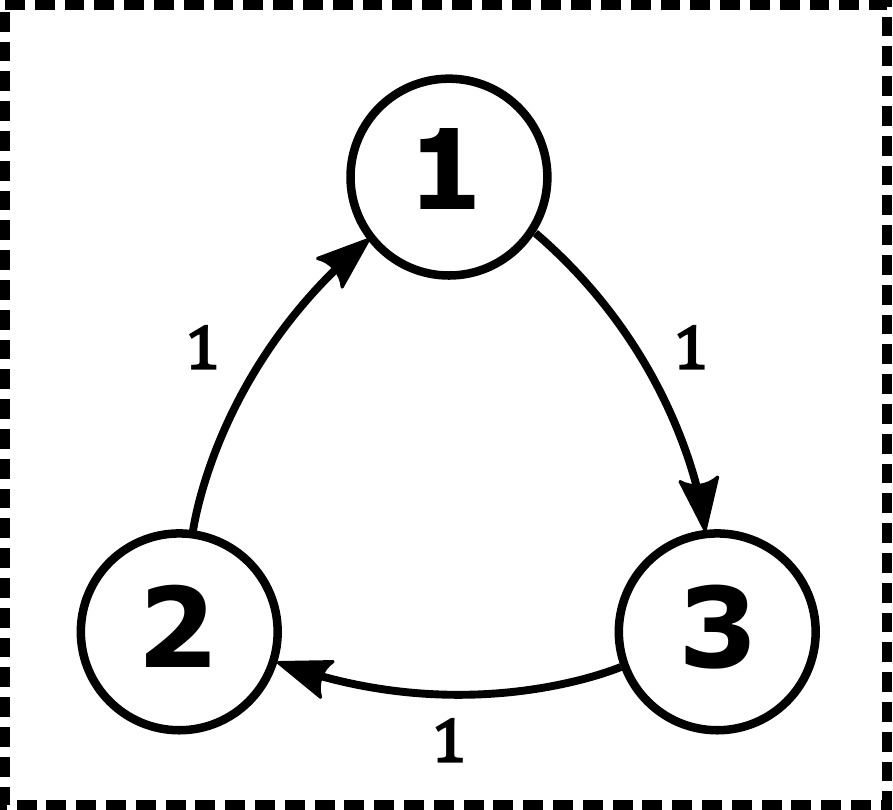}
         \caption{$d=3$}
     \end{subfigure}
     \hfill
     \begin{subfigure}[b]{0.2\textwidth}
         \centering
         \includegraphics[width=\textwidth]{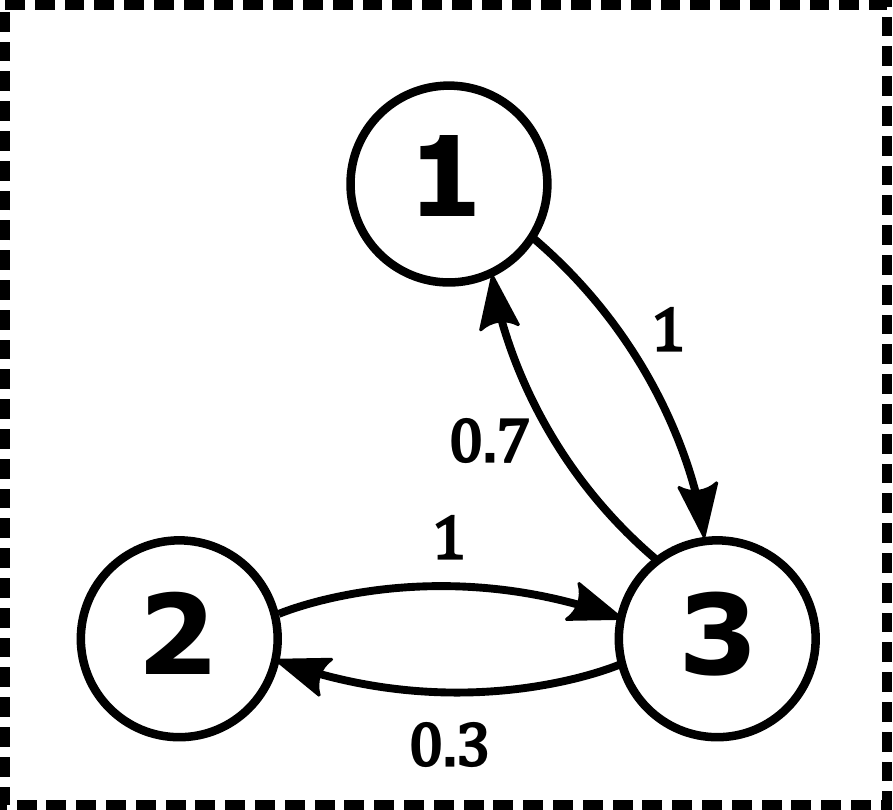}
         \caption{$d=2$}
     \end{subfigure}
     \hfill
     \begin{subfigure}[b]{0.2\textwidth}
         \centering
         \includegraphics[width=\textwidth]{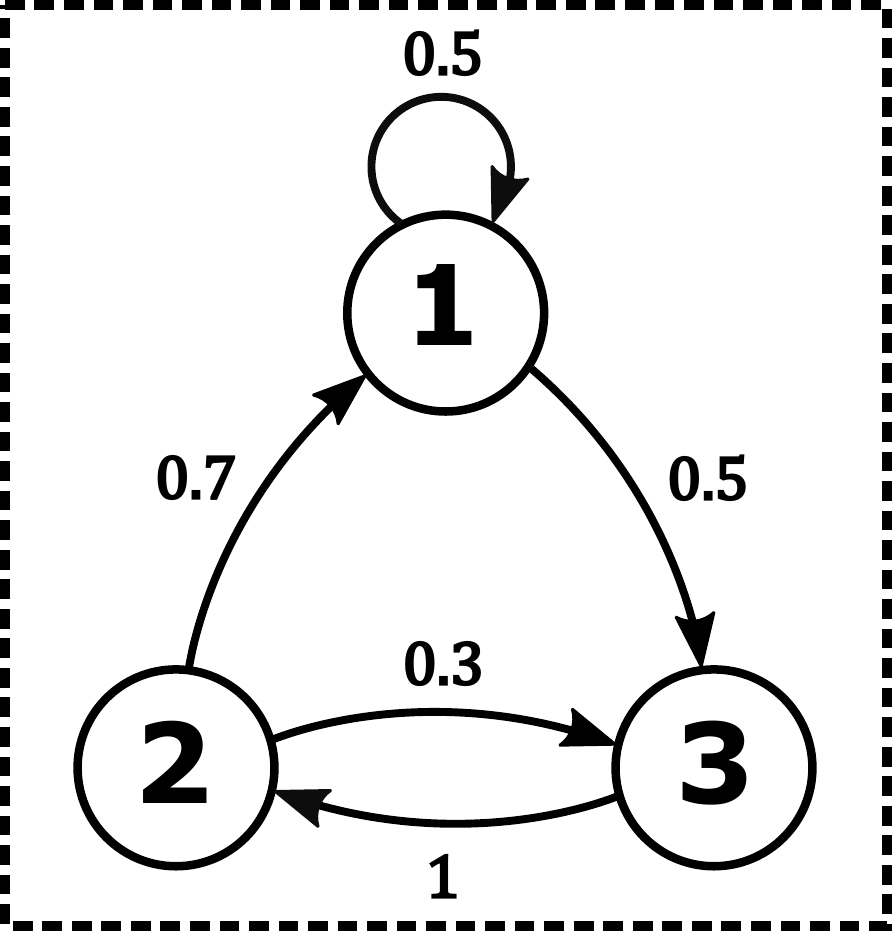}
         \caption{$d=1$}
     \end{subfigure}
        \caption{Four Markov chains with differing periodicity, each with $N=3$ states.}
        \label{fig:Period}
\end{figure}

Consider the example shown in \cref{fig:Period}(a). This chain has two communicating classes, the red one being a transient class with $d=2$ and the blue one being a recurrent class with $d=1$, meaning that this chain has mixed periodicity. In \cref{fig:Period}(b-d), we show three irreducible Markov chains, with (b) and (c) having period $3$ and $2$, respectively, and (d) being aperiodic.

\bigskip
Markov processes are a broad class of models, and even under the restricted settings considered in this tutorial (discrete time, homogeneous and finite state spaces), there are many distinct types of chains. In the following sections, we concentrate on three particular types that are relevant in applied domains.

\subsection{Ergodic chains}
\label{MCs-Erg}
When modeling a system that evolves over time, it is important to ask what can be said, if anything, about its long term behavior. For a Markov chain, this question can be phrased in two ways. On one hand, we can sample a single trajectory starting from some initial state and ask what the average behavior is over time, i.e.\ how \textit{often} is it found in each state $s_i\in\mathcal{S}$ for a trajectory of length $t$? On the other hand, we can describe our starting conditions as a distribution $\vec{\mu}(0)$ and ask what this evolves to in the future, i.e.\ what is the probability of being in each state $s_i$ at a later time $t$? We refer to these two notions of long term behavior as the \textit{trajectory} and \textit{distribution} perspectives, respectively. While the analyses given so far predominantly use the latter perspective, we remind readers that in \cref{MCs-def} we have introduced the idea of a distribution over $\mathcal{S}$ by taking the limit of an infinitely large ensemble of trajectories, meaning that the two concepts are closely related.

This two-way view originates from the field of statistical physics, where physical processes can either be analyzed with temporal averages (i.e.\ the trajectory perspective) or ensemble averages (i.e.\ the distribution perspective). One class of systems that has received a lot of study in this field are those for which these two types of averaging yield the same result as $t\to\infty$. Such systems are known as \textit{ergodic systems}, and this equivalence means that a statistical description of their long term behavior can be described simply by a single, sufficiently long sample. One implication of this is that initial conditions are forgotten over time, which makes ergodic systems particularly attractive from a simulation or modeling perspective. Finally, with this in mind we can define an ergodic Markov chain as follows:

\begin{defin}[Ergodic Markov chain]
\label{def:Erg}
An ergodic Markov chain is one that is guaranteed to converge to a unique stationary distribution.
\end{defin}

Clearly, in order for a chain to be ergodic it must have a unique stationary distribution. Therefore, by virtue of \cref{thm:IrredSD}, a necessary condition for a chain to be ergodic is that it is irreducible. However, there is no guarantee that an irreducible chain converges, which is the second condition of \cref{def:Erg}. The convergence of a Markov chain is related to its periodicity, as explained by the following result:

\begin{theorem}[Convergence of a Markov chain]
\label{thm:ConvMC}
The evolution of a Markov chain with period $d$ can lead to a permanent, repeating sequence of $d$ distributions, i.e.
\begin{equation}
    \vec{\mu}(k)\to\vec{\mu}(k+1)\to ... \to\vec{\mu}(k+d-1)\to\vec{\mu}(k)
\end{equation}
which for $d=2$ and $d>2$ correspond to persistent oscillations and persistent cycles, respectively. In the case of an aperiodic Markov chain ($d=1$), only sequences of length $1$ are allowed, which means that one of the stationary distributions is guaranteed to be reached.
\end{theorem}

We can apply this result to the irreducible Markov chain in \cref{fig:Period}(c). This chain has a unique stationary distribution $\vec{\pi}=(0.35, 0.15, 0.5)^T$ and a period of $d=2$. Therefore, there is no guarantee that the chain converges to $\vec{\pi}$ since it can get trapped in persistent oscillations. To observe this, we can try out different initial conditions and iteratively applying the update rule in \cref{eq:Pmult}. For example, starting with the distribution $\vec{\mu}(0)=(0.25,0.5,0.25)^T$ we get a persistent oscillation between the following two distributions: $\vec{\mu}(k)=(0.175, 0.075, 0.75)^T$ and  $\vec{\mu}(k+1)=(0.525, 0.225, 0.25)^T$. However, this is not the only persistent oscillation possible for this chain, which can be observed by trying out different initial conditions. Lastly, in \cref{Graphs-Eig} we gain more insight on \cref{thm:ConvMC} by using tools from graph theory to describe the eigenvectors of $\mat{P}$ with $|\lambda|=1$.

A key insight from \cref{thm:ConvMC} is that a Markov chain is guaranteed to converge only if it is aperiodic. Together with irreducibility, this therefore provides the conditions under which a chain is ergodic:

\begin{theorem}[Conditions for Ergodicity]
A Markov chain is ergodic if and only if it is both irreducible and aperiodic, which respectively ensure that (i) there is a unique distribution $\vec{\pi}$, and (ii) the chain always converges to this distribution. Furthermore, the distribution $\vec{\pi}$ is said to be the limiting distribution of the chain.
\end{theorem}

Ergodic Markov chains have a number of beneficial properties. Firstly, as with any ergodic system, the initial conditions are eventually forgotten, which means that when studying the statistical behavior of an ergodic chain it is not necessary to explore different starting states. This advantageous property underlies a number of methods in machine learning and beyond, with Markov chain Monte Carlo methods being a particularly well known example \cite{MatthewRichey2010}. Secondly, the equivalence between the trajectory and distribution perspectives allows us to interpret the limiting probabilities $\pi_i$ as the long run fraction of steps that the chain spends in each state $s_i$ \cite{Porod2021}. Lastly, analyzing the evolution of a Markov chain in terms of the eigenvectors of its transition matrix becomes somewhat simpler in the case of an ergodic chain. We have already seen that when the transition matrix of a Markov chain is diagonalizable, we can vastly simplify the computation of the $k$-step evolution of the Markov chain (\cref{eq:EveigkPart}). When the Markov chain is also ergodic, we know that its persistent behavior is fully described by $\vec{l}_1=\vec{\pi}$, meaning that this expression simplifies to:
\begin{equation}
\label{eq:ErgEig}
    \vec{\mu}(t+k)^T=\vec{\pi}^T+\sum_{\omega=2}^Nc_\omega \lambda_\omega^k \vec{l}_\omega^T
\end{equation}

\subsection{Reversible chains}
\label{MCs-Rev}
One of the defining characteristics of Markov chains is that the future ($X$) is conditionally independent of the past ($Z$), given the present ($Y$). A simple calculation demonstrates that the relationship of conditional independence is symmetric. Assuming that $\text{Pr}(X|Y,Z)=\text{Pr}(X|Y)$, we find that:

\begin{align}
    \text{Pr}(Z|Y,X)&=\frac{\text{Pr}(X,Y,Z)}{\text{Pr}(Y,X)}\\
    &=\frac{\text{Pr}(X|Y,Z)\text{Pr}(Y,Z)}{\text{Pr}(X|Y)\text{Pr}(Y)}\\
    &=\frac{\text{Pr}(X|Y)\text{Pr}(Y,Z)}{\text{Pr}(X|Y)\text{Pr}(Y)}\quad\text{(by assumption)}\\
    &=\frac{\text{Pr}(Y,Z)}{\text{Pr}(Y)}\\
    &=\text{Pr}(Z|Y)
\end{align}

Assuming that we have an initial Markov chain $\mathcal{X}$ described by a transition matrix $\mat{P}$, this symmetry tells us that reversing the direction of time produces a new process which itself satisfies the Markov property. Hence, we can define this new Markov chain as the \textit{time-reversal} of $\mathcal{X}$, which we denote as $\mathcal{\tilde{X}}$. A natural question we can ask is how the transition matrix of $\mathcal{\tilde{X}}$ is related to $\mat{P}$. In order to answer this, we make the assumption that at time $t$ the chain is described by one of its stationary distributions $\vec{\pi}$. Therefore:
\begin{equation}
   \text{Pr}(X_t=s_i)=\pi_i \;\forall s_i\in\mathcal{S}\; \text{and} \; t\geq 0
\end{equation}
We can make use of this, together with Bayes' theorem, to work out the transition probabilities $(\mat{P}_{\text{rev}})_{ij}$ of $\mathcal{\tilde{X}}$ \cite{Bremaud1999}:

\begin{equation}
\label{eq:BayesRev}
    (\mat{P}_{\text{rev}})_{ij}=\text{Pr}(X_t=s_j|X_{t+1}=s_i)=\frac{\text{Pr}(X_{t+1}=s_i|X_{t}=s_j)\text{Pr}(X_t=s_j)}{\text{Pr}(X_{t+1}=s_i)}=\frac{P_{ji}\; \pi_j}{\pi_i}
\end{equation}
A couple of reflections can be made about the result above. Firstly, since we initialized $\mathcal{X}$ to one of its stationary distributions, the time reversal relationship between $\mathcal{X}$ and $\mathcal{\tilde{X}}$ only holds once they have converged. Secondly, since we have $\pi_i$ in the denominator of \cref{eq:BayesRev}, the time reversal of a Markov chain is only valid for a starting distribution with $\pi_i>0\; \forall s_i\in\mathcal{S}$. In \cref{MCs-CommClass}, we have established that only recurrent chains have stationary distributions of this type, meaning that the time reversal of a Markov chain is only well-defined if the chain is recurrent. Lastly, using \cref{eq:BayesRev} it is possible to define $\mat{P}_{\text{rev}}$ in matrix notation as follows:

\begin{defin}[Time reversal]
\label{def:TimeRev}
Let $\mathcal{X}$ be a recurrent Markov chain with transition matrix $\mat{P}$ and $\vec{\pi}>0$ one of its stationary distributions. Then, the transition matrix of its time reversal $\mathcal{\tilde{X}}$ is given by:
\begin{equation}
\label{eq:RevTrans}
   \mat{P}_{\textup{rev}}:=\mat{\Pi}^{-1}\mat{P^T}\mat{\Pi}
\end{equation}
\end{defin}

It is worth emphasizing that since no assumption is made in \cref{def:TimeRev} about the number communicating classes, it also applies in the case of reducible recurrent chains where there are an infinite number of distributions $\vec{\pi}>0$ to choose from. However, in such cases the choice of $\vec{\pi}$ makes no difference:

\begin{prop}[Time reversal of reducible chains]
\label{prop:RevRedRec}
For a reducible recurrent Markov chain $\mathcal{X}$, the time reversal $\mathcal{\tilde{X}}$ is uniquely defined, with $\mat{P}_{\textup{rev}}$ being independent of which stationary distribution $\vec{\pi}>0$ is used (proof: see Appendix A).
\end{prop}

\noindent Moreover, the set of stationary distributions $\vec{\pi}>0$ belonging to a recurrent Markov chain is the same as the set belonging to the corresponding time reversal:

\begin{prop}[Stationary distributions of time reversal]
\label{prop:SDRev}
Let $\mathcal{X}$ be a recurrent Markov chain. Then $\vec{\pi}>0$ is a stationary distribution of $\mathcal{\tilde{X}}$ if and only if it is a stationary distribution of $\mathcal{X}$.\footnote{For continuous time Markov chains, this result is known as Kelly's lemma.}
\end{prop}
 
We now consider the special case where a Markov chain $\mathcal{X}$ is indistinguishable from its time reversal $\mathcal{\tilde{X}}$. Such processes are known as \textit{reversible Markov chains}, since in any stationary distribution the forward and backwards dynamics of the chain are statistically equivalent, i.e.\ any trajectory $X_1$, $X_2$, ..., $X_{k-1}$, $X_k$ occurs with equal probability as the corresponding reversed trajectory $X_k$, $X_{k-1}$, ..., $X_2$, $X_1$. The stationary dynamics of such chains therefore has no inherent arrow of time. Furthermore, since $\mathcal{X}$ and $\mathcal{\tilde{X}}$ are indistinguishable, they have the same transition matrix, i.e.\ $\mat{P}=\mat{P}_{\text{rev}}$, and for any stationary distribution $\vec{\pi}>0$ the forward and backwards flow matrices are the same. Therefore:
\begin{align}
    \mat{F}^{\vec{\pi}}&=\mat{F}^{\vec{\pi}}_{\text{rev}}\\
    \mat{\Pi}\mat{P}&=\mat{\Pi}\mat{P}_{\text{rev}}\\
    \mat{\Pi}\mat{P}\overset{(\ref{eq:RevTrans})}&{=}\mat{P^T}\mat{\Pi} \label{eq:FlowSym}
\end{align}
By expressing \cref{eq:FlowSym} in component form, we arrive at the following theorem which is used throughout the rest of the tutorial:

\begin{theorem}[Detailed balance]
\label{thm:RevChain}
A recurrent Markov chain is reversible if and only if it for any stationary distribution $\vec{\pi}>0$:
\begin{equation}
    \label{eq:DB}
    \pi_i P_{ij}=\pi_j P_{ji} \quad \forall s_i,s_j \in \mathcal{S}
\end{equation}
which are known as the equations of detailed balance.
\end{theorem}
A number of observations can be made about this definition. Firstly, the left (right) terms represent the flow of probability from $s_i$ to $s_j$ ($s_j$ to $s_i$), given that the chain is described by distribution $\vec{\pi}$. Thus, for a reversible Markov chain in one of its stationary distributions, the flow from one state to another is completely balanced by the flow in the reverse direction, meaning that the flow matrix $\mat{F}^{\vec{\pi}}$ is always symmetric for such chains. By comparison with \cref{eq:GB2}, we see that detailed balance is a stronger condition than global balance, since in the latter case there is only an equivalence between the \textit{total} flow in and out of each state. Secondly, since $\pi_i$ and $\pi_j$ are non-zero, it follows that $P_{ij}\neq0$ if and only if $P_{ji}\neq0$. Thus, the transition structure of a reversible Markov chain always permits the return to the previous state, and because of this the period of a reversible Markov chain can be at most $2$. Thirdly, while some sources assume irreducibility as a precondition of reversibility, we instead base our definition on the weaker condition of recurrence \cite{Porod2021}. This is due to the fact that we only need recurrence in order to define the time reversal of a Markov chain. Furthermore, with this convention \cref{thm:RevChain} applies more broadly to reducible Markov chains, which lets us make a closer comparison between reversible Markov chains and undirected graphs in \cref{RWs}. Lastly, \cref{thm:RevChain} implies that there are two distinct ways in which Markov chains can be non-reversible: (i) they can be recurrent without satisfying detailed balance, or (ii) they can be non-recurrent. In the case of (i), $\mat{\Pi P}\neq\mat{P^T}\mat{\Pi}$ for any distribution $\vec{\pi}>0$, meaning that the flow matrix $\mat{F}^{\vec{\pi}}$ is asymmetric, and in the case of (ii) no positive stationary distribution exists. Lastly, note that for a non-recurrent chain, there exists the possibility that $\mat{F}^{\vec{\pi}}$ is symmetric for all stationary distributions despite none of those distributions being strictly positive. For such chains, removing all transient states from $\mathcal{S}$ produces a reversible chain. We therefore refer to such chains as \textit{semi-reversible}.

\begin{figure}
     \centering
     \begin{subfigure}[c]{0.25\textwidth}
         \centering
         \includegraphics[width=\textwidth]{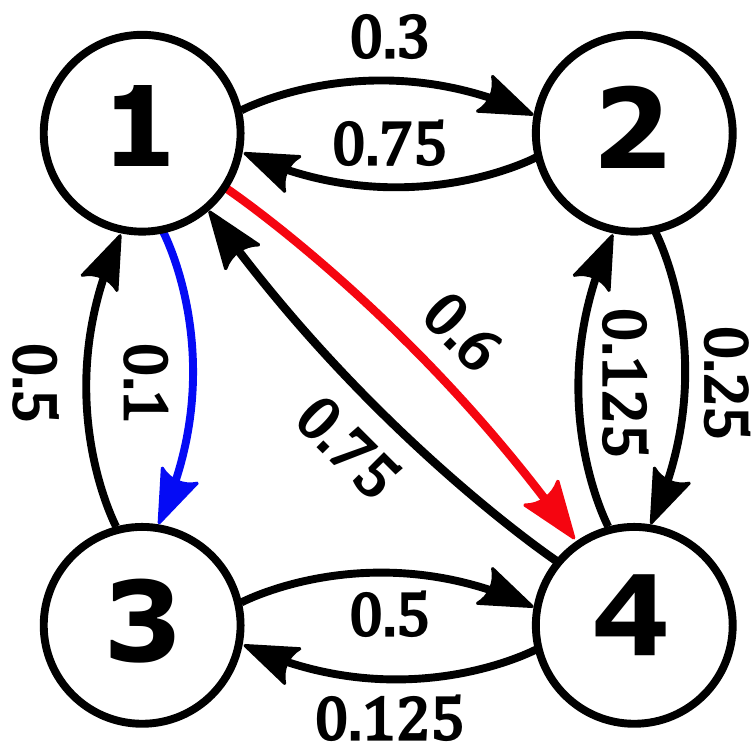}
         \caption{}
     \end{subfigure}
     \hfill
     \begin{subfigure}[c]{0.35\textwidth}
         \centering
         \begin{footnotesize}
         $\mat{P}_{\text{1}}=\left(
         \begin{array}{cccc}
         0 & 0.3 & \textcolor{blue}{0.1} & \textcolor{red}{0.6}\\
         0.75 & 0 & 0 & 0.25\\
         0.5 & 0 & 0 & 0.5\\
         0.75 & 0.125 & 0.125 & 0
         \end{array} \right)$
         \par\bigskip
         $\vec{\pi}_{\text{1}}=(0.417, 0.167, 0.083, 0.333)^T$
         \par\bigskip
         $\mat{F}_{\text{1}}^{\vec{\pi}}=\left(
         \begin{array}{cccc}
         0 & 0.125 & 0.042 & 0.25\\
         0.125 & 0 & 0 & 0.042\\
         0.042 & 0 & 0 & 0.042\\
         0.25 & 0.042 & 0.042 & 0
         \end{array} \right)$
         \end{footnotesize}
         \caption{}
     \end{subfigure}
     \hfill
     \begin{subfigure}[c]{0.35\textwidth}
         \centering
         \includegraphics[width=\textwidth]{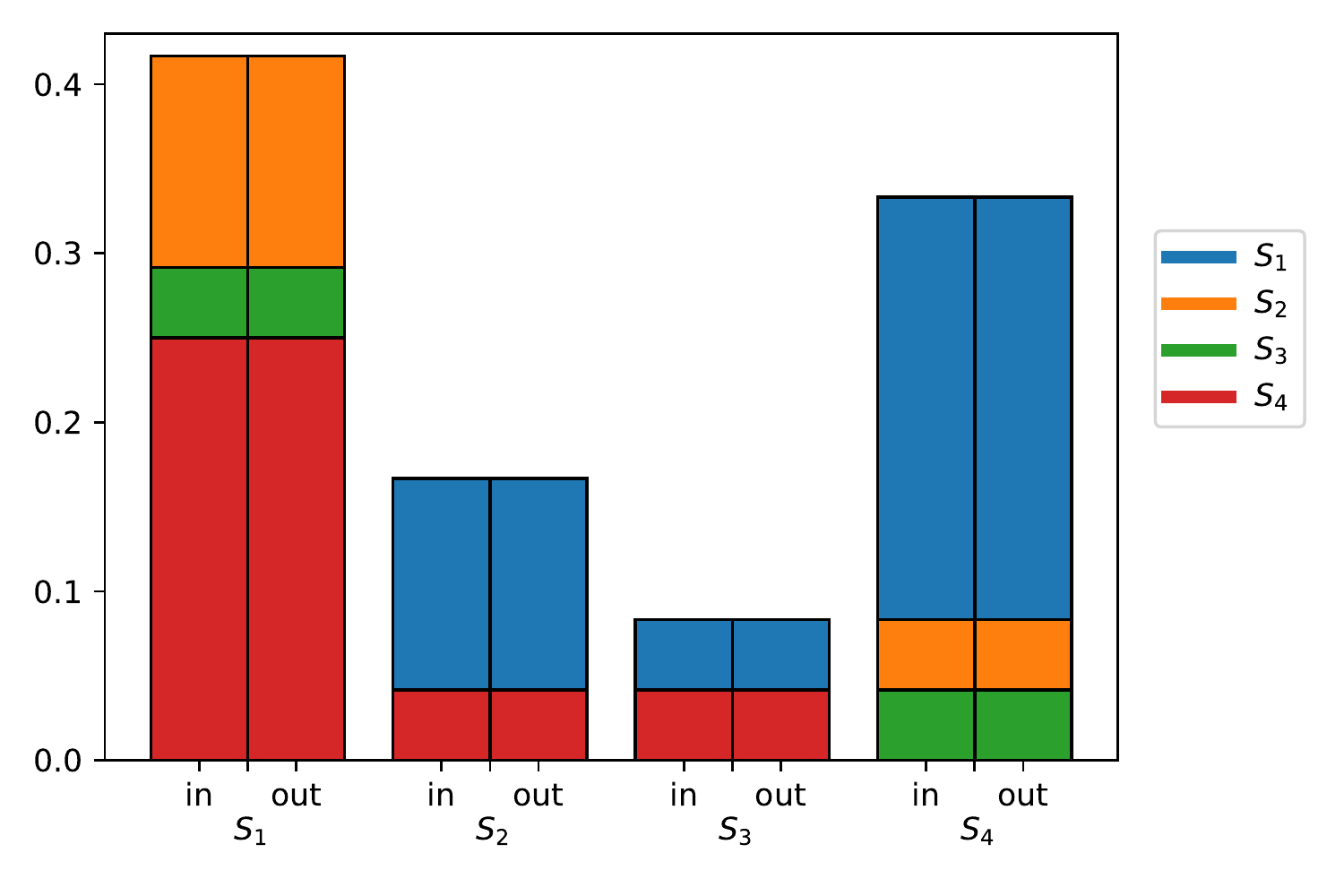}
         \caption{}
     \end{subfigure}
     \par\smallskip
     \begin{subfigure}[c]{0.25\textwidth}
         \centering
         \includegraphics[width=\textwidth]{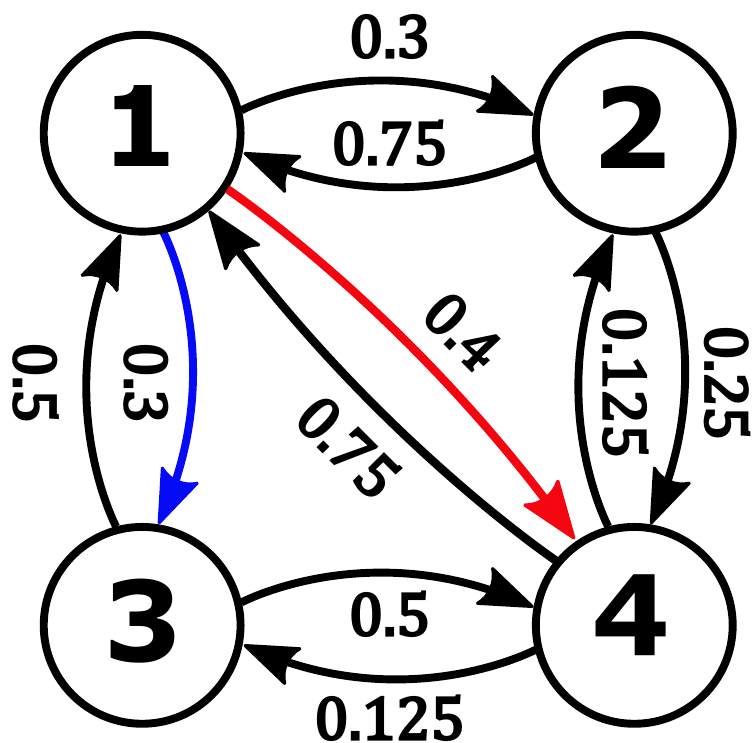}
         \caption{}
     \end{subfigure}
     \hfill
     \begin{subfigure}[c]{0.35\textwidth}
        \centering
        \begin{footnotesize}
         $\mat{P}_{\text{2}}=\left(
         \begin{array}{cccc}
         0 & 0.3 & \textcolor{blue}{0.3} & \textcolor{red}{0.4}\\
         0.75 & 0 & 0 & 0.25\\
         0.5 & 0 & 0 & 0.5\\
         0.75 & 0.125 & 0.125 & 0
         \end{array} \right)$
         \par\bigskip
         $\vec{\pi}_{\text{2}}=(0.406, 0.157, 0.157, 0.280)^T$
         \par\bigskip
         $\mat{F}_{\text{2}}^{\vec{\pi}}=\left(
         \begin{array}{cccc}
         0 & 0.122 & 0.122 & 0.165 \\
         0.118 & 0 & 0 & 0.039 \\
         0.078 & 0 & 0 & 0.078 \\
         0.210 & 0.035 & 0.035 & 0 \\
         \end{array} \right)$
         \end{footnotesize}
         \caption{}
    \end{subfigure}
    \hfill
    \begin{subfigure}[c]{0.35\textwidth}
         \centering
         \includegraphics[width=\textwidth]{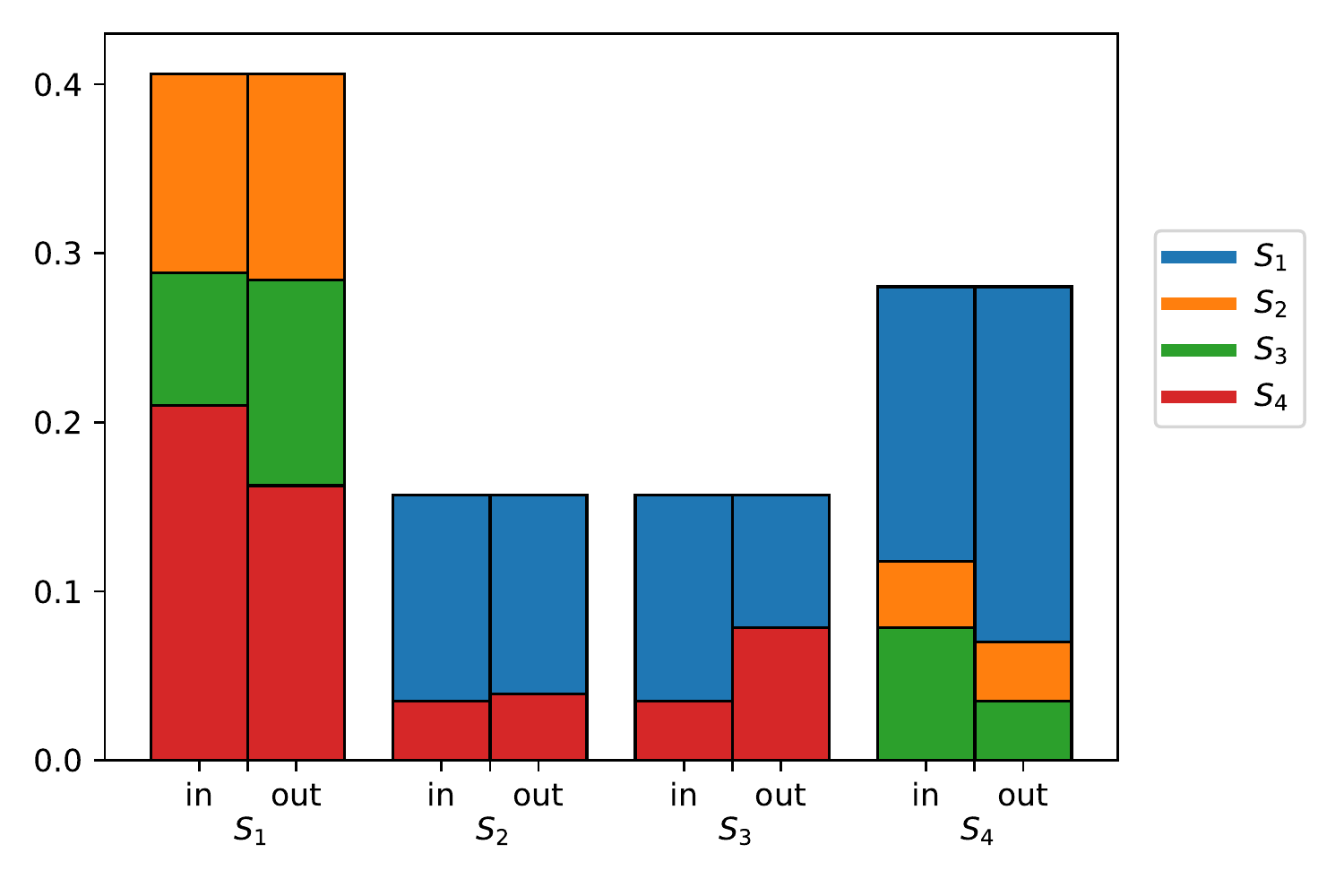}
         \caption{}
    \end{subfigure}
    \par\smallskip
    \begin{subfigure}[c]{0.25\textwidth}
         \centering
         \includegraphics[width=\textwidth]{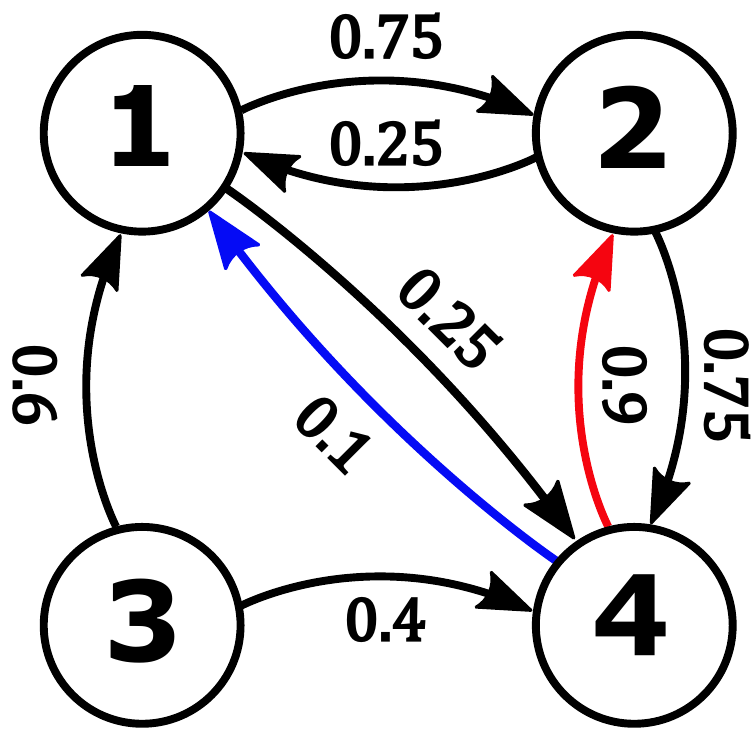}
         \caption{}
    \end{subfigure}
    \hfill
     \begin{subfigure}[c]{0.35\textwidth}
        \centering
        \begin{footnotesize}
         $\mat{P}_{\text{3}}=\left(
         \begin{array}{cccc}
         0 & 0.75 & 0 & 0.25 \\
         0.25 & 0 & 0 & 0.75 \\
         0.6 & 0 & 0 & 0.4 \\
         \textcolor{blue}{0.1} & \textcolor{red}{0.9} & 0 & 0 \\
         \end{array} \right)$
         \par\bigskip
         $\vec{\pi}_{\text{3}}=(0.154, 0.462, 0, 0.385)$
         \par\bigskip
         $\mat{F}_{\text{3}}^{\vec{\pi}}=\left(
         \begin{array}{cccc}
         0 & 0.115 & 0 & 0.038 \\
         0.115 & 0 & 0 & 0.346 \\
         0 & 0 & 0 & 0 \\
         0.038 & 0.346 & 0 & 0 \\
         \end{array} \right)$
         \end{footnotesize}
         \caption{}
    \end{subfigure}
    \hfill
    \begin{subfigure}[c]{0.35\textwidth}
         \centering
         \includegraphics[width=\textwidth]{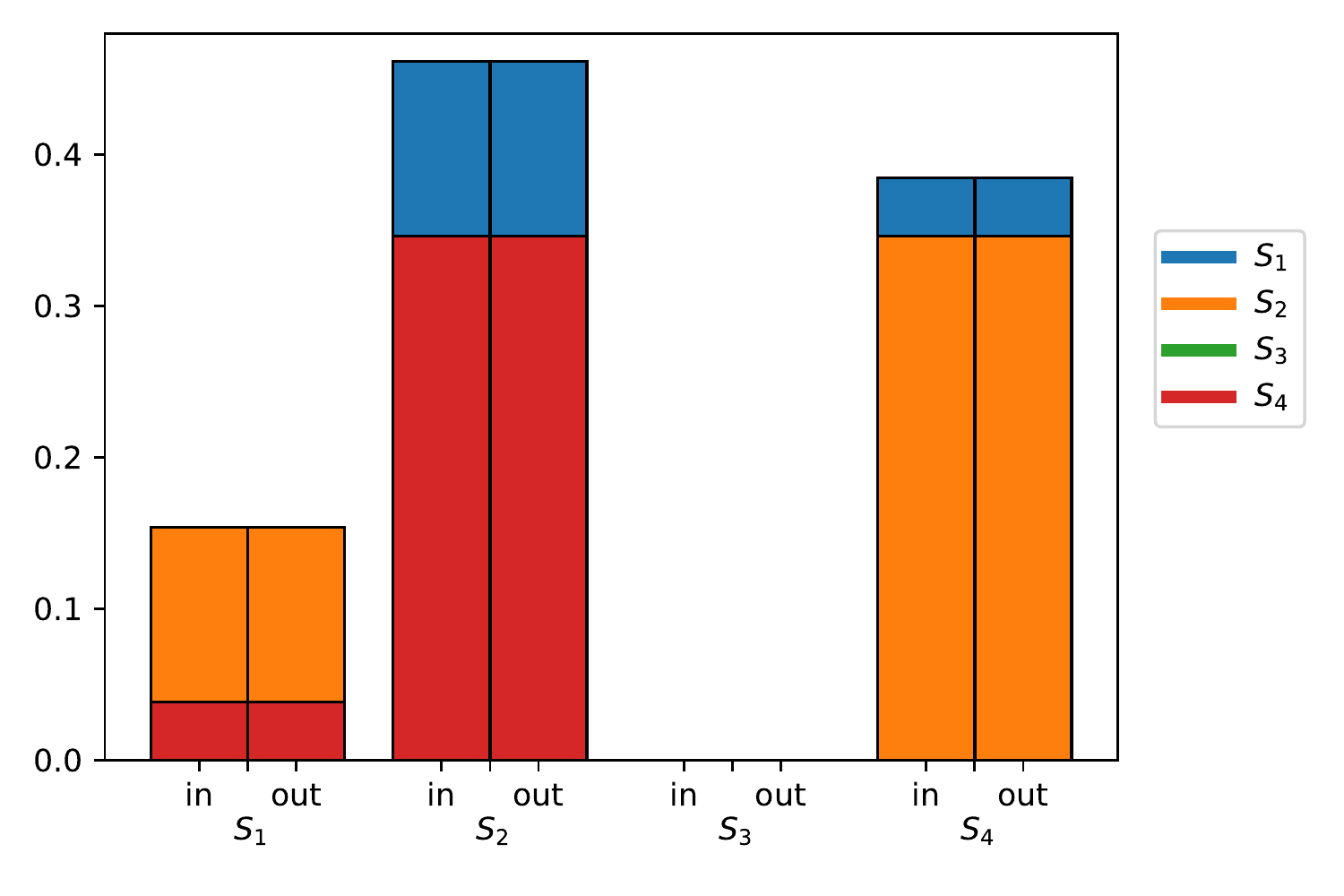}
         \caption{}
    \end{subfigure}
    \par\smallskip
    \begin{subfigure}[c]{0.25\textwidth}
         \centering
         \includegraphics[width=\textwidth]{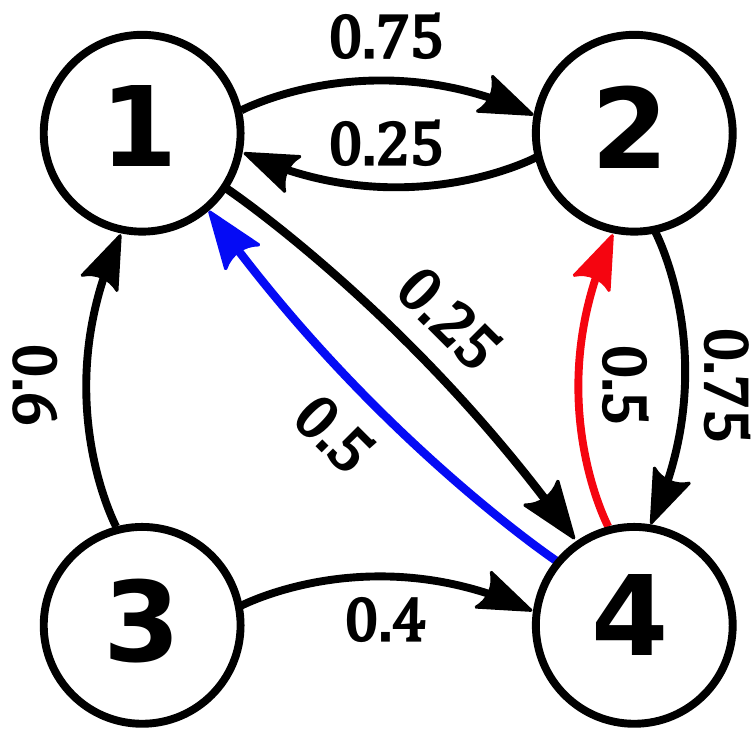}
         \caption{}
    \end{subfigure}
    \hfill
     \begin{subfigure}[c]{0.35\textwidth}
        \centering
        \begin{footnotesize}
         $\mat{P}_{\text{4}}=\left(
         \begin{array}{cccc}
         0 & 0.75 & 0 & 0.25 \\
         0.25 & 0 & 0 & 0.75 \\
         0.6 & 0 & 0 & 0.4 \\
         \textcolor{blue}{0.5} & \textcolor{red}{0.5} & 0 & 0 \\
         \end{array} \right)$
         \par\bigskip
         $\vec{\pi}_{\text{4}}=(0.270, 0.378, 0, 0.352)^T$
         \par\bigskip
         $\mat{F}_{\text{4}}^{\vec{\pi}}=\left(
         \begin{array}{cccc}
         0 & 0.203 & 0 & 0.068 \\
         0.095 & 0 & 0 & 0.284 \\
         0 & 0 & 0 & 0 \\
         0.176 & 0.176 & 0 & 0 \\
         \end{array} \right)$
         \end{footnotesize}
         \caption{}
    \end{subfigure}
    \hfill
    \begin{subfigure}[c]{0.35\textwidth}
         \centering
         \includegraphics[width=\textwidth]{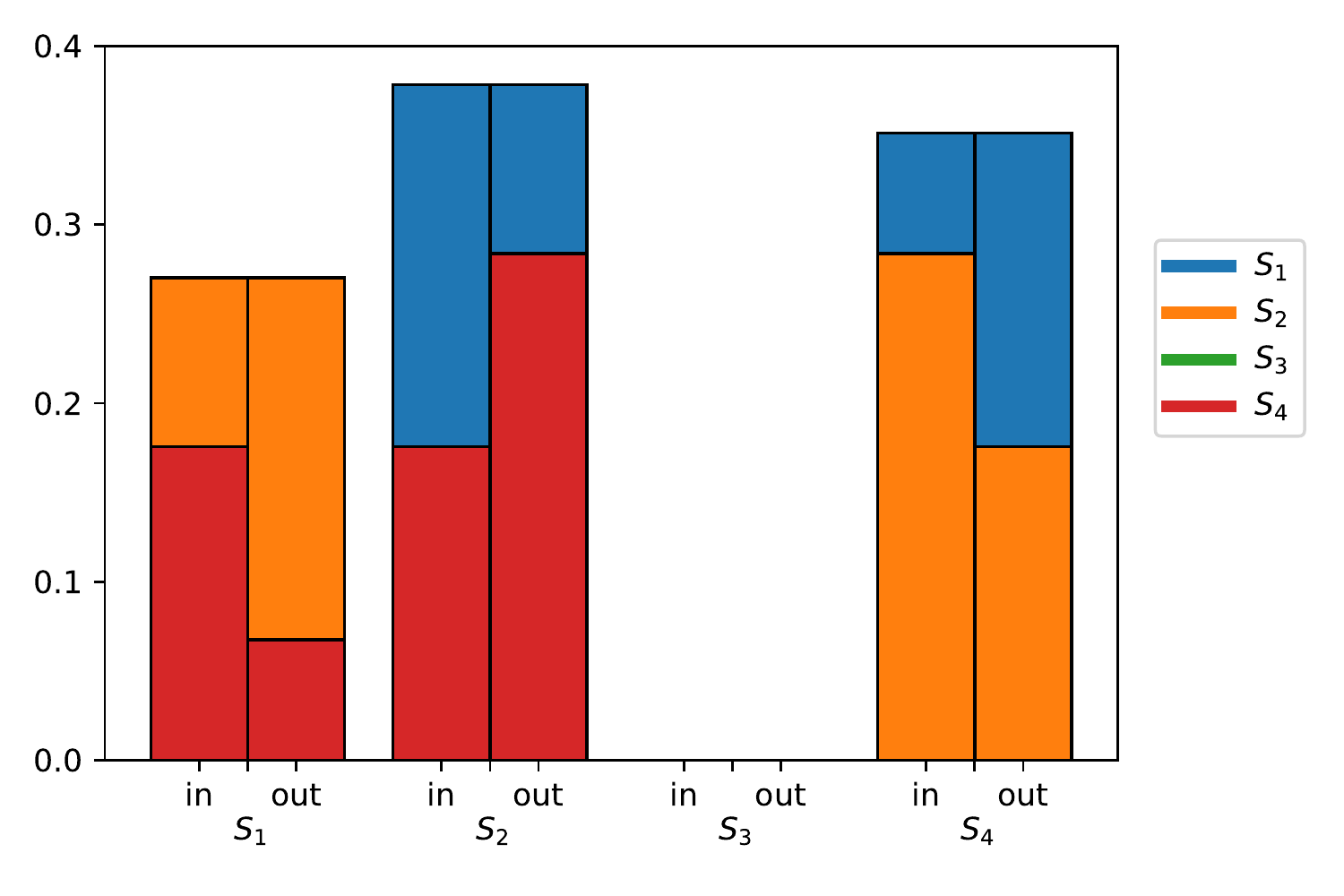}
         \caption{}
    \end{subfigure}
    \caption{Three example Markov chains: (a) reversible, (d) recurrent and non-reversible, (g) semi-reversible, (j) non-recurrent and non-reversible. In (b, e, h, k), the transition matrices, stationary distributions and flow matrices of each example are shown, with numerical values rounded to 3 decimal places. In (c, f, i, l), the stationary distributions are visualized by a bar plot, with states represented along the $x$-axis. The right (left) portion of each bar represents the outgoing (incoming) flow of probability mass to (from) all other states. The color of each segment indicates which state the probability mass is flowing to or from, with $s_1$, $s_2$, $s_3$ and $s_4$ depicted in blue, orange, green and red, respectively.}
    \label{fig:GBDB}
\end{figure}

To illustrate some of these points, we consider three Markov chains in \cref{fig:GBDB}, with (a, d, g, j) showing the transition graphs of each example, and (b, e, h, k) showing the respective transition matrix, stationary distribution and flow matrix (for simplicity, each example has a single recurrent class, so that both the stationary distribution and the associated flow matrix are unique). As a visual illustration of the pairwise stationary flow between states in each example, in (c, f, i, l) the stationary distribution is represented as a bar plot, with the portions of probability mass flowing in (left) and out (right) of each state shown as portions of each bar. The example in (a) is reversible, as can be seen by the symmetry of $\mat{F}_{\text{1}}^{\vec{\pi}}$ in (b), or equivalently by the matching between left and right portions of all bars in (c). The example in (d) is almost equivalent to the one in (a), except that the outgoing transition probabilities from $s_1$ have been slightly modified (indicated by the colored arrows in (a, d) and the colored entries of $\mat{P}_{\text{1}}$ and $\mat{P}_{\text{2}}$ in (b, e)). This modification is enough to violate detailed balance, as can be seen by the asymmetry of $\mat{F}_{\text{2}}^{\vec{\pi}}$ or the bar plot in (f). Lastly, the chains
depicted in (g, j) are non-recurrent since in both cases state $s_3$ has only outgoing transitions. Therefore, $\pi_3=0$ and both examples are non-reversible. However, a quick check of $\mat{F}_{\text{3}}^{\vec{\pi}}$ or the bar plot in (i) reveals that the example in (g) is semi-reversible. Conversely, the example in (h) is identical to the one in (g) except for the outgoing transitions from $s_4$ (again indicated by the colored arrows in (g, j) and the colored entries of $\mat{P}_{\text{3}}$ and $\mat{P}_{\text{4}}$ in (h,k)), which leads to an asymmetric stationary flow between the recurrent states.

It is worth pointing out that in our analysis above, reversibility was checked by inspecting the stationary distributions and the corresponding flow matrices of each example. However, since reversibility is a property associated to Markov chains and not to distributions, one might wonder whether there is an alternative way to formalize it based purely on the transition probabilities $P_{ij}$. Clearly, \cref{eq:DB} prohibits one way transitions (i.e.\ $P_{ij}>0$ and $P_{ji}=0$), but this is only a necessary condition of reversibility - can we offer anything more precise? Fortunately, the answer is yes, and it is given by Kolmogorov's criterion \cite{Kolmogoroff1936}:

\begin{theorem}[Kolmogorov's criterion]
\label{thm:KolCrit} A recurrent Markov chain is reversible if and only if the product of one-step transition probabilities along any finite closed path of length more than two is the same as the product of one-step transition probabilities along the reversed path. In other words:
\begin{equation}
\label{eq:KolmEqn}
    P_{i_1,i_2}P_{i_2,i_3}\cdots P_{i_{n-1},i_n}P_{i_n,i_1}=P_{i_1,i_n}P_{i_n,i_{n-1}}\cdots P_{i_3,i_2}P_{i_2,i_1}
\end{equation}
for any $n\geq 2$ and any sequence of states $s_{i_{1}}$, $s_{i_{2}}$, $s_{i_{3}}$, ..., $s_{i_{n-1}}$, $s_{i_{n}}\in\mathcal{S}$.
\end{theorem}
\noindent One way to understand this theorem is that for reversible Markov chains, the probability for traversing any closed path in the state space $\mathcal{S}$ is independent of the direction of traversal. Hence, reversible Markov chains can be thought of as having \textit{zero net circulation}. By contrast, recurrent Markov chains that are non-reversible have at least one path that violates \cref{eq:KolmEqn}, over which there is a higher probability to traverse in one direction than the other. For the example in \cref{fig:GBDB}(a), the relevant closed paths are (up to a cyclic permutation): (i) $s_1\leftrightarrow s_2 \leftrightarrow s_4 \leftrightarrow s_3 \leftrightarrow s_1$, (ii) $s_1\leftrightarrow s_2 \leftrightarrow s_4 \leftrightarrow s_1$, (iii) $s_1\leftrightarrow s_4 \leftrightarrow s_3 \leftrightarrow s_1$. In any of these cases, going around clockwise is equally probable as going around anticlockwise, which is to be expected since this Markov chain is reversible. The example in  \cref{fig:GBDB}(d) has the same closed paths available, except that the outgoing transition probabilities from $s_1$ have been changed. This small adjustment is enough to introduce circulation on all the closed paths: for both (i) and (iii) the anticlockwise direction is more probable since $P_{13}P_{34}P_{42}P_{21}>P_{12}P_{24}P_{43}P_{31}$ and $P_{13}P_{34}P_{41}>P_{14}P_{43}P_{31}$, respectively, and for (ii) the clockwise direction is more probable since $P_{12}P_{24}P_{41}>P_{14}P_{42}P_{21}$. Therefore, by virtue of having at least one path with net circulation, \cref{eq:KolmEqn} confirms that this chain is indeed non-reversible.

The foregoing analyses illustrate how \cref{thm:RevChain} and \cref{thm:KolCrit} provide two alternative but equivalent definitions of reversibility. Something common to both of these interpretations is that reversible Markov chains satisfy a type of equilibrium, either between the exchange of probability mass between pairs of states or the circulation along closed paths, respectively. In fact, the concept of detailed balance stems from early work in the field of statistical mechanics aimed at formalizing the notion of thermodynamic equilibrium on a microscopic level \cite{Gorban2014}. More recently, Markov chain Monte-Carlo methods, which are predominantly based on reversible ergodic chains, have received widespread application in the natural sciences as a way to model systems that are in thermodynamic equilibrium \cite{MatthewRichey2010}. Conversely, Markov chains that violate detailed balance, or equivalently those with net circulation, have been applied to the less well understood case of systems which are out of equilibrium \cite{Jiang2004,Zhang2012,Ge2012}. Furthermore, their stationary distributions have been referred to as \textit{non-equilibrium steady states} (or NESS) \cite{Jiang2004,Zhang2012, Ge2012,Conrad2016, Witzig2018}, which reflects the fact that such distributions are kept fixed over time via unequal flows of probability mass between states ($\vec{\pi}_2$ is an example of a NESS, as can be seen in the bar plot in \cref{fig:GBDB}(f)).

Reversible Markov chains are significantly easier to treat both analytically and numerically than non-reversible chains. Because of this, there exist various procedures for modifying a non-reversible Markov chain so that it becomes reversible, which is sometimes referred to as \textit{reversibilization} \cite{Fill1991,Bremaud1999}. For a recurrent chain, this can be done by taking an average of the forward and backwards transition probabilities, $\mat{P}$ and $\mat{P}_{\text{rev}}$, that describe the chain and its time reversal, respectively. This averaging process can be either additive or multiplicative, leading to the following two definitions:

\begin{defin}[Additive Reversibilization]
\label{defin:AddRev}
Let $\mathcal{X}$ be a recurrent non-reversible Markov chain with transition matrix $\mat{P}$ and a stationary distribution $\vec{\pi}>0$. Then the additive reversibilization of $\mathcal{X}$ is a chain with the following transition matrix:
\begin{align}
    \mat{P}_{\textup{A}}&:=\frac{\mat{P}+\mat{P}_{\textup{rev}}}{2}\\
    &\;=\frac{\mat{P}+\mat{\Pi}^{-1}\mat{P^T}\mat{\Pi}}{2}
\end{align}
and for which $\vec{\pi}$ is also a stationary distribution.
\end{defin}

\begin{defin}[Multiplicative Reversibilization]
For a non-reversible Markov chain with transition matrix $\mat{P}$ and a strictly positive stationary distribution $\vec{\pi}$, the multiplicative reversibilization produces a Markov chain described by the following transition matrix:
\begin{align}
    \mat{P}_{\textup{M}}&:=\mat{P}\mat{P}_{\textup{rev}}\\
    &\;=\mat{P}\mat{\Pi}^{-1}\mat{P^T}\mat{\Pi}
\end{align}
and for which $\vec{\pi}$ is also a stationary distribution.
\end{defin}

Since both definitions produce chains that have the same set of stationary distributions as the starting chain, a simple way to verify their reversibility is to calculate the flow matrix for the distribution $\vec{\pi}$. For the additive reversibilization this gives:
\begin{align}
\mat{F}_{\text{A}}^{\vec{\pi}}&=\mat{\Pi}\mat{P}_{\textup{A}}\\
&=\frac{\mat{\Pi}\mat{P}+\mat{P^T}\mat{\Pi}}{2}\\
&=\frac{\mat{\Pi}\mat{P}+(\mat{\Pi}\mat{P})^T}{2}\\
&=\frac{\mat{F}^{\vec{\pi}}+(\mat{F}^{\vec{\pi}})^T}{2} \label{eq:AdRevFlow}
\end{align}
and for the multiplicative reversibilization:
\begin{align}
    \mat{F}_{\text{M}}^{\vec{\pi}}&=\mat{\Pi}\mat{P}_{\textup{M}}\\
    &=\mat{\Pi}\mat{P}\mat{\Pi}^{-1}\mat{P^T}\mat{\Pi}\\
    &=(\mat{\Pi}\mat{P}\mat{\Pi}^{-\frac{1}{2}})(\mat{\Pi}^{-\frac{1}{2}}\mat{P}^T\mat{\Pi})\\
    &=(\mat{\Pi}\mat{P}\mat{\Pi}^{-\frac{1}{2}})(\mat{\Pi}\mat{P}\mat{\Pi}^{-\frac{1}{2}})^T \label{eq:MultRevFlow}
\end{align}
both of which are by definition symmetric. While \cref{eq:MultRevFlow} does not admit a simple interpretation, \cref{eq:AdRevFlow} says that for the additive reversibilization the flow matrix is symmetric because it corresponds to an average of the forwards flow, $\mat{F}^{\vec{\pi}}$, and backwards flow, $(\mat{F}^{\vec{\pi}})^T$, of the starting chain $\mathcal{X}$. This interpretation is used when we consider random walks on directed graphs in \cref{RWs-dir}.

\subsection{Absorbing chains}
\label{MCs-Abs}
Finally, one concept in the theory of Markov chains that is particularly relevant to applied domains is absorption. A state $s_i\in \mathcal{S}$ is called \textit{absorbing} if it is possible to transition into the state but not out of it, meaning that $P_{ii}=1$ and the chain stays in $s_i$ for all future time steps. An \textit{absorbing Markov chain} is one for which from every state $s_i\in \mathcal{S}$ there exists some path to an absorbing state. Since it is possible to start in a non-absorbing state and never return, all non-absorbing states are transient, and the presence of such states means that absorbing chains can be neither reversible nor ergodic. Absorbing chains often occur in Markov Decision Processses (MDPs), which are central to the field of reinforcement learning \cite{Sutton2018}.

The possible transitions in an absorbing chain can be partitioned into three types: (i) transient $\to$ transient, (ii) transient $\to$ absorbing, and (iii) absorbing $\to$ absorbing. Although the assignment of indices to states in $\mathcal{S}$ is arbitrary, an assignment based on this partitioning simplifies the analysis of absorbing Markov chains. 

\begin{defin}[Canonical Form]
For an absorbing Markov chain with $r$ absorbing and $t$ transient states, the transition matrix $\mat{P}$ can be arranged to have the following block structure, known as the \textit{canonical form}:
\begin{equation}
\label{eq:CanForm}
    \mat{P}=\left(
    \begin{array}{c|c}
    \mat{Q} & \mat{R} \\ \hline
    \mat{0} & \mat{\mathbbm{1}} \\
    \end{array}
    \right)
\end{equation}
where $\mat{Q}\in\mathbb{R}^{t\times t}$, $\mat{R}\in\mathbb{R}^{t\times r}$ and $\mat{\mathbbm{1}}\in\mathbb{R}^{r\times r}$ describe transitions of type (i), (ii), and (iii), respectively, and $\mat{0}\in \mathbb{R}^{r \times t}$ is a matrix of zeros. Thus, matrix $\mat{Q}$ is what remains of $\mat{P}$ when we remove any absorbing states from $\mathcal{S}$.
\end{defin}

\begin{figure}[h]
     \centering
     \begin{subfigure}[b]{0.35\textwidth}
         \centering
         \includegraphics[width=\textwidth]{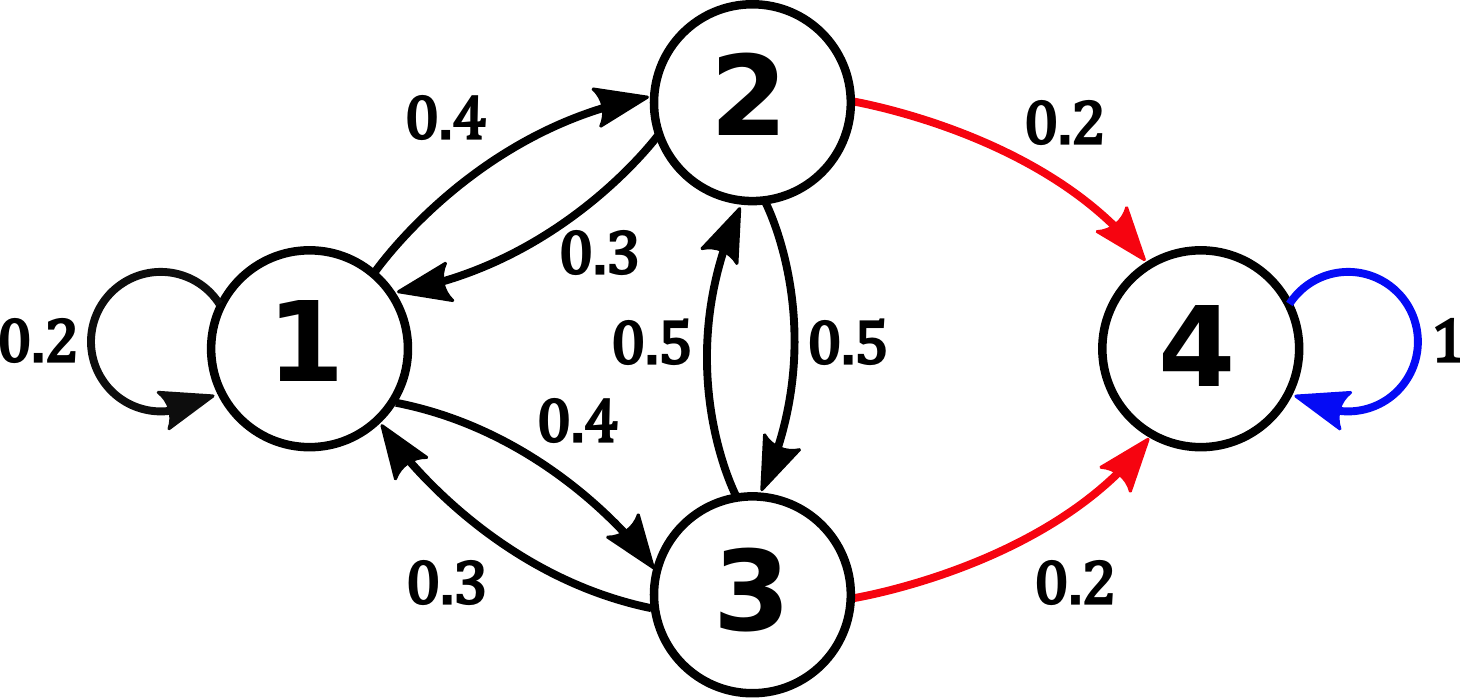}
         \caption{}
     \end{subfigure}
     \quad
     \begin{subfigure}[b]{0.3\textwidth}
         \centering
         $\mat{Q}=\left(
         \begin{array}{cccc}
         0.2 & 0.4 & 0.4 \\
         0.3 & 0 & 0.5 \\
         0.3 & 0.5 & 0
         \end{array} \right)$\\
         \bigskip
         $\mat{R}=\left(
         \begin{array}{cccc}
         0 \\
         0.2 \\
         0.2 
         \end{array} \right)$
         \caption{}
     \end{subfigure}
      \caption{(a) The transition graph of an absorbing Markov chain with $4$ states, where $s_4$ is an absorbing state, and (b) the matrices $\mat{Q}$ and $\mat{R}$ associated to this chain.}
    \label{fig:AbsChain}
\end{figure}

We depict this partitioning of transition probabilities in \cref{fig:AbsChain}(a). An absorbing chain with one absorbing state is shown, with the transitions belonging to $\mat{Q}$, $\mat{R}$ and $\mat{\mathbbm{1}}$ colored black, red and blue, respectively. Furthermore, in \cref{fig:AbsChain}(b) we show the matrices $\mat{Q}$ and $\mat{R}$.

Since any transient state can reach an absorbing state in a finite number of steps, the probability that the chain ends up in an absorbing state at some future time is 1. For this reason, in the infinite time limit we can expect to see no transitions taking place between transient states, i.e.\ $\lim_{n\to\infty}\mat{Q}^n=0$. This is an advantageous property, since it means that if we sum up all powers of $\mat{Q}$, known as the Neumann series of $\mat{Q}$, then the contributions for larger powers get progressively smaller and the sum converges to $(\mat{\mathbbm{1}}-\mat{Q})^{-1}$ (see \cite{Meyer2000} p.\ 618). Calculating this sum for $\mat{Q}$ leads to the following useful quantity which relates transient states in $\mathcal{S}$ \cite{Porod2021}:

\begin{defin}[Fundamental Matrix]
\label{defin:FundMat}
For any absorbing Markov chain, the Neumann series of matrix $\mat{Q}$ is given by:
\begin{align}
    \mat{N}&:=\sum_{k=0}^{\infty} \mat{Q}^k\\
    &\;=\mat{\mathbbm{1}}+\mat{Q}+\mat{Q}^2+\mat{Q}^3+...\\
    &\;=(\mat{\mathbbm{1}}-\mat{Q})^{-1}
\end{align}
and is known as the fundamental matrix of the Markov chain. The elements of this matrix $N_{ij}$ give the expected number of times the chain visits a transient state $s_j$ before absorption, given that the chain started in a transient state $s_i$.
\end{defin}

When analyzing an absorbing chain, it is very handy to have access to the fundamental matrix. By taking into account all non-negative powers of $\mat{Q}$, it contains information about all possible paths available between pairs of transient states. Because of this, it is a useful predictive tool that allows several properties of the Markov chain to be deduced \cite{Porod2021}. Furthermore, in the field of reinforcement learning, it is closely related to the successor representation \cite{Dayan1993}.

\subsection{Summary}
\begin{wrapfigure}[20]{r}{0.55\textwidth}
\vspace{-25pt}
  \centering
   \includegraphics[width=0.45\textwidth]{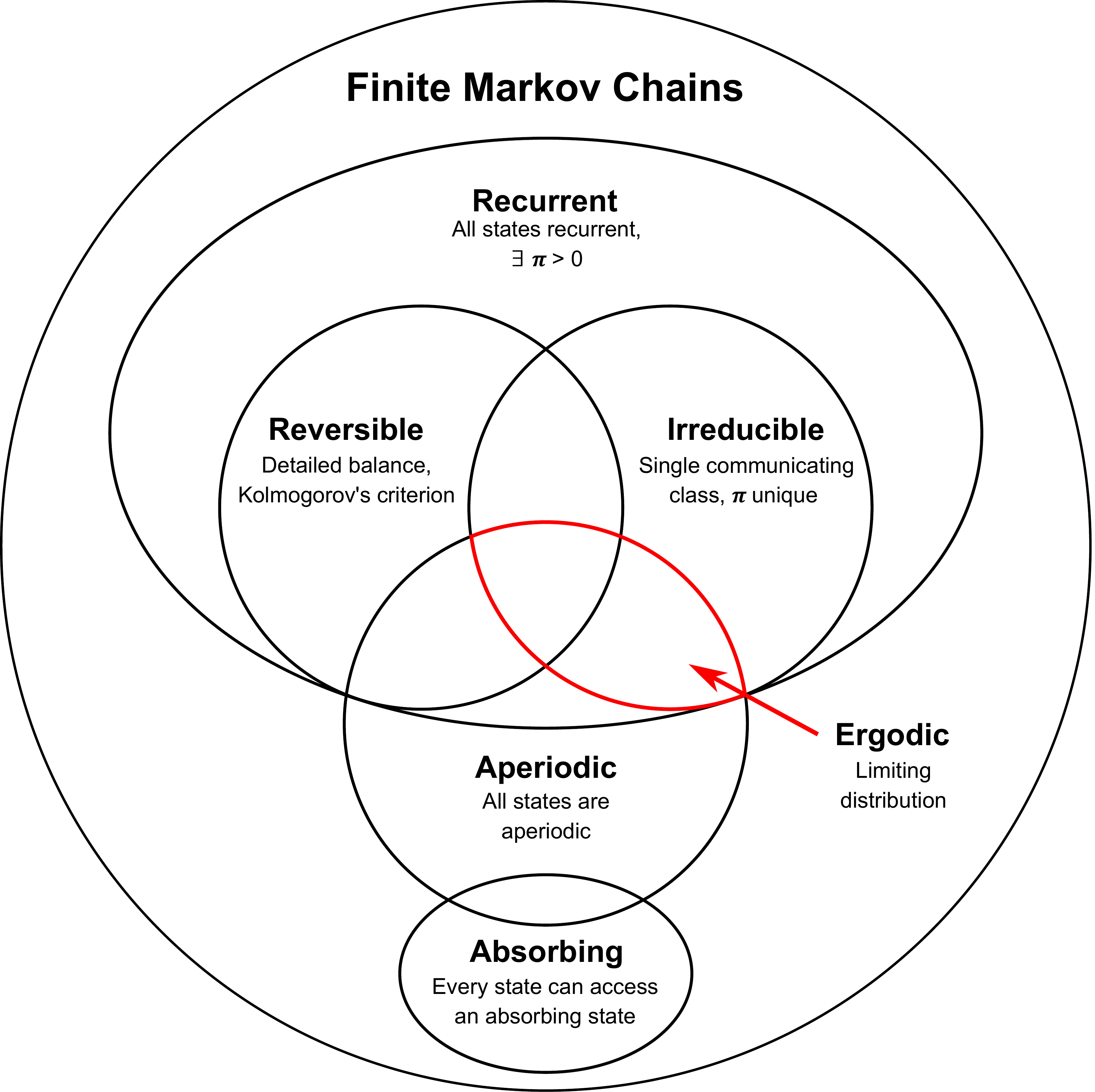}
  \caption{A Venn diagram composed of the different types of Markov chains introduced in this section.}
  \label{fig:MCVenn}
\end{wrapfigure}
This concludes our exploration of different types of Markov chains. In \cref{fig:MCVenn}, we provide a summary of the material presented in this section in the form of a Venn diagram. In this diagram, each type of Markov chain is drawn as a circle or ellipse, with defining properties/results listed in each case. Take a moment to look at this image and pay attention to the overlapping regions which indicate how different types of chains are related. Furthermore, for a more in-depth presentation of the material in this section, we recommend \cite{Porod2021} and \cite{Bremaud1999}.

In the next section, we introduce graphs as an alternative way to describe Markov chains and summarize insights that emerge from this description. Then, in \cref{RWs} the connection between graphs and Markov chains is explored in more depth using the notion of random walks, which allows various relationships to be made between specific types of graphs and some types of Markov chains introduced in this section.
\section{Graphs}
\label{Graphs}

So far, we have implicitly been interpreting Markov chains as graphs whenever we draw a transition graph. In the current section, we formally introduce the concept of graphs, which provides a foundation to the material on random walks in \cref{RWs}. Readers should note that definitions in graph theory often vary between different sources. Here we use a convention that can encompass a wider variety of graphs, thereby offering greater generality.

\subsection{Definition}
\begin{figure}[h]
     \centering
     \begin{subfigure}[t]{0.3\textwidth}
         \centering
         \includegraphics[width=0.6\textwidth]{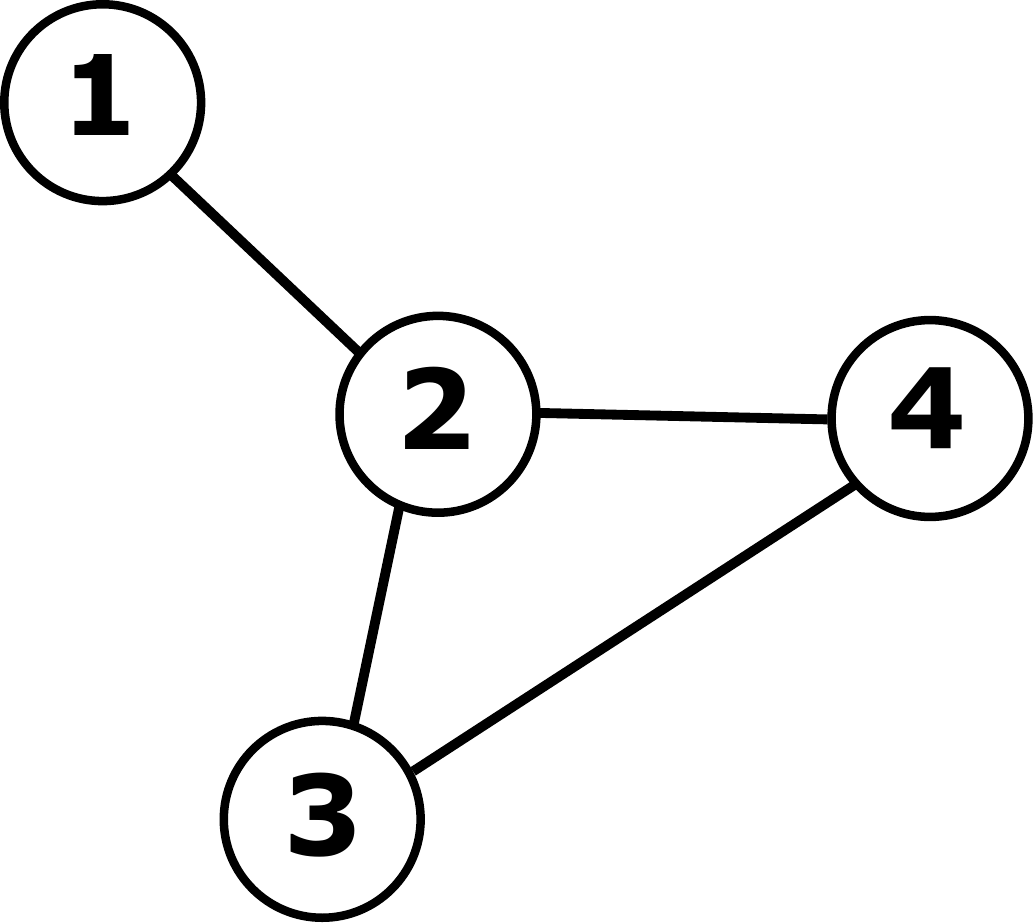}
         \par\bigskip
         $\mat{A}=\left(
         \begin{array}{cccc}
         0 & 1 & 0 & 0 \\
         1 & 0 & 1 & 1 \\
         0 & 1 & 0 & 1 \\
         0 & 1 & 1 & 0 \\
         \end{array} \right)$
         \par\smallskip
         \caption{}
     \end{subfigure}
     \hfill
     \begin{subfigure}[t]{0.3\textwidth}
         \centering
         \includegraphics[width=0.6\textwidth]{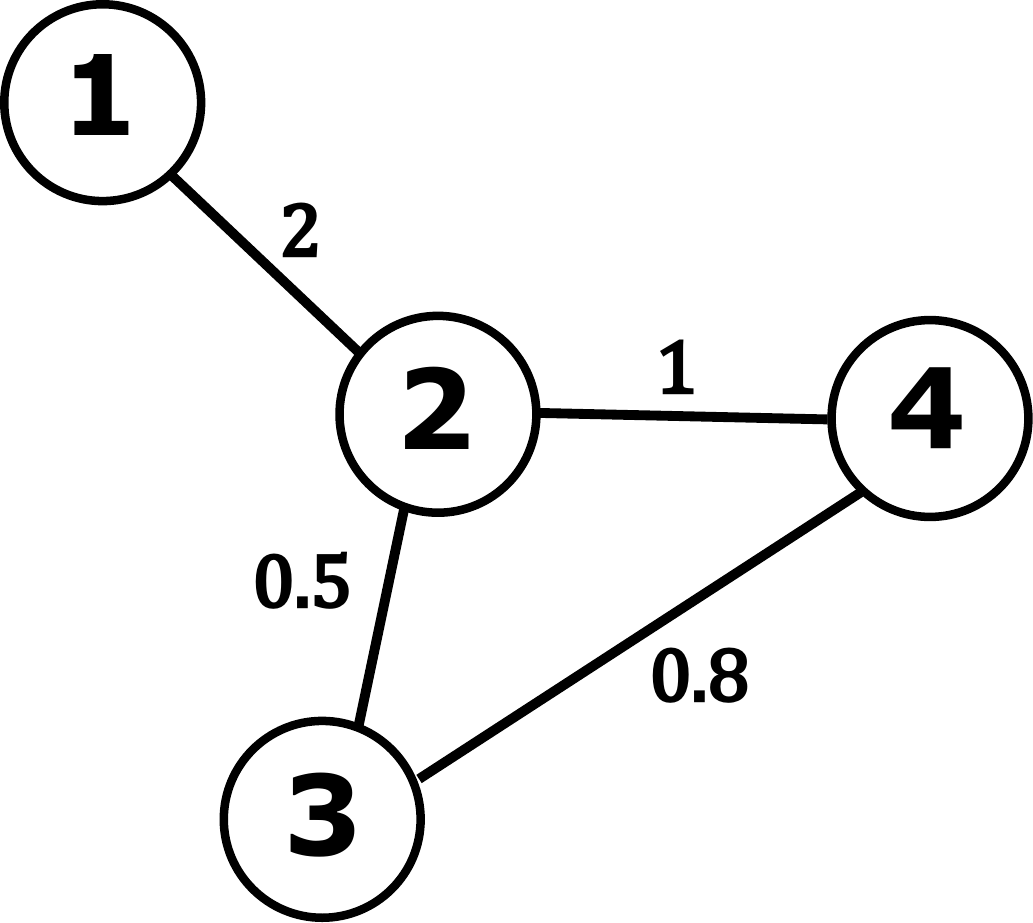}
         \par\bigskip
         $\mat{W}=\left(
         \begin{array}{cccc}
         0 & 2 & 0 & 0 \\
         2 & 0 & 0.5 & 1 \\
         0 & 0.5 & 0 & 0.8 \\
         0 & 1 & 0.8 & 0 \\
         \end{array} \right)$
         \par\smallskip
         \caption{}
     \end{subfigure}
     \hfill
     \begin{subfigure}[t]{0.3\textwidth}
     \centering         
     \includegraphics[width=0.6\textwidth]{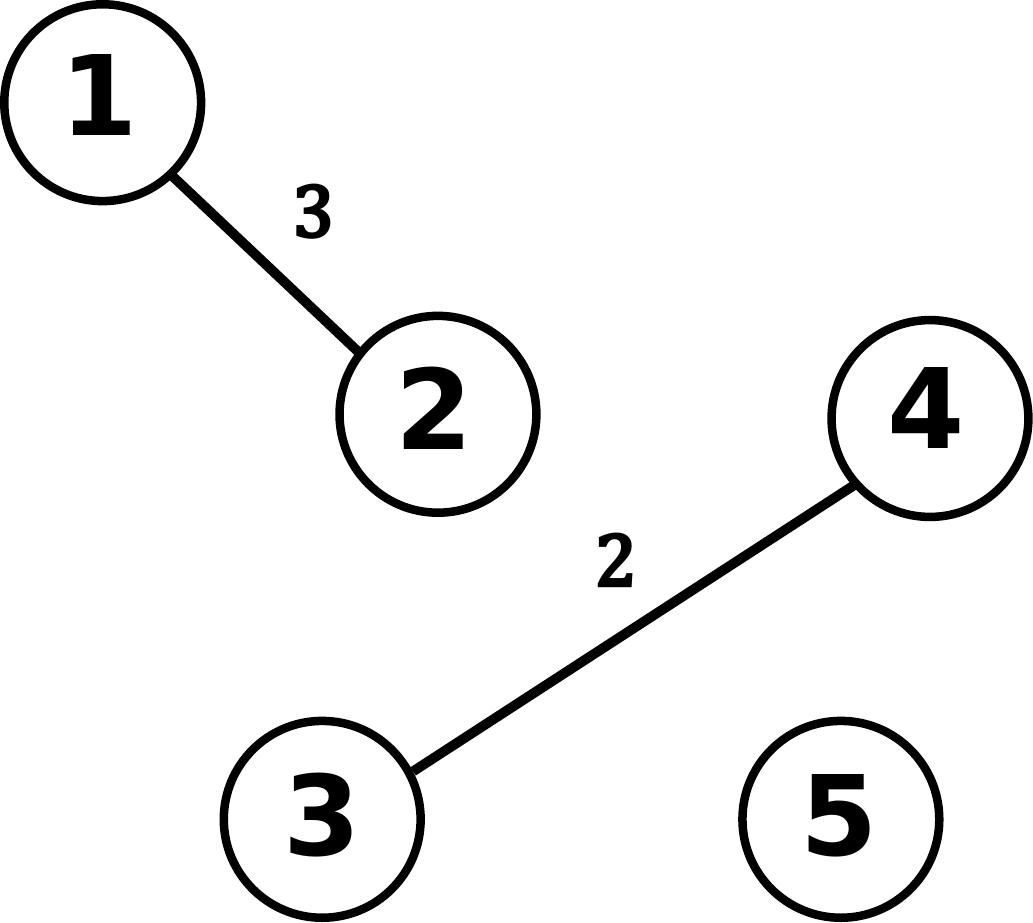}
     \par\bigskip
         $\mat{W}=\left(
         \begin{array}{cccc}
         0 & 3 & 0 & 0 \\
         3 & 0 & 0 & 0 \\
         0 & 0 & 0 & 2 \\
         0 & 0 & 2 & 0 \\
         \end{array} \right)$
         \par\smallskip
     \caption{}
     \end{subfigure}
     \par\bigskip
     \begin{subfigure}[t]{0.3\textwidth}
         \centering
         \includegraphics[width=0.6\textwidth]{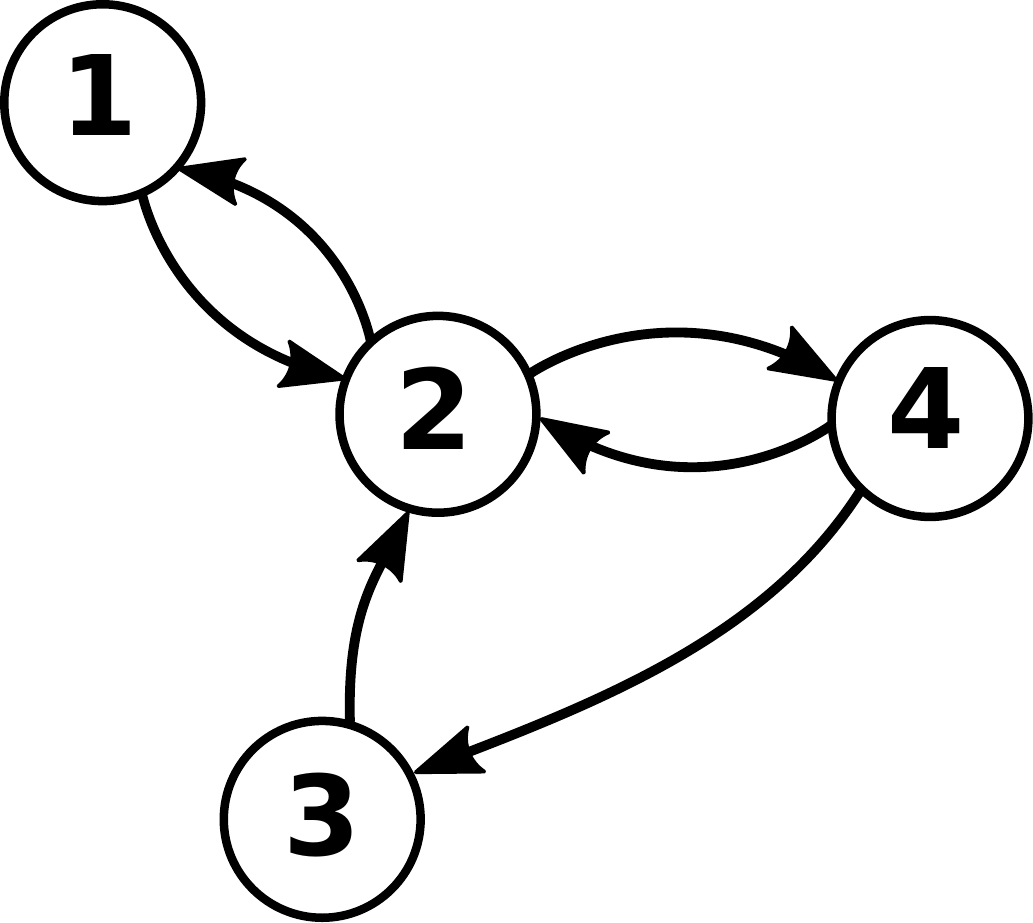}
         \par\bigskip
         $\mat{A}=\left(
         \begin{array}{cccc}
         0 & 1 & 0 & 0 \\
         1 & 0 & 0 & 1 \\
         0 & 1 & 0 & 0 \\
         0 & 1 & 1 & 0 \\
         \end{array} \right)$
         \par\smallskip
         \caption{}
     \end{subfigure}
     \hfill
     \begin{subfigure}[t]{0.3\textwidth}
         \centering
         \includegraphics[width=0.6\textwidth]{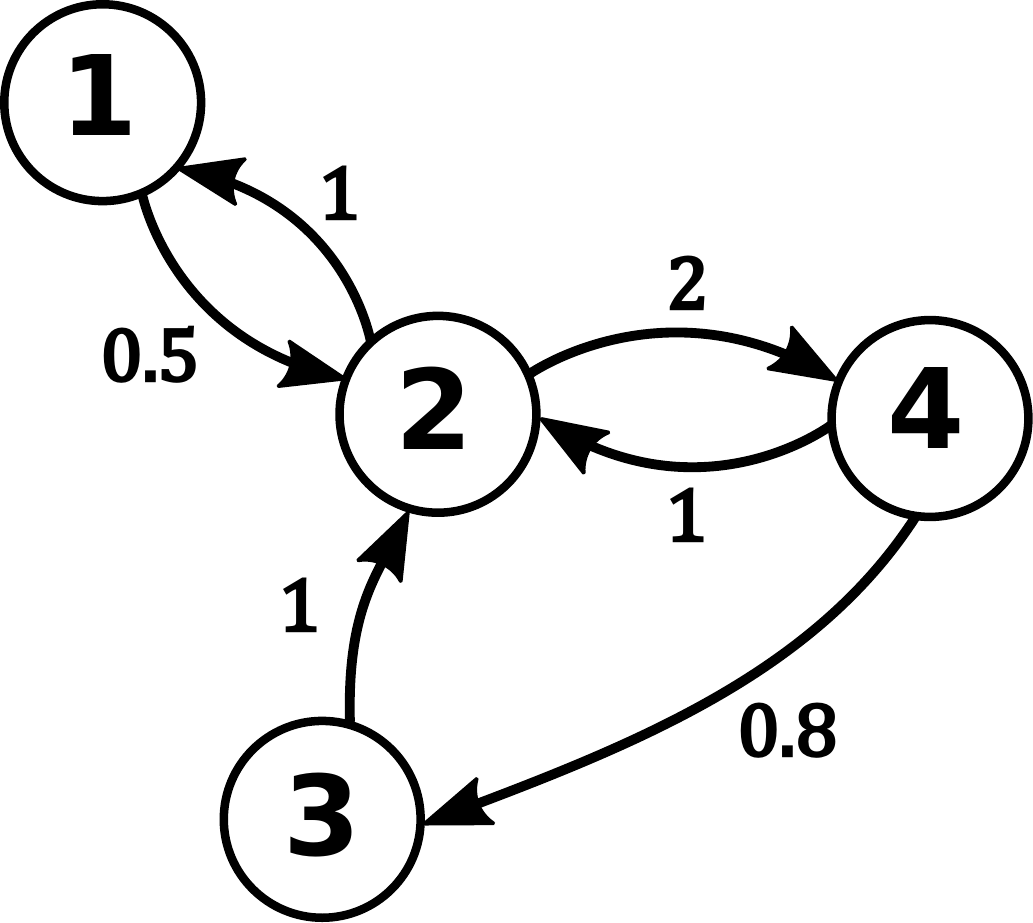}
         \par\bigskip
         $\mat{W}=\left(
         \begin{array}{cccc}
         0 & 0.5 & 0 & 0 \\
         1 & 0 & 0 & 2 \\
         0 & 1 & 0 & 0 \\
         0 & 1 & 0.8 & 0 \\
         \end{array} \right)$
         \par\smallskip
         \caption{}
     \end{subfigure}
     \hfill
     \begin{subfigure}[t]{0.3\textwidth}
     \centering         
     \includegraphics[width=0.6\textwidth]{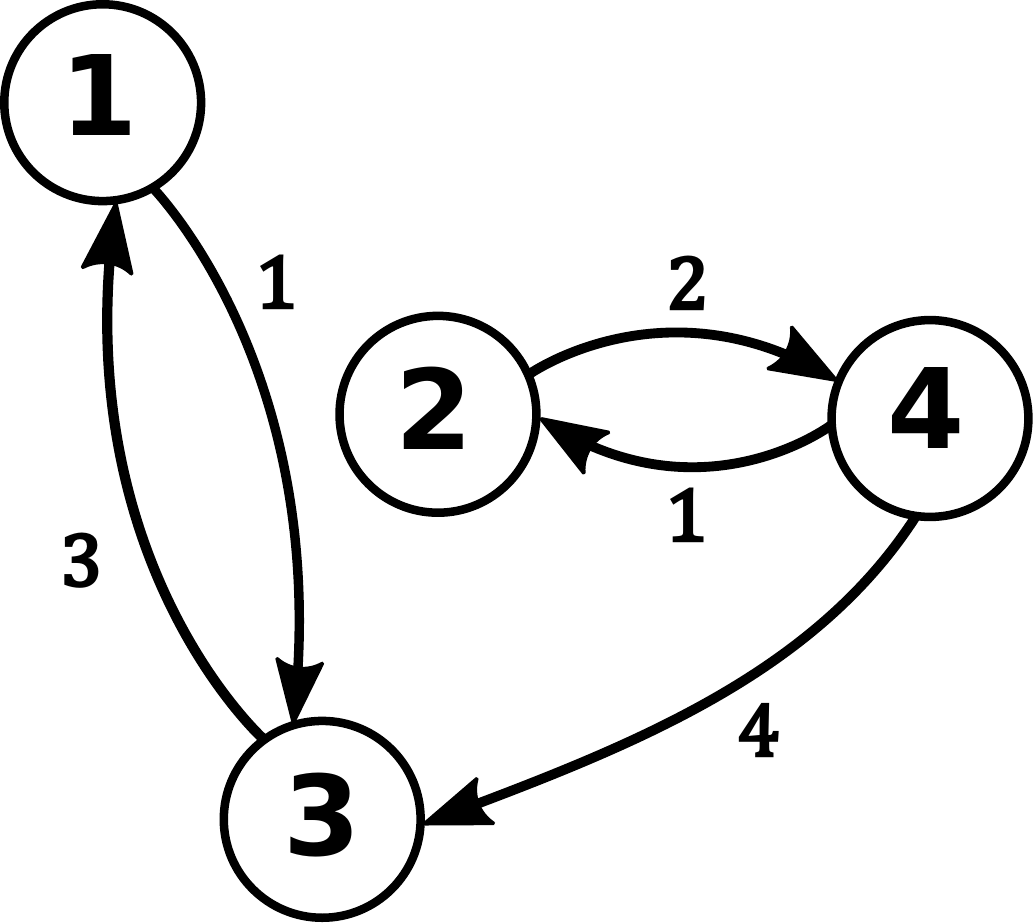}
     \par\bigskip
     $\mat{W}=\left(
     \begin{array}{cccc}
     0 & 0 & 1 & 0 \\
     0 & 0 & 0 & 2 \\
     3 & 0 & 0 & 0 \\
     0 & 1 & 4 & 0 \\
     \end{array} \right)$
     \par\smallskip
     \caption{}
     \end{subfigure}
        \caption{Six examples of graphs and their corresponding matrices. (a-c) are undirected whereas (d-f) are directed. Furthermore, (a,d) are unweighted whereas (b,c,e,f) are weighted.}
        \label{fig:Graph+Mat}
\end{figure}

A graph $G=(V,E)$ is a set of $N$ vertices $V=\{v_1,v_2, \cdots, v_N\}$ together with an edge set $E$ containing pairs of vertices in $V$. Conceptually, $V$ might represent a collection of objects, and $E$ a specification of how some pairs in this collection are related to one another. A natural way to categorize graphs is based on the way in which edges are defined. For instance, in an \textit{undirected graph} each edge has no direction and is typically denoted as $(v_i, v_j)\in E$, whereas in a \textit{directed graph} each edge has a specified starting and ending vertex and is usually denoted as $(v_i\to v_j)\in E$. Examples of undirected and directed graphs can be seen in the first and second rows of \cref{fig:Graph+Mat}, respectively. Unless otherwise stated, we depict undirected edges as straight lines and directed edges as curved lines with arrowheads indicating the direction. A second distinction we can make is between \textit{unweighted graphs}, in which one only cares about whether two vertices are related or not, and \textit{weighted graphs}, in which each edge has a positive weight $w_{ij}$ describing the strength of the relationship.\footnote{We restrict edge weights to be positive in order to maintain this notion of strength, however it is worth noting that some conventions in graph theory allow negative weights.} In \cref{fig:Graph+Mat}, the examples in the first column are unweighted and all other graphs are weighted, with weights indicated by numbers next to each edge. The type of edges that a graph has is often chosen based on the type of relationship that one wants to describe. For example, assume that we have a graph $G$ where vertices represent PhD students. Then, if we want to represent the relationship of being in the same research group, undirected unweighted edges are a natural choice (such as \cref{fig:Graph+Mat}(a)). Conversely, if we want edges to describe whether one student has participated on a main project of another student, then this clearly requires directed unweighted edges (such as \cref{fig:Graph+Mat}(d)). If we now consider variants of the first and second examples, instead focusing on how similar the research topics of two students are, or how much work has one student has contributed to another student's project, then we now need undirected weighted and directed weighted edges, respectively (such as \cref{fig:Graph+Mat}(b,c) and \cref{fig:Graph+Mat}(e,f)). It is worth noting that in order to assign weights to edges, one needs to specify a scale on which to measure the strength of relationships between vertices.

One can also describe graphs based on their connectivity. In an undirected graph, if there exists a path between each pair of vertices then the graph is said to be \textit{connected}, otherwise it is \textit{disconnected}. The notion of connectivity can generalize to disconnected graphs if we instead consider subsets of vertices in $G$, which are known as \textit{subgraphs}. Any subgraph that is connected but is not part of any larger connected subgraph is called a \textit{connected component}. Both of the undirected graphs in \cref{fig:Graph+Mat}(a, b) are connected, whereas \cref{fig:Graph+Mat}(c) shows an example that is disconnected, with two connected components. In particular, this latter example even has a vertex that does not have any edges at all, which is known as an \textit{isolated vertex}. For a directed graph, if there are directed paths running from $v_i$ to $v_j$ and from $v_j$ to $v_i$ for all pairs of vertices $v_i, v_j \in V$ then the graph is said to be \textit{strongly connected}. Alternatively, a directed graph is \textit{weakly connected} if for all pairs of vertices $v_i, v_j \in V$ it is possible to get from $v_i$ to $v_j$ and from $v_j$ to $v_i$ by any path, regardless of the direction of edges. Clearly, a directed graph is weakly connected if it is strongly connected, but not vice versa. Furthermore, strongly or weakly connected subgraphs that are not part of any larger such subgraphs are referred to as \textit{strongly} or \textit{weakly connected components}, respectively. The directed graphs in \cref{fig:Graph+Mat}(d,e) are strongly connected, whereas the one in \cref{fig:Graph+Mat}(f) is only weakly connected and has a two strongly connected components  (take a moment to verify this).

\subsection{Matrix representation}
A natural way to numerically represent graphs with $|V|=N$ vertices is with a $N\times N$ matrix. In the unweighted case, this matrix is a binary matrix $\mat{A}$, with entries:
\begin{equation}
\label{eq:AdjMat}
A_{ij}:=\left\{
	\begin{array}{ll}
		1 & \mbox{if } (v_i \sim v_j)\in E\\
		0 & \mbox{otherwise}
	\end{array}
\right.
\end{equation}
where $(v_i \sim v_j)=(v_i,v_j)$ when $G$ is undirected, and $(v_i \sim v_j)=(v_i\to v_j)$ when it is directed. Furthermore, the matrix $\mat{A}$ is usually referred to as the \textit{adjacency matrix} of $G$. This extends easily to the weighted case, where instead we have a non-negative matrix $\mat{W}$, with entries:
\begin{equation}
\label{eq:WMat}
W_{ij}:=\left\{
	\begin{array}{ll}
		w_{ij} & \mbox{if } (v_i \sim v_j)\in E\\
		0 & \mbox{otherwise}
	\end{array}
\right.
\end{equation}
which is often referred to as the \textit{weight matrix} of $G$. In \cref{fig:Graph+Mat}, the relevant matrix is shown below each graph, and we encourage readers to verify that they match in each case. Furthermore, something worth pointing out is that undirected graphs always have corresponding matrices that are symmetric, as can be seen in \cref{fig:Graph+Mat}(a-c).

\begin{wrapfigure}[12]{r}{0.37\textwidth}
\vspace{-10pt}
  \centering
 \includegraphics[width=0.2\textwidth]{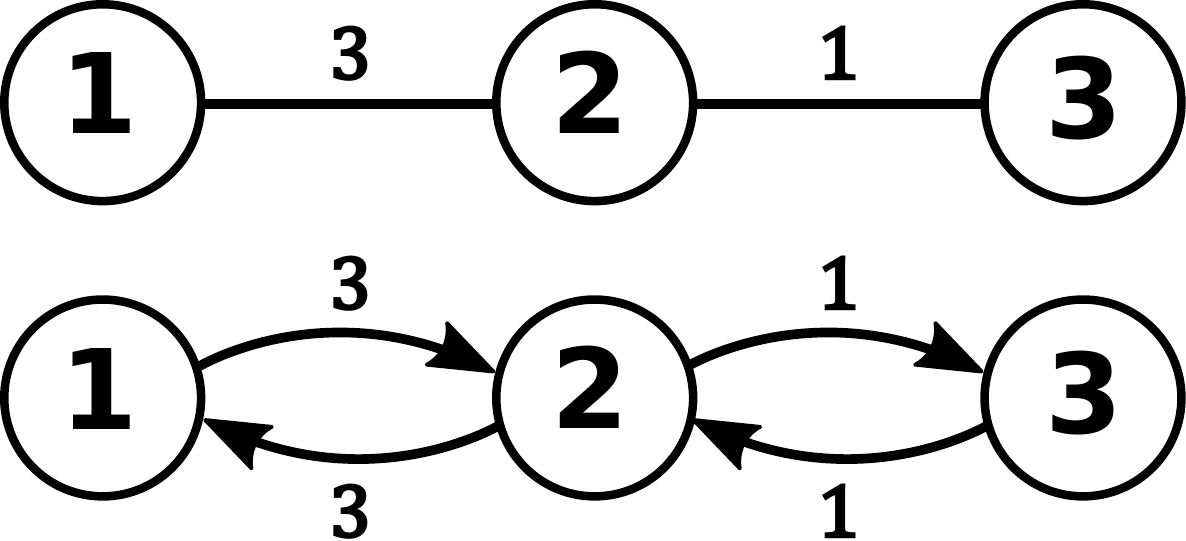}
 \par\bigskip
 $\mat{W}=\left(
 \begin{array}{ccc}
 0 & 3 & 0 \\
 3 & 0 & 1 \\
 0 & 1 & 0
 \end{array} \right)$
 \caption{An undirected graph represented both with undirected edges (top) and pairs of directed edges (middle), as well as its weight matrix (bottom).}
 \label{fig:GraphEdges}
\end{wrapfigure}

For the remainder of this tutorial, we assume that the graphs we deal with are both weighted and directed. The reason we choose this convention is that it is more general. On one hand, any unweighted graph can be considered as a special case of a weighted graph where the weights are all set to $1$. Thus, we henceforth only talk about weight matrices as opposed to adjacency matrices when describing graphs numerically. On the other hand, there is one sense in which directed graphs can be thought of as a generalization of undirected graphs. If $v_i$ and $v_j$ are two distinct vertices that share an edge in an undirected graph, then the weight of this edge is guaranteed to appear twice in the weight matrix $\mat{W}$ by virtue of it being symmetric. If instead these vertices belong to a directed graph and there is an edge $(v_i\to v_j)$, then this edge appears only once in $\mat{W}$. Therefore, it is possible to interpret an undirected edge between $v_i$ and $v_j$ as being equivalent to a pair of directed edges of the same weight, with one connecting $v_i$ to $v_j$ and the other connecting $v_j$ to $v_i$. This is the interpretation we use throughout the rest of the tutorial whenever we refer to undirected graphs. As an example, in \cref{fig:GraphEdges} we show two equivalent depictions of an undirected graph, with the top image drawn in the usual way, and the bottom image drawn using pairs of directed edges. Below these two drawings, the weight matrix of this graph is shown. One must note that interpreting undirected graphs in this way is somewhat atypical, however it allows us a greater level of generality when dealing with different types of graphs in \cref{RWs}. Furthermore, this interpretation only applies to edges between distinct vertices, and edges that connect vertices to themselves are discussed in \cref{Graphs-self}.

\subsection{Vertex degrees}

Once the weight matrix of a graph is known, it is easy to calculate the total weight coming in and out of each vertex. The total incoming weight of a vertex $v_i$ can be found by summing over the $i$-th column of $\mat{W}$, and is known as the \textit{in-degree} of $v_i$, i.e.\ $d_i^-:=\sum_{j=1}^NW_{ji}$. Conversely, the total outgoing weight of $v_i$ is calculated by the sum over the $i$-th row of $\mat{W}$, i.e.\ $d_i^+:=\sum_{j=1}^NW_{ij}$, and is known as the \textit{out-degree} of $v_i$. Since undirected graphs always have bidirectional edges and symmetric weight matrices, the in- and out-degrees of such graphs are always equal, and are simply referred to as vertex \textit{degrees}, denoted by $d_i$.\footnote{Since in graph theory unweighted graphs are a more commonly studied than weighted graphs, the degree quantities we have defined are sometimes referred to as \textit{weighted degrees} \cite{Chapman2011}, but for simplicity we just use the term degree.} \begin{wrapfigure}[15]{r}{0.37\textwidth}
 \centering
 \includegraphics[width=0.2\textwidth]{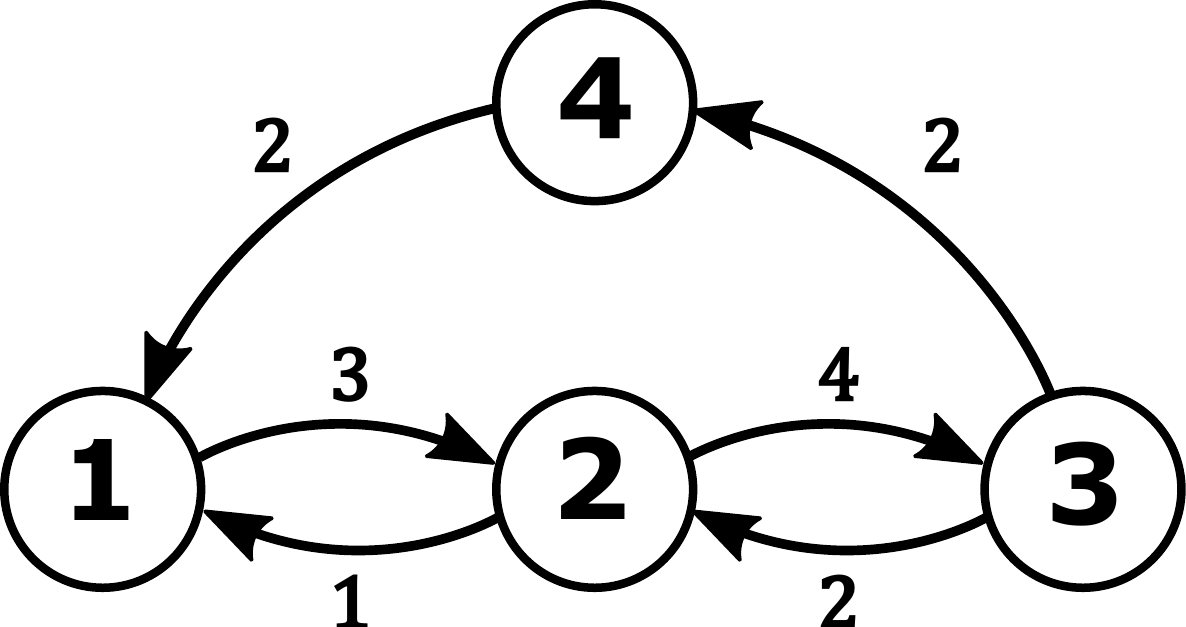}
 \par\bigskip
 $\mat{W}=\left(
 \begin{array}{cccc}
 0 & 3 & 0 & 0 \\
 1 & 0 & 4 & 0 \\
 0 & 2 & 0 & 2 \\
 2 & 0 & 0 & 0
 \end{array} \right)$
 \caption{A balanced graph (top) and its weight matrix (bottom).}
 \label{fig:BalancedGraph}
\end{wrapfigure} For example, in the graph of \cref{fig:GraphEdges} the degrees are $d_1=3$, $d_2=4$ and $d_3=1$. In the more general case of directed graphs, there is no guarantee that $d_i^+=d_i^-$. However, summing over all in- or out- degrees for any graph always produces the same number, i.e.\ $\text{vol}(G)=\sum_{i=1}^Nd_i^+=\sum_{i=1}^Nd_i^-=\sum_{i=1}^N\sum_{j=1}^NW_{ij}$, which is sometimes referred to as the \textit{volume} of $G$. As an example, consider the directed graph in \cref{fig:Graph+Mat}(e): each vertex has a different in- and out-degrees, i.e.\ $d_1^-=1$, $d_1^+=0.5$, $d_2^-=2.5$, $d_2^+=3$, $d_3^-=0.8$, $d_3^+=1$, $d_4^-=2$ and $d_4^+=1.8$, but summing over either of the degree types yields $\text{vol}(G)=6.3$. Nonetheless, some directed graphs can have $d_i^+=d_i^-$ for each vertex, and such cases are known as \textit{balanced graphs} \cite{Banderier2000,Aldous2002}. In keeping with the notation of undirected graphs, we denote the vertex degrees of a balanced graph as $d_i=d_i^+=d_i^-$. An example of a balanced graph along with its corresponding weight matrix is shown in \cref{fig:BalancedGraph}, and a quick check reveals that summing over the rows or columns of $\mat{W}$ indeed yields the same values. Just as a balanced graph is a special case of a directed graph, we can similarly say that an undirected graph is a special case of a balanced graph, and this interpretation is important in \cref{RWs}.

\subsection{Self-loops}
\label{Graphs-self}
In the examples considered so far, all edges connect pairs of vertices that are distinct, i.e.\ $v_i\neq v_j$. While this is sometimes enforced as a rule, some conventions also allow edges to connect vertices to themselves, which are known as \textit{self-loops}. For undirected graphs, the standard convention is that a self-loop at vertex $v_i$ counts doubly to the vertex degree $d_i$, while other edges only count singly, i.e.\ $d_i=2\times W_{ii} +\sum_{j\neq i}W_{ij}$. This somewhat counter-intuitive property is typically demonstrated using the \textit{degree sum formula} \cite{West2001}. For undirected graphs, this states that each edge contributes twice its weight to the volume. Since self-loops only involve a single vertex, the only way that this rule can be respected is if they count twice as much to the vertex degrees as other edges. A property that we require when dealing with undirected graphs in \cref{RWs} is that the vertex degrees are calculated by the row sums of $\mat{W}$.\begin{wrapfigure}[20]{r}{0.37\textwidth}
\vspace{-10pt}
     \centering
     \begin{subfigure}[b]{0.2\textwidth}
         \centering
         \includegraphics[width=0.9\textwidth]{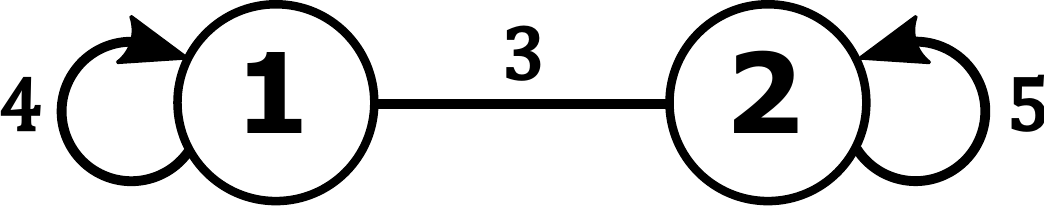}
         \par\bigskip
         $\mat{W}=\left(
         \begin{array}{cc}
         4 & 3 \\
         3 & 5 \\
         \end{array} \right)$
         \par\smallskip
         \caption{}
     \end{subfigure}
     \par\bigskip
     \begin{subfigure}[b]{0.2\textwidth}
         \centering
         \includegraphics[width=0.9\textwidth]{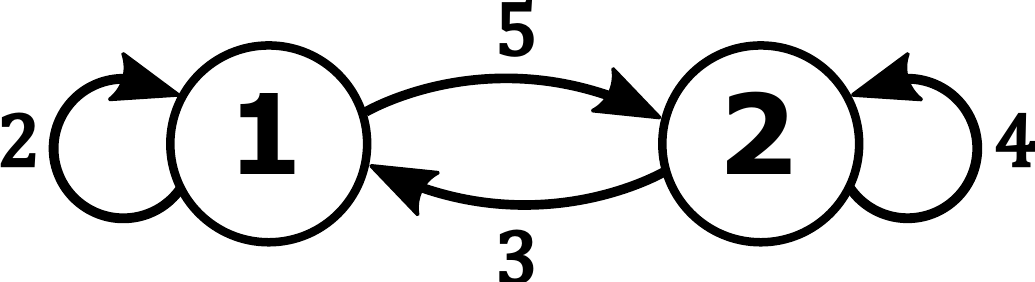}
         \par\bigskip
         $\mat{W}=\left(
         \begin{array}{cc}
         2 & 5 \\
         3 & 4 \\
         \end{array} \right)$
         \par\smallskip
         \caption{}
     \end{subfigure}
        \caption{(a) An undirected graph containing self-loops, and its weight matrix (bottom). (b) A directed graph containing self-loops (top), and its weight matrix (bottom).}
        \label{fig:GraphLoops}
\end{wrapfigure} Clearly, this property is violated by the factor of $2$ that applies to undirected self-loops. As a result, in this tutorial we assume that self-loops are always directed, regardless of whether they occur in undirected or directed graphs. This is an atypical definition, since undirected graphs typically are not allowed to have directed edges. However, as can be seen from the examples in \cref{fig:GraphLoops}, this preserves the fact that undirected graphs have symmetric weight matrices, whereas directed graphs have non-symmetric weight matrices, which is sufficient for the scope of this tutorial.

\bigskip

We close this section by noting some similarities between our definitions of Markov chains and  graphs. Firstly, the transition matrices of Markov chains, like the weight matrices of graphs, are non-negative. Secondly, in a directed graph any entry $W_{ij}\neq 0$ of the weight matrix describes an outgoing edge from vertex $v_i$ to $v_j$, and analogously any entry $P_{ij}\neq 0$ of a transition matrix describes an outgoing transition probability from $s_i$ to $s_j$. Putting these together, we see that in the most general sense any Markov chain can be thought of as a directed graph, with $\mat{P}$ being the associated weight matrix. Indeed, this interpretation is precisely what justifies us in visualizing a Markov chain by its transition graph. In the next section, we present some useful results that emerge as a result of this way of thinking about a Markov chain. Lastly, for a comprehensive text on graph theory that covers much of the material in this section, we recommend \cite{West2001}.

\subsection{Eigenspaces of transition matrices}
\label{Graphs-Eig}
Non-negative matrices have received widespread attention in mathematics, and in particular their eigenvalues and eigenvectors are the focus of \textit{spectral graph theory} \cite{Chung1997}. In this section, we apply some results from this field to transition matrices, considering first irreducible chains and then subsequently exploring the generalization to reducible chains.

\subsubsection{Irreducible chains}

A fundamental result used in spectral graph theory is the \textit{Perron-Frobenius theorem}, and while a full treatment of it is beyond the scope of this tutorial, we now summarize its key implications for transition matrices of irreducible Markov chains.

\begin{theorem}[Perron Frobenius theorem for irreducible Markov chains]
\label{thm:PeigsIrred}
If $\mat{P}$ is the transition matrix of an irreducible Markov chain, then:
\begin{itemize}
    \item $\lambda=1$ is guaranteed to be an eigenvalue.
    \item $\lambda=1$ is simple eigenvalue, meaning that it occurs only once.
    \item Upon suitable normalization, the eigenvalue $\lambda=1$ has a left eigenvector equal to the unique stationary distribution $\vec{\pi}=(\pi_1, \pi_2, ..., \pi_N)^T$ and a right eigenvector equal to $\vec{\eta}=(1, 1, ..., 1)^T$.
    \item All other eigenvalues have $|\lambda|\leq 1$, where $|\cdot|$ is the complex modulus, meaning that the spectral radius of $\mat{P}$ is $1$.\footnote{We remind readers that the spectral radius of a square matrix $\mat{X}$ is its largest eigenvalue in absolute value, and is denoted $\rho(\mat{X})$. The name relates to the fact that all eigenvalues are contained within a disk of radius $\rho(\mat{X})$ centered at the origin of the complex plane.}
\end{itemize}
\end{theorem}

\begin{figure}[h]
     \centering
     \begin{subfigure}[b]{0.4\textwidth}
         \centering
         \includegraphics[width=\textwidth]{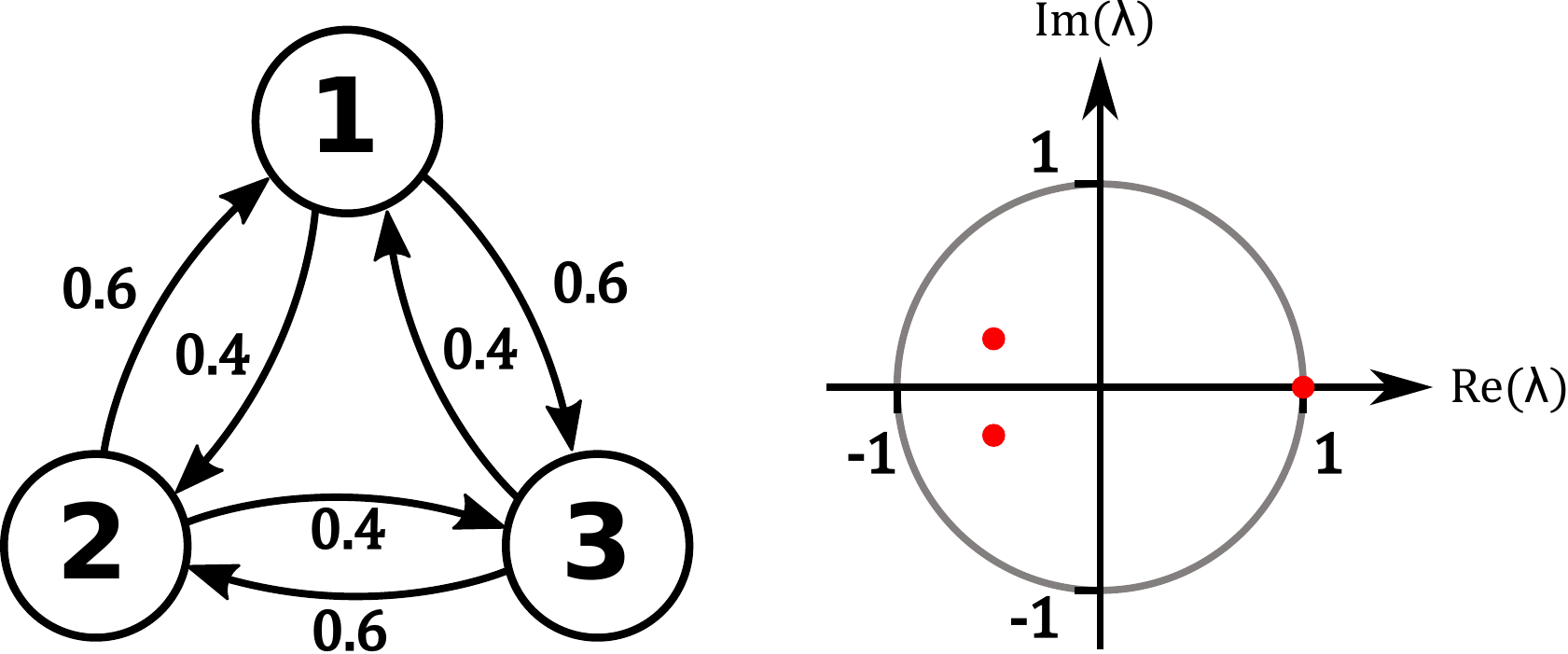}
         \caption{irreducible, $d=1$}
     \end{subfigure}
     \quad
     \begin{subfigure}[b]{0.4\textwidth}
         \centering
         \includegraphics[width=\textwidth]{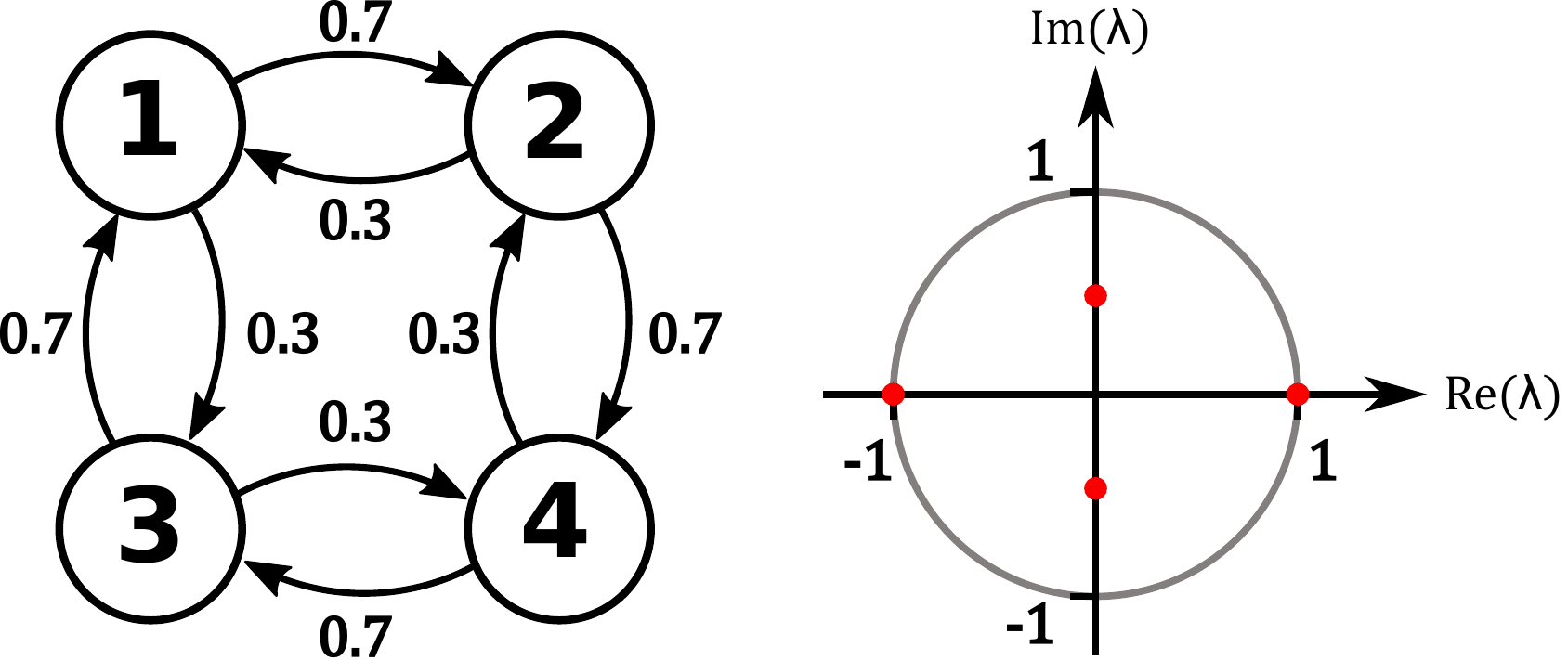}
         \caption{irreducible, $d=2$}
     \end{subfigure}
     \par\bigskip
     \begin{subfigure}[b]{0.4\textwidth}
         \centering
         \includegraphics[width=\textwidth]{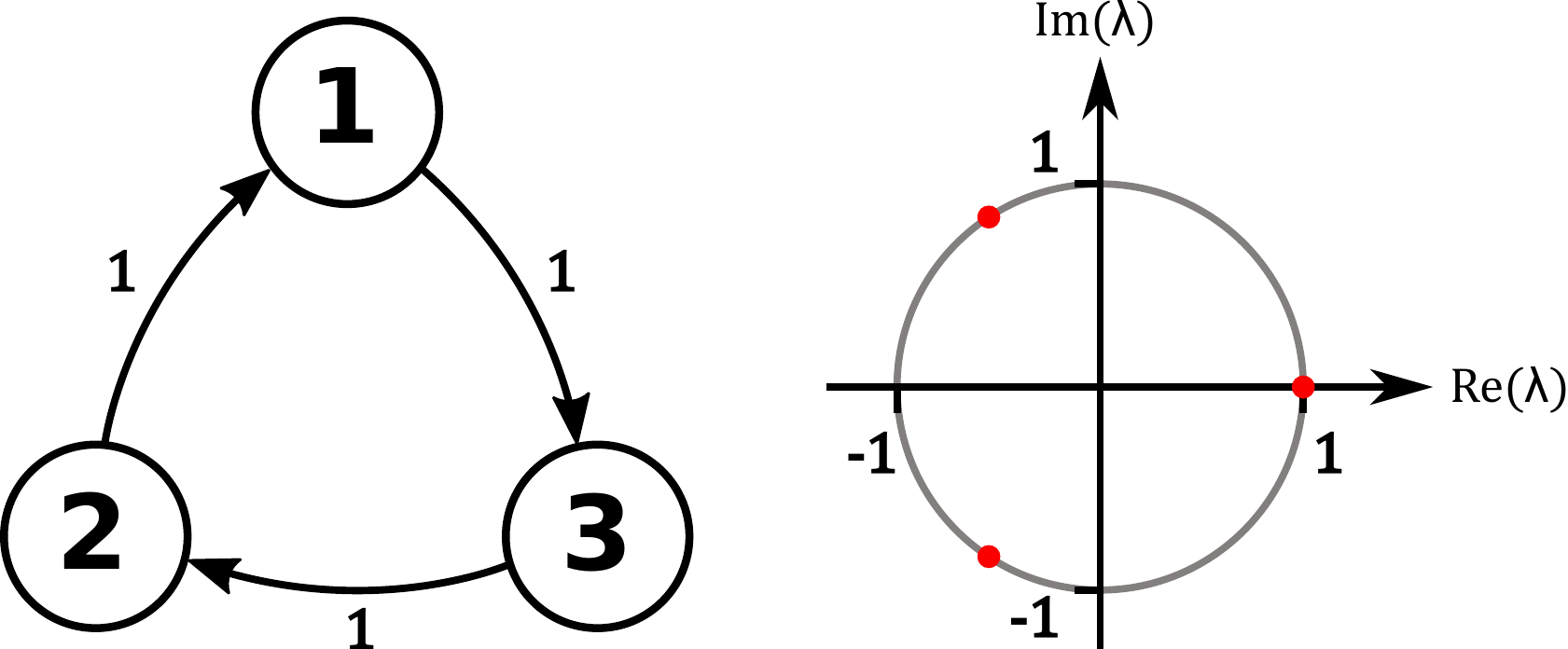}
         \caption{irreducible, $d=3$}
     \end{subfigure}
     \quad
     \begin{subfigure}[b]{0.5\textwidth}
         \centering
         \includegraphics[width=\textwidth]{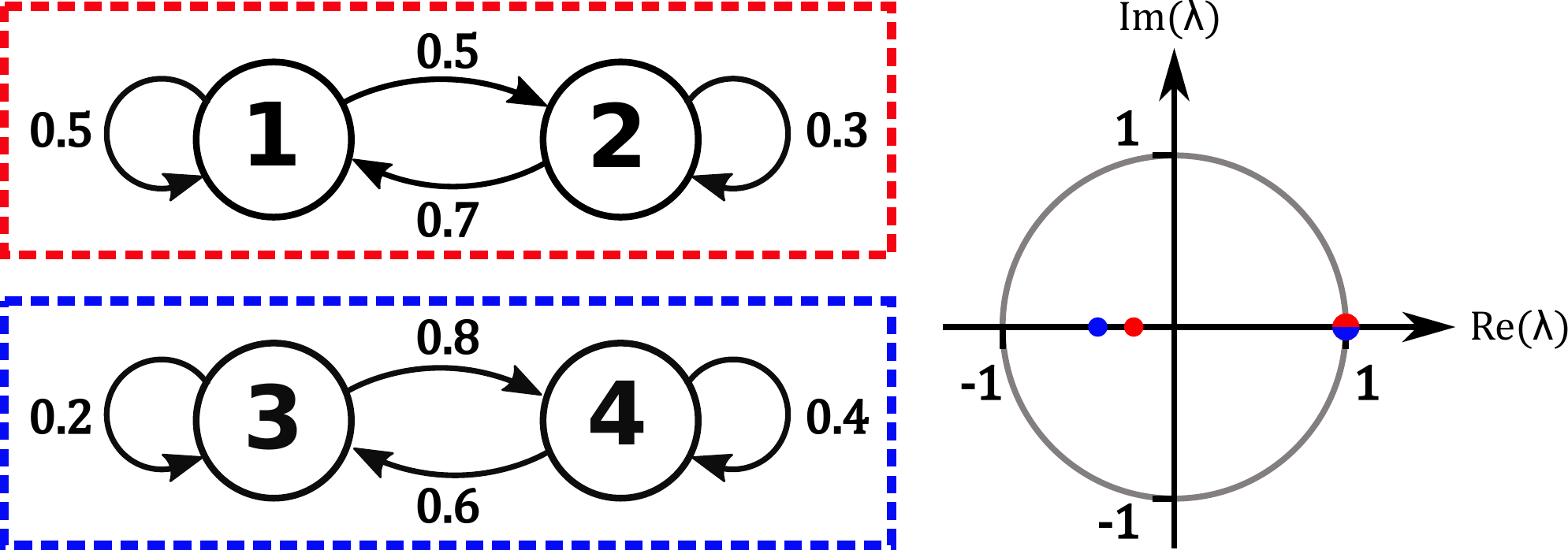}
         \caption{reducible, $d=1$}
     \end{subfigure}
        \caption{Four Markov chains, together with plots showing the eigenvalues of their transition matrices.}
    \label{fig:MC+EigPlot}
\end{figure}

To illustrate the above theorem, in \cref{fig:MC+EigPlot}(a-c) we show the transition graphs and eigenvalue plots of three irreducible Markov chains. The first observation to make is that, in agreement with \cref{thm:PeigsIrred}, $\lambda=1$ is an eigenvalue in each case and occurs only once. Furthermore, as a quick exercise we encourage readers to find the eigenvectors of $\lambda=1$ for each example and normalize them to obtain $\vec{\pi}$ and $\vec{\eta}$. Lastly, the eigenvalue plots show that in each example all eigenvalues indeed lie either on the unit circle ($|\lambda|=1$), or within it ($|\lambda|<1$).

Other than $\lambda=1$, \cref{thm:PeigsIrred} implies that the following two types of eigenvalues are also possible: (i) those that lie on other points of the unit circle (i.e.\ $|\lambda|=1$, persistent), and (ii) those that lie within the unit circle (i.e.\ $|\lambda|<1$, transient). In either case, \cref{eq:LeftRightOrthog} tells us that if $\vec{l}_\omega$ is a corresponding left eigenvector, then it must be orthogonal to the unique right eigenvector with $\lambda=1$, which is $\vec{\eta}$. Therefore:
\begin{equation}
\label{eq:LeigSum}
    \vec{l}_\omega^T\vec{\eta}= \sum_{i=1}^N l_{\omega,i} = 0
\end{equation}
where $l_{\omega,i}$ denotes the $i$-th component of $\vec{l}_\omega$. Consequently, left eigenvectors with $\lambda\neq 1$ sum to zero, meaning that unlike stationary distributions they are not probability vectors.

Using our terminology from \cref{MCs-EV}, case (i) includes transient structures, transient oscillations and transient cycles. Of the irreducible chains in \cref{fig:MC+EigPlot}, only (a) and (b) have eigenvalues of this type, and in both cases they are complex conjugate pairs describing transient cycles. Looking at the transition probabilities in each example, it is clear that these transient cycles flow clockwise around the state space.

Case (ii) on the other hand includes persistent oscillations and persistent cycles. \cref{thm:ConvMC} tells us that these are only possible when a chain is periodic. The following result sheds light on this by relating the eigenvalues with $|\lambda|=1$ to the period of a chain \cite{Gebali2008}:

\begin{prop}
\label{prop:PeriodicEigIrred}
If $\mat{P}$ is the transition matrix of an irreducible Markov chain with period $d$, then there are $d$ distinct eigenvalues with modulus $1$, given by:
\begin{equation}
\lambda_1=1, \lambda_2=\phi, \lambda_3=\phi^2, ..., \lambda_d=\phi^{d-1}
\end{equation}
where $\phi=e^{\frac{2\pi i}{d}}$.
\end{prop}

\noindent In simple terms, \cref{prop:PeriodicEigIrred} says that the eigenvalues of $\mat{P}$ with modulus $1$ are always $d$-th roots of unity. We can verify this by checking the periodic examples in \cref{fig:MC+EigPlot}(b,c). In both case the number of eigenvalues on the unit circle is indeed equal to the period of the chain and they are also equally spaced. Furthermore, \cref{prop:PeriodicEigIrred} offers an alternative perspective on how the periodicity affects the persistent behavior of a Markov chain. For example, the chain in \cref{fig:MC+EigPlot}(a) has only a single vector on the unit circle, corresponding to its unique stationary distribution. It is therefore guaranteed to end up in this distribution since all other eigenvalues have $|\lambda|<1$. This is equivalent to the statement that this chain is ergodic, which a quick check of the transition graph confirms. Conversely, the chain in \cref{fig:MC+EigPlot}(b) has an additional eigenvalue $\lambda=e^{\pi i}=-1$ on the unit circle, by virtue of the fact that it has period $2$. Therefore, its persistent behavior can only be fully described using both the unique stationary distribution $\vec{\pi}$ and the eigenvector $\vec{y}$ associated to $\lambda=-1$. For example, we know from \cref{thm:ConvMC} that such a chain can get trapped in a persistent oscillation (i.e.\ $\vec{\mu}_1\to\vec{\mu}_2\to\vec{\mu}_1\to\vec{\mu}_2\to...$). For any such oscillation, $\vec{\mu}_1$ and $\vec{\mu}_2$ can always be expressed as a linear combination of $\vec{\pi}$ and $\vec{y}$, meaning that this sequence indeed \textit{oscillates} between two points in the space spanned by these eigenvectors. While this example only involves real eigenvalues and therefore only real eigenvectors, the interpretation extends to $d>2$, for which \cref{prop:PeriodicEigIrred} tells us that there must be complex eigenvalues with $|\lambda|=1$. For example, the chain in \cref{fig:MC+EigPlot}(c) has period $d=3$, and it has the following $3$ eigenvalues on the unit circle: $\lambda_1=1$, $\lambda_2=e^{\frac{2\pi i}{3}}$ and $\lambda_3=e^{\frac{4\pi i}{3}}$. Analogous to the $d=2$ case, any persistent cycle of this chain can be expressed using the three corresponding eigenvectors, which in the case of $\lambda_2$ and $\lambda_3$ must have complex entries. Rather interestingly, this means that for chains with period $d>2$, persistent cycles are cycles in a complex space despite being sequences of real valued distributions.

\subsubsection{Reducible chains}

Applying spectral graph theory to the transition matrices of reducible Markov chains produces a weaker set of results. For example, the generalization of \cref{thm:PeigsIrred} to the reducible chains is the following:

\begin{theorem}[Perron Frobenius theorem for reducible Markov chains]
\label{thm:PeigsRed}
If $\mat{P}$ is the transition matrix of a reducible Markov chain, then:
\begin{itemize}
    \item $\lambda=1$ is guaranteed to be an eigenvalue.
    \item The number of linearly independent eigenvectors with $\lambda=1$ is equal to the number r of recurrent communicating classes in the Markov chain.
    \item There are many choices of left and right eigenvectors for $\lambda=1$. However, a convenient choice which mirrors the irreducible case is to choose $\vec{\pi}_k$ and $\vec{\eta}_k$ as a pair of left and right eigenvectors for each of the recurrent communicating classes, with $\vec{\pi}_k$ being the unique stationary distribution associated to each class and $\vec{\eta}_k$ being an indicator vector with entry $1$ for states in this class and zeros elsewhere.
    \item All other eigenvalues have $|\lambda|\leq 1$, where $|\cdot|$ is the complex modulus, meaning that the spectral radius of $\mat{P}$ is $1$.
\end{itemize}
\end{theorem}

To understand \cref{thm:PeigsRed} in more depth, consider the example in \cref{fig:MC+EigPlot}(d). This Markov chain has $2$ recurrent communicating classes, which means that $\lambda=1$ has a multiplicity of $2$. We indicate this on the eigenvalue plot by a larger size circle. Furthermore, we color this circle half red and half blue to reflect the fact that we can choose the eigenvectors for $\lambda=1$ based on the two recurrent communicating classes, i.e.\ $\vec{\pi}_1=(0.58, 0.42, 0, 0)^T$ and $\vec{\eta}_1=(1, 1, 0, 0)^T$ as a pair of left and right eigenvectors for the red class, and $\vec{\pi}_2=(0, 0, 0.43, 0.57)^T$ and $\vec{\eta}_2=(0, 0, 1, 1)^T$ as a pair of left and right eigenvectors for the blue class. Then, $\vec{\pi}_1$ and $\vec{\pi}_2$ together span the left eigenspaces of $\lambda=1$ (including all possible stationary distributions), whereas $\vec{\eta}_1$ and $\vec{\eta}_2$ span the right eigenspaces of $\lambda=1$. It is worth emphasizing that while there are an infinite number of other ways to choose the basis vectors for $\lambda=1$, this is a convenient choice since it is the only one for which all basis vectors have strictly non-negative entries. For example, consider the choice of $\tilde{\vec{\pi}}_1=\frac{1}{3}\vec{\pi}_1+\frac{2}{3}\vec{\pi}_2=(0.19, 0.14, 0.29, 0.38)^T$ and $\tilde{\vec{\eta}}_1=\vec{\eta}=(1, 1, 1, 1)^T$ as a first pair of eigenvectors. If we then choose a second pair $\tilde{\vec{\pi}}_2$ and $\tilde{\vec{\eta}}_2$ that satisfy biorthogonality (\cref{eq:Biorthog}) within this space, they are guaranteed to contain negative entries, for example $\tilde{\vec{\pi}}_2=\vec{\pi}_1-\vec{\pi}_2=(0.58, 0.42, -0.43, -0.57)^T$ and $\tilde{\vec{\eta}}_2=3\vec{\eta}_1-\frac{3}{2}\vec{\eta}_2=(3, 3, -\frac{3}{2}, -\frac{3}{2})^T$. Thus, the choice of eigenvectors stated in \cref{thm:PeigsRed} is in some sense special since it is the only one that preserves our intuition that left eigenvectors with $\lambda=1$ correspond to distributions over the state space $\mathcal{S}$. For this reason, we henceforth assume this convention when referring to eigenvectors with $\lambda=1$.

Unfortunately, there is no general analogue of \cref{eq:LeigSum} for reducible chains. One reason for this is that, like the $\lambda=1$ case, there are many choices of eigenvectors for $\lambda\neq 1$. However, for a recurrent reducible chain the transition matrix can be written in block diagonal form (see proof of \cref{prop:RevRedRec}), which means we can mirror the $\lambda=1$ case by choosing eigenvectors with $\lambda\neq 1$ to have non-zero entries only in a single recurrent class. If the $k$-th recurrent class has $n$ states, then the corresponding block is a $n\times n$ matrix, meaning that there are a total of $n$ pairs of left and right eigenvectors with non-zero entries for states in this class, i.e.\ one pair with $\lambda=1$ ($\vec{\pi}_k$ and $\vec{\eta}_k$), and another $n-1$ pairs with $\lambda\neq 1$. We can therefore apply an equivalent argument to \cref{eq:LeigSum} for the $k$-th class, but instead using the vector $\vec{\eta}_k$. Thus, for each recurrent class, the left eigenvectors with $\lambda\neq 1$ can also be chosen such that they all sum to zero. Conversely, non-recurrent chains cannot be written in block diagonal form, which means that this argument cannot be applied. Therefore, some eigenvectors will not sum to zero for such chains, although to our knowledge this case has not received significant attention so far in the literature. 

Looking at the eigenvalue plot of \cref{fig:MC+EigPlot}(d), we see that there are two eigenvalues with $|\lambda|<1$, both of which are real and negative. Using the terminology from \cref{MCs-EV}, they therefore correspond to transient oscillations of the chain. Furthermore, since the chain is recurrent we can apply the procedure described above and choose one eigenvector to have non-zero entries only in the red class, and the other eigenvector to have non-zero entries only in the blue class. With this choice, we see that each transient oscillation takes place on a distinct communicating class, which we indicate on the plot by coloring the eigenvalues red and blue.

In the case of \cref{prop:PeriodicEigIrred}, the extension to reducible chains is straightforward since one can simply apply this result individually to each recurrent communicating class of a reducible chain. Therefore, for each class of period $d$, there are $d$ eigenvalues of modulus $1$ that satisfy the same properties as in the irreducible case.

Lastly, a couple of similarities between \cref{thm:PeigsRed} and \cref{thm:PeigsIrred} can be pointed out. Firstly, in both theorems $\lambda=1$ is guaranteed to be an eigenvalue. Since every eigenvalue has at least one eigenvector, this means that we can always find a left eigenvector with $\lambda=1$. Provided that we choose an eigenvector with non-negative entries and normalize it to one, it is a stationary distribution of the chain. This therefore justifies our claim from \cref{MCs-Evol} that every finite Markov chain has at least one stationary distribution. Secondly, in both theorems the eigenvalues cannot have absolute value greater than one, which is one of the assumptions we made when studying the evolution of a chain in terms of its eigenvectors and eigenvalues in \cref{MCs-EV}, and which justified our partitioning of \cref{eq:EveigkPart} into persistent and transient terms.

\bigskip

The results of this section emerge by treating Markov chains as graphs. However, in most graphs the outgoing edges from each vertex do not sum up to $1$, meaning that they cannot be interpreted as transition probabilities. Because of this, Markov chains can be more precisely interpreted as a type of normalized graph. This idea is formalized in the next section, where we introduce a well-known method for transforming any graph $G$ into a Markov chain.

\section{Random walks}
\label{RWs}
\subsection{Definition}
Suppose we have a graph $G$ with weight matrix $\mat{W}$ that we want to normalize such that the outgoing weights from each vertex $v_i$ sum up to $1$.\footnote{For this definition to work, we require that all vertices have $d_i^+>0$, meaning that isolated vertices or vertices with only incoming edges are forbidden.} The most obvious way to do this is to divide all entries in the $i$-th row of $\mat{W}$ by $d_i^+$. By scaling each edge weight $W_{ij}$ by the out-degree of the starting vertex $v_i$, we obtain the transition probabilities $P_{ij}=\frac{W_{ij}}{d_i^+}$. In order to write this in matrix notation, we first define a \textit{degree matrix} $\mat{D}$ whose elements are given by:
\begin{equation}
\label{eq:DegMat}
D_{ij}:=\left\{
	\begin{array}{ll}
		d_i^+& \mbox{if } i=j \\
		0 & \mbox{otherwise}
	\end{array}
\right.
\end{equation}
Since $d_i^+>0$, we can invert $\mat{D}$ to produce a diagonal matrix with the entries $\frac{1}{d_i^+}$ on the diagonals. Using this inverse, $\mat{W}$ can be row normalized by multiplying with $\mat{D}^{-1}$ from the left:
\begin{equation}
    \mat{P}:=\mat{D}^{-1}\mat{W} \label{eq:RWmat}
\end{equation}
The Markov chain represented by \cref{eq:RWmat} is called the \textit{random walk} on $G$ \cite{Gobel1974, Lovasz93}. This is a fitting name, since if we imagine an agent walking between vertices in $G$, and randomly choosing where to go at each time step based on the weights of outgoing edges, then \cref{eq:RWmat} would describe the resulting Markov chain. 

Qualitatively, we can say that the transformation in \cref{eq:RWmat} is useful when we have a starting graph $G$ that we would like to describe in probabilistic and/or temporal terms. Conversely, if we have a starting Markov chain and transition matrix $\mat{P}$, knowing some matrix $\mat{W}$ for which \cref{eq:RWmat} holds can offer insight into the type of relationships between states that give rise to the chain. However, the latter of these two perspectives is partially complicated by the fact that the mapping from $\mat{P}$ to $\mat{W}$ is one-to-many, and there are in fact an infinite number of different graphs $G$ which produce the same Markov chain. As an example, in \cref{fig:RW} we show two distinct weight matrices $\mat{W}_1$ and $\mat{W}_2$ that get transformed to the same transition matrix $\mat{P}$ using \cref{eq:RWmat}. How can we describe the infinite set of graphs corresponding to a single Markov chain? In principle, it involves undoing the row normalization of \cref{eq:RWmat}. Thus, for a given Markov chain $\mathcal{X}$ with transition matrix $\mat{P}$, we consider all possible scalings of the rows of $\mat{P}$ by positive constants. Every such scaling can be described by a diagonal matrix $\mat{A}\in\mathbb{R}_{>0}^{N\times N}$ that multiplies $\mat{P}$ from the left to produce a single corresponding weight matrix, i.e.\ $\mat{W}=\mat{A}\mat{P}$. As an illustration, in \cref{fig:RW}(d, e) we show the two scaling matrices that undo the row normalization of the transition matrix in \cref{fig:RW}(c) and transform it back into the weight matrices $\mat{W}_1$ and $\mat{W}_2$, respectively. The following definition generalizes this to the set of all such weight matrices that can be realized in this way:

\begin{figure}[h]
     \centering
     \begin{subfigure}[c]{0.3\textwidth}
     \centering
     $\mat{W}_1=\left(
         \begin{array}{cccc}
         0 & 0 & 0 & 1.5 \\
         2 & 0 & 6 & 0 \\
         1 & 3 & 0 & 0 \\
         2 & 0 & 8 & 0 \\
         \end{array} \right)$
         \caption{}
     \end{subfigure}
     \quad
     \begin{subfigure}[c]{0.3\textwidth}
     \centering
     $\mat{W}_2=\left(
         \begin{array}{cccc}
         0 & 0 & 0 & 6 \\
         20 & 0 & 60 & 0 \\
         3 & 9 & 0 & 0 \\
         10 & 0 & 40 & 0 \\
         \end{array} \right)$
         \caption{}
     \end{subfigure}
     \quad
     \begin{subfigure}[c]{0.3\textwidth}
     \centering
     $\mat{P}=\left(
         \begin{array}{cccc}
         0 & 0 & 0 & 1 \\
         0.25 & 0 & 0.75 & 0 \\
         0.25 & 0.75 & 0 & 0 \\
         0.2 & 0 & 0.8 & 0 \\
         \end{array} \right)$
         \caption{}
     \end{subfigure}
     \par\bigskip
     \begin{subfigure}[c]{0.3\textwidth}
     \centering
     $\mat{A}_1=\left(
         \begin{array}{cccc}
         1.5 & 0 & 0 & 0 \\
         0 & 8 & 0 & 0 \\
         0 & 0 & 4 & 0 \\
         0 & 0 & 0 & 10 \\
         \end{array} \right)$
         \caption{}
     \end{subfigure}
     \quad
     \begin{subfigure}[c]{0.3\textwidth}
     \centering
     $\mat{A}_2=\left(
         \begin{array}{cccc}
         6 & 0 & 0 & 0 \\
         0 & 80 & 0 & 0 \\
         0 & 0 & 12 & 0 \\
         0 & 0 & 0 & 50 \\
         \end{array} \right)$
         \caption{}
     \end{subfigure}
     \caption{(a, b) The weight matrices of two graphs $G_1$ and $G_2$ that have edges connecting the same pairs of states but with different weights, (c) the transition matrix of the Markov chain produced via a random walk on either of these graphs, (d, e) the scaling matrices that transform $\mat{P}$ into $\mat{W}_1$ and $\mat{W}_2$, respectively.}
     \label{fig:RW}
\end{figure}

\begin{defin}
\label{defin:RWset}
For a Markov chain $\mathcal{X}$ defined on a state space with $|\mathcal{S}|=N$ states and with a transition matrix $\mat{P}$, the following defines the set of all graphs from which the chain can be generated via a random walk:
\begin{equation}
    \mathcal{RW}(\mathcal{X})=\{G : \mat{W}=\mat{A}\mat{P},\; \textup{where}\;\mat{A}\in\mathbb{R}_{>0}^{N\times N}\;\textup{is a positive diagonal matrix} \} \label{eq:RWsetscaling}
\end{equation}
which we call the random walk set of $\mathcal{X}$.
\end{defin}

\noindent A few details are worth noting about \cref{defin:RWset}. Firstly, since the trivial scaling $\mat{A}=\mathbbm{1}$ is allowed, $\mat{P}\in \mathcal{RW}(\mathcal{X})$ for any Markov chain. Secondly, the fact that $\mat{A}\in\mathbb{R}_{>0}^{N\times N}$ means that each of these matrices is invertible, with $\mat{A}^{-1}$ also being diagonal and having entries equal to the reciprocals of the diagonals of $\mat{A}$. Consequently, if $\mat{W}_1=\mat{A}_1\mat{P}$ and $\mat{W}_2=\mat{A}_2\mat{P}$ are two weight matrices in the random walk set of a given Markov chain, we can always write $\mat{W}_1=\mat{A}_1\mat{A}_2^{-1}\mat{W}_2=\mat{A}_3\mat{W}_2$, meaning that $\mat{W}_1$ and $\mat{W}_2$ are also related simply by a row scaling with positive constants, with $\mat{A}_3=\mat{A}_1\mat{A}_2^{-1}$ describing this scaling. Therefore, the row scaling defined in \cref{eq:RWsetscaling} effectively partitions the set of all non-negative matrices into equivalence classes. Lastly, since \cref{defin:RWset} allows any non-zero scaling of the rows of $\mat{P}$, the random walk set of any Markov chain predominantly consists of graphs which are neither undirected nor balanced. In fact, only certain types of Markov chains have random walk sets that \textit{contain} undirected or balanced graphs, as explained by the following two results:

\begin{theorem}
\label{thm:RWequivClassRec}
A Markov chain $\mathcal{X}$ is recurrent if and only if $\mathcal{RW}(\mathcal{X})$ contains balanced graphs (proof: see Appendix A).
\end{theorem}

\begin{theorem}
\label{thm:RWequivClassRev}
A recurrent Markov chain $\mathcal{X}$ is reversible if and only if the balanced graphs in $\mathcal{RW}(\mathcal{X})$ are undirected (proof: see Appendix A).
\end{theorem}

\begin{sidewaysfigure}
     \centering
     \begin{subfigure}[c]{0.48\textheight}
         \centering
         \includegraphics[width=\textwidth]{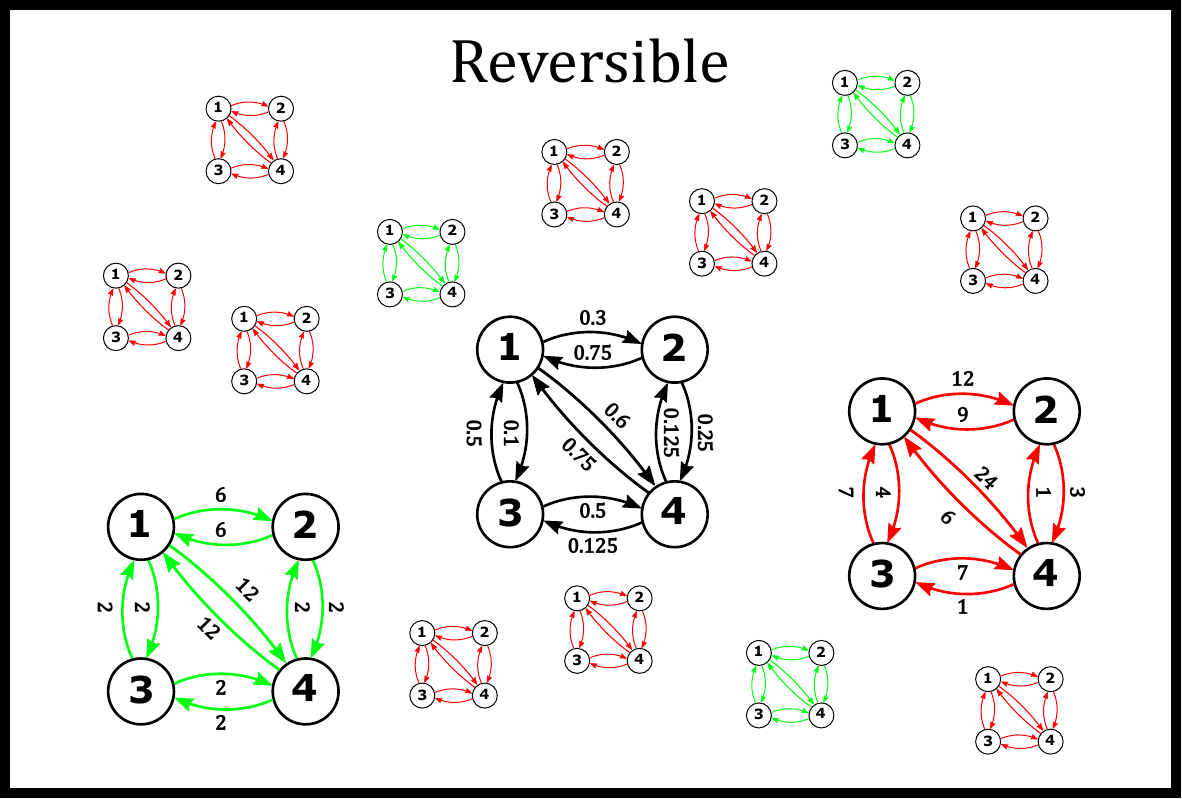}
         \caption{}
     \end{subfigure}
     \hfill
     \begin{subfigure}[c]{0.48\textheight}
         \centering
         \includegraphics[width=\textwidth]{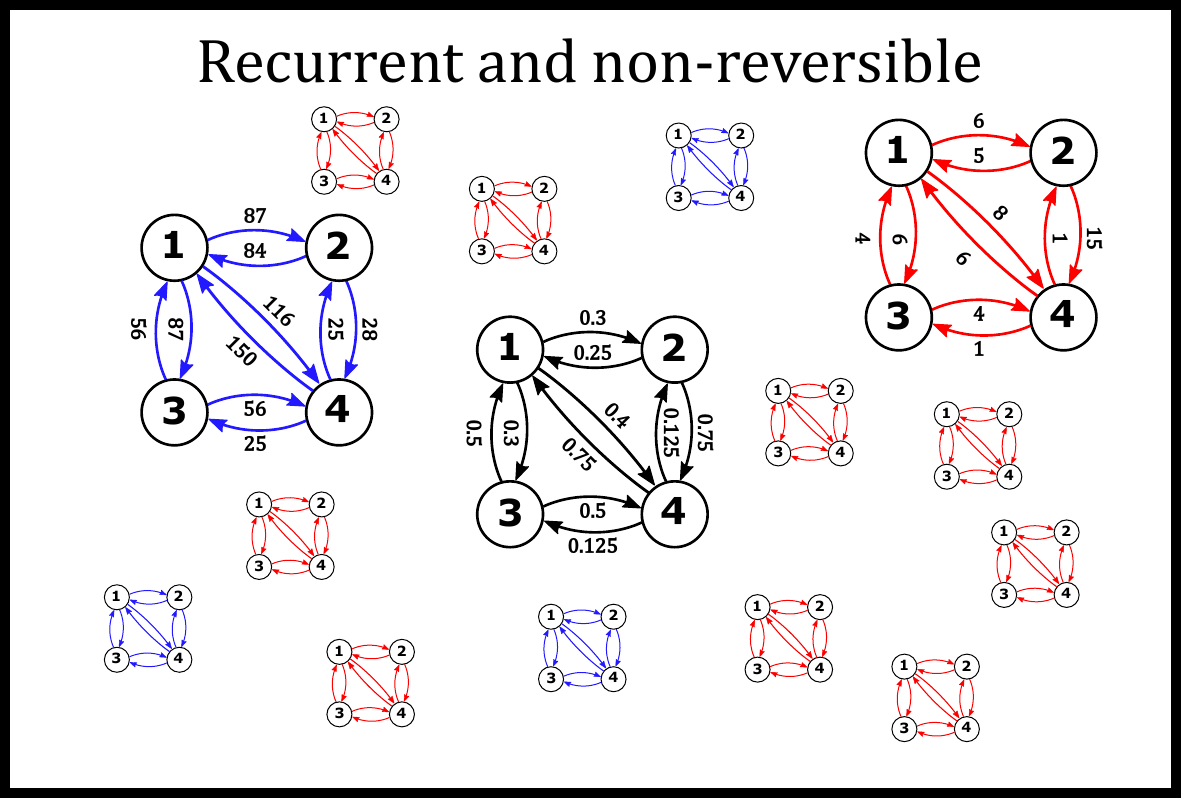}
         \caption{}
     \end{subfigure}
     \par\bigskip
     \begin{subfigure}[c]{0.48\textheight}
         \centering
         \includegraphics[width=\textwidth]{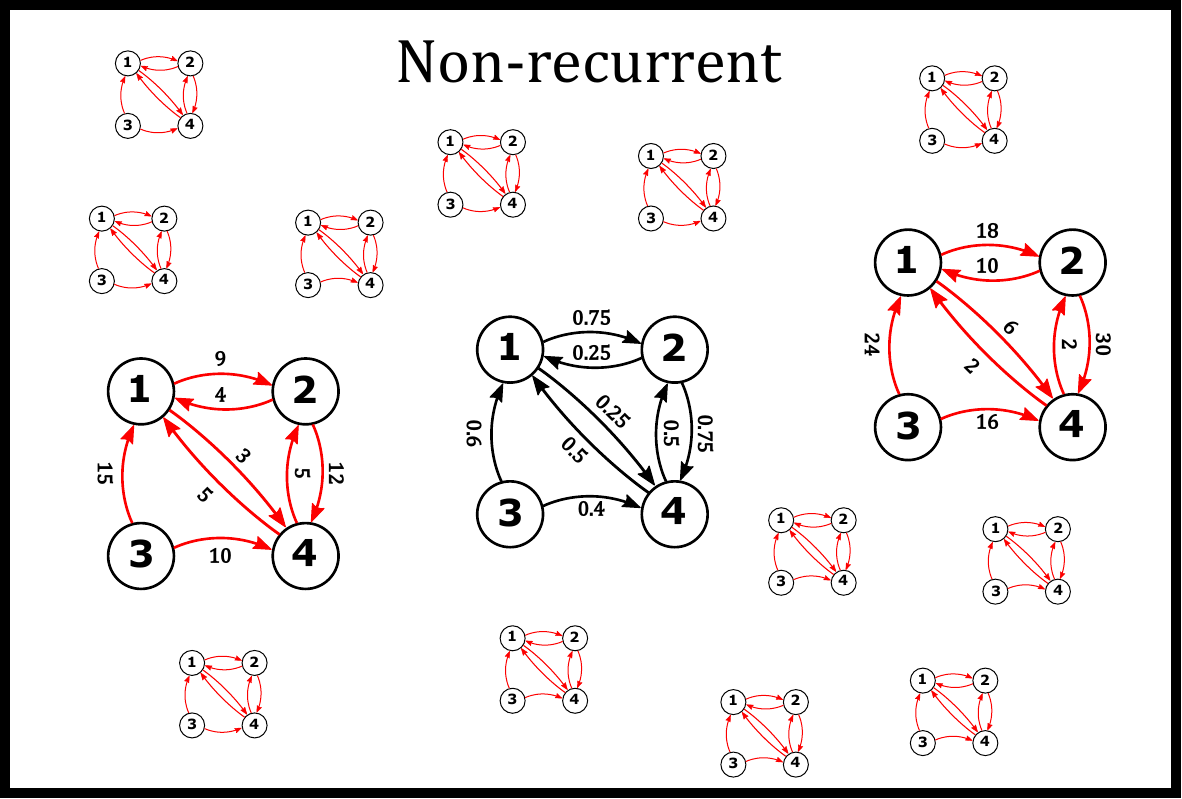}
         \caption{}
    \end{subfigure}
    \hfill
    \begin{subfigure}[c]{0.24\textheight}
         \centering
         \includegraphics[width=\textwidth]{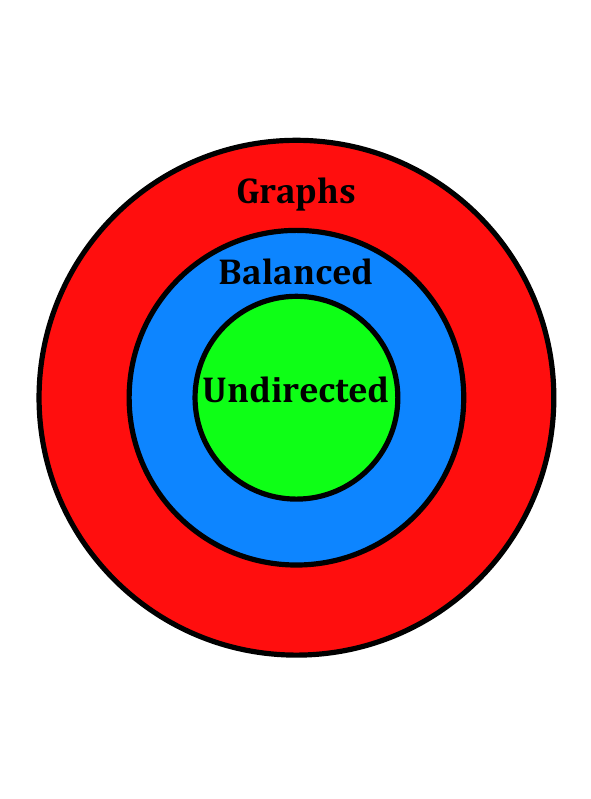}
         \caption{}
    \end{subfigure}
    \begin{subfigure}[c]{0.24\textheight}
         \centering
         \includegraphics[width=\textwidth]{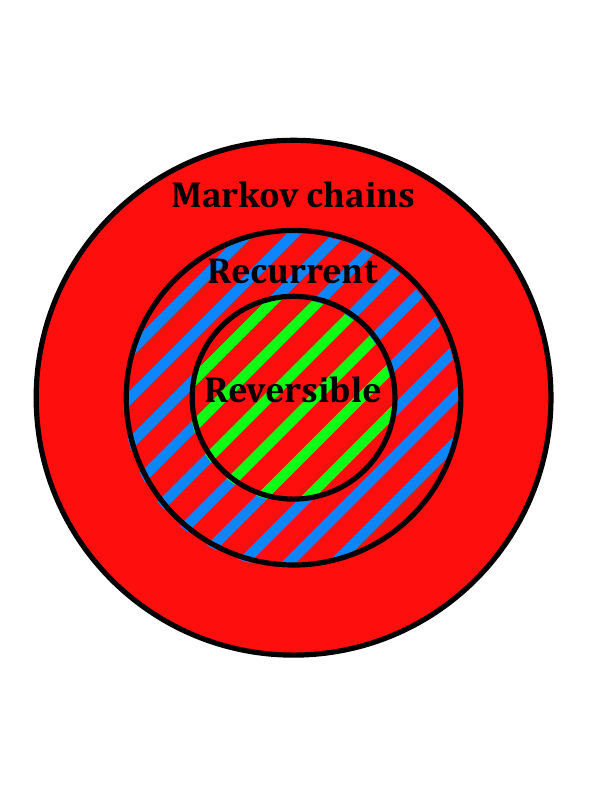}
         \caption{}
    \end{subfigure}
    \caption{(a-c): The random walk sets of the three Markov chains in \cref{fig:GBDB}(a, d, j), including undirected (green), balanced directed (blue), and unbalanced (red) graphs, as well as the chains themselves (black) which are reversible (a), recurrent and non-reversible (b), and non-recurrent (c). Venn diagrams in (d, e) illustrate the relationships between the three types of graph, and the three types of Markov chain, respectively. In (d) the colors indicate the type of graph, whereas in (e) they indicate the types of graphs that exist in the random walk sets of each type of Markov chain. For both Venn diagrams, the color scheme is in accordance with (a-c).}
    \label{fig:RWequivClass}
\end{sidewaysfigure}

As an illustration, in \cref{fig:RWequivClass}(a-c) we show the random walk sets for the three Markov chains that were studied in \cref{fig:GBDB}(a, d, j) of \cref{MCs-Rev}.\footnote{While the examples here each have a single communicating class, the analysis would apply equally to chains with multiple classes.} In each case, the Markov chain is located in the center and is colored in black. Other graphs in the random walk sets are colored based on the graph type (undirected=green, balanced directed=blue, unbalanced=red), and are depicted as miniature graphs without edge weights (except for one representative example of each type). The Markov chains of \cref{fig:RWequivClass}(a, b) are both recurrent, and we indeed see that their random walk sets contain balanced graphs (\cref{thm:RWequivClassRev}). In both figures, notice that more unbalanced graphs are drawn to reflect the fact that they are more numerous than the balanced cases. Furthermore, the chain in \cref{fig:RWequivClass}(a) is reversible, whereas the chain in \cref{fig:RWequivClass}(b) is non-reversible, meaning that in the former case all balanced graphs in $\mathcal{RW}(\mathcal{X})$ are undirected, and in the latter case they are directed (\cref{thm:RWequivClassRev}). The Markov chain in \cref{fig:RWequivClass}(c) is non-recurrent, and in agreement with \cref{thm:RWequivClassRec} it contains only unbalanced graphs. Lastly, note that we do not include a corresponding diagram for the semi-reversible example in \cref{fig:GBDB}(g). However, since such chains can be made reversible by removing non-recurrent states, a simple extension of \cref{thm:RWequivClassRev} ensures that for such chains there exist graphs in $\mathcal{RW}(\mathcal{X})$ for which the edges between recurrent states are undirected.

To summarize these observations, in \cref{fig:RWequivClass}(d,e) we show two Venn diagrams that illustrate the relationships between the different types of Markov chains and graphs considered. In \cref{fig:RWequivClass}(d), graphs are shown as an outer circle, with balanced graphs as a particular case, and undirected graphs as a special type of balanced graph, i.e.\ graphs $\supset$ balanced graphs $\supset$ undirected graphs (colored red, blue and green, respectively). In \cref{fig:RWequivClass}(e), Markov chains are organized in a similar way, i.e.\ Markov chains $\supset$ recurrent chains $\supset$ reversible chains. Moreover, the colors in the Markov chain diagram are based on the types of graphs allowed in $\mathcal{RW}(\mathcal{X})$ for each type of chain. For example, reversible chains are shaded in red and green since they correspond to random walks on either undirected or unbalanced graphs.

The balanced graphs belonging to a recurrent Markov chain's random walk set are in some sense special, since the vertex degrees have a simple relationship to the stationary probabilities:

\begin{prop}
\label{thm:RWequivClassRecSD}
Let $\mathcal{X}$ be a recurrent Markov chain and $G$ one of the balanced graphs in $\mathcal{RW}(\mathcal{X})$. Then, the degrees of this graph are related to one of the stationary distributions $\vec{\pi}>0$ of $\mathcal{X}$, via:
\begin{equation}
\label{eq:DegSD}
    \frac{d_i}{z}=\pi_i
\end{equation}
where $z=\sum_i d_i$ is the volume of $G$ (proof: see Appendix A).
\end{prop}

For example, evaluating $\frac{d_i}{z}$ for $v_1$ in the balanced graphs in \cref{fig:RWequivClass}(a) and \cref{fig:RWequivClass}(b) yields $\frac{20}{48}\approx 0.417$ and $\frac{290}{714}\approx 0.406$, respectively, which are indeed the stationary probabilities of state $s_1$ for each of the corresponding Markov chains (see \cref{fig:GBDB}(b, e)). This highlights something useful about balanced graphs, which is that the weight matrix $\mat{W}$ allows direct calculation of one of the stationary distributions without having to simulate the random walk. Conversely, for an unbalanced graph there is no universally valid expression relating stationary probabilities of the random walk to the vertex degrees.
\bigskip

Perhaps the most important conclusion to draw from this section is that one can always describe a reversible Markov chain as a random walk on \textit{some} undirected graph. Since undirected graphs have symmetric weight matrices, and since matrices of this type have received a large amount of study in mathematics, this interpretation provides a number of tools for describing reversible chains in more detail. This is the focus of the next section. Directed graphs, on the other hand, are as of yet far less understood, meaning that the same level of description for non-reversible chains is not possible. However, in \cref{RWs-dir} we explore some cases where concepts can be extended to the directed/non-reversible case. Since balanced directed graphs are less common objects in graph theory, we do not dedicate a section to them and instead consider them briefly as a special case in \cref{RWs-dir}.

\subsection{Random walks on undirected graphs}
\label{RWs-undir}
\subsubsection{Relationship to symmetric matrices}
In this section, we explore in more depth the connections between real symmetric matrices and the transition matrices of reversible chains. We start by providing the following two results for real symmetric matrices:

\begin{theorem}
\label{thm:RealSym1}
A real matrix $\mat{A}\in\mathbb{R}^{n\times n}$ is symmetric if and only if:
\begin{equation}
\langle \mat{A}\vec{x},\vec{x}'\rangle=\langle \vec{x}, \mat{A}\vec{x}'\rangle\quad \forall\vec{x},\vec{x}'\in\mathbb{R}^n
\end{equation}
where $\langle\cdot, \cdot\rangle$ denotes the standard Euclidean inner product.
\end{theorem}

\begin{theorem}
\label{thm:RealSym2}
A real matrix $\mat{A}\in\mathbb{R}^{n\times n}$ is symmetric if and only if it is orthogonally diagonalizable, meaning that there exists a orthogonal matrix $\mat{Y}$ for which:
\begin{align}
\mat{\Delta}&=\mat{Y}^{-1}\mat{A}\mat{Y} \label{eq:OrthogDiag1}\\
&=\mat{Y}^T\mat{A}\mat{Y} \label{eq:OrthogDiag2}
\end{align}
where $\mat{\Delta}$ is a diagonal matrix.
\end{theorem}

A couple of details can be pointed out about \cref{thm:RealSym2}. Firstly, by comparing \cref{eq:OrthogDiag1} to our analysis of \cref{MCs-EV}, we see that the columns of $\mat{Y}$ are a basis of right eigenvectors of $\mat{A}$ and the rows of $\mat{Y}^T$ are the corresponding dual basis of left eigenvectors. Secondly, both of these bases are orthonormal since $\mat{Y}$ is orthogonal. Thirdly, the matrix $\mat{\Delta}$ contains the eigenvalues of $\mat{A}$, which are guaranteed to be real since all other matrices in \cref{eq:OrthogDiag1,eq:OrthogDiag2} are real. Lastly, it is worth emphasizing the existential condition of \cref{thm:RealSym2}, since not all choices of eigenvectors of a symmetric matrix obey this result. On one hand, \cref{eq:OrthogDiag1} requires that the sets of left and right eigenvectors are chosen together to be a biorthogonal system. On the other hand, even if we assume this property, when a symmetric matrix has repeated eigenvalues the corresponding eigenvectors can be non-orthogonal. However, even in this case it is always possible to apply the Gramm-Schmidt procedure to make them orthonormal, and provided that we find such a basis we can relate the left and right eigenvectors in the following way:

\begin{prop}
\label{prop:RealSym3}
Let $\mat{A}$ be a real symmetric matrix. Then, if $\mat{Y}$ is an orthonormal basis of right eigenvectors of $\mat{A}$ and $\mat{Y}^T$ is the corresponding dual basis, for each eigenvalue $\lambda_\omega$ the left and right eigenvectors are equal, i.e.\ $\vec{l}_\omega=\vec{r}_\omega$.
\end{prop}

Clearly, for any Markov chain, reversible or otherwise, the transition matrix is itself rarely symmetric. Therefore, the results above do not directly apply to $\mat{P}$. However, reversible Markov chains can be related in a number of ways to symmetric matrices, and by virtue of these relations they satisfy variants of the results above \cite{Bremaud1999}. For example, from \cref{MCs-Rev} we know that any flow matrix of a reversible Markov is symmetric. This fact allows us to establish the following analogue of \cref{thm:RealSym1}:

\begin{theorem}
\label{thm:Rev-selfadj}
Let $\mathcal{X}$ be a Markov chain with transition matrix $\mat{P}$. Then $\mathcal{X}$ is reversible if and only if there exists a stationary distribution $\vec{\pi}>0$, such that:
\begin{equation}
\label{eq:Pselfadj}
     \langle \vec{x}, \mat{P}\vec{x}'\rangle_{\mat{\Pi}}=\langle \mat{P}\vec{x}, \vec{x}'\rangle_{\mat{\Pi}} \quad \forall \vec{x},\vec{x}'\in\mathbb{R}^n
\end{equation}
where
\begin{equation}
\label{eq: SDip}
\langle \vec{a},\vec{b}\rangle_{\mat{\Pi}}:=\vec{a}^T\mat{\Pi} \vec{b}=\sum_i \pi_i a_i b_i
\end{equation}
defines an inner product normalized by the stationary probabilities $\pi_i$ (proof: see Appendix A).
\end{theorem}

Furthermore, \cref{thm:RWequivClassRev} tells us that there exist certain row scalings which transform the transition matrix $\mat{P}$ of a reversible Markov chain into symmetric matrices. In fact, this is not the only scaling operation that transforms $\mat{P}$ into a matrix of this type. The next result shows that such a matrix is also formed when we scale both the rows \textit{and} columns of $\mat{P}$ as follows:
\begin{theorem}
\label{thm:KmatOrthog}
Let $\mathcal{X}$ be a Markov chain with transition matrix $\mat{P}$. Then $\mathcal{X}$ is reversible if and only if for any stationary distribution $\vec{\pi}>0$:
\begin{equation}
\label{eq:Kmat}
    \mat{K}:=\mat{\Pi}^{\frac{1}{2}}\mat{P\Pi}^{-\frac{1}{2}}
\end{equation}
is a symmetric matrix. Furthermore, regardless of which stationary distribution is used, the matrix $\mat{K}$ is unique (proof: see Appendix A).
\end{theorem}

The first thing to note about \cref{thm:KmatOrthog} is that the output of the scaling operation is somewhat different to that in \cref{thm:RWequivClassRev}, since in the former case there is only a single symmetric matrix, whereas in the latter there are an infinite number. Additionally, \cref{eq:Kmat} implies that $\mat{P}$ and $\mat{K}$ are similar matrices (\cref{MCs-EV}). Since similar matrices have the same eigenvalues and related sets of eigenvectors, we can use this to establish the following generalizations of \cref{thm:RealSym2} and \cref{prop:RealSym3}:

\begin{theorem}
\label{thm:Rev-Eig}
Let $\mathcal{X}$ be a reversible Markov chain with transition matrix $\mat{P}$. Then $\mathcal{X}$ is reversible if and only if it is diagonalizable with real eigenvalues, and there exists a basis of right eigenvectors that are orthogonal w.r.t.\ $\langle \cdot , \cdot \rangle_{\mat{\Pi}}$ and a corresponding dual basis of left eigenvectors that are orthogonal w.r.t.\ $\langle \cdot , \cdot \rangle_{\mat{\Pi}^{-1}}$, where $\vec{\pi}>0$ is a stationary distribution of the chain (proof: see Appendix A).
\end{theorem}

\begin{prop}
\label{prop:Rev-LR}
Let $\mathcal{X}$ be a reversible Markov chain with transition matrix $\mat{P}$, and $\vec{\pi}>0$ one of its stationary distributions. Furthermore, let $\mat{Y}_R$ be a set of right eigenvectors of $\mat{P}$ and $\mat{Y}_L$ its dual basis, both of which obeying the orthogonality relations of \cref{thm:Rev-Eig}. Then, if $\vec{r}_\omega$ and $\vec{l}_\omega$ are right and left eigenvectors of the same eigenvalue $\lambda_\omega$, they are related via $\vec{l}_\omega=\mat{\Pi}\vec{r}_\omega$ (proof: see Appendix A).
\end{prop}

\cref{thm:Rev-Eig} is particularly important for practical reasons. Firstly, the diagonalizability of $\mat{P}$ implies a full set of linearly independent eigenvectors. Linearly independent feature spaces are often desirable from a computational perspective since they (i) reduce the overall redundancy, (ii) can express any function in $\mathbb{R}^N$, and (iii) ensure that certain matrix operations are well-defined. Furthermore, as already explained in \cref{MCs}, having a diagonalizable transition matrix means that evolving a Markov chain becomes computationally cheaper. Secondly, since $\mat{P}$ is a real matrix with real eigenvalues, we can always choose an eigenbasis consisting only of real valued vectors. This property is useful because real spaces are often more intuitive to deal with than complex spaces. Furthermore, in many applications of Markov chains the underlying vector space is required to be real, either due to the semantic nature of the problem, or because the algorithm being used is not suited to complex spaces.

While a linearly independent set of eigenvectors is useful, having a basis that is pairwise orthogonal w.r.t.\ the standard Euclidean product offers further analytical and numerical benefits. From \cref{thm:Rev-Eig}, it is clear that this is not the case for transition matrices of reversible Markov chains. However, the matrix $\mat{K}$ is a normalized symmetric version of $\mat{P}$, and from the proof of \cref{thm:Rev-Eig} we know that the eigenvalues of these two matrices are the same and their eigenvectors are related simply by a multiplication of $\mat{\Pi}^{\pm\frac{1}{2}}$. Thus, they contain similar information regarding the relationship between pairs of states in $\mathcal{S}$, and so $\mat{K}$ can be used as a surrogate for $\mat{P}$ in situations where orthogonal eigenvectors are required. 

When using the matrix $\mat{K}$, it is sometimes useful to express it purely in terms related to a graph $G$. Starting with \cref{eq:Kmat}, this can be done by swapping stationary probabilities for degree vertices:
\begin{align}
    \mat{K}\overset{(\ref{eq:Kmat},\ref{eq:DegSD})}&{=}\mat{D}^{\frac{1}{2}}z^{-\frac{1}{2}}\mat{P}z^{\frac{1}{2}}\mat{D}^{-\frac{1}{2}}\\
    &=\hspace{8.5pt}\mat{D}^{\frac{1}{2}}\mat{PD}^{-\frac{1}{2}}\label{eq:Kmat-P}\\
    \overset{(\ref{eq:RWmat})}&{=}\hspace{5pt}\mat{D}^{-\frac{1}{2}}\mat{WD}^{-\frac{1}{2}}\label{eq:Kmat-W}
\end{align}
This expression appears in the next section, where we use $\mat{K}$ to define a positive semi-definite matrix which has the same eigenvectors.

Collectively, the results of this section demonstrate that transition matrices of reversible chains satisfy similar properties to symmetric matrices, but subject to a different type of normalization. This alternative normalization is important for the next section, and in order to make our analyses more concise we end this section by defining the following two coordinate transformations:

\begin{defin}[Left and right coordinate transformations]
\label{defin:LR-coord-trans}
Let $G$ be an undirected graph with $N$ vertices and degree matrix $\mat{D}$. If $\vec{x}\in\mathbb{R}^N$ is a vector defined over these vertices, we define the following related vectors:
\begin{align}
    \vec{x}_L&=\mat{D}^{\frac{1}{2}}\vec{x}\\
    \vec{x}_R&=\mat{D}^{-\frac{1}{2}}\vec{x}
\end{align}
which we call the left and right coordinate transformations of $\vec{x}$, respectively.
\end{defin}

\subsubsection{Normalized Graph Laplacian}
\label{RWs-Lap}
A symmetric matrix $\mat{A}\in\mathbb{R}^{N\times N}$ is positive semi-definite if $\vec{x}^T\mat{A}\vec{x}\geq 0\;\forall\vec{x}\in\mathbb{R}^N$, or equivalently if all its eigenvalues are non-negative. Such matrices have a number of numerical properties that make them useful for solving optimization problems, and because of this they often appear in computational applications. The matrix $\mat{K}$ is not positive-semidefinite, since it has the same eigenvalues as $\mat{P}$ (which can be negative). However, it is straightforward to define a variant of $\mat{K}$ which does have this property, while also having the same eigenvectors as $\mat{K}$:

\begin{defin}
\label{defin:Lnorm}
For an undirected graph $G$ with weight matrix $\mat{W}$, the normalized graph Laplacian is given by:
\begin{align}
    \mat{\mathcal{L}}&\hspace{0.5pt}:=\hspace{3.5pt}\mat{\mathbbm{1}}-\mat{K}\label{eq:NLapMat}\\
    &\overset{(\ref{eq:Kmat-P})}{=}\mat{\mathbbm{1}}-\mat{D}^{\frac{1}{2}}\mat{PD}^{-\frac{1}{2}} \label{eq:NLapMatP}\\
    &\overset{(\ref{eq:Kmat-W})}{=}\mat{\mathbbm{1}}-\mat{D}^{-\frac{1}{2}}\mat{WD}^{-\frac{1}{2}} \label{eq:NLapMatW}
\end{align}
with elements equal to:
\begin{equation}
    \mathcal{L}_{ij}=\left\{
    \begin{array}{ll}
		1 - \frac{W_{ii}}{d_i}& \mbox{if } i=j \\
		\\
		-\frac{W_{ij}}{\sqrt{d_id_j}} & \mbox{if i $\neq$ j and $(v_i,v_j)\in E$}\\
		\\
		0 & \mbox{otherwise}
	\end{array}
	\right.
\end{equation}
and for which the following properties hold:
\begin{itemize}
    \item $\mat{\mathcal{L}}$ is symmetric and positive semi-definite, with eigenvalues in the interval $[0,2]$.
    \item $\lambda=0$ is guaranteed to be an eigenvalue and its multiplicity is equal to the number of connected components in $G$.
    \item If $C$ is a connected component of $G$, then there is an eigenvector $\vec{y}_C$ with eigenvalue $\lambda=0$ whose entries are $d_i^{\frac{1}{2}}$ for $v_i\in C$ and $0$ otherwise, or equivalently $\vec{y}_C=\mat{D}^{\frac{1}{2}}\vec{1}_{C}=\vec{1}_{C,L}$ where $\vec{1}_C$ is an indicator vector for vertices in $C$.
    \item Therefore, for fully connected graphs $\vec{y}=\mat{D}^{\frac{1}{2}}\vec{1}=\vec{1}_{L}$ is the unique eigenvector with $\lambda=0$, where $\vec{1}$ is a vector of ones.
\end{itemize}
\end{defin}

There exist other types of graph Laplacians for undirected graphs, such as the unnormalized Laplacian $\mat{L}:=\mat{D}-\mat{W}$, and the random walk Laplacian $\mat{L}_{\text{RW}}:=\mathbbm{1}-\mat{D}^{-1}\mat{W}$. Many useful properties of a graph $G$ can be obtained from graph Laplacians, and they form the basis of spectral graph theory \cite{Chung1997}. Due to its close relationship to $\mat{P}$, in this tutorial we predominantly focus $\mat{\mathcal{L}}$. In the case of $\mat{L}$, there is a weaker connection to $\mat{P}$ which we only discuss briefly. Furthermore, to avoid redundancy we do not discuss $\mat{L}_{\text{RW}}$ at all, since it is the same as $-\mat{P}$ but shifted by the identity. For a concise review of each of these objects, see \cite{Wiskott2019}.

As the name suggests, graph Laplacians are analogues of the Laplace operator. In particular, a number of studies have found links between graph Laplacians and a variant of the Laplace operator on manifolds, known as the Laplace-Beltrami operator \cite{Belkin2008, Hein2007}. Broadly speaking, Laplace operators measure how much the value of a function at a point varies from its local average, which is informally related to the notion of mean curvature \cite{Reilly1982}. Furthermore, the eigenvalues of this operator are non-negative real numbers and are related to how much the corresponding eigenfunctions vary over the domain of the function \cite{Grebenkov2013}. All of these properties have analogues in the case of $\mat{\mathcal{L}}$, which justifies the term given to this object. To see this, we start with the following result \cite{vonLuxburg2007}:

\begin{theorem}
\label{thm:NLapQuad}
Let $G$ be an undirected graph $G$ with weight matrix $\mat{W}$ and normalized Laplacian $\mat{\mathcal{L}}$. Then for any vector $\vec{x}$ defining values on the vertices of $G$:
\begin{align}
\vec{x}^T\mat{\mathcal{L}}\vec{x}&=\frac{1}{2}\sum_{i, j=1}^NW_{ij}\bigg(\frac{x_i}{d_i^{\frac{1}{2}}}-\frac{x_j}{d_j^{\frac{1}{2}}}\bigg)^2\label{eq:NLapQuad}
\end{align}
which is guaranteed to produce a real non-negative number since $\mat{\mathcal{L}}$ is positive semi-definite (proof: see Appendix A).
\end{theorem}

\noindent In \cref{eq:NLapQuad}, the quadratic term measures the difference between the entries of a vector that is related to $\vec{x}$ by normalization with $\mat{D}^{-\frac{1}{2}}$, i.e.\ the vector $\vec{x}_R$. Since this is minimized when the entries of this vector are similar for pairs of vertices with large edge weight $W_{ij}$, it can be interpreted to describe the smoothness of $\vec{x}_R$ with respect to the connectivity of $G$. Furthermore, since this measure describes the smoothness of $\vec{x}_R$ as opposed to that of $\vec{x}$, it is useful for dealing with graphs that have a very heterogeneous degree structure. 

If we evaluate the left-hand side of \cref{thm:NLapQuad} using a normalized eigenvector $\vec{y}_\omega$ of $\mat{\mathcal{L}}$, we get:
\begin{align}
\label{eq:NLapQuadEig}
    \vec{y}_\omega^T\mat{\mathcal{L}}\vec{y}_\omega&=\lambda_\omega\langle \vec{y}_\omega, \vec{y}_\omega\rangle\\
    &=\lambda_\omega
\end{align}
meaning that $\lambda_\omega$ describes precisely this notion of smoothness for $\vec{y}_\omega$. Now assume that we have an orthonormal set of eigenvectors of $\mat{\mathcal{L}}$. From \cref{defin:Lnorm}, we know that the corresponding eigenvalues are real numbers in the interval $[0,2]$, and that there is guaranteed to be at least one eigenvalue $\lambda=0$. Thus, they can be ordered as follows:
\begin{equation}
\label{eq:NLapEigOrder}
    0=\lambda_1\leq\lambda_2 \leq \cdots \leq \lambda_N\leq 2
\end{equation} 
Putting these findings together, we can say that given an orthonormal set of eigenvectors of $\mat{\mathcal{L}}$, there is a natural way to order them based on their smoothness across $G$, according to \cref{eq:NLapQuad}. For example, taking an eigenvector with $\lambda=0$ (\cref{defin:Lnorm}) and plugging it into \cref{eq:NLapQuad} indeed yields zero. These are therefore the eigenvectors of $\mat{\mathcal{L}}$ that have the lowest smoothness score in \cref{eq:NLapQuad}. Similarly, the eigenvectors of $\mat{\mathcal{L}}$ with second-smallest eigenvalue are those that, subject to orthonormality, score the second lowest in \cref{eq:NLapQuad}.

As indicated by \cref{eq:NLapEigOrder}, we can apply this ordering to all eigenvalues of $\mat{\mathcal{L}}$, meaning that the resulting set of eigenvectors gets progressively less smooth, or equivalently more oscillatory, as $\lambda_\omega$ increases. Since this basis is orthonormal, it can express any signal $\vec{x}\in\mathbb{R}^N$ on $G$, with the resulting components describing the projection of $\vec{x}$ onto each of these oscillatory modes. Due to its correspondence with Fourier analysis, this decomposition has been referred to as the \textit{Graph Fourier Transform} of $\vec{x}$ \cite{Shuman2013}. Moreover, a useful result in spectral graph theory is that a basis of eigenvectors corresponding to the $k$ leading eigenvalues of $\mat{\mathcal{L}}$ is one which, subject to orthonormality, has the lowest possible total score in \cref{eq:NLapQuad}. In other words, if $\mat{Y}\in\mathbb{R}^{N\times k}$ represents such a basis of right eigenvectors, it solves the following optimization problem \cite{Belkin2001,Belkin2003,Wiskott2019}:
\begin{align}
    \text{minimize}\quad & \text{tr}(\mat{Y}^T\mat{\mathcal{L}}\mat{Y})\label{eq:NLapOpt1}\\
    \text{subject to}\quad &\mat{Y^T}\mat{Y}=\mathbbm{1}\label{eq:NLapOpt2}
\end{align}
This property underlies several Laplacian-based methods in unsupervised learning \cite{Ghojogh2021,Sprekeler2011}, where typically the eigenvectors of $\mat{\mathcal{L}}$ with the smallest eigenvalues are those that capture the most important information relating to the learning objective, and so discarding eigenvectors with higher eigenvalues can provide a description of a data set that has lower dimensionality, while also being more interpretable.

These observations can be related to the random walk on $G$. Since $\mat{\mathcal{L}}$ shares an eigenbasis with $\mat{K}$, and the eigenbasis of $\mat{K}$ is related to that of $\mat{P}$ by \cref{thm:KmatOrthog}, we can relate the eigenvectors of $\mat{\mathcal{L}}$ to those of $\mat{P}$ as follows:

\begin{prop}
\label{prop:Lnorm}
If $\vec{y}_\omega$ is an eigenvector of $\mat{\mathcal{L}}$ with eigenvalue $\lambda_\omega$, then $\vec{l}_\omega=\mat{D}^{\frac{1}{2}}\vec{y}_\omega=\vec{y}_{\omega,L}$ and $\vec{r}_\omega=\mat{D}^{-\frac{1}{2}}\vec{y}_\omega=\vec{y}_{\omega,R}$ are left and right eigenvectors of $\mat{P}$, respectively, with eigenvalue $1-\lambda_\omega$.
\end{prop}

\noindent Because of this, the eigenvectors of $\mat{P}$ have a similar form to those of $\mat{\mathcal{L}}$, but subject to the coordinate transformations of \cref{defin:LR-coord-trans}. Therefore, they can also be interpreted using the notion of smoothness, with $\lambda=1$ in some sense being the smoothest case.

\begin{figure}[h]
     \centering
     \begin{subfigure}[c]{0.7\textwidth}
         \centering
         \includegraphics[width=\textwidth]{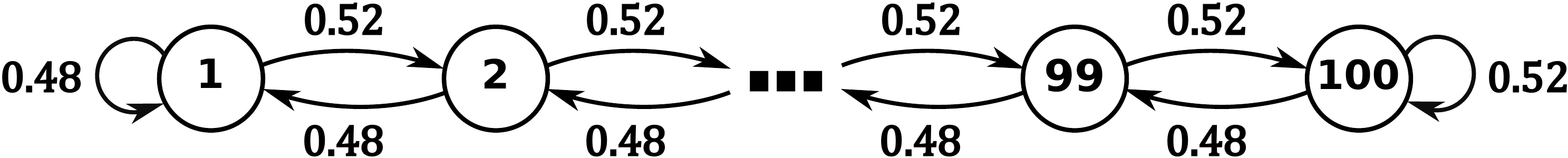}
         \caption{}
     \end{subfigure}
     \par\smallskip
     \begin{subfigure}[c]{0.4\textwidth}
         \centering
         \includegraphics[width=\textwidth]{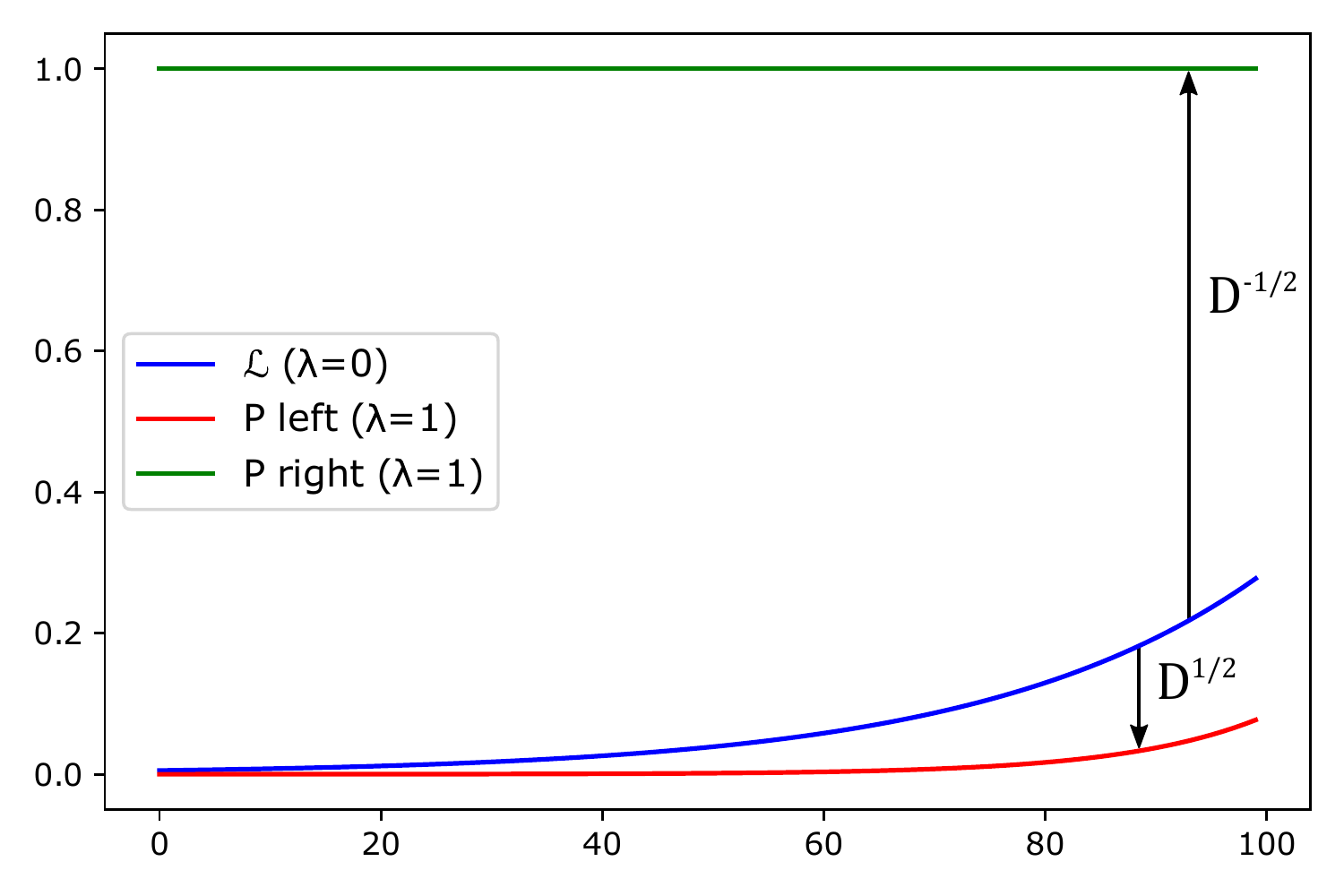}
         \caption{}
     \end{subfigure}
     \quad
     \begin{subfigure}[c]{0.4\textwidth}
         \centering
         \includegraphics[width=\textwidth]{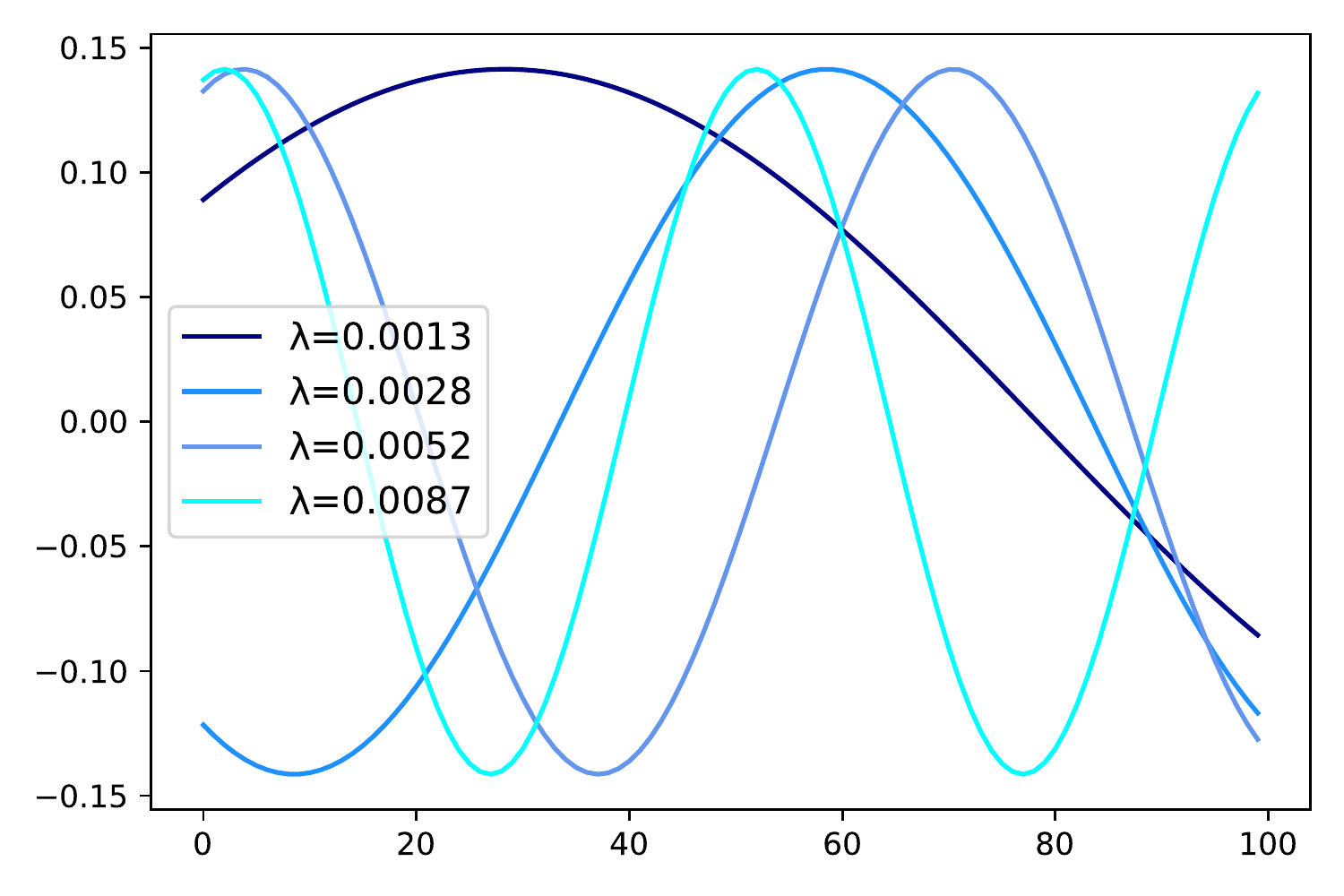}
         \caption{}
     \end{subfigure}
     \par\smallskip
     \begin{subfigure}[c]{0.4\textwidth}
         \centering
         \includegraphics[width=\textwidth]{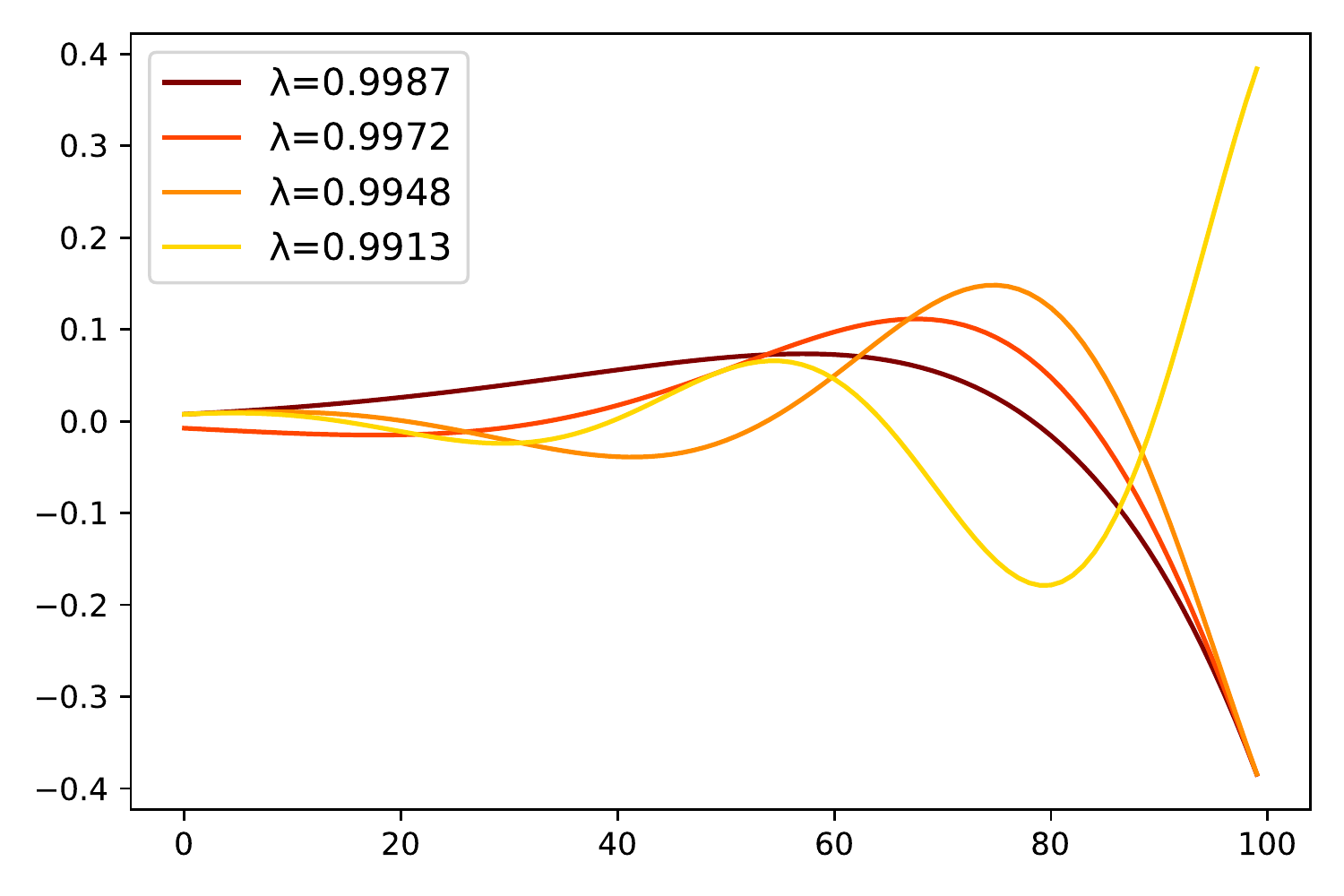}
         \caption{}
     \end{subfigure}
     \quad
     \begin{subfigure}[c]{0.4\textwidth}
         \centering
         \includegraphics[width=\textwidth]{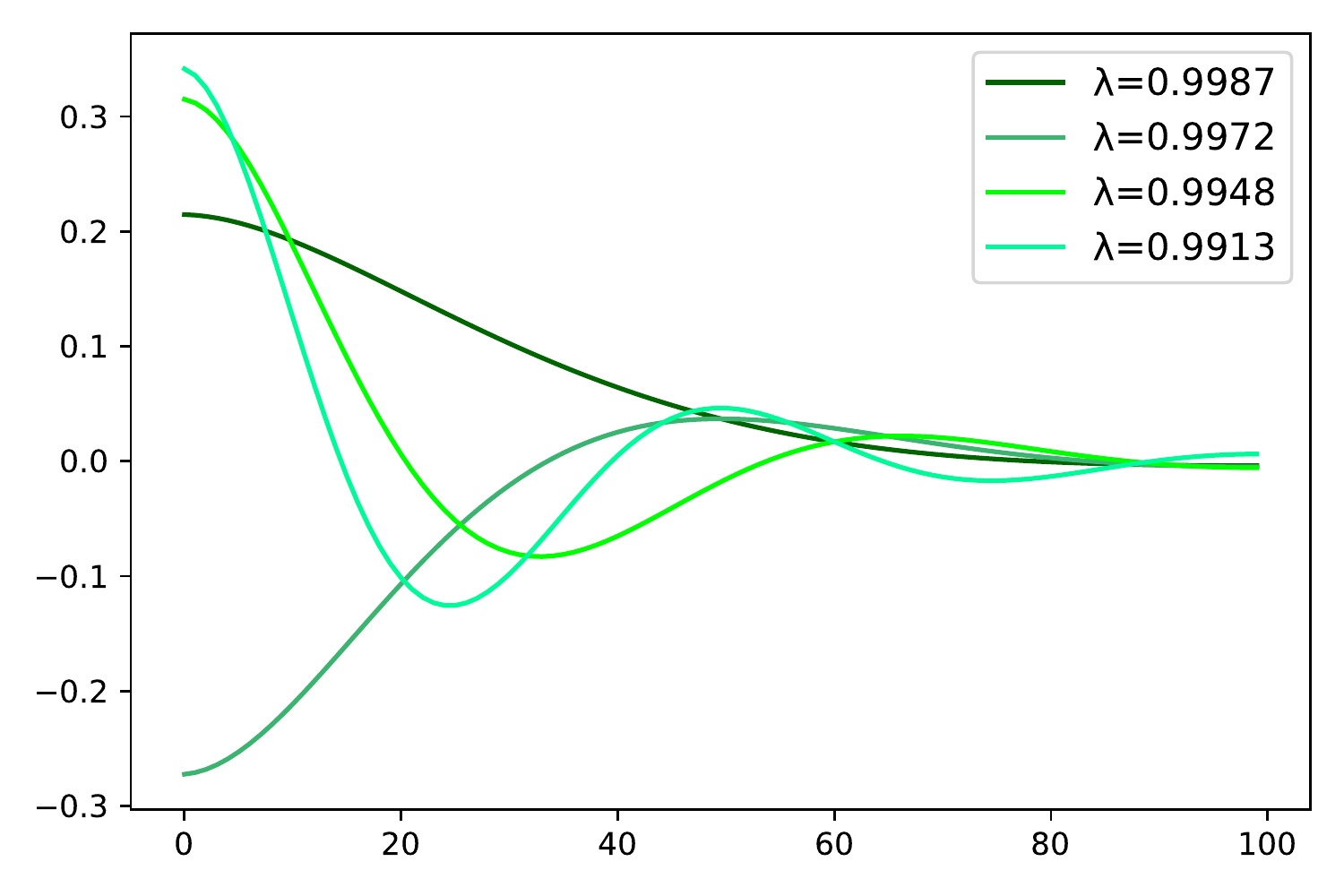}
         \caption{}
     \end{subfigure}
     \caption{(a) A Markov chain with $N=100$ states arranged in a line, with left and right transitions happening with probability $0.48$ and $0.52$, respectively. (b) The eigenvector of $\mat{\mathcal{L}}$ with $\lambda=0$ (blue), as well as the left (red) and right (green) eigenvectors of $\mat{P}$ for this model. (c) The next four eigenvectors of $\mat{\mathcal{L}}$, with the corresponding eigenvectors indicated in the legend, as well as similar plots for the next four left (d) and right (e) eigenvectors of $\mat{P}$. All eigenvalues are rounded to 4dp.}
    \label{fig:1DChains}
\end{figure}

To illustrate the relationship between the eigenvectors of $\mat{\mathcal{L}}$ and $\mat{P}$, we consider a simple Markov chain consisting of a linear arrangement of $100$ states, where only nearest neighbor transitions are allowed (\cref{fig:1DChains}(a)). Transitions to the left and right occur with probability 0.48 and 0.52, respectively, and the self loops at $s_1$ and $s_{100}$ mean that staying fixed is allowed in these states. Take a moment to verify that this chain is both ergodic and reversible (hint: in the latter case, try \cref{thm:KolCrit}). Once we know the stationary probabilities $\pi_i$ of this chain, we can calculate the corresponding matrices $\mat{K}$ and $\mat{\mathcal{L}}$ by using \cref{eq:Kmat} and then \cref{eq:NLapMat}. In \cref{fig:1DChains}(b), we plot the left and right eigenvector of $\mat{P}$ with $\lambda=1$, which are the stationary distribution $\vec{\pi}$ and $\vec{\eta}=(1, 1,..., 1)^T$, respectively. The stationary probabilities are larger for states near the right end of the chain by virtue of the tendency for rightward transitions in this chain. Additionally, we plot the corresponding eigenvector of $\mat{\mathcal{L}}$ with $\lambda=0$, which is $\vec{y}=(\pi_1^{\frac{1}{2}}, \pi_2^{\frac{1}{2}}, ..., \pi_{100}^{\frac{1}{2}})^T$. This vector is in some sense an intermediary between $\vec{\pi}$ and $\vec{\eta}$ since, in agreement with \cref{prop:Lnorm}, $\vec{\pi}=\mat{\Pi}^{\frac{1}{2}}\vec{y}=\mat{D}^{\frac{1}{2}}\vec{y}=\vec{y}_L$ and $\vec{\eta}=\mat{\Pi}^{-\frac{1}{2}}\vec{y}=\mat{D}^{-\frac{1}{2}}\vec{y}=\vec{y}_R$ (as indicated by the black arrows in \cref{fig:1DChains}(b)).

In \cref{fig:1DChains}(c), we show four more eigenvectors of $\mat{\mathcal{L}}$ having eigenvalues closest to $0$. In agreement with our foregoing discussion, the eigenvectors get less smooth as $\lambda$ increases, becoming more oscillatory and resembling trigonometric functions over the state space. In \cref{fig:1DChains}(d, e), we do the same for the left and right eigenvectors of $\mat{P}$, respectively. The smoothness of these eigenvectors also depends on the size of $\lambda$, however this time we are interested in those with eigenvalues closest to $1$, and the smoothness decreases as $\lambda$ decreases. Furthermore, in comparison to \cref{fig:1DChains}(c) these eigenvectors have an additional weighting effect across the state space. The left eigenvectors have larger amplitudes for states on the right-hand side, which is intuitive since these states have higher stationary probabilities $\pi_i$, and the eigenvectors can be obtained from those of $\mat{\mathcal{L}}$ by the left coordinate transformation of \cref{defin:LR-coord-trans}. For the right eigenvectors, the weighting is the opposite, with states on the left-hand side having larger amplitudes. This is explained by an equivalent argument, except that this time we apply the right coordinate transformation to the eigenvectors of $\mat{\mathcal{L}}$. It should be noted that for the purpose of visualization, all eigenvectors in \cref{fig:1DChains}(c-e) are normalized to have Euclidean norm $1$, even though for the eigenvectors of $\mat{P}$ this is not the natural normalization (\cref{thm:Rev-Eig}).

\begin{figure}[h]
\centering
\begin{subfigure}[c]{0.75\textwidth}
         \centering
         \includegraphics[width=\textwidth]{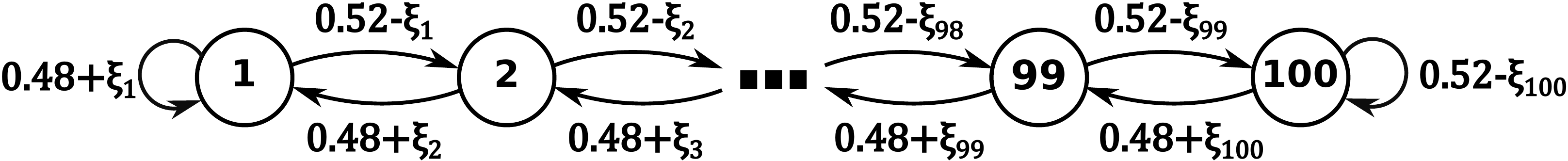}
         \caption{}
     \end{subfigure}
     \par\smallskip
     \begin{subfigure}[c]{0.4\textwidth}
         \centering
         \includegraphics[width=\textwidth]{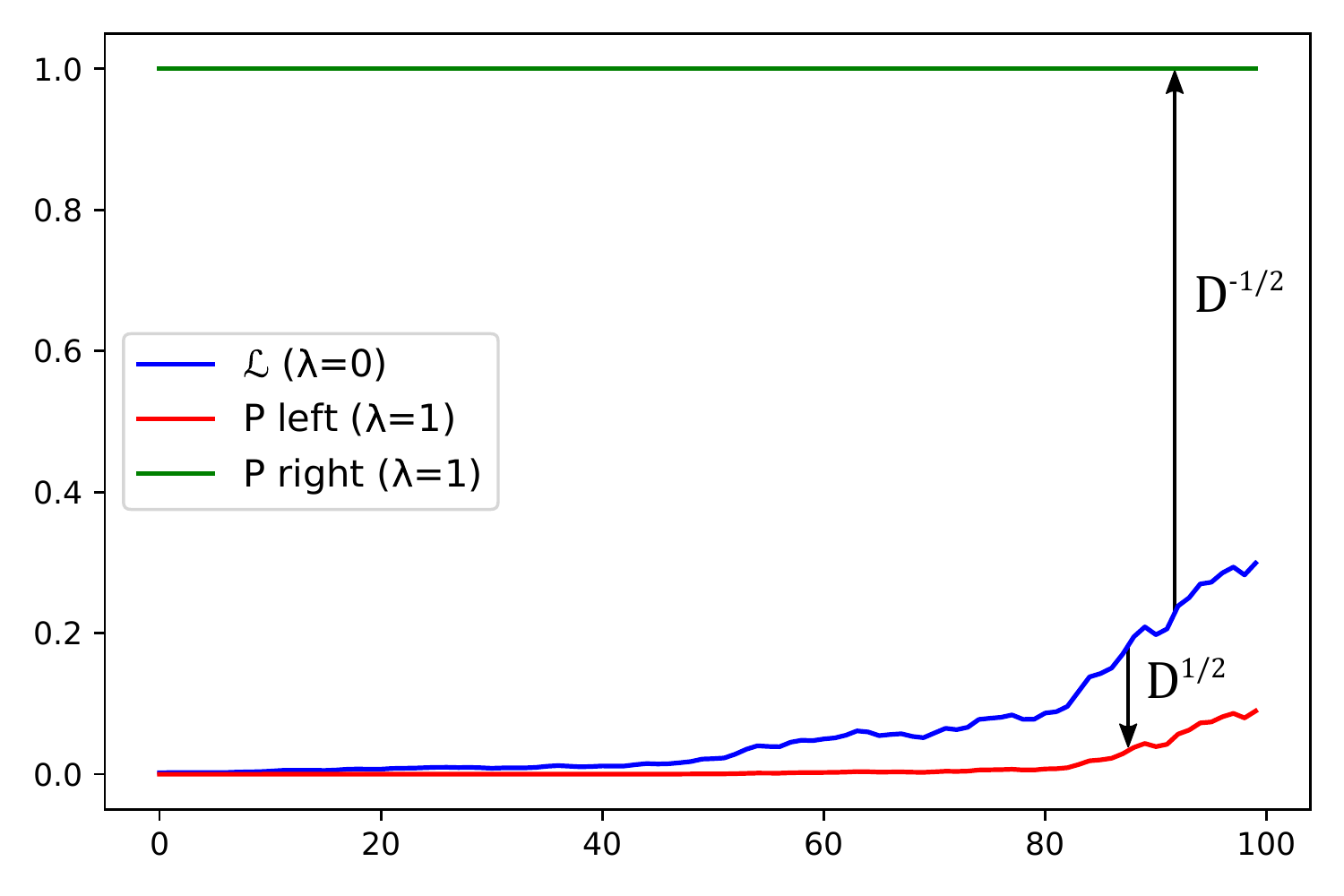}
         \caption{}
     \end{subfigure}
     \quad
     \begin{subfigure}[c]{0.4\textwidth}
         \centering
         \includegraphics[width=\textwidth]{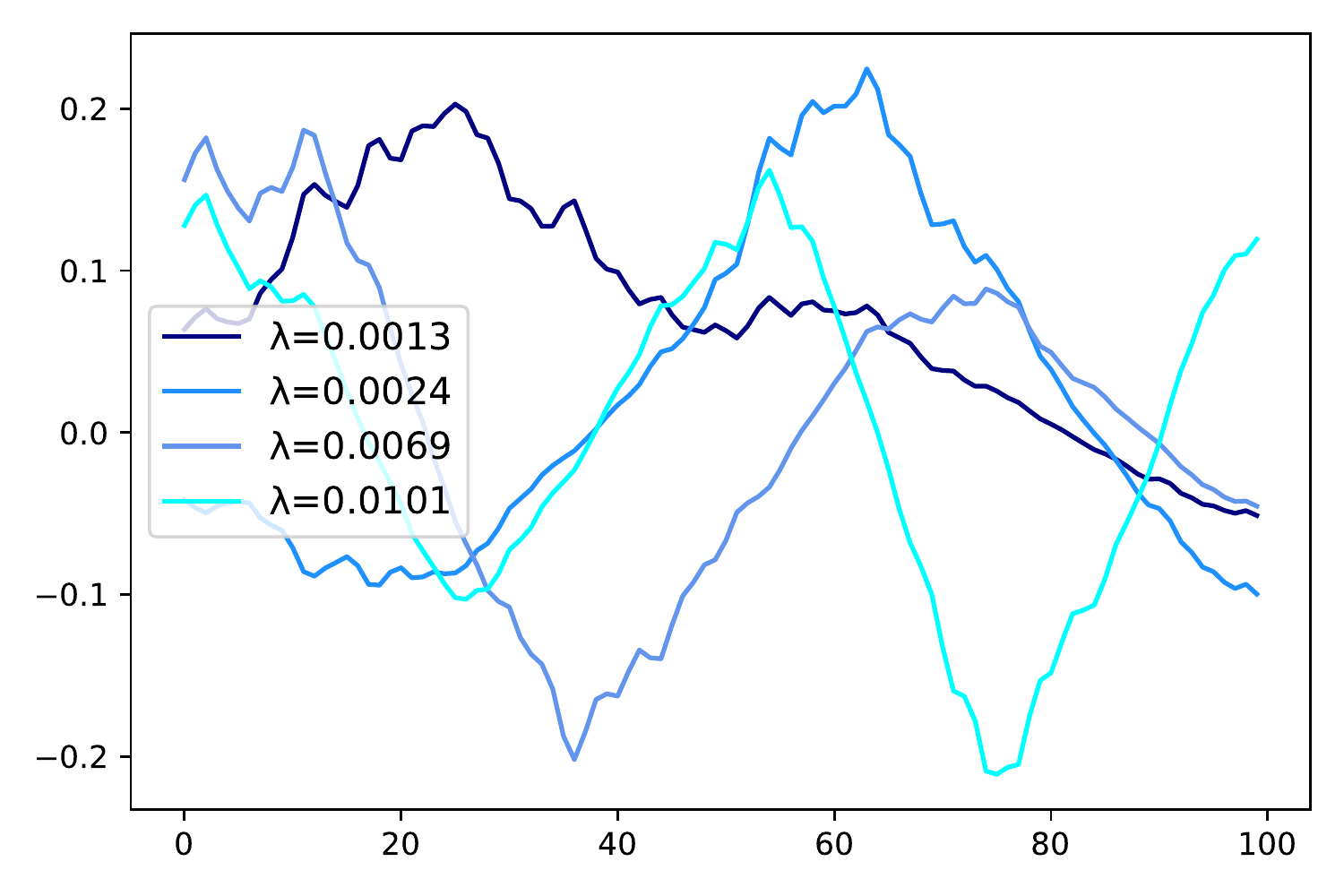}
         \caption{}
     \end{subfigure}
     \par\bigskip
     \begin{subfigure}[c]{0.4\textwidth}
         \centering
         \includegraphics[width=\textwidth]{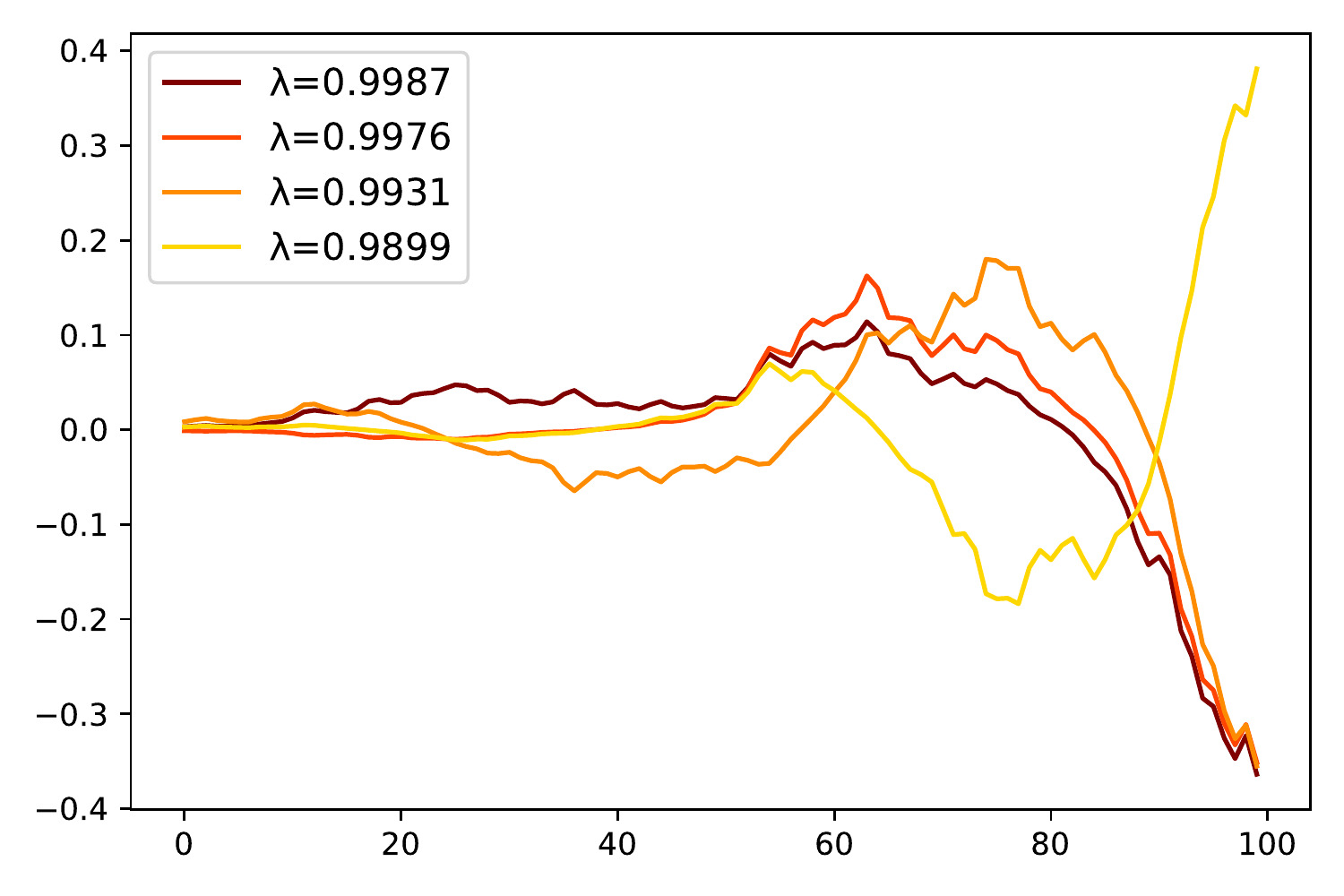}
         \caption{}
     \end{subfigure}
     \quad
     \begin{subfigure}[c]{0.4\textwidth}
         \centering
         \includegraphics[width=\textwidth]{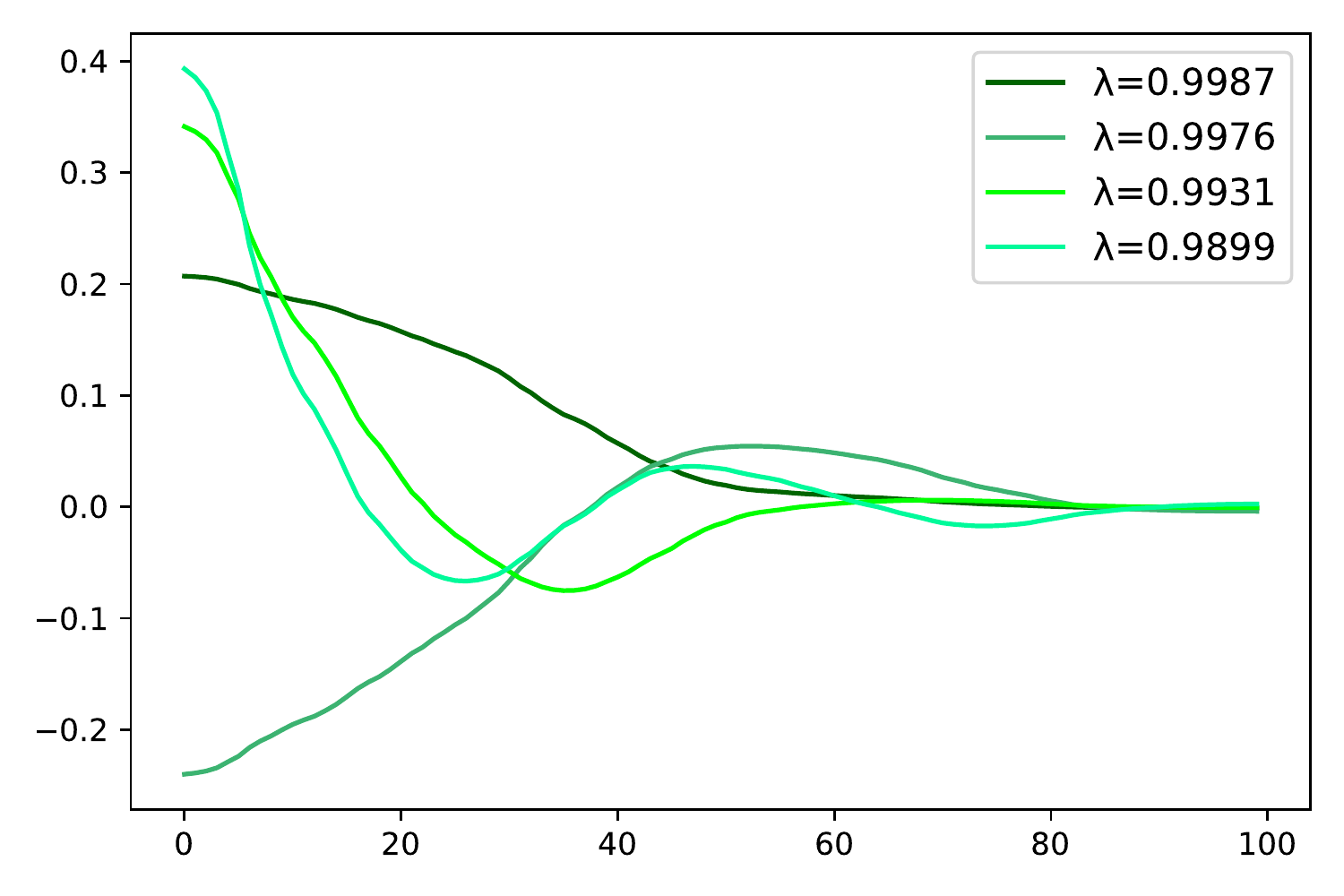}
         \caption{}
     \end{subfigure}
     \caption{(a) A modified version of the chain in \cref{fig:1DChains}(a), whereby the transition probabilities of $0.48$ and $0.52$ are perturbed by random numbers $\xi_i\in[-0.1,0.1]$. (c-e) The equivalent of the plots in \cref{fig:1DChains}(c-e), where again all eigenvalues are rounded to 4dp.}
    \label{fig:1DChainsAlt}
\end{figure}

The example chain in \cref{fig:1DChains}(a) is somewhat artificial since the transition probabilities are homogeneous, which is why the stationary distribution in \cref{fig:1DChains}(b) is such appears to be so smooth visually. This in turn means that the underlying graph $G$ has a degree structure that varies somewhat smoothly from left to right, which is partly responsible for the other eigenfunctions also being so smooth. To see how this generalizes as we make the transition probabilities less homogeneous, we consider a modified chain which we show in \cref{fig:1DChainsAlt}(a). In this model, left transitions from a state $s_i$ happen with probability $0.52+\xi_i$ and right transitions with probability $0.48-\xi_i$, where $\xi_i$ is a number sampled from a uniform distribution on the interval $[-0.1,0.1]$. Therefore, this chain is like the previous example, but with a random perturbation at each state. In \cref{fig:1DChainsAlt}(b-e), we show the corresponding versions of \cref{fig:1DChains}(b-e). As one can see, the eigenvectors of $\mat{\mathcal{L}}$ as well as the left eigenvectors of $\mat{P}$ have similar shapes to the previous case, but they appear more rugged due to the inhomogoneity of the transition probabilities. Given that the perturbations made to the previous model are small in size, this is an illustration of the fact that these sets of eigenvectors are very sensitive to local information. Rather interestingly, the same is not true for the right eigenvectors of $\mat{P}$: in \cref{fig:1DChainsAlt}(b) the right eigenvector with $\lambda=1$ is the same as in the previous example, and in \cref{fig:1DChainsAlt}(e) the other right eigenvectors are partially distorted but still appear somewhat smooth visually. This can be understood in the following way. Since the right eigenvectors of $\mat{P}$ are the right transformations of the eigenvectors of $\mat{\mathcal{L}}$, we can reformulate the optimization problem in \cref{eq:NLapOpt1,eq:NLapOpt2} in terms of the basis $\mat{Y}_R=\mat{D}^{-\frac{1}{2}}\mat{Y}$:
\begin{align}
    \text{minimize}\quad & \text{tr}(\mat{Y}_R^T\mat{D}^{\frac{1}{2}}\mat{\mathcal{L}}\mat{D}^{\frac{1}{2}}\mat{Y}_R)=\text{tr}(\mat{Y}_R^T\mat{L}\mat{Y}_R)\label{eq:PrOpt1}\\
    \text{subject to}\quad &\mat{Y}_R^T\mat{D}^{\frac{1}{2}}\mat{D}^{\frac{1}{2}}\mat{Y}_R=\mat{Y}_R^T\mat{D}\mat{Y}_R=\mathbbm{1}\label{eq:PrOpt2}
\end{align}
where we have used the fact that $\mat{D}^{\frac{1}{2}}\mat{\mathcal{L}}\mat{D}^{\frac{1}{2}}=\mat{D}^{\frac{1}{2}}(\mat{\mathbbm{1}}-\mat{D}^{-\frac{1}{2}}\mat{WD}^{-\frac{1}{2}})\mat{D}^{\frac{1}{2}}=\mat{D}-\mat{W}=\mat{L}$. Therefore, for right eigenvectors of $\mat{P}$, the relevant smoothness measure is related to $\mat{L}$ as opposed to $\mat{\mathcal{L}}$. Furthermore, writing \cref{eq:NLapQuad} in terms of $\mat{L}$ and $\vec{x}_R$ gives:
\begin{equation}
    \vec{x}_R^T\mat{L}\vec{x}_R=\frac{1}{2}\sum_{i, j=1}^NW_{ij}(x_{R,i}-x_{R,j})^2
\end{equation}
which we can interpret to mean that the vectors which optimize this measure, i.e.\ the right eigenvectors of $\mat{P}$ with eigenvalue close to $1$, have entries that are close for vertices with large $W_{ij}$. In the context of the example in \cref{fig:1DChainsAlt}, this means that such eigenvectors have similar values for neighboring vertices, and a quick check reveals that this is indeed the case in \cref{fig:1DChainsAlt}(d). Conversely, for the eigenvectors in \cref{fig:1DChainsAlt}(b,c) the values on average vary more between neighbouring vertices.

Lastly, we note that the ordering of eigenvalues used in this section is somewhat different to that in \cref{MCs-EV}. In the current section, the eigenvalues of $\mat{\mathcal{L}}$ are ordered from $0$ up to $2$, which corresponds to the eigenvalues of $\mat{P}$ being ordered from $1$ to $-1$. To reflect our interpretation, we call this \textit{ordering by smoothness}. In \cref{MCs}, however, the eigenvalues of a transition matrix are ordered by their absolute value, which describes how long the contribution from the corresponding eigenvector persists as the chain evolves. We therefore call this choice \textit{ordering by persistence}. The suitability of either of these types of ordering depends on the problem domain in which a Markov chain is being used. For example, in the case of spectral clustering they correspond to distinct objectives, as demonstrated in \cite{Liu2011}.

This concludes our treatment of random walks on undirected graphs. As a summary of the results given, in \cref{tab:UndirMatrices} we compare the mathematical properties and relationships of $\mat{\mathcal{L}}$, $\mat{K}$ and $\mat{P}$. The material presented in this section is particularly important for applications in machine learning and data mining in which the data set can be formulated as a graph. In particular, it underlies work that has been done on problems such as spectral clustering  \cite{Meila2000,Meila2001,Tishby2001,Saerens2004,Liu2011}, manifold learning/graph embedding \cite{Coifman2005,Coifman2006}, graph-based classification \cite{Kamvar2003,Szummer2001,Joachims2003}, and value function approximation in reinforcement learning \cite{Mahadevan2005a,Mahadevan2007,Petrik2007,Stachenfeld2014,Stachenfeld2017}. In the next section we consider how, if at all, the material presented in this section generalizes to directed graphs.

\begin{table}[h]
\centering
\begin{tabular}{ |0c|0c|0c|0c|}
 \hline
  & $\mat{P}$ & $\mat{K}$ & $\mat{\mathcal{L}}$ \\[1.5ex]
 \hline
 Relationship to $\mat{W}$ & $\mat{D}^{-1}\mat{W}$ & $\mat{D}^{-\frac{1}{2}}\mat{WD}^{-\frac{1}{2}}$ & $\mat{\mathbbm{1}}-\mat{D}^{-\frac{1}{2}}\mat{WD}^{-\frac{1}{2}}$ \\[1.5ex]
 \hline
 Diagonalizable & \ding{51} & \ding{51} & \ding{51} \\[1ex]
 \hline
 Symmetric & \ding{55} & \ding{51} & \ding{51} \\[1ex]
 \hline
 Positive semi-definite & \ding{55} & \ding{55} & \ding{51} \\[1ex]
 \hline
 Eigenvalues & $\lambda_\omega\in [-1, 1]$ & $\lambda_\omega\in [-1, 1]$ & $\lambda_\omega\in [0,2]$ \\[1ex]
 \hline
 Eigenvectors & lin.\ indep.\ & orthogonal & orthogonal \\[1ex]
 \hline
 Left eigenvectors & $\vec{y}_{\omega,L}$ & $\vec{y}_\omega$ & $\vec{y}_\omega$ \\[1ex]
 \hline
 Right eigenvectors & $\vec{y}_{\omega,R}$ & $\vec{y}_\omega$ & $\vec{y}_\omega$ \\[1ex]
 \hline
\end{tabular}
\caption{A summary of the properties established for the three matrices associated to random walks on undirected graphs: $\mat{P}$, $\mat{K}$, and $\mat{\mathcal{L}}$.}\label{tab:UndirMatrices}
\end{table}

\subsection{Random walks on directed graphs}
\label{RWs-dir}
Broadening our consideration to directed graphs is necessary if we want to describe non-reversible Markov chains. However, since many of the guarantees established in  \cref{RWs-undir} do not hold for directed graphs, this case is a lot harder to treat analytically. In \cref{RWs-dir-diff} we explore the main challenges that occur when applying spectral graph theory to the transition matrices of non-reversible Markov chains, and in \cref{RWs-dir-methods} we describe methods for circumventing these issues. Finally, in \cref{RWs-dir-Lap} we define a generalization of $\mat{\mathcal{L}}$ to directed graphs, and in \cref{RWs-dir-Erg} we present a method for enforcing ergodicity on random walks of directed graphs.

\subsubsection{Key difficulties}
\label{RWs-dir-diff}
\begin{figure}[h!]
     \centering
     \begin{subfigure}[b]{0.45\textwidth}
        \centering         
        \includegraphics[width=3.7cm]{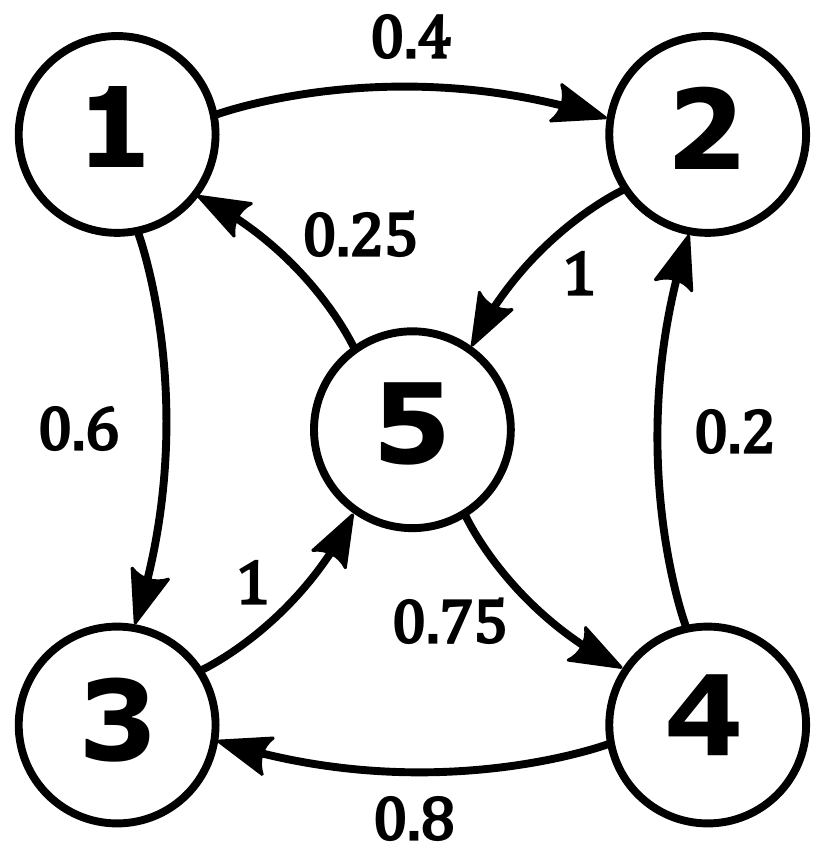}
        \par\bigskip
        $\mat{P}=\left(
            \begin{array}{ccccc}
            0 & 0.4 & 0.6 & 0 & 0 \\
            0 & 0 & 0 & 0 & 1 \\
            0 & 0 & 0 & 0 & 1 \\
            0 & 0 & 0.8 & 0 & 0.2 \\
            0.25 & 0 & 0 & 0.75 & 0 
            \end{array} \right)$
        \par\bigskip
     \caption{Non-diagonalizble with complex eigenvalues}
     \end{subfigure}
     \qquad
     \begin{subfigure}[b]{0.45\textwidth}
        \centering         
        \includegraphics[width=3.4cm]{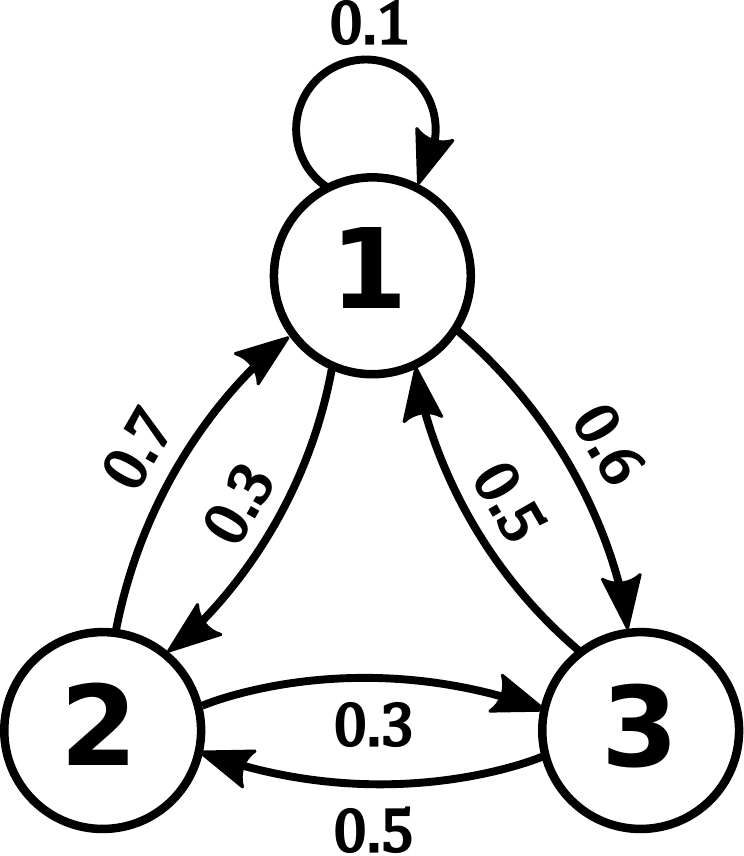}
        \par\bigskip\bigskip
         $\mat{P}=\left(
            \begin{array}{ccc}
            0.1 & 0.3 & 0.6 \\
            0.7 & 0 & 0.3 \\
            0.5 & 0.5 & 0
            \end{array} \right)$
        \par\bigskip\bigskip
        \caption{Diagonalizable with complex eigenvalues}
     \end{subfigure}
     \par\bigskip
     \begin{subfigure}[b]{0.45\textwidth}
        \centering         
        \includegraphics[width=5.2cm]{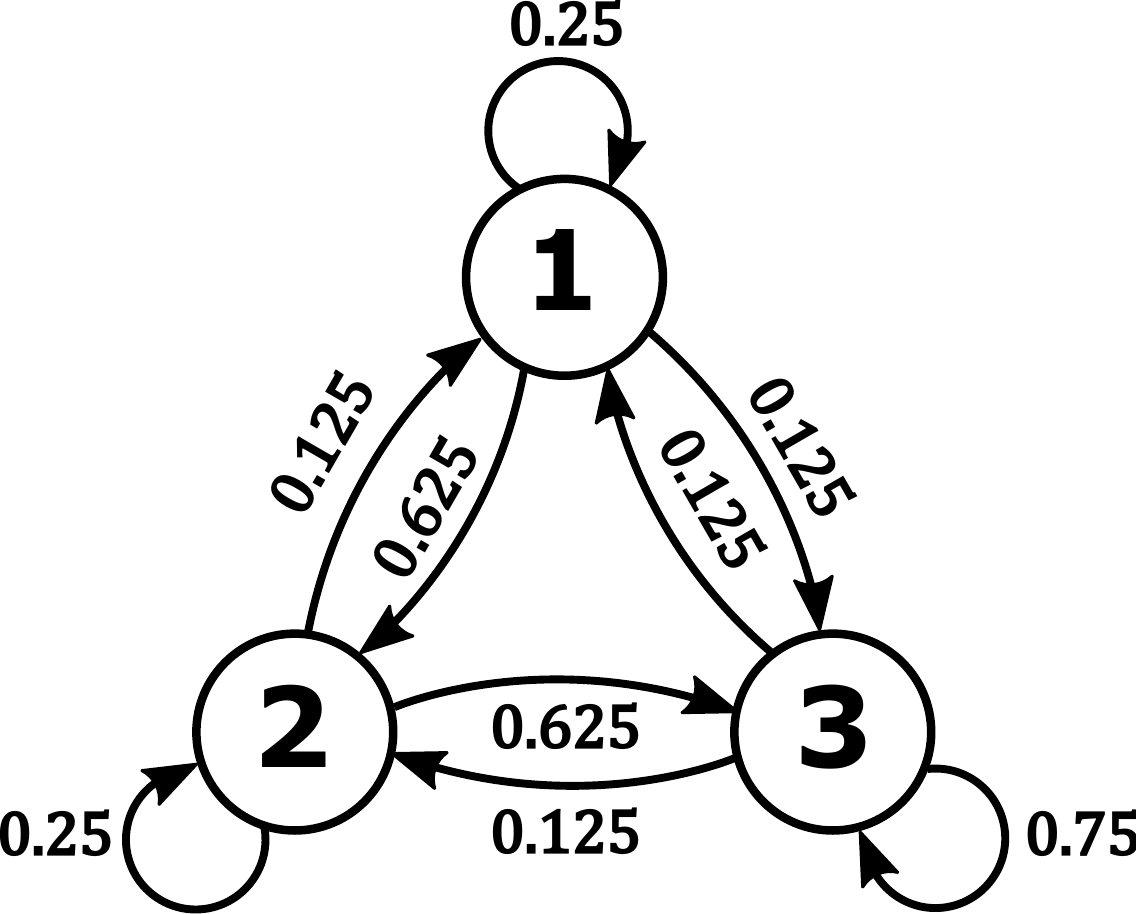}
        \par\bigskip
        $\mat{P}=\left(
            \begin{array}{cccc}
            0.25 & 0.625 & 0.125 \\
            0.125 & 0.25 & 0.625 \\
            0.125 & 0.125 & 0.75
            \end{array} \right)$
        \par\bigskip
        \caption{Non-diagonalizable with real eigenvalues}
     \end{subfigure}
     \qquad
     \begin{subfigure}[b]{0.45\textwidth}
        \centering
        \includegraphics[width=5cm]{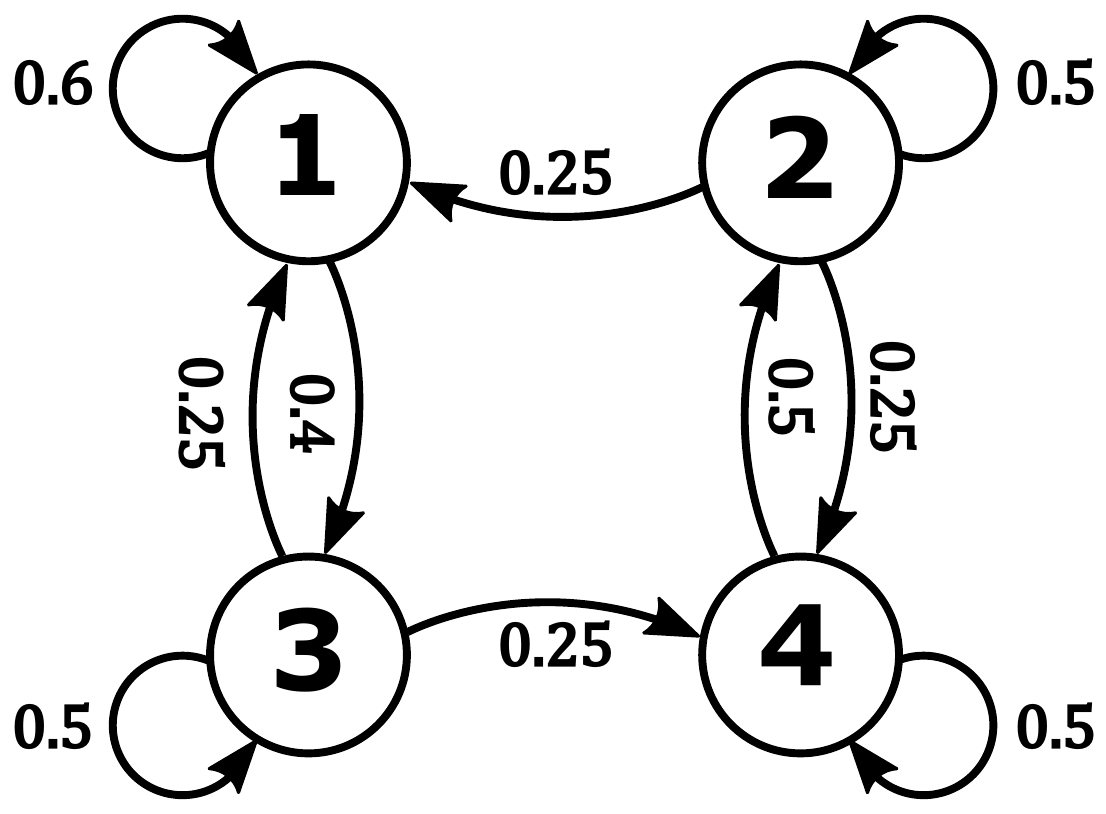}
        \par\bigskip
        $\mat{P}=\left(
            \begin{array}{cccc}
            0.6 & 0 & 0.4 & 0 \\
            0.25 & 0.5 & 0 & 0.25 \\
            0.25 & 0 & 0.5 & 0.25 \\
            0 & 0.5 & 0 & 0.5
            \end{array} \right)$
        \par\bigskip
        \caption{Diagonalizable with real eigenvalues}
     \end{subfigure}
     \caption{(a-d) Four irreducible non-reversible Markov chains (top), that have transition (bottom) with different properties.}
     \label{fig:RW-dir-Eig}
\end{figure}

Since many of the guarantees established in  \cref{RWs-undir} do not hold for directed graphs, they are a lot harder to treat analytically. Perhaps most importantly, the transition matrices of non-reversible chains are neither guaranteed to be diagonalizable, nor to have real eigenvalues \cite{Weber2017}, which can be observed even for simple cases such as those shown in  \cref{fig:RW-dir-Eig}(a-c). There are nonetheless some cases for which both properties hold \cite{Weber2017} (as shown by the example in \cref{fig:RW-dir-Eig}(d)), but it is still not fully understood to what degree, if at all, the transition structure of a non-reversible chain determines either the diagonalizability of $\mat{P}$ or whether its eigenvalues are real/complex.\footnote{Although some studies have considered complex eigenspaces in specific settings \cite{Meyn2008,Andrieux2011, Mieghem2018, Conrad2016}.} When $\mat{P}$ is non-diagonalizable, it does not have a set of $N$ linearly independent eigenvectors, which can cause numerical issues since in this case some matrix operations are computationally more expensive or not well-defined. Moreover, if the eigenvalues of $\mat{P}$ are complex, then so are its eigenvectors. As in the real case, we can choose to order these eigenvectors based on persistence, or smoothness. In the former case, the generalization is somewhat straightforward since $|\lambda|$ still describes how long each eigenvector typically persists. In the latter case, however, the question of how to generalize the concept of smoothness to complex eigenvectors is non-trivial, and is still an actively researched topic in the literature \cite{Sevi2018,Marques2020}. These factors make analyzing the transition matrices of non-reversible Markov chains more challenging than in the reversible case.

\subsubsection{Alternative methods}
\label{RWs-dir-methods}
One general technique for treating a non-diagonalizable matrix $\mat{X}$ is to add a perturbation so that it becomes diagonalizable. This is based on the notion that diagonalizable matrices densely fill the set of all matrices \cite{Golub2013}, meaning that it is always possible to find some nearby matrix $\mat{X}'$ that is diagonalizable. In one study, a method along these lines was developed for dealing with non-diagonalizable transition matrices \cite{Pauwelyn2021}. In particular, for a starting transition matrix $\mat{P}$, a perturbation matrix $\mat{E}$ is found such that $\mat{P}'=\mat{P}+\mat{E}$ preserves a number of the spectral properties of $\mat{P}$ and is diagonalizable. However, two limitations of this method are (i) that it has computational complexity $\mathcal{O}(N^8)$ for $N\times N$ matrices, and (ii) the resulting transition matrix can still have complex eigenvalues.

Other lines of work have attempted to circumvent these issues by using alternative matrix decompositions, with a prominent example being the real Schur decomposition \cite{Stewart1994,Conrad2016,Weber2017,Fackeldey2018,Ghosh2020}.\footnote{It should be noted that most of these studies primarily consider non-reversible chains that are ergodic.} This decomposition provides a set of real orthogonal basis vectors, known as Schur vectors, that spans the eigenspaces of $\mat{X}$ \cite{Golub2013}. This basis is not unique and corresponds to some ordering of the eigenvalues of $\mat{X}$, with the first $k$ Schur vectors spanning the eigenspaces of the first $k$ eigenvectors in this ordering. Therefore, given some Schur decomposition of $\mat{X}$, if $U_k=(\vec{u}_1, \vec{u}_2, ..., \vec{u}_k)$ is the set of first $k$ Schur vectors, then for any linear combination $\tilde{\vec{u}}=\sum_{\gamma=1}^k c_\gamma\vec{u}_\gamma$ it is guaranteed that $\mat{X}\tilde{\vec{u}}\in \text{span}(U_k)$. For this reason, $U_k$ is said to be an \textit{invariant subspace} of $\mat{X}$ \cite{Golub2013}, and when $k<<N$ it provides a low dimensional description of the transformation that $\mat{X}$ represents. The real Schur decomposition is therefore a useful alternative to the eigendecomposition. However, in order that the basis captures the most important information about $\mat{X}$, a reordering algorithm is needed to specify \textit{which} eigenspaces of $\mat{X}$ it should span, and various methods for this have been developed \cite{Ng1987,Dongarra1992,Bai1993,Granat2009,Brandts2002}. Furthermore, it is worth emphasizing that in contrast to the eigendecomposition, the real Schur decomposition is guaranteed to exist for any real square matrix $\mat{X}$, meaning that it sidesteps the issues of non-diagonalizability and complex feature spaces that can occur with transition matrices of non-reversible Markov chains. In the field of machine learning, the real Schur decomposition of transition matrices has been used as a tool for clustering \cite{Fackeldey2018} as well as for building state representations in reinforcement learning \cite{Ghosh2020}.

\subsubsection{Directed Normalized Graph Laplacian}
\label{RWs-dir-Lap}
In \cref{RWs-Lap}, the normalized graph Laplacian $\mat{\mathcal{L}}$ was introduced as a way to get a more precise description of the left and right eigenvectors belonging to transition matrices of reversible Markov chains. Generalizing $\mat{\mathcal{L}}$ to directed graphs is challenging since two of its defining features are that it is symmetric and positive semi-definite, neither of which can be satisfied by \cref{eq:NLapMatW} if $\mat{W}$ is non-symmetric. However, various definitions for directed graphs exist, and while some loosen the constraint that $\mat{\mathcal{L}}$ should be positive semi-definite \cite{Agaev2005,Caughman2006,Li2012,Singh2016}, others strictly enforce this via a type of symmetrization \cite{Chung2005}. We here focus on the latter type, and demonstrate connections that this has to some of the material in \cref{MCs}.

Perhaps the simplest method along these lines is to symmetrize the weight matrix $\mat{W}$ of a directed graph $G$ to get an alternative weight matrix $\mat{W}_\text{sym}$, e.g.\ $\mat{W}_\text{sym}=\mat{W}+\mat{W}^T$ or $\mat{W}_\text{sym}=\mat{W}^T\mat{W}$, and then use the regular definition of the normalized Laplacian using this new matrix, i.e.\ $\mat{D}^{-\frac{1}{2}}\mat{W}_\text{sym}\mat{D}^{-\frac{1}{2}}$. Since $\mat{W}_\text{sym}$ describes an undirected graph, the resulting object can be interpreted in the same way as \cref{RWs-Lap}. However, a major drawback of this approach is that the graphs described by $\mat{W}$ and $\mat{W}_\text{sym}$ can have very different structural properties. For instance, there is no guarantee that the random walks on these two graphs have stationary distributions that bear any resemblance to one another. Indeed, various studies in machine learning have indicated that symmetrizing $\mat{W}$ leads to a significant erasure of structural information from a directed graph \cite{Pentney2005,Mahadevan2006,Meila2007,Johns2007}.

A more principled approach was given in \cite{Chung2005}, in which the directed normalized graph Laplacian is defined as:
\begin{equation}
\label{eq:DNLap}
    \mat{\mathcal{L}}_{\text{dir}}:=\mat{\mathbbm{1}}- \frac{\mat{\Pi}^{\frac{1}{2}}\mat{P\Pi}^{-\frac{1}{2}}+\mat{\Pi}^{-\frac{1}{2}}\mat{P}^T\mat{\Pi}^{\frac{1}{2}}}{2}
\end{equation}
where $\mat{P}$ and $\mat{\Pi}$ are defined as usual. In the original paper, restrictions are placed on the underlying graph so that $\mat{P}$ is ergodic. Among other things, this means that the stationary probabilities are all non-zero, which is needed for $\mat{\Pi}^{-\frac{1}{2}}$ to be well-defined. In the next section, we present a method for enforcing this property for any directed graph $G$, but for now we simply assume that $\vec{\pi}>0$. \cref{eq:DNLap} can be simplified as follows:
\begin{align}
    \mat{\mathcal{L}}_{\text{dir}}&=\mat{\mathbbm{1}}- \frac{\mat{\Pi}^{\frac{1}{2}}\mat{P\Pi}^{-\frac{1}{2}}+\mat{\Pi}^{\frac{1}{2}}(\mat{\Pi}^{-1} \mat{P}^T\mat{\Pi})\mat{\Pi}^{-\frac{1}{2}}}{2}\\
    &=\mat{\mathbbm{1}}-\mat{\Pi}^{\frac{1}{2}}\underbrace{\bigg(\frac{\mat{P}+\mat{\Pi}^{-1}\mat{P}^T\mat{\Pi}}{2}\bigg)}_{=\mat{P}_{\text{A}}}\mat{\Pi}^{-\frac{1}{2}}\label{eq:DNLap2}
\end{align}
where $\mat{P}_{\text{A}}$ is the additive reversibilization of the random walk (\cref{defin:AddRev}). Thus, by comparing \cref{eq:DNLap2} to \cref{eq:NLapMatP}, we see that $\mat{\mathcal{L}}_{\text{dir}}$ is a variant of $\mat{\mathcal{L}}$ but where $\mat{P}$ has been exchanged for $\mat{P}_{\text{A}}$. Thus, this corresponds to a symmetrization of the stationary flow between states (\cref{eq:AdRevFlow}). While this transformation still destroys some information about the underlying graph $G$, it has been observed empirically that this effect is less severe in comparison to methods that symmetrize $\mat{W}$ itself \cite{Pentney2005,Mahadevan2006,Meila2007,Johns2007}. For example, note that $\mat{P}$ and $\mat{P}_{\text{A}}$ describe random walks that have the same stationary distributions (\cref{defin:AddRev}). 

The only time that symmetrizing $\mat{W}$ and using $\mat{P}_{\text{A}}$ yields the same result is in the special case of a balanced directed graph. In this case,  $\mat{\Pi}^{\frac{1}{2}}=z^{-\frac{1}{2}}\mat{D}^{\frac{1}{2}}$ (\cref{thm:RWequivClassRecSD}), which allows us to simplify \cref{eq:DNLap2}:
\begin{align}
    \mat{\mathcal{L}}_{\text{dir}}&=\mat{\mathbbm{1}}-z^{-\frac{1}{2}}\mat{D}^{\frac{1}{2}}\bigg(\frac{\mat{D}^{-1}\mat{W}+\mat{D}^{-1}(\mat{D}^{-1}\mat{W})^T\mat{D}}{2}\bigg)z^{\frac{1}{2}}\mat{D}^{-\frac{1}{2}}\\
    &=\mat{\mathbbm{1}}-\mat{D}^{\frac{1}{2}}\bigg(\frac{\mat{D}^{-1}\mat{W}+\mat{D}^{-1}\mat{W}^T\mat{D}^{-1}\mat{D}}{2}\bigg)\mat{D}^{-\frac{1}{2}}\\
    &=\mat{\mathbbm{1}}-\mat{D}^{\frac{1}{2}}\bigg(\frac{\mat{D}^{-1}\mat{W}+\mat{D}^{-1}\mat{W}^T}{2}\bigg)\mat{D}^{-\frac{1}{2}}\\
    &=\mat{\mathbbm{1}}-\mat{D}^{-\frac{1}{2}}\bigg(\frac{\mat{W}+\mat{W}^T}{2}\bigg)\mat{D}^{-\frac{1}{2}} \label{eq:DNLap3}
\end{align}
meaning that $\mat{\mathcal{L}}_{\text{dir}}$ is equivalent to an additive symmetrization of $\mat{W}$. However, in most practical applications directed graphs are not balanced, and it is in these cases that using $\mat{P}_{\text{A}}$ preserves more information about a directed graph than simply symmetrizing $\mat{W}$.

The directed normalized Laplacian has been used in various contexts of machine learning as a way to generalize methods that are restricted to undirected graphs. It has been applied to problems such as spectral clustering \cite{Meila2007,Huang2006}, graph embedding \cite{Chen2007,Joncas2011}, graph-based classification \cite{Zhou2005}, and value function approximation in reinforcement learning \cite{Johns2007}. In most of these applications, ergodicity is enforced on the random walk described by $\mat{P}$, and in the next section we introduce the standard method for doing this.

\subsubsection{Random surfer model}
\label{RWs-dir-Erg}
As explained in \cref{MCs-Erg}, ergodic Markov chains have the useful property that they are guaranteed to converge to a unique stationary distribution, and various reasons were given for why this is desirable in a general context. For directed graphs, it is a particularly beneficial property, since without it a random walk can get trapped in a small cluster of states, or even a single absorbing state. Because of this, sometimes ergodicity is enforced for random walks on directed graphs \cite{Page1999,Zhou2005,Huang2006,Meila2007,Johns2007}. We remind readers that one effect this has is that $\vec{\pi}>0$, meaning that the directed normalized Laplacian is well-defined. 

The typical method used for enforcing ergodicity can be thought of as a variation on the random walk process. Given a directed graph $G$ with weight matrix $\mat{W}$, at each time step there are two possible outcomes: either a regular random walk is performed with probability $\alpha$, or the process teleports randomly to any vertex with probability $1-\alpha$. This is known as a \textit{random surfer model} or \textit{teleporting random walk}, and it appeared for the first time in the PageRank algorithm \cite{Page1999}. If $\mat{P}$ and $\mat{P}_{\text{tel}}$ represent the transition probabilities of the two possible outcomes at each time point, then the overall process is described by the following transition matrix:

\begin{equation}
\label{eq:GoogMat1}
    \mat{P}':=\alpha\mat{P}+(1-\alpha)\mat{P}_{\text{tel}}
\end{equation}
which is sometimes referred to as a \textit{Google matrix} \cite{Franceschet2011}.

It is worth noting that there exist many variants of the random surfer model that differ in the assumptions they make about $\mat{P}_{\text{tel}}$ \cite{Berkhin2005}. For example, teleporting transitions can either be uniformly random, or biased towards certain vertices through a set of weights. Furthermore, the parameter $\alpha\in [0,1]$ is known as the \textit{damping factor} and determines how close the process is to a regular random walk. Typically, it is set close to $1$, so that the process still accurately reflects the structure of the underlying graph $G$. 

\begin{figure}[h]
     \centering
     \begin{subfigure}[b]{0.25\textwidth}
         \centering
         \includegraphics[width=0.8\textwidth]{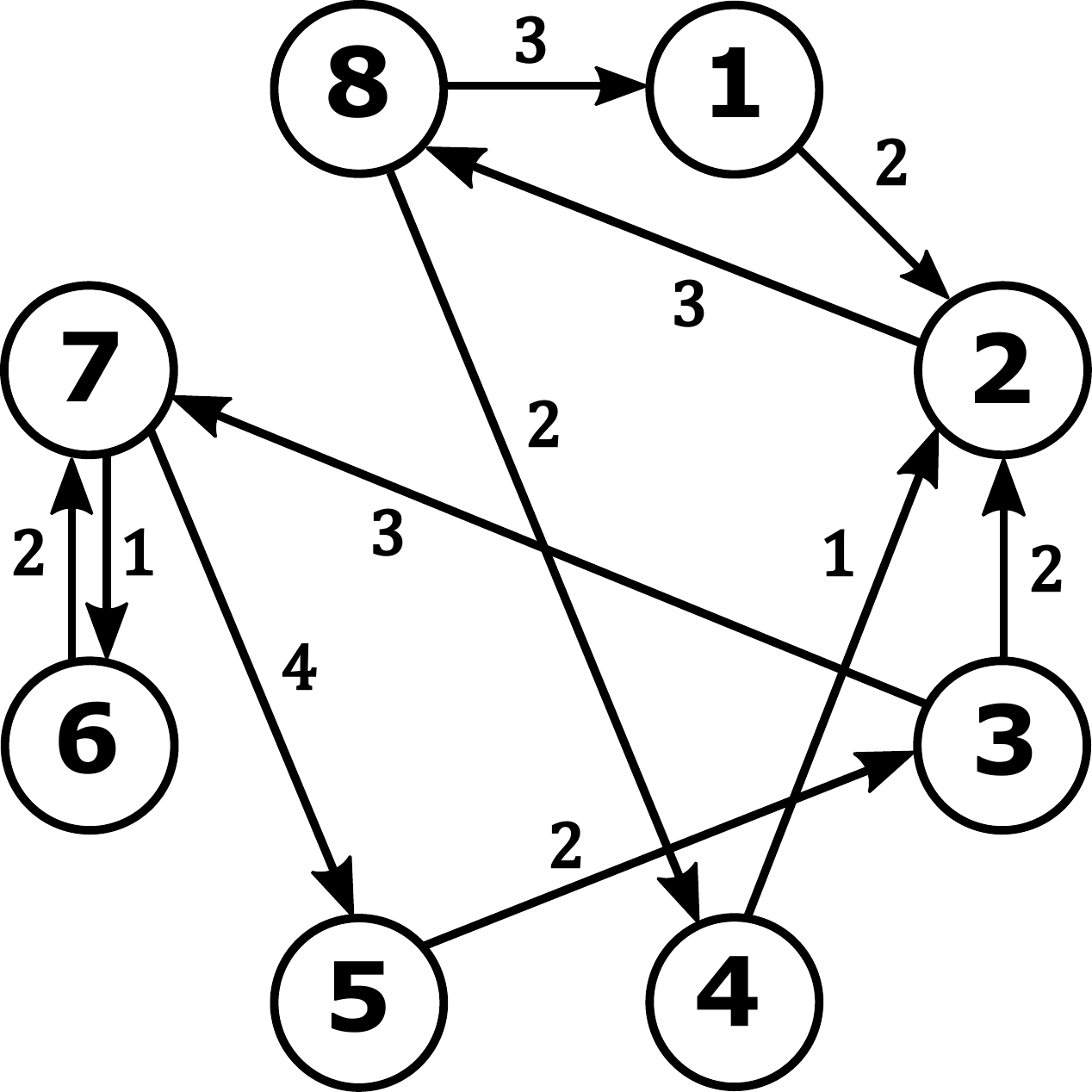}
         \vspace{9pt}
        \caption{}
     \end{subfigure}
     \hfill
     \begin{subfigure}[b]{0.25\textwidth}
         \centering
         \includegraphics[width=0.95\textwidth]{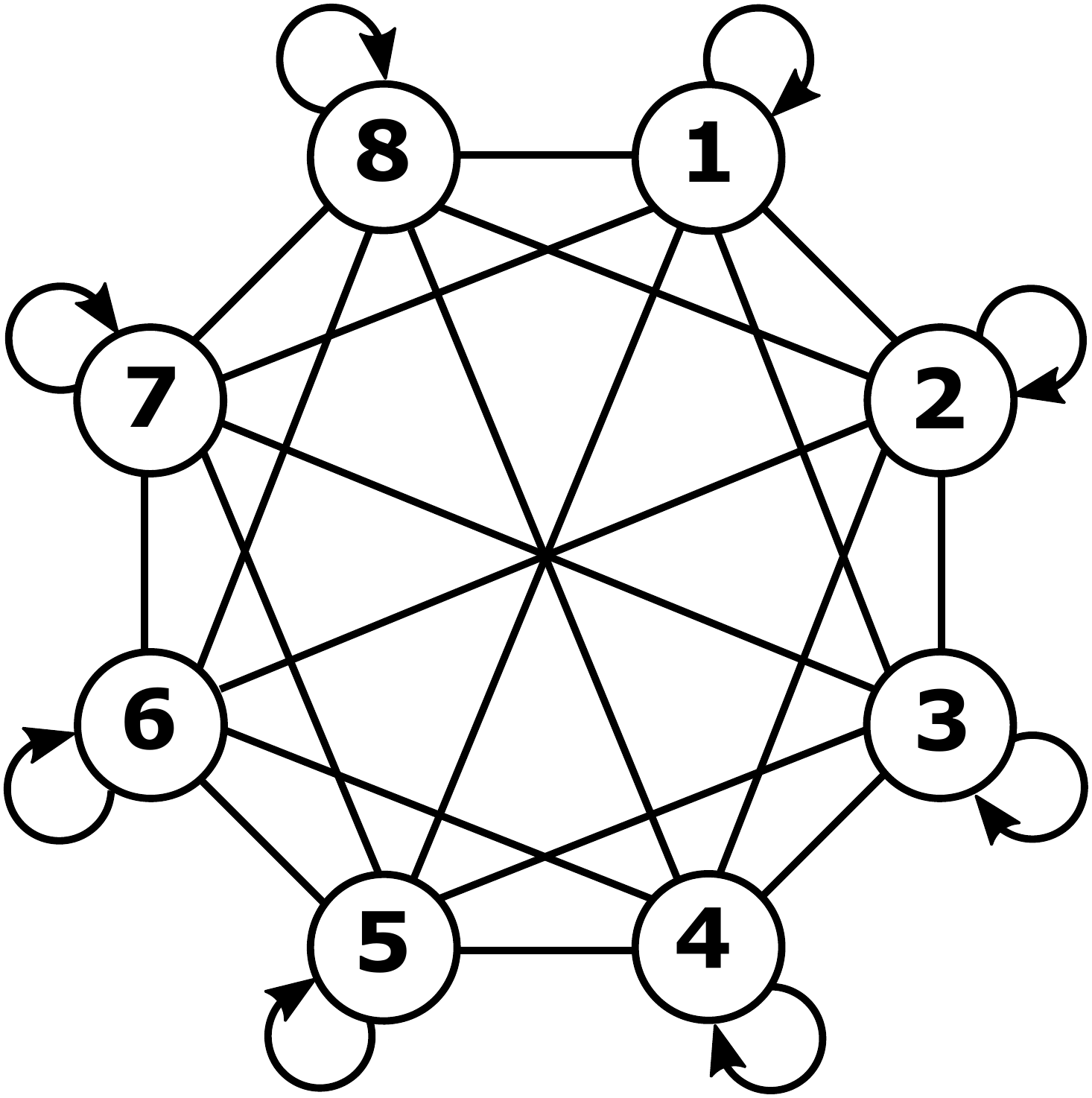}
        \caption{}
     \end{subfigure}
     \hfill
     \begin{subfigure}[b]{0.45\textwidth}
     \centering
        $\mat{P}=\left(
         \begin{array}{cccccccc}
         0 & 1 & 0 & 0 & 0 & 0 & 0 & 0\\
         0 & 0 & 0 & 0 & 0 & 0 & 0 & 1\\
         0 & 0.4 & 0 & 0 & 0 & 0 & 0.6 & 0\\
         0 & 1 & 0 & 0 & 0 & 0 & 0 & 0\\
         0 & 0 & 1 & 0 & 0 & 0 & 0 & 0\\
         0 & 0 & 0 & 0 & 0 & 0 & 1 & 0\\
         0 & 0 & 0 & 0 & 0.8 & 0.2 & 0 & 0\\
         0.6 & 0 & 0 & 0.4 & 0 & 0 & 0 & 0\\
         \end{array} \right)$
         \caption{}
     \end{subfigure}
     \par\bigskip
     \begin{subfigure}[b]{0.38\textwidth}
     \centering
         $\mat{P}_{\text{tel}}=\frac{1}{8}\left(
         \begin{array}{cccccccc}
         1 & 1 & 1 & 1 & 1 & 1 & 1 & 1\\
         1 & 1 & 1 & 1 & 1 & 1 & 1 & 1\\
         1 & 1 & 1 & 1 & 1 & 1 & 1 & 1\\
         1 & 1 & 1 & 1 & 1 & 1 & 1 & 1\\
         1 & 1 & 1 & 1 & 1 & 1 & 1 & 1\\
         1 & 1 & 1 & 1 & 1 & 1 & 1 & 1\\
         1 & 1 & 1 & 1 & 1 & 1 & 1 & 1\\
         1 & 1 & 1 & 1 & 1 & 1 & 1 & 1\\
         \end{array} \right)$
         \caption{}
     \end{subfigure}
     \hfill
     \begin{subfigure}[b]{0.6\textwidth}
     \centering
         $\mat{P}'=\left(
         \begin{array}{cccccccc}
         0.02 & 0.87 & 0.02 & 0.02 & 0.02 & 0.02 & 0.02 & 0.02\\
         0.02 & 0.02 & 0.02 & 0.02 & 0.02 & 0.02 & 0.02 & 0.87\\
         0.02 & 0.36 & 0.02 & 0.02 & 0.02 & 0.02 & 0.53 & 0.02\\
         0.02 & 0.87 & 0.02 & 0.02 & 0.02 & 0.02 & 0.02 & 0.02\\
         0.02 & 0.02 & 0.87 & 0.02 & 0.02 & 0.02 & 0.02 & 0.02\\
         0.02 & 0.02 & 0.02 & 0.02 & 0.02 & 0.02 & 0.87 & 0.02\\
         0.02 & 0.02 & 0.02 & 0.02 & 0.7 & 0.19 & 0.02 & 0.02\\
         0.53 & 0.02 & 0.02 & 0.36 & 0.02 & 0.02 & 0.02 & 0.02\\
         \end{array} \right)$
         \caption{}
     \end{subfigure}
    \caption{(a) a directed graph $G$, (b) the graph $G_\text{C}$, (c-d) the corresponding transition matrices $\mat{P}$ and $\mat{P}_{\text{tel}}$, respectively, (e) the random surfer transition matrix $\mat{P}'$ for $\alpha=0.85$, with transition probabilities accurate to $2$ decimal places.}
    \label{fig:TeleportingGraphs}
\end{figure}

In the case of uniform teleportation, it is straightforward to specify the teleportation probabilities since they are all equal to $\frac{1}{N}$. Therefore, \cref{eq:GoogMat1} becomes:

\begin{equation}
\label{eq:GoogMat2}
    \mat{P}'=\alpha\mat{P}+(1-\alpha)\frac{\mathbf{1}}{N}
\end{equation}
where $\mathbf{1}\in\mathbb{R}^{N\times N}$ denotes a matrix of ones. While $\mat{P}$ represents the random walk on $G$, we can interpret $\frac{\mathbf{1}}{N}$ as the transition matrix of a random walk on a graph $G_\text{C}$ that has the same number of vertices as $G$ but where each vertex $v_i$ is connected to all others including itself.\footnote{Sometimes objects like $G_\text{C}$ are referred to as \textit{complete graphs} or \textit{fully connected networks}. However, such terms typically do not include the possibility of self-loops, which we by definition need since we consider uniform teleportation.} An example is shown in \cref{fig:TeleportingGraphs}, with (a) showing the starting graph $G$ and (b) showing $G_\text{C}$. In (c, d) we show the transition matrices $\mat{P}$ and $\mat{P}_{\text{tel}}$ of the random walks on $G$ and $G_\text{C}$, respectively. Finally, in (e) we show the transition matrix of the overall teleporting random walk with $\alpha=0.85$ is shown.

Due to the teleportation term $\mat{P}_{\text{tel}}$, from any given state $s_i$ there is always a non-zero probability to access any other state, or to stay in the same state. By virtue of this, such processes are guaranteed to be both irreducible and aperiodic, and therefore ergodic.

\subsection{Summary}
This concludes our treatment of random walks. The material presented in this section forms a useful framework that connects Markov chains and graphs. On the one hand, describing a Markov chain as a process taking place on a graph is a useful interpretation since it provides intuition about underlying relationships between states. Furthermore, it allows one to apply the toolkit of spectral graph theory to Markov chains. On the other hand, graphs by themselves represent only static relationships between entities, and performing a random walk is one way to describe a graph in terms that are dynamic/temporal. Moreover, the fact that a transition matrix can be easily exponentiated, i.e $\mat{P}^k$, means that a random walk provides information about a graph $G$ at multiple time scales, which is a property that has been exploited in the field of manifold learning \cite{Coifman2005,Coifman2006}. As a final note, in the current section many concepts and results from linear algebra are required, for which we recommend \cite{Meyer2000} as a general resource and \cite{Stewart1994} as a more specific summary of the application to transition matrices.
\section{Conclusion}
The key motivation of this tutorial is to provide a single introductory text on the spectral theory of Markov chains. By bringing together concepts and results from different areas of mathematics, this work is a useful resource for readers hoping to gain a broad, yet concise, overview of this topic. Since we only assume minimal exposure to concepts from linear algebra and probability theory, and since focus is placed on providing intuition rather than rigorous results, the material of this tutorial is accessible to researchers and students in a variety of quantitative disciplines. For those working in fields related to machine learning and data mining, this work is particularly relevant due to the applications discussed at various points. Although the material mostly consists of known results, two novel contributions are the categorization of eigenvalues given in \cref{tab:Eigs} and \cref{fig:EigsUnitCircle}, as well as the notion of random walk sets (\cref{defin:RWset}). 

Our presentation involved two different paradigms for interpreting and analyzing Markov chains. In \cref{MCs} we presented a categorization based on the transition structure and asymptotic behavior, and in \cref{Graphs} we instead formalized the idea of a Markov chain as a type of graph. In \cref{RWs}, we connected these two perspectives by introducing the idea of a random walk, and in doing so provided a number of parallels between some categories of Markov chains and certain types of graphs. In particular, one theme that aligns the two perspectives is the distinction between reversible/non-reversible Markov chains on the one hand, and undirected/directed graphs on the other, where in both cases the former option is easier to treat than the latter. With the additional use of results from linear algebra, this provided us with an in-depth description of the eigenvalues and eigenvectors of transition matrices in the reversible case. Finally, we discussed various attempts that have been made to generalize spectral methods to the non-reversible case.

\section{Acknowledgements}
We would like to thank Jonathan Hermon for some useful discussions on Markov chain theory, as well as Josu\'e Tonelli‑Cueto for his insight on the Perron-Frobenius theorem and various concepts in linear algebra.

\begin{appendices}
\label{App:Proofs}
\section{Proofs}
\addtocontents{toc}{\protect\setcounter{tocdepth}{0}}
\label{AppendixA}
\subsection{\cref{prop:RevRedRec}}
\begin{proof}
Assume that $\mat{P}$ is the transition matrix of a reducible recurrent Markov chain with $r$ communicating classes. Therefore, for a suitable indexing of states in $\mathcal{S}$ the transition matrix of this chain has a block diagonal form:
\begin{equation}
\label{eq:Pblock}
\mat{P}=\left(
    \begin{array}{cccc}
       \fbox{$\mat{P}_1$} & 0 & \cdots & 0 \\
       0 & \fbox{$\mat{P}_2$} & \cdots & 0 \\
       \vdots & \vdots & \ddots & \vdots \\
       0 & 0 & \cdots & \fbox{$\mat{P}_r$}
    \end{array}
    \right)
\end{equation}
where $\mat{P}_k$ is a transition matrix composed of the transition probabilities for states in the $k$-th class. Furthermore, if $\vec{\pi}>0$ is one of its stationary distributions, it can be written as $\vec{\pi}=\sum_{k=1}^r \alpha_k \vec{\pi}_k$ where each $\vec{\pi}_k$ has non-zero entries only in the $k$-th class and $\alpha_k>0$ (\cref{prop:RedSD}). Thus, given the same indexing of states the matrix $\mat{\Pi}$ and its inverse $\mat{\Pi}^{-1}$ have the following forms:
\begin{equation}
\label{eq:SDblock}
\mat{\Pi}=\left(
    \begin{array}{cccc}
       \fbox{$\alpha_1\mat{\Pi}_1$} & 0 & \cdots & 0 \\
       0 & \fbox{$\alpha_2\mat{\Pi}_2$} & \cdots & 0 \\
       \vdots & \vdots & \ddots & \vdots \\
       0 & 0 & \cdots & \fbox{$\alpha_r\mat{\Pi}_r$}
    \end{array}
    \right)\quad
\mat{\Pi}^{-1}=\left(
    \begin{array}{cccc}
       \fbox{$(\alpha_1\mat{\Pi}_1)^{-1}$} & 0 & \cdots & 0 \\
       0 & \fbox{$(\alpha_2\mat{\Pi}_2)^{-1}$} & \cdots & 0 \\
       \vdots & \vdots & \ddots & \vdots \\
       0 & 0 & \cdots & \fbox{$(\alpha_1\mat{\Pi}_r)^{-1}$}
    \end{array}
    \right) 
\end{equation}
where $\mat{\Pi}_k$ is the diagonal block matrix formed from the non-zero entries of $\vec{\pi}_k$. Hence, because $\mat{P}_{\textup{rev}}$ is a product of three block diagonal matrices, it too has the same form:
\begin{align}
\label{eq:Prevblock}
\mat{P}_{\textup{rev}}&=\mat{\Pi}^{-1}\mat{P^T}\mat{\Pi}\\
&=\left(
    \begin{array}{cccc}
       \fbox{$\mat{P}_{\textup{rev},1}$} & 0 & \cdots & 0 \\
       0 & \fbox{$\mat{P}_{\textup{rev},2}$} & \cdots & 0 \\
       \vdots & \vdots & \ddots & \vdots \\
       0 & 0 & \cdots & \fbox{$\mat{P}_{\textup{rev},r}$}
    \end{array}
    \right)
\end{align}
where $\mat{P}_{\textup{rev},k}$ contains the transition probabilities of states in the $k$-th class for the time reversed Markov chain. Furthermore, we can easily evaluate these block matrices:
\begin{align}
\label{eq:Prevblocksingle}
\mat{P}_{\textup{rev},k}&=(\alpha_k\mat{\Pi}_k)^{-1}(\mat{P}_k)^T(\alpha_k\mat{\Pi}_k)\\
&=\frac{\alpha_k}{\alpha_k}(\mat{\Pi}_k)^{-1}(\mat{P}_k)^T(\mat{\Pi}_k)\\
&=(\mat{\Pi}_k)^{-1}(\mat{P}_k)^T(\mat{\Pi}_k)
\end{align}
Thus, the matrix $\mat{P}_{\textup{rev}}$ does not depend on the $\alpha_k$ terms that parameterize the stationary distribution, meaning that the time reversed Markov chain the same regardless of which distribution is considered.
\end{proof}

\subsection{\cref{prop:SDRev}}
\begin{proof}
Assume that $\mat{P}$ is the transition matrix of a recurrent Markov chain and $\vec{\pi}>0$ is one of its stationary distributions. First note that summing over the $j$-th row or column of the flow matrix $\mat{F}^{\vec{\pi}}$ gives the same result:
\begin{align}
    \sum_{i=1}^N (\mat{F}^{\vec{\pi}})_{ji}&=\sum_{i=1}^N\pi_jP_{ji}=\pi_j\sum_{i=1}^NP_{ji}=\pi_j\\
    \sum_{i=1}^N (\mat{F}^{\vec{\pi}})_{ij}&=\sum_{i=1}^N \pi_iP_{ij}\overset{(\ref{eq:GB2})}{=}\pi_j\\
\end{align}
which in vector notation can be written as $\vec{\pi}^T=\vec{1}^T\mat{F}^{\vec{\pi}}=\vec{1}^T(\mat{F}^{\vec{\pi}})^T$, where the symbol $\vec{1}$ denotes a vector of ones. Using this we see that:
\begin{align}
    & & \vec{\pi}^T&=\vec{\pi}^T\mat{P}\\
    \Longleftrightarrow& & &=\vec{1}^T\mat{\Pi}\mat{P}\\
    \Longleftrightarrow& & &=\vec{1}^T\mat{F}^{\vec{\pi}}\\
    \Longleftrightarrow& & &=\vec{1}^T(\mat{F}^{\vec{\pi}})^T\\
    \Longleftrightarrow& & &=\vec{1}^T\mat{P}^T\mat{\Pi}\\
    \Longleftrightarrow& & &=\vec{1}^T\mat{\Pi}\mat{\Pi}^{-1}\mat{P}^T\mat{\Pi} \label{eq:SDrev}\\
    \Longleftrightarrow& & &=\vec{\pi}^T\mat{\Pi}^{-1}\mat{P}^T\mat{\Pi}\\
    \Longleftrightarrow& & &=\vec{\pi}^T\mat{P}_{\text{rev}}
\end{align}
Therefore, $\vec{\pi}$ is a stationary distribution of $\mathcal{X}$ if and only if it is a stationary distribution of $\tilde{\mathcal{X}}$.
\end{proof}

\subsection{\cref{thm:RWequivClassRec}}
\begin{proof}
($\Rightarrow$) Assume that $\mat{P}$ is the transition matrix of a recurrent Markov chain. Then, for any stationary distribution $\vec{\pi}>0$ the flow matrix $\mat{F}^{\vec{\pi}}=\mat{\Pi P}$ corresponds to one of the allowed graphs in $\mathcal{RW}(\mathcal{X})$. Therefore, using the same argument given in the proof of \cref{prop:SDRev}, the row and column sums of $\mat{F}^{\vec{\pi}}$ are the same, meaning that this matrix describes a balanced graph.\\
\linebreak
($\Leftarrow$) We adapt this direction of the proof from \cite{Banderier2000}, where the unweighted case is considered. Assume that $G$ is a balanced graph with weight matrix $\mat{W}$, and that $\mathcal{X}$ is the Markov chain realised by a random walk on this graph. Now consider the distribution $\vec{\pi}=\frac{1}{z}(d_1, d_2, ..., d_N)^T$, where $z=\sum_{j=1}^N d_j=\text{vol}(G)$ is needed in order for $\vec{\pi}$ to sum to $1$. We can then easily verify that the equations of global balance hold for this distribution (\cref{eq:GB2}):
\begin{equation}
    \sum_{i=1}^N \pi_iP_{ij}\overset{(\ref{eq:RWmat})}{=}\sum_{i=1}^N \pi_{i} \frac{W_{ij}}{d_i}=\sum_{i=1}^N \frac{d_i}{z}\frac{W_{ij}}{d_i}=\sum_{i=1}^N\frac{W_{ij}}{z}=\frac{1}{z}\underbrace{\sum_{i=1}^NW_{ij}}_{=d_j^-}=\frac{d_j}{z}=\pi_j
\end{equation}
and so $\vec{\pi}^T\mat{P}=\vec{\pi}^T$ which means that $\vec{\pi}$ is a stationary distribution of the chain. Note that the summation in the fourth expression defines the in-degree of vertex $v_j$, but since the graph $G$ is balanced we only have one degree $d_j$ associated to each vertex. Lastly, since isolated vertices are not allowed, each degree must be bigger than zero, meaning similarly that $\pi_i>0$ $\forall s_i\in\mathcal{S}$. Hence, there are no transient states, and $\mathcal{X}$ is recurrent.
\end{proof}

\subsection{\cref{thm:RWequivClassRev}}
\begin{proof}
Assume that $\mathcal{X}$ is a recurrent Markov chain and $G$ is one of the balanced graphs in $\mathcal{RW}(\mathcal{X})$. Using the same argument from the second part of the proof of \cref{thm:RWequivClassRec}, the degrees of $G$ are related to a stationary distribution $\vec{\pi}>0$ of the chain via $\pi_i=\frac{d_i}{z}$. From \cref{thm:RevChain} we know that $\mathcal{X}$ is reversible if and only if the flow matrix associated $\vec{\pi}$ is symmetric, and it is straightforward to show that this equivalent to $G$ being undirected:
\begin{align}
    & & \mat{F}^{\vec{\pi}}&=(\mat{F}^{\vec{\pi}})^T\\
    \Longleftrightarrow & &\mat{\Pi}\mat{P}&=\mat{P}^T\mat{\Pi}\\
    \Longleftrightarrow & &\mat{\Pi}\mat{D}^{-1}\mat{W}&=\mat{W}^T\mat{D}^{-1}\mat{\Pi}\\
    \Longleftrightarrow & &\frac{1}{z}\mat{D}\mat{D}^{-1}\mat{W}&=\mat{W}^T\mat{D}^{-1}\frac{1}{z}\mat{D}\\
    \Longleftrightarrow & &\mat{W}&=\mat{W}^T
\end{align}
and since $G$ was chosen arbitrarily, this applies to any balanced graph in $\mathcal{RW}(\mathcal{X})$.
\end{proof}

\subsection{\cref{thm:RWequivClassRecSD}}
\begin{proof}
See the second part of the proof of \cref{thm:RWequivClassRec}.
\end{proof}

\subsection{\cref{thm:Rev-selfadj}}
\begin{proof}
($\Rightarrow$) Let $\vec{x},\vec{x}'\in\mathbb{R}^n$ be two arbitrary vectors. Then, if $\mat{P}$ is the transition matrix of a reversible chain and $\vec{\pi}>0$ is one of its stationary distributions, it is guaranteed that $\mat{\Pi}\mat{P}=\mat{P}^T\mat{\Pi}$ (\cref{thm:RevChain}). Using this, we can show that $\mat{P}$ satisfies \cref{eq:Pselfadj}:
\begin{align}
    \langle \vec{x}, \mat{P}\vec{x}'\rangle_{\mat{\Pi}}\overset{(\ref{eq: SDip})}&{=}\vec{x}^T\mat{\Pi}\mat{P}\vec{x}'\\
    &=\vec{x}^T\mat{P}^T \mat{\Pi} \vec{x}'\\
    &=(\mat{P}\vec{x})^T\mat{\Pi} \vec{x}'\\
    \overset{(\ref{eq: SDip})}&{=}\langle \mat{P}\vec{x}, \vec{x}'\rangle_{\mat{\Pi}}
\end{align}\\
\linebreak
($\Leftarrow$) Let $\mat{P}$ be the transition matrix of a reversible Markov chain, and $\vec{\pi}>0$ one of its stationary distributions. \cref{eq:Pselfadj} can be written as:
\begin{equation}
    \vec{x}^T\mat{\Pi}\mat{P}\vec{x}'=\vec{x}^T\mat{P}^T\mat{\Pi}\vec{x}\quad \forall \vec{x},\vec{x}'\in\mathbb{R}^n
\end{equation}
If we choose $\vec{x}=\vec{e}_i$ to be a vector with $i$-th entry equal to $1$ and zeros elsewhere, this reduces to:
\begin{align}
    \sum_{j'}\pi_iP_{ij'}x'_{j'}&=\sum_{j''}(\mat{P}^T)_{ij''}\pi_{j''}x'_{j''}\\
    \pi_i\sum_{j'}P_{ij'}x'_{j'}&=\sum_{j''}P_{j''i}\pi_{j''}x'_{j''}
\end{align}
If we then choose $\vec{x}'=\vec{e}_j$ to be a vector with $j$-th entry equal to $1$ and zeros elsewhere, we get:
\begin{equation}
\label{eq:DBfromSelfAdj}
    \pi_iP_{ij}=\pi_jP_{ji}
\end{equation}
Since the indices $i$ and $j$ were chosen arbitrarily, \cref{eq:DBfromSelfAdj} between any pair of states. Therefore, the chain is reversible (\cref{thm:RevChain}).
\end{proof}

\subsection{\cref{thm:KmatOrthog}}
\begin{proof}
Let $\mathcal{X}$ be a reversible Markov chain with transition matrix $\mat{P}$ and a stationary distribution $\vec{\pi}>0$, and let $\mat{K}=\mat{\Pi}^{\frac{1}{2}}\mat{P} \mat{\Pi}^{-\frac{1}{2}}$. Then the $(i,j)$-th element of $\mat{K}$ is:
\begin{align}
    K_{ij}&=\hspace{3pt}\sqrt{\pi_{i}}P_{ij}\frac{1}{\sqrt{\pi_j}}\\
    \overset{(\ref{eq:DB})}&{=}\sqrt{\pi_i} P_{ji}\frac{\pi_j}{\pi_i}\frac{1}{\sqrt{\pi_j}}\\
    &=\hspace{3pt}\sqrt{\pi_j}P_{ji}\frac{1}{\sqrt{\pi_i}}\\
    &=\hspace{3pt}K_{ji}
\end{align}
meaning that $\mat{K}$ is a symmetric matrix. Note that in the third line we made use of detailed balance which holds for $\vec{\pi}>0$ if and only if $\mathcal{X}$ is reversible. Therefore, this concludes both directions of the proof.

In order to establish the uniqueness of $\mat{K}$, we make use of the fact that a reversible Markov chain must be recurrent, meaning that any stationary distribution $\vec{\pi}>0$ is of the form $\vec{\pi}=\sum_{k=1}^r \alpha_k \vec{\pi}_k$ with $\alpha_k>0$ (\cref{prop:RedSD}). Since all entries of $\vec{\pi}$ are greater than zero, $K_{ij}=\sqrt{\pi_i}P_{ij}\frac{1}{\sqrt{\pi_j}}\neq 0$ iff $P_{ij}\neq 0$. For a recurrent chain, the latter can only be true for pairs of states in the same communicating class. If $s_i$ and $s_j$ belong to the $k$-th communicating class, then:
\begin{align}
    K_{ij}&=\sqrt{\pi_{i}}P_{ij}\frac{1}{\sqrt{\pi_j}}\\
    &=\sqrt{\alpha_k\pi_{k,i}}P_{ij}\frac{1}{\sqrt{\alpha_k\pi_{k,j}}}\\
    &=\sqrt{\pi_{k,i}}P_{ij}\frac{1}{\sqrt{\pi_{k,j}}}
\end{align}
where $\pi_{k,i}$ denotes the $i$-th component of the stationary distribution associated to the $k$-th class. Hence, the non-zero entries of $\mat{K}$ do not depend on the values of $\alpha_k$, i.e.\ they are irrespective of the stationary distribution used.
\end{proof}

\subsection{\cref{thm:Rev-Eig}}
\begin{proof}
Let $\mathcal{X}$ be a reversible Markov chain with transition matrix $\mat{P}$. The proof of \cref{thm:KmatOrthog} establishes that this is equivalent to $\mat{K}=\mat{\Pi}^{\frac{1}{2}}\mat{P} \mat{\Pi}^{-\frac{1}{2}}$ being symmetric, where $\vec{\pi}>0$ is a stationary distribution of the chain. By virtue of \cref{thm:RealSym2}, this in turn is equivalent to the existence of an orthogonal matrix $\mat{Y}$ for which:
\begin{align}
    \mat{D}&=\mat{Y}^T\mat{K}\mat{Y}\\
    &=\mat{Y}^T\mat{\Pi}^{\frac{1}{2}}\mat{P}\mat{\Pi}^{-\frac{1}{2}}\mat{Y} \label{eq:Peigdecomp}
\end{align}
\cref{eq:Peigdecomp} says that $\mat{P}$ has the same eigenvalues as $\mat{K}$, which are real. It also tells us that $\mat{P}$ is diagonalizable by sets of right and left eigenvectors that are related to the eigenvectors of $\mat{K}$ by $\mat{\Pi}^{-\frac{1}{2}}$ and $\mat{\Pi}^{\frac{1}{2}}$, respectively. Hence, if $\vec{y}_\omega$ is an eigenvector of $\mat{K}$ with eigenvalue $\lambda_\omega$, then $\vec{r}_\omega=\mat{\Pi}^{-\frac{1}{2}}\vec{y}_\omega$ and $\vec{l}_\omega=\mat{\Pi}^{\frac{1}{2}}\vec{y}_\omega$ are a pair of corresponding right and left eigenvectors of $\mat{P}$ with the same eigenvalue. Using this we see that if $\vec{r}_\omega$ and $\vec{r}_\gamma$ are a pair of right eigenvectors of $\mat{P}$, then:
\begin{equation}
    \langle \vec{r}_\omega , \vec{r}_\gamma \rangle_{\mat{\Pi}}=\langle \mat{\Pi}^{-\frac{1}{2}} \vec{y}_\omega , \mat{\Pi}^{-\frac{1}{2}} \vec{y}_\gamma \rangle_{\mat{\Pi}}=\vec{y}_\omega^T\underbrace{\mat{\Pi}^{-\frac{1}{2}}\mat{\Pi}\mat{\Pi}^{-\frac{1}{2}}}_{=\mathbbm{1}} \vec{y}_\gamma=\vec{y}_\omega^T\vec{y}_\gamma=\delta_{\omega\gamma}
\end{equation}
where we have used the fact that the basis $\mat{Y}$ is orthonormal. Similarly, if $\vec{l}_\omega$ and $\vec{l}_\gamma$ are a pair of left eigenvectors of $\mat{P}$,  then:
\begin{equation}
    \langle \vec{l}_\omega , \vec{l}_\gamma \rangle_{\mat{\Pi}^{-1}}=\langle \mat{\Pi}^{\frac{1}{2}} \vec{y}_\omega , \mat{\Pi}^{\frac{1}{2}} \vec{y}_\gamma \rangle_{\mat{\Pi}^{-1}}=\vec{y}_\omega^T\underbrace{\mat{\Pi}^{\frac{1}{2}}\mat{\Pi}^{-1}\mat{\Pi}^{\frac{1}{2}}}_{=\mathbbm{1}} \vec{y}_\gamma=\delta_{\omega\gamma} \tag*{\qedhere} 
\end{equation}
\end{proof}

\subsection{\cref{prop:Rev-LR}}
\begin{proof}
From the proof of \cref{thm:Rev-Eig}, we know that if $\mat{Y}$ is an orthonormal basis of $\mat{K}$ and $\vec{y}_\omega$ is a basis vector with eigenvalue $\lambda_\omega$, then $\vec{r}_\omega=\mat{\Pi}^{-\frac{1}{2}}\vec{y}_\omega$ and $\vec{l}_\omega=\mat{\Pi}^{\frac{1}{2}}\vec{y}_\omega$ are right end left eigenvectors of $\mat{P}$, respectively, with the same eigenvalue. Therefore:
\begin{equation}
    \vec{l}_\omega=\mat{\Pi}^{\frac{1}{2}}(\mat{\Pi}^{\frac{1}{2}}\vec{r}_\omega)=\mat{\Pi}\vec{r}_\omega
\end{equation}
\end{proof}

\subsection{\cref{thm:NLapQuad}}
\begin{proof}
Assume that $\mat{\mathcal{L}}$ is the normalized Laplacian of an undirected graph $G$ and $\vec{x}\in\mathbb{R}^N$. Then:
\begin{align}
    \vec{x}^T\mat{\mathcal{L}}\vec{x}\overset{(\ref{eq:NLapMat})}&{=}\vec{x}^T\vec{x}-\vec{x}^T\mat{D}^{-\frac{1}{2}}\mat{WD}^{-\frac{1}{2}}\vec{x}\\
    &=\bigg(\sum_{i=1}^N x_i^2\bigg) - \bigg(\sum_{i,j=1}^NW_{ij}\frac{x_ix_j}{d_i^{\frac{1}{2}}d_j^{\frac{1}{2}}}\bigg)\\
    &=\frac{1}{2}\bigg(\sum_{i=1}^N x_i^2\bigg) + \frac{1}{2}\bigg(\sum_{j=1}^N x_j^2\bigg) - 2\bigg(\sum_{i,j=1}^NW_{ij}\frac{x_ix_j}{d_i^{\frac{1}{2}}d_j^{\frac{1}{2}}}\bigg)\\
    &=\frac{1}{2}\bigg(\sum_{i=1}^N x_i^2\underbrace{\frac{d_i}{d_i}}_{=1}\bigg)+ \frac{1}{2}\bigg(\sum_{j=1}^N x_j^2\underbrace{\frac{d_j}{d_j}}_{=1}\bigg) - 2\bigg(\sum_{i,j=1}^NW_{ij}\frac{x_ix_j}{d_i^{\frac{1}{2}}d_j^{\frac{1}{2}}}\bigg)\\
    &=\frac{1}{2}\bigg(\sum_{i=1}^N x_i^2\frac{\sum_{j=1}^N W_{ij}}{d_i}\bigg) + \frac{1}{2}\bigg(\sum_{j=1}^N x_j^2\frac{\sum_{i=1}^N W_{ji}}{d_j}\bigg) - 2\bigg(\sum_{i,j=1}^NW_{ij}\frac{x_ix_j}{d_i^{\frac{1}{2}}d_j^{\frac{1}{2}}}\bigg)\\
    &=\frac{1}{2}\bigg(\sum_{i,j=1}^N x_i^2\frac{W_{ij}}{d_i}\bigg)+\frac{1}{2}\bigg(\sum_{i,j=1}^N x_j^2\frac{W_{ji}}{d_j}\bigg)- 2\bigg(\sum_{i,j=1}^NW_{ij}\frac{x_ix_j}{d_i^{\frac{1}{2}}d_j^{\frac{1}{2}}}\bigg)\\
    &=\frac{1}{2}\sum_{i,j=1}^N\bigg(x_i\bigg(\frac{W_{ij}}{d_i}\bigg)^{\frac{1}{2}}-x_j\bigg(\frac{W_{ji}}{d_j}\bigg)^{\frac{1}{2}}\bigg)^2\\
    &=\frac{1}{2}\sum_{i,j=1}^NW_{ij}\bigg(\frac{x_i}{d_i^{\frac{1}{2}}}-\frac{x_j}{d_j^{\frac{1}{2}}}\bigg)^2\hspace{40pt}\text{($W_{ij}=W_{ji}$ since $G$ is undirected)}\end{align}
\end{proof}
\end{appendices}

\bibliographystyle{unsrt}
\bibliography{eigen}

\begin{thebibliography}{10}

\bibitem{Pardoux2010}
E~Pardoux.
\newblock {\em Markov Processes and Applications: Algorithms, Networks, Genome
  and Finance}.
\newblock {Wiley/Dunod}, {Chichester, U.K; Hoboken, NJ}, 2010.

\bibitem{Aggarwal2015}
Charu~C. Aggarwal.
\newblock {\em Data {{Mining}}}.
\newblock {Springer International Publishing}, {Cham}, 2015.

\bibitem{Belkin2001}
Mikhail Belkin and Partha Niyogi.
\newblock Laplacian eigenmaps and spectral techniques for embedding and
  clustering.
\newblock In {\em Advances in Neural Information Processing Systems 14}, pages
  585--591. {MIT Press}, 2001.

\bibitem{Belkin2003}
Mikhail Belkin and Partha Niyogi.
\newblock Laplacian {{Eigenmaps}} for {{Dimensionality Reduction}} and {{Data
  Representation}}.
\newblock {\em Neural Computation}, 15(6):1373--1396, June 2003.

\bibitem{Weiss1999}
Y.~Weiss.
\newblock Segmentation using eigenvectors: A unifying view.
\newblock In {\em Proceedings of the {{Seventh IEEE International Conference}}
  on {{Computer Vision}}}, pages 975--982 vol.2, {Kerkyra, Greece}, 1999.
  {IEEE}.

\bibitem{Ng2001}
Andrew~Y. Ng, Michael~I. Jordan, and Yair Weiss.
\newblock On spectral clustering: {{Analysis}} and an algorithm.
\newblock In {\em Proceedings of the 14th International Conference on Neural
  Information Processing Systems: {{Natural}} and Synthetic}, {{NIPS}}'01,
  pages 849--856, {Cambridge, MA, USA}, 2001. {MIT Press}.

\bibitem{vonLuxburg2007}
Ulrike {von Luxburg}.
\newblock A tutorial on spectral clustering.
\newblock {\em Statistics and Computing}, 17(4):395--416, December 2007.

\bibitem{Meila2000}
Marina Meila and Jianbo Shi.
\newblock Learning segmentation by random walks.
\newblock In T.~Leen, T.~Dietterich, and V.~Tresp, editors, {\em Advances in
  Neural Information Processing Systems}, volume~13. {MIT Press}, 2000.

\bibitem{Meila2001}
Marina Meila and Jianbo Shi.
\newblock A random walks view of spectral segmentation.
\newblock 2001.

\bibitem{Tishby2001}
Naftali Tishby and Noam Slonim.
\newblock Data clustering by markovian relaxation and the information
  bottleneck method.
\newblock {\em Advances in Neural Information Processing Systems 13: 2000;
  Denver}, November 2001.

\bibitem{Saerens2004}
Marco Saerens, Francois Fouss, Luh Yen, and Pierre Dupont.
\newblock The {{Principal Components Analysis}} of a {{Graph}}, and {{Its
  Relationships}} to {{Spectral Clustering}}.
\newblock In {\em Machine {{Learning}}: {{ECML}} 2004}, volume 3201, pages
  371--383. {Springer Berlin Heidelberg}, {Berlin, Heidelberg}, 2004.

\bibitem{Liu2011}
Ning Liu.
\newblock Markov {{Chains}} and {{Spectral Clustering}}.
\newblock In {\em Performance {{Evaluation}} of {{Computer}} and
  {{Communication Systems}}. {{Milestones}} and {{Future Challenges}}}, volume
  6821, pages 87--98. {Springer Berlin Heidelberg}, {Berlin, Heidelberg}, 2011.

\bibitem{Meila2007}
Marina Meil{\u a} and William Pentney.
\newblock Clustering by weighted cuts in directed graphs.
\newblock In {\em Proceedings of the 2007 {{SIAM International Conference}} on
  {{Data Mining}}}, pages 135--144. {Society for Industrial and Applied
  Mathematics}, April 2007.

\bibitem{Huang2006}
Jiayuan Huang, Tingshao Zhu, and Dale Schuurmans.
\newblock Web {{Communities Identification}} from {{Random Walks}}.
\newblock In {\em Knowledge {{Discovery}} in {{Databases}}: {{PKDD}} 2006},
  volume 4213, pages 187--198. {Springer Berlin Heidelberg}, {Berlin,
  Heidelberg}, 2006.

\bibitem{Coifman2005}
R.~R. Coifman, S.~Lafon, A.~B. Lee, M.~Maggioni, B.~Nadler, F.~Warner, and
  S.~W. Zucker.
\newblock Geometric diffusions as a tool for harmonic analysis and structure
  definition of data: {{Diffusion}} maps.
\newblock {\em Proceedings of the National Academy of Sciences},
  102(21):7426--7431, May 2005.

\bibitem{Coifman2006}
Ronald~R. Coifman and St{\'e}phane Lafon.
\newblock Diffusion maps.
\newblock {\em Applied and Computational Harmonic Analysis}, 21(1):5--30, July
  2006.

\bibitem{Ghojogh2021}
Benyamin Ghojogh, Ali Ghodsi, Fakhri Karray, and Mark Crowley.
\newblock Laplacian-based dimensionality reduction including spectral
  clustering, laplacian eigenmap, locality preserving projection, graph
  embedding, and diffusion map: {{Tutorial}} and survey, June 2021.

\bibitem{Wiskott2002}
Laurenz Wiskott and Terrence~J. Sejnowski.
\newblock Slow {{Feature Analysis}}: {{Unsupervised Learning}} of
  {{Invariances}}.
\newblock {\em Neural Computation}, 14(4):715--770, April 2002.

\bibitem{Sprekeler2011}
Henning Sprekeler.
\newblock On the {{Relation}} of {{Slow Feature Analysis}} and {{Laplacian
  Eigenmaps}}.
\newblock {\em Neural Computation}, 23(12):3287--3302, December 2011.

\bibitem{Kamvar2003}
Sepandar~D. Kamvar, Dan Klein, and Christopher~D. Manning.
\newblock Spectral learning.
\newblock In {\em Proceedings of the 18th International Joint Conference on
  Artificial Intelligence}, {{IJCAI}}'03, pages 561--566, {San Francisco, CA,
  USA}, 2003. {Morgan Kaufmann Publishers Inc.}

\bibitem{Szummer2001}
Martin Szummer and Tommi Jaakkola.
\newblock Partially labeled classification with markov random walks.
\newblock In {\em Proceedings of the 14th International Conference on Neural
  Information Processing Systems: {{Natural}} and Synthetic}, {{NIPS}}'01,
  pages 945--952, {Cambridge, MA, USA}, 2001. {MIT Press}.

\bibitem{Joachims2003}
Thorsten Joachims.
\newblock Transductive learning via spectral graph partitioning.
\newblock In {\em Proceedings of the Twentieth International Conference on
  International Conference on Machine Learning}, {{ICML}}'03, pages 290--297,
  {Washington, DC, USA}, 2003. {AAAI Press}.

\bibitem{Zhou2005}
Dengyong Zhou, Jiayuan Huang, and Bernhard Sch{\"o}lkopf.
\newblock Learning from labeled and unlabeled data on a directed graph.
\newblock In {\em Proceedings of the 22nd International Conference on
  {{Machine}} Learning - {{ICML}} '05}, pages 1036--1043, {Bonn, Germany},
  2005. {ACM Press}.

\bibitem{Mahadevan2005a}
Sridhar Mahadevan.
\newblock Proto-value functions: Developmental reinforcement learning.
\newblock In Luc~De Raedt and Stefan Wrobel, editors, {\em Machine Learning,
  Proceedings of the Twenty-Second International Conference ({{ICML}} 2005),
  Bonn, Germany, August 7-11, 2005}, volume 119 of {\em {{ACM}} International
  Conference Proceeding Series}, pages 553--560. {ACM}, 2005.

\bibitem{Mahadevan2007}
Sridhar Mahadevan and Mauro Maggioni.
\newblock Proto-value functions: {{A}} laplacian framework for learning
  representation and control in markov decision processes.
\newblock {\em Journal of Machine Learning Research}, 8:2169--2231, December
  2007.

\bibitem{Johns2007}
Jeff Johns and Sridhar Mahadevan.
\newblock Constructing basis functions from directed graphs for value function
  approximation.
\newblock In {\em Proceedings of the 24th International Conference on
  {{Machine}} Learning - {{ICML}} '07}, pages 385--392, {Corvalis, Oregon},
  2007. {ACM Press}.

\bibitem{Stachenfeld2014}
Kimberly~L Stachenfeld, Matthew Botvinick, and Samuel~J Gershman.
\newblock Design principles of the hippocampal cognitive map.
\newblock In {\em Advances in Neural Information Processing Systems},
  volume~27. {Curran Associates, Inc.}, 2014.

\bibitem{Stachenfeld2017}
Kimberly~L Stachenfeld, Matthew~M Botvinick, and Samuel~J Gershman.
\newblock The hippocampus as a predictive map.
\newblock {\em Nature Neuroscience}, 20(11):1643--1653, November 2017.

\bibitem{Petrik2007}
Marek Petrik.
\newblock An analysis of laplacian methods for value function approximation in
  {{MDPs}}.
\newblock In {\em Proceedings of the 20th International Joint Conference on
  Artifical Intelligence}, {{IJCAI}}'07, pages 2574--2579, {San Francisco, CA,
  USA}, 2007. {Morgan Kaufmann Publishers Inc.}

\bibitem{Wu2019}
Yifan Wu, George Tucker, and Ofir Nachum.
\newblock The {{Laplacian}} in {{RL}}: {{Learning}} representations with
  efficient approximations.
\newblock In {\em 7th International Conference on Learning Representations,
  {{ICLR}} 2019, New Orleans, {{LA}}, {{USA}}, May 6-9, 2019}.
  {OpenReview.net}, 2019.

\bibitem{Stewart1994}
William~J. Stewart.
\newblock {\em Introduction to the Numerical Solution of {{Markov}} Chains}.
\newblock {Princeton University Press}, {Princeton, N.J}, 1994.

\bibitem{Sutton2018}
Richard~S. Sutton and Andrew~G. Barto.
\newblock {\em Reinforcement Learning: An Introduction}.
\newblock Adaptive Computation and Machine Learning Series. {The MIT Press},
  {Cambridge, Massachusetts}, second edition edition, 2018.

\bibitem{Denton2021}
Peter~B. Denton, Stephen~J. Parke, Terence Tao, and Xining Zhang.
\newblock Eigenvectors from eigenvalues: {{A}} survey of a basic identity in
  linear algebra.
\newblock {\em Bulletin of the American Mathematical Society}, 59(1):31--58,
  February 2021.

\bibitem{Conrad2016}
Nata{\v s}a~Djurdjevac Conrad, Marcus Weber, and Christof Sch{\"u}tte.
\newblock Finding {{Dominant Structures}} of {{Nonreversible Markov
  Processes}}.
\newblock {\em Multiscale Modeling \& Simulation}, 14(4):1319--1340, January
  2016.

\bibitem{Meyer2000}
C.~D. Meyer.
\newblock {\em Matrix Analysis and Applied Linear Algebra}.
\newblock {Society for Industrial and Applied Mathematics}, {Philadelphia},
  2000.

\bibitem{MatthewRichey2010}
{Matthew Richey}.
\newblock The {{Evolution}} of {{Markov Chain Monte Carlo Methods}}.
\newblock {\em The American Mathematical Monthly}, 117(5):383, 2010.

\bibitem{Porod2021}
Ursula Porod.
\newblock {\em Dynamics of {{Markov Chains}} for {{Undergraduates}}}.
\newblock {Preprint:
  https://www.math.northwestern.edu/documents/book-markov-chains.pdf}, February
  2021.

\bibitem{Bremaud1999}
Pierre Br{\'e}maud.
\newblock {\em Markov Chains: {{Gibbs}} Fields, {{Monte Carlo}} Simulation, and
  Queues}.
\newblock Number~31 in Texts in Applied Mathematics. {Springer}, {New York},
  1999.

\bibitem{Kolmogoroff1936}
A.~Kolmogoroff.
\newblock {Zur Theorie der Markoffschen Ketten}.
\newblock {\em Mathematische Annalen}, 112(1):155--160, December 1936.

\bibitem{Gorban2014}
A.N. Gorban.
\newblock Detailed balance in micro- and macrokinetics and
  micro-distinguishability of macro-processes.
\newblock {\em Results in Physics}, 4:142--147, 2014.

\bibitem{Jiang2004}
Da-Quan Jiang, Min Qian, and Min-Ping Qian.
\newblock {\em Mathematical {{Theory}} of {{Nonequilibrium Steady States}}},
  volume 1833 of {\em Lecture {{Notes}} in {{Mathematics}}}.
\newblock {Springer Berlin Heidelberg}, {Berlin, Heidelberg}, 2004.

\bibitem{Zhang2012}
Xue-Juan Zhang, Hong Qian, and Min Qian.
\newblock Stochastic theory of nonequilibrium steady states and its
  applications. {{Part I}}.
\newblock {\em Physics Reports}, 510(1-2):1--86, January 2012.

\bibitem{Ge2012}
Hao Ge, Min Qian, and Hong Qian.
\newblock Stochastic theory of nonequilibrium steady states. {{Part II}}:
  {{Applications}} in chemical biophysics.
\newblock {\em Physics Reports}, 510(3):87--118, January 2012.

\bibitem{Witzig2018}
Jakob Witzig, Isabel Beckenbach, Leon Eifler, Konstantin Fackeldey, Ambros
  Gleixner, Andreas Grever, and Marcus Weber.
\newblock Mixed-{{Integer Programming}} for {{Cycle Detection}} in
  {{Nonreversible Markov Processes}}.
\newblock {\em Multiscale Modeling \& Simulation}, 16(1):248--265, January
  2018.

\bibitem{Fill1991}
James~Allen Fill.
\newblock Eigenvalue {{Bounds}} on {{Convergence}} to {{Stationarity}} for
  {{Nonreversible Markov Chains}}, with an {{Application}} to the {{Exclusion
  Process}}.
\newblock {\em The Annals of Applied Probability}, 1(1), February 1991.

\bibitem{Dayan1993}
Peter Dayan.
\newblock Improving {{Generalization}} for {{Temporal Difference Learning}}:
  {{The Successor Representation}}.
\newblock {\em Neural Computation}, 5(4):613--624, July 1993.

\bibitem{Chapman2011}
Airlie Chapman and Mehran Mesbahi.
\newblock Advection on graphs.
\newblock In {\em {{IEEE Conference}} on {{Decision}} and {{Control}} and
  {{European Control Conference}}}, pages 1461--1466, {Orlando, FL, USA},
  December 2011. {IEEE}.

\bibitem{Banderier2000}
Cyril Banderier and Robert~P. Dobrow.
\newblock A {{Generalized Cover Time}} for {{Random Walks}} on {{Graphs}}.
\newblock In {\em Formal {{Power Series}} and {{Algebraic Combinatorics}}},
  pages 113--124. {Springer Berlin Heidelberg}, {Berlin, Heidelberg}, 2000.

\bibitem{Aldous2002}
David Aldous and James~Allen Fill.
\newblock Reversible markov chains and random walks on graphs.
\newblock 2002.

\bibitem{West2001}
Douglas~Brent West.
\newblock {\em Introduction to Graph Theory}.
\newblock {Prentice Hall}, {Upper Saddle River, N.J}, 2nd ed edition, 2001.

\bibitem{Chung1997}
Fan R.~K. Chung.
\newblock {\em Spectral Graph Theory}.
\newblock Number no. 92 in Regional Conference Series in Mathematics.
  {Published for the Conference Board of the mathematical sciences by the
  American Mathematical Society}, {Providence, R.I}, 1997.

\bibitem{Gebali2008}
Fayez Gebali.
\newblock {\em Periodic {{Markov Chains}}}, pages 1--39.
\newblock {Springer US}, {Boston, MA}, 2008.

\bibitem{Gobel1974}
F.~G{\"o}bel and A.A. Jagers.
\newblock Random walks on graphs.
\newblock {\em Stochastic Processes and their Applications}, 2(4):311--336,
  October 1974.

\bibitem{Lovasz93}
L{\'a}szl{\'o} Lov{\'a}sz.
\newblock Random walks on graphs: {{A}} survey, 1993.

\bibitem{Wiskott2019}
Laurenz Wiskott and Fabian Sch{\"o}nfeld.
\newblock Laplacian {{Matrix}} for {{Dimensionality Reduction}} and
  {{Clustering}}.
\newblock 2019.

\bibitem{Belkin2008}
Mikhail Belkin and Partha Niyogi.
\newblock Towards a theoretical foundation for {{Laplacian-based}} manifold
  methods.
\newblock {\em Journal of Computer and System Sciences}, 74(8):1289--1308,
  December 2008.

\bibitem{Hein2007}
M.~Hein, J-Y. Audibert, and U.~{von Luxburg}.
\newblock Graph laplacians and their convergence on random neighborhood graphs.
\newblock {\em Journal of Machine Learning Research}, 8:1325--1370, June 2007.

\bibitem{Reilly1982}
Robert~C. Reilly.
\newblock Mean {{Curvature}}, the {{Laplacian}}, and {{Soap Bubbles}}.
\newblock {\em The American Mathematical Monthly}, 89(3):180--198, March 1982.

\bibitem{Grebenkov2013}
D.~S. Grebenkov and B.-T. Nguyen.
\newblock Geometrical {{Structure}} of {{Laplacian Eigenfunctions}}.
\newblock {\em SIAM Review}, 55(4):601--667, January 2013.

\bibitem{Shuman2013}
D.~I. Shuman, S.~K. Narang, P.~Frossard, A.~Ortega, and P.~Vandergheynst.
\newblock The emerging field of signal processing on graphs: {{Extending}}
  high-dimensional data analysis to networks and other irregular domains.
\newblock {\em IEEE Signal Processing Magazine}, 30(3):83--98, May 2013.

\bibitem{Weber2017}
Marcus Weber.
\newblock Eigenvalues of non-reversible {{Markov}} chains - {{A}} case study.
\newblock Technical Report 17-13, {ZIB}, {Takustr. 7, 14195 Berlin}, 2017.

\bibitem{Meyn2008}
Sean Meyn, Gregory Hagen, George Mathew, and Andrzej Banasuk.
\newblock On complex spectra and metastability of {{Markov}} models.
\newblock In {\em 2008 47th {{IEEE Conference}} on {{Decision}} and
  {{Control}}}, pages 3835--3839, {Cancun, Mexico}, 2008. {IEEE}.

\bibitem{Andrieux2011}
David Andrieux.
\newblock Spectral signature of nonequilibrium conditions, 2011.

\bibitem{Mieghem2018}
P.~V. Mieghem.
\newblock Directed graphs and mysterious complex eigenvalues.
\newblock 2018.

\bibitem{Sevi2018}
Harry Sevi, Gabriel Rilling, and Pierre Borgnat.
\newblock Harmonic analysis on directed graphs and applications: From
  {{Fourier}} analysis to wavelets, November 2018.

\bibitem{Marques2020}
Antonio~G. Marques, Santiago Segarra, and Gonzalo Mateos.
\newblock Signal {{Processing}} on {{Directed Graphs}}: {{The Role}} of {{Edge
  Directionality When Processing}} and {{Learning From Network Data}}.
\newblock {\em IEEE Signal Processing Magazine}, 37(6):99--116, November 2020.

\bibitem{Golub2013}
Gene~H. Golub and Charles~F. Van~Loan.
\newblock {\em Matrix Computations}.
\newblock Johns {{Hopkins}} Studies in the Mathematical Sciences. {The Johns
  Hopkins University Press}, {Baltimore}, fourth edition edition, 2013.

\bibitem{Pauwelyn2021}
P.~J. Pauwelyn and M.~A. Guerry.
\newblock Perturbations of non-diagonalizable stochastic matrices with
  preservation of spectral properties.
\newblock {\em Linear and Multilinear Algebra}, pages 1--31, March 2021.

\bibitem{Fackeldey2018}
K.~Fackeldey, A.~Sikorski, and M.~Weber.
\newblock Spectral clustering for non-reversible {{Markov}} chains.
\newblock {\em Computational and Applied Mathematics}, 37(5):6376--6391,
  November 2018.

\bibitem{Ghosh2020}
Dibya Ghosh and Marc~G. Bellemare.
\newblock Representations for stable off-policy reinforcement learning.
\newblock In Hal~Daum{\'e} III and Aarti Singh, editors, {\em Proceedings of
  the 37th International Conference on Machine Learning}, volume 119 of {\em
  Proceedings of Machine Learning Research}, pages 3556--3565. {PMLR}, July
  2020.

\bibitem{Ng1987}
K~C Ng and Beresford~N. Parlett.
\newblock Programs to swap diagonal blocks.
\newblock 1987.

\bibitem{Dongarra1992}
Jack~J. Dongarra, Sven Hammarling, and James~H. Wilkinson.
\newblock Numerical {{Considerations}} in {{Computing Invariant Subspaces}}.
\newblock {\em SIAM Journal on Matrix Analysis and Applications},
  13(1):145--161, January 1992.

\bibitem{Bai1993}
Zhaojun Bai and James~W. Demmel.
\newblock On swapping diagonal blocks in real {{Schur}} form.
\newblock {\em Linear Algebra and its Applications}, 186:75--95, June 1993.

\bibitem{Granat2009}
R.~Granat, B.~K{\aa}gstr{\"o}m, and D.~Kressner.
\newblock Parallel eigenvalue reordering in real {{Schur}} forms.
\newblock {\em Concurrency and Computation: Practice and Experience},
  21(9):1225--1250, June 2009.

\bibitem{Brandts2002}
J.~H. Brandts.
\newblock Matlab code for sorting real {{Schur}} forms.
\newblock {\em Numerical Linear Algebra with Applications}, 9(3):249--261,
  April 2002.

\bibitem{Agaev2005}
Rafig Agaev and Pavel Chebotarev.
\newblock On the spectra of nonsymmetric {{Laplacian}} matrices.
\newblock {\em Linear Algebra and its Applications}, 399:157--168, April 2005.

\bibitem{Caughman2006}
J.~S. Caughman and J.~J.~P. Veerman.
\newblock Kernels of {{Directed Graph Laplacians}}.
\newblock {\em The Electronic Journal of Combinatorics}, 13(1):R39, April 2006.

\bibitem{Li2012}
Yanhua Li and Zhi-Li Zhang.
\newblock Digraph {{Laplacian}} and the {{Degree}} of {{Asymmetry}}.
\newblock {\em Internet Mathematics}, 8(4):381--401, December 2012.

\bibitem{Singh2016}
Rahul Singh, Abhishek Chakraborty, and B.~S. Manoj.
\newblock Graph {{Fourier}} transform based on directed {{Laplacian}}.
\newblock In {\em 2016 {{International Conference}} on {{Signal Processing}}
  and {{Communications}} ({{SPCOM}})}, pages 1--5, {Bangalore, India}, June
  2016. {IEEE}.

\bibitem{Chung2005}
Fan Chung.
\newblock Laplacians and the {{Cheeger Inequality}} for {{Directed Graphs}}.
\newblock {\em Annals of Combinatorics}, 9(1):1--19, April 2005.

\bibitem{Pentney2005}
William Pentney and Marina Meila.
\newblock Spectral clustering of biological sequence data.
\newblock In {\em Proceedings of the 20th National Conference on Artificial
  Intelligence - Volume 2}, {{AAAI}}'05, pages 845--850, {Pittsburgh,
  Pennsylvania}, 2005. {AAAI Press}.

\bibitem{Mahadevan2006}
Sridhar Mahadevan, Mauro Maggioni, Kimberly Ferguson, and Sarah Osentoski.
\newblock Learning representation and control in continuous markov decision
  processes.
\newblock In {\em Proceedings of the 21st National Conference on Artificial
  Intelligence - Volume 2}, {{AAAI}}'06, pages 1194--1199, {Boston,
  Massachusetts}, 2006. {AAAI Press}.

\bibitem{Chen2007}
Mo~Chen, Qiong Yang, and Xiaoou Tang.
\newblock Directed graph embedding.
\newblock {{IJCAI International Joint Conference}} on {{Artificial
  Intelligence}}, pages 2707--2712, January 2007.

\bibitem{Joncas2011}
Dominique~C. {Perrault-Joncas} and Marina Meil{\u a}.
\newblock Directed graph embedding: {{An}} algorithm based on continuous limits
  of laplacian-type operators.
\newblock In {\em Proceedings of the 24th International Conference on Neural
  Information Processing Systems}, {{NIPS}}'11, pages 990--998, {Red Hook, NY,
  USA}, 2011. {Curran Associates Inc.}

\bibitem{Page1999}
Lawrence Page, S.~Brin, R.~Motwani, and T.~Winograd.
\newblock The {{PageRank}} citation ranking : {{Bringing}} order to the web.
\newblock In {\em {{WWW}} 1999}, 1999.

\bibitem{Franceschet2011}
Massimo Franceschet.
\newblock {{PageRank}}: Standing on the shoulders of giants.
\newblock {\em Communications of the ACM}, 54(6):92--101, June 2011.

\bibitem{Berkhin2005}
Pavel Berkhin.
\newblock A {{Survey}} on {{PageRank Computing}}.
\newblock {\em Internet Mathematics}, 2(1):73--120, January 2005.

\end{thebibliography}


\end{document}